\documentclass[times,review,10pt]{elsarticle}

\biboptions{sort&compress}

\usepackage{amsmath}
\usepackage{amssymb}
\usepackage{amsthm}
\usepackage{array}
\usepackage{booktabs}
\usepackage{graphicx}
\usepackage{xcolor}
\usepackage{stfloats}
\usepackage{algorithm}
\usepackage{algorithmic}
\usepackage{url}
\usepackage{verbatim}
\usepackage{textcomp}
\usepackage{placeins}
\usepackage[caption=false,font=normalsize,labelfont=sf,textfont=sf]{subfig}
\usepackage[hidelinks]{hyperref}
\renewcommand{\topfraction}{0.85}
\renewcommand{\bottomfraction}{0.6}
\renewcommand{\textfraction}{0.1}
\renewcommand{\floatpagefraction}{0.75}
\renewcommand{\dbltopfraction}{0.85}
\renewcommand{\dblfloatpagefraction}{0.75}
\usepackage{etoolbox}
\AtBeginEnvironment{algorithm}{\linespread{1}\selectfont}

\journal{Pattern Recognition}

\newcommand{\var}{\mathsf{Var}}

\begin{document}

\begin{frontmatter}

\title{Unsupervised Feature Selection Using Local Intrinsic Dimensionality -- Extended Version}

\author[1]{Oussama Chelly}
\author[2]{Michael E. Houle}
\author[3,4]{Ken-ichi Kawarabayashi}
\author[5]{Muhammad Rajabinasab\corref{cor1}}
\ead{rajabinasab@imada.sdu.dk}
\author[5]{Arthur Zimek}

\cortext[cor1]{Corresponding author}

\address[1]{Oratio Technologies}
\address[2]{New Jersey Institute of Technology, United States}
\address[3]{National Institute of Informatics, Japan}
\address[4]{University of Tokyo, Japan}
\address[5]{University of Southern Denmark, Denmark}

\begin{abstract}
High dimensionality degrades both the efficiency and the effectiveness of learning algorithms, and in many applications no class labels are available to guide dimensionality reduction. This paper proposes LIDFS, an unsupervised feature selection method based on Local Intrinsic Dimensionality (LID). The method builds on the formal equivalence between LID and the indiscriminability of distance measures. The data matrix is transposed so that each feature is treated as a point, and the LID of each feature is estimated by maximum likelihood from the distances to its $k$ nearest neighboring features. Features with lower LID exhibit greater local discriminability and receive higher importance scores. A refinement step then retains one representative per feature neighborhood and demotes locally redundant features. The result is a complete ranking from which any number of features can be selected without rerunning the procedure. LIDFS does not use any information about labels or the number of classes, and runs in $\mathcal{O}(d^2 n)$ time for $n$ instances and $d$ features. LIDFS is compared with classic and state-of-the-art methods and a random baseline on 23 high-dimensional datasets, the classic benchmark collection. The evaluation uses classification accuracy and AUC with five downstream classifiers, as well as clustering accuracy and normalized mutual information, over both full and aggressive selection ranges. LIDFS ranks first in three of the four supervised evaluations and second in the remaining one, and no competitor is significantly better in the clustering-based evaluations. It belongs to the top statistical group together with methods that require the number of classes, and it scales considerably better than its closest competitors with the number of instances. Several recent methods, by contrast, fail to outperform the random baseline. A sensitivity analysis shows that small neighborhood sizes yield the best results.
\end{abstract}


\end{frontmatter}

\section{Introduction}
\label{s:intro}
Advances in computer technology
are enabling the handling of increasingly complex data.
The scale of data can be measured with respect to  
volume (the number of data instances) 
and dimensionality (the number of features that describe these instances).
When dealing with colossal data volume 
on high-capacity computing platforms, 
volume reduction techniques such as
sampling, 
together with parallelization, 
can allow the exploration and analysis of data with millions or even
billions of entries.

In many pattern recognition applications, such as image and speech recognition, gene expression analysis, and text categorization, each instance is described by thousands of features, while labeled examples are scarce or unavailable.
Feature reduction, the dimensional analogue of volume reduction, 
has several motivations~\citep{guyon}:
reducing the costs of data collection, storage, and processing;
improving the model generalizability and interpretability;
and improving the discriminative power of
feature-based distance measures. 
In general, high dimensionality is associated
with a degradation in the efficiency
and effectiveness of fundamental data mining and machine learning tasks
such as clustering and anomaly detection~\citep{enns}, an effect
often referred to as the `curse of dimensionality'.
As the dimensionality rises,
distance values between points tend to concentrate 
around their mean values 
due to the effects of noise and other sources of error, 
which drastically diminishes the
discriminative power of the distance measure. 

Feature reduction methods fall into two main categories:
selection and extraction.
Feature selection, as the name suggests, retains a
subset of the original features,
chosen with the aim of improving the performance and efficiency
of data mining and machine learning tasks \citep{rajabinasab2024}.
In feature extraction, 
the attributes are transformed into a smaller artificial feature space, with each new feature depending
on many (or even all) of the original features.
Many feature extraction algorithms are based on
PCA~\citep{Cam03} 
or LDA.

Where extraction methods are appropriate, extracted features are likely to outperform an equal number of selected features, due to the greater flexibility of the data transformation. However, in some contexts where feature semantics are of major importance and where artificial features are meaningless, feature extraction cannot be used. Such applications are common in bioinformatics~\citep{fsinbioinformatics} (gene annotation, micro-array analysis) and chemistry~\citep{fsinbioinformatics} (mass spectra analysis, interpretability of molecular signatures), among others. In such contexts, feature selection can still be applied.

Even when feature extraction methods are applicable, 
they are generally not suitable for very large datasets,
as the transformation of the full dataset 
to a new full-dimensional basis can incur a prohibitively high computational cost.  
This computational complexity, together with a lack of robustness  in the presence of noise~\citep{robustPCA},
makes feature extraction less useful when the instances
outnumber the features.
Furthermore, feature extraction has limited use
when the goal is to remove irrelevant features, since all of the original
features must be retained in order to compute values within the transformed feature space. Feature extraction also reduces the explainability of the downstream task, as results that rely on a transformed space are difficult to interpret.


In this paper, we consider the case of unsupervised feature selection, where class label information is unavailable.  Unsupervised feature selection 
is a more difficult problem than supervised feature selection, due to the difficulty in evaluating the quality of features. In the absence of ground truth information, there are two main types of quality criteria that are commonly  used to guide the selection process:
redundancy elimination, where an attribute is discarded whenever it can be entirely or partially inferred from other attributes, and similarity conservation, in which a feature is selected according to the degree
to which it helps to conserve the similarity between data points. Of the two, similarity conservation has received less attention in the literature.

Depending on their evaluation strategy,
feature selection frameworks can be categorized as 
either filters or wrappers (or embedded).
Wrapper methods execute a predictive learning task over the data for many choices of feature subsets,
and retain candidate features for which the predictive model achieves high performance~\citep{guyon}.
For data of even moderate volume or dimension, wrappers quickly become prohibitively expensive. A more affordable alternative is that of filter methods, which evaluate feature quality 
without resorting to an external learning machine.
Although filters generally cannot match the quality of wrappers on small-scale datasets, on datasets of larger scale filters generally require far less computation time,
and are less susceptible to overfitting.
Many unsupervised feature selection algorithms  (such as LapAOFS / LapDOFS~\citep{LapAOFS}) require that a target number of features be supplied as an input parameter, while others automatically determine the number of selected features. With the former, increasing the number of features usually requires rerunning the entire selection; with the latter, the user has no control over the number of selected features. For large datasets, a more convenient alternative is to rank all features, so that downstream processes can select the number of features most appropriate to the task and the available computational resources. The most widely known unsupervised feature ranking framework is that of spectral feature selection~\citep{spectral}, which includes Laplacian Score for Feature Selection (LS)~\citep{laplacianscore}. LS attempts to produce a reduced feature set that preserves the original neighborhood information to the greatest possible extent.
Further work on filters for unsupervised feature ranking includes Multi-Cluster Feature Selection (MCFS)~\citep{mcfs},
which attempts to preserve the cluster structure of the data.
Recent embedded methods combine graph learning, subspace learning, and sparsity-inducing norms. Wang et al.~\citep{wang2026dualgraph} integrate manifold information through a dual-graph clustering framework and constrain the feature transformation matrix with the $\ell_{2,1-2}$-norm and a minimum-redundancy term, while Miao et al.~\citep{miao2026redundancy} propose a redundancy-aware sparse subspace learning approach that uses a variance-covariance distance as the loss function and the $\ell_{2,1}$-norm to promote sparsity. Both approaches learn the feature weights through iterative optimization, in contrast to the filter strategy followed in this paper.

Most existing filter methods are suitable
only for datasets of relatively small volume and dimensionality,
such as those in bioinformatics.
In this work, we target the problem of unsupervised feature
ranking at higher scales, where the data volume and data dimension
together preclude the use of wrapper methods
and computationally expensive filters.
Such feature selection could be of particular benefit for pattern recognition tasks such as similarity search,
neighborhood-based classification, and clustering.

As a criterion for guiding the selection process, we propose that
features be assessed according to their ability to discriminate between
the distances encountered in the neighborhoods of data points, 
in an attempt to alleviate
the effects of the curse of dimensionality.
Accordingly, we develop a feature ranking method that assesses the discriminative power of individual features
according to a model of the local intrinsic dimensionality
(LID) of continuous distance distributions~\citep{5d}, which has an extreme-value-theoretic foundation~\citep{LID1sisap17} and can be estimated efficiently from neighborhood distances~\citep{kdd2015estimation}.
While most feature selection methods employ indirect strategies
in order to (implicitly) improve the discriminability of distance measures,
the use of LID as guiding criterion allows for a more explicit approach,
due to a formal equivalence \citep{5d} between
the discriminability of distance measures and
low intrinsic dimensionality.
Taking into account larger data scales in both volume and dimensionality,
we develop a forward filter algorithm applicable to continuous-valued datasets.
The resulting method, LIDFS, requires neither labels nor the number of classes, produces a complete feature ranking in a single run, from which any number of features can be selected, and runs in $\mathcal{O}(d^2 n)$ time for $n$ instances and $d$ features.

The specific contributions of this paper include:
\begin{itemize}
\item
a greedy unsupervised feature selection algorithm based on LID\footnote{Python implementation and benchmark code are available at \url{https://github.com/mrajabinasab/LIDFS/}.};
\item
a theoretical analysis and discussion of the proposed method; 
\item
an experimental framework setting 
random feature selection as a baseline 
for feature selection,
and showing the advantage of using our method 
with 23 high-dimensional feature selection benchmark datasets, in comparison with nine competitors, including recent methods from the pattern recognition literature, using five classifiers, two clustering metrics, two selection scenarios, and runtime and sensitivity analyses.
\end{itemize}

The experiments show that LIDFS ranks first in most supervised evaluations, is not significantly outperformed by any competitor in the clustering-based evaluations, and scales considerably better than its closest competitors with the number of instances, while several recent methods fail to outperform the random baseline.

The paper is organized as follows.
In the next section we present a survey of existing methods for feature
selection.
In Section~\ref{s:id}, we introduce the LID model 
upon which our feature ranking method is based, 
as well as its relationship
to the discriminability of distance measures.
In Section~\ref{s:method}, 
we present the details of the proposed unsupervised feature selection method.
The experimental framework is presented in Section~\ref{s:framework},
in which our method is compared against state-of-the-art unsupervised feature selection algorithms.
The experimental results are discussed in Section~\ref{s:res}.
Finally, in Section~\ref{s:conclusion} we conclude by summarizing
the advantages and the limitations
of our feature selection strategy,
and propose possible extensions as future work.
\section{Feature selection}
\label{s:fs}
Feature selection consists of choosing a subset of the original features
so as to best satisfy a quality criterion.
The ultimate goal is to improve model generalizability and interpretability,
as well as to lower the computational costs of learning algorithms.

Formally, let us assume that we are given
a set of $n$ points in a space of dimension $d$
represented by the matrix $X \in M_{n,d}(\mathbb{R})$, $X = ( x_1, \ldots , x_n )^\top = ( f_1, \ldots , f_d )$, where $ x_i \in \mathbb{R}^d $ are points and $ f_i \in \mathbb{R}^n $ are features. 
Feature selection finds $\Omega' = \{f'_1,\ldots,f'_{d'}\}$, a subset of $\Omega = \{f_1,\ldots,f_d\}$ with $d' < d$ that best satisfies a quality criterion.

Feature selection algorithms can be differentiated 
in several ways. They can be
\begin{itemize}
\item wrappers or filters,
according to whether or not they employ a 
learning task as part of their feature evaluation strategy
(with embedded methods combining the characteristics of both);
\item forward or backward
depending on their search direction,
whether they build a feature set incrementally starting from an empty set,
or cull features incrementally starting from the full set;
\item univariate or multivariate,
according to whether they assess features individually
or in groups.
\end{itemize}

Wrapper methods apply a predictive learning machine to the data 
using the candidate feature subsets; 
the objective function is defined by the performance
of this predictive model on a task for which the feature
selection is targeted~\citep{guyon}.
Wrappers can achieve high performance with the learning machine
used in the selection process.
However, they are model-specific~\citep{guyon},
as features selected using a given learning model are often inadequate for others.
Wrapper methods are also
computationally expensive~\citep{guyon}:
in a wrapper selection process,
assessing the quality of each candidate subset
requires one run of the learning machine,
and one measurement of the quality of its output.
Moreover, wrappers present a risk of overfitting
not only to the predictive model,
but also to the data. In contrast, filter methods use properties of the data to
compare the candidate subsets:
they evaluate candidate feature subsets
in terms of some quality criterion, often
information-theoretic
(MRMR~\citep{mrmr},
Information Gain~\citep{infogain}),
correlation-based
(Correlation-based Feature Selection~\citep{cfs}),
statistical
(t-test and chi-squared~\citep{chi2}),
or based on interclass distances
(Fisher Score~\citep{fisher},
supervised Spectral Feature Selection~\citep{spectral}).
Although they cannot match the performance of wrappers on the learning tasks
used to guide the feature selection process,
filters are not specific to any learning model,
generally require far less computation time,
and are less susceptible to overfitting.
In the context of complex data, the advantages of filters are crucial.

Feature selection algorithms can also be categorized according to the
order in which candidate feature subsets are considered.
Backward elimination starts from the full set of features
and incrementally removes features
that underperform with respect to some criterion. 
The opposite approach, forward selection,
consists of starting with an empty set,
and building up candidate feature subsets through the incremental 
introduction of individual features.
When the number of features is very high, 
examining candidate subsets of smaller cardinality is more affordable. 
In the context of high-dimensional data, 
this argument is largely in favor of forward selection.

Feature selection methods also differ in the way they group features
for assessment:
univariate methods assess features individually and independently,
while multivariate methods consider sets of features 
during the selection process.
Since combinations of features are not considered,
the univariate approach cannot account for redundancy of features.
However, univariate methods have the general advantage of being faster: since they consider attributes separately, 
their complexity is usually linear in the number of features.
Although the multivariate approach is usually more computationally expensive,
in the context of high-dimensional data,
multivariate methods that strictly limit the number of candidates
can be more affordable than univariate ones.

In some cases, a vector
$L = ( l_1, l_2, ... , l_n )$ of class labels associated with $X$ may
also be available.
Methods that use such labels as ground truth are said to be `supervised', and otherwise `unsupervised'.
In general, wrapper methods tend to be supervised (due to the 
need to assess the result on a specific learning task).

In the context of big data, 
where ground truth label information 
is rarely available, 
the complexity argument favors unsupervised feature selection filters
which operate in a forward direction.
Among all existing candidates that fit these criteria,
scalability concerns prevent the use of computationally expensive algorithms
such as LapDOFS and LapAOFS~\citep{LapAOFS}, both of which require time more than quadratic in the number of points
and number of features.
Among all algorithms with 
reasonable scalability, 
the unsupervised general-purpose forward-filter 
Laplacian Score (LS)
feature selection method~\citep{laplacianscore}
is perhaps the most popular.
It belongs to the general framework of spectral
feature selection~\citep{spectral},
and has a computational complexity of $\mathcal{O}(dn^{2})$.
The spectral framework offers three different approaches to scoring features,
with the objective always being to preserve the graph structure 
of the dataset described by a spectral matrix $S$.
Of the three scoring functions available,
the second is used by LS:
\begin{eqnarray}
\mathit{Score}_{\mathrm{LS}}(f_t)
& = &
\sum_{x_i,x_j \in X} \frac{(x_{i,t} - x_{j,t})^2}{\var(f_t)} s_{i,j}, 
\\
\textrm{ where } \:\:\: s_{i,j} 
& = & \delta_{i,j} e^{-\frac{d_{i,j}^2}{\theta}} ;
\end{eqnarray}
$\var(f_t)$ is the variance of the feature $f_t$;
$d_{i,j}$ denotes the Euclidean distance between $x_i$ and $x_j$;
$\delta_{i,j}$ is $1$ when $x_i$ or $x_j$ is among the $k$-nearest neighbors
of the other, and 0 otherwise; 
and $\theta$ represents the variance 
within a set of distance values,
but is not explicitly defined in the original work.
Features are ranked in ascending order of $\mathit{Score}_{\mathrm{LS}}$.

Feature selection can also be guided according to 
how well the new features preserve the cluster structure of the data.
Given a similarity graph represented as a
matrix $W$ with $(W)_{i,j} = \delta_{i,j}$,
and a target number of clusters $c$,
Multi-Cluster Feature Selection (MCFS)~\citep{mcfs}
computes the $c$ eigenvectors $(y_1,\ldots,y_c)$ with the smallest eigenvalues of the generalized eigenproblem
$Ly = \lambda D y$,
where $D$ is the diagonal matrix $(D)_{i,i} = \sum_{j} (W)_{i,j}$, and $L=D-W$.
Least Angle Regression (LAR) is then used to solve the $c$ regression problems
\begin{equation}
\min_{a_i} \| y_i - X a_i \|^2, \:\:\: \textrm{ subject to } \:\:\: \|a_i\|_1 \leq \gamma.
\end{equation}
Finally, features are scored by $\mathit{Score}_{\mathrm{MCFS}}(f_t) = \max_i |a_{i,t}|$.
Features are ranked in descending order of $\mathit{Score}_{\mathrm{MCFS}}$. MCFS has limited use in practice due to its high computational complexity:
$\mathcal{O}(dn^{2}+cd^{3}+cnd^{2})$.


Structured Optimal Graph Feature Selection (SOGFS) \citep{DBLP:conf/aaai/NieZL16} jointly learns a prob\-a\-bil\-is\-tic-neighbor similarity graph and a row-sparse projection matrix, rather than fixing the graph from the raw data beforehand. To keep the graph well-structured, it constrains the Laplacian $\mathbf{L_S}$ to have exactly $c$ connected components, relaxed via Ky Fan's theorem into a trace-minimization term,
\begin{equation}
\sum_{i=1}^{c}\sigma_i(\mathbf{L_S})=\min_{\mathbf{F}\in\mathbb{R}^{n\times c},\,\mathbf{F}^\top\mathbf{F}=\mathbf{I}}\mathrm{Tr}(\mathbf{F}^\top\mathbf{L_S}\mathbf{F}),
\end{equation}
solved by alternating updates of the projection, $\mathbf{F}$, and the graph, at complexity $\mathcal{O}(d^3+n^2m)$, where $m$ is the projection dimensionality.

Local structure Learning and exponentially weighted Sparse Regression Feature Selection (LLSRFS) \citep{DBLP:journals/prl/WANG2024110183} counters the tendency of sparse mapping matrices to overweight already-high-scoring features by reweighting each feature's contribution to the pairwise distance by $w_l^q$,
\begin{equation}
\|\mathbf{x}_i-\mathbf{x}_j\|_{ew}^2=\sum_{l=1}^{d}w_l^{q}(x_{il}-x_{jl})^2,
\end{equation}
where a smaller exponent $q\in(0,1]$ flattens the weight distribution across more features. This is equivalent to $\ell_{2,p}^2$-norm sparse regression under $q=1/p-1/2$, generalizing earlier rescaled-least-squares selectors, at complexity $\mathcal{O}(d^3+n^2d+nd^2)$.

Subspace Clustering unsupervised Feature Selection (SCFS) \citep{DBLP:journals/eaai/ParsaZG20} implicitly learns sample similarity through a self-expressive representation rather than an explicit graph,
\begin{equation}
\min_{\mathbf{G}} \|\mathbf{X}-\mathbf{G}\mathbf{G}^\top\mathbf{X}\|_F^2, \quad \text{s.t. } \mathbf{G}\ge 0,\ \mathbf{G}\mathbf{G}^\top\mathbf{1}=\mathbf{1},
\end{equation}
where the nonnegative $\mathbf{G}$ doubles as a soft clustering assignment, so $\mathbf{G}\mathbf{G}^\top$ encodes pairwise similarity implicitly. This is linked to the features by a sparse regression $\min_{\mathbf{W}} \|\mathbf{X}\mathbf{W}-\mathbf{G}\|_F^2+\beta\|\mathbf{W}\|_{2,1}$, optimized jointly with the first objective, at complexity $\max\{\mathcal{O}(d^3),\mathcal{O}(n^2d)\}$ when $c\ll d$.

Bi-Level Unsupervised Feature Selection (BLUFS)~\citep{liu2027blufs} combines a clustering level and a feature level. At the clustering level, spectral clustering on a normalized $k$-nearest-neighbor similarity matrix $\hat{\mathbf{S}}$ yields continuous pseudo-labels $\mathbf{Y}\in\mathbb{R}^{n\times c}$, and at the feature level, a projection matrix $\mathbf{W}$ is learned by regression onto $\mathbf{Y}$, together with an adaptive similarity graph $\mathbf{P}$ in the projected space,
\begin{equation}
\begin{aligned}
\min_{\mathbf{P},\mathbf{Y},\mathbf{W}}\ & \|\mathbf{X}\mathbf{W}-\mathbf{Y}\|_F^2+\lambda\|\mathbf{W}\|_F^2-\alpha\,\mathrm{Tr}(\mathbf{Y}^\top\hat{\mathbf{S}}\mathbf{Y})+\beta\sum_{i,j=1}^{n}\|\mathbf{W}^\top\mathbf{x}_i-\mathbf{W}^\top\mathbf{x}_j\|_2^2\,p_{ij}+\mu\|\mathbf{P}\|_F^2,\\
\text{s.t. }\ & \|\mathbf{W}\|_{2,0}\le s,\ \mathbf{P}^\top\mathbf{1}=\mathbf{1},\ \mathbf{P}\ge 0,\ \mathbf{Y}^\top\mathbf{Y}=\mathbf{I},
\end{aligned}
\end{equation}
where the $\ell_{2,0}$-norm constraint limits the number of nonzero rows of $\mathbf{W}$, and thus the number of selected features, to $s$. The problem is solved by proximal alternating minimization at complexity $\max\{\mathcal{O}(n^2c),\mathcal{O}(d^3),\mathcal{O}(cdn)\}$.


Variance–Covariance Subspace Distance Feature Selection (VCSDFS) \citep{DBLP:journals/nn/KaramiSTMV23} selects features by minimizing a proper metric distance between the full feature matrix and a candidate sub-matrix,
\begin{equation}
\mathrm{dist}(\mathbf{X},\mathbf{X}_I)=\|\mathbf{XX}^\top-\mathbf{X}_I\mathbf{X}_I^\top\|_F,
\end{equation}
which equals the Frobenius norm of the covariance matrix of the discarded features, so minimizing it discards the least-variance features. This avoids the degeneracy of principal-angle-based distances, which are zero between any matrix and its sub-matrices. The feature weights are learned via a matrix-factorization objective with an inner-product regularizer, at complexity $\mathcal{O}(td^2)$, where $t$ is the number of iterations.

Feature selection with Minkowski weighted $k$-means++ (FS-MWK++) \citep{zhang2025scalableunsupervisedfeatureselection} selects centroids probabilistically using feature relevance estimates derived from the data, and then aggregates feature weights across a range of Minkowski exponents to identify informative features. A scalable variant of the method, namely SFS-MWK++, is based on subsampling. Robust, Adaptive and Flexible Graph (RAFG) \citep{Jiang2024} is a graph-learning framework for unsupervised feature selection. The $L_{2,1}$-norm is imposed on the flexible regression term to alleviate the adverse effects of both noisy features and outliers, and an $L_{2,1}$-norm regularization term is incorporated to ensure that the selected transformation matrix is sufficiently sparse. Several other methods proposed in recent years focus on the unsupervised \emph{multi-view} feature selection problem, for instance by learning a consensus partition from diverse view-specific graphs~\citep{cao2024multi}, or by combining collaborative and discriminative subspace learning across views~\citep{wu2024collaborative}.
\section{Indiscriminability and Local ID}
\label{s:id}


The success of learning algorithms on high-dimensional data is often attributed to a substantial gap between the high representational dimension of the data and its comparatively low intrinsic dimensionality (ID)~\citep{5d,LID1sisap17}. Many feature extraction and feature selection methods implicitly exploit this gap in order to improve discriminability. However, intrinsic dimensionality may vary considerably across different regions or clusters of a dataset. Such local variation in ID is seldom exploited in feature selection.


\subsection{Expansion Dimension and Local ID}
\label{s:expdim}

Many models of ID have been proposed over the last few decades:
for the theoretical analysis and characterization of datasets,
the most popular of these have been 
topological models, 
such as the well-known PCA~\citep{Cam03};
fractal models~\citep{Man67};
covering models~\citep{CG99};
and expansion models~\citep{ged,KargerR02}.
Most of the measures of ID associated with these models
are very expensive to compute, and
are thus of limited use in practice.
These models can be used `globally' across the entire dataset,
or adapted `locally' so as to characterize the local dimensionality
in the vicinity of reference points. 

Global estimates of ID are typically higher than local estimates,
since the number of degrees of freedom required to describe a dataset
is (as a rule) higher than that required to describe its components.
This gap between global and local ID measurements
indicates an opportunity for the latter in guiding 
feature selection to improve the performance of similarity search.
Improvements in the performance of search would have a direct impact
on applications of machine learning, data mining, and databases. 
Among the models of ID mentioned above, expansion models 
have seen use in the design and analysis of similarity search
algorithms, including nearest-neighbor search in growth-restricted metrics~\citep{KargerR02}, dimensional testing for multi-step~\citep{xiguo}, reverse $k$-nearest-neighbor search~\citep{DBLP:journals/pvldb/CasanovaEHKNSZ17},
and outlier detection~\citep{Anderberg}.


The intrinsic dimensionality $m$ in a Euclidean vector space 
can be determined using the ratios of two volume and
two distance measurements.
Consider the concentric balls $B(x,r_1)$ and $B(x,r_2)$
centered at $x$ with radii
$0 < r_1 < r_2$.
Let $\lambda$ be a (Lebesgue) volume measure. The ratio
of volumes leads to a simple closed-form expression for the dimensionality
$d$ of Euclidean spaces:
\begin{equation}
\begin{split}
&\frac{\lambda(B(x,r_2))}{\lambda(B(x,r_1))}
= {\Bigg(\frac{r_2}{r_1}\Bigg)}^m \\
\Longrightarrow\,\,\,\,
&m = \frac{ \ln \lambda(B(x,r_2)) - \ln \lambda(B(x,r_1)) }{ \ln r_2 - \ln r_1 }\,.
\end{split}
\end{equation}
An oracle that accepts two concentric balls as a query, 
and returns the dimension of the space, 
is not immediately useful.
However, if volume is estimated by the number of points
within distance $r$ from a reference point $x$,
then the returned value can serve as a measure of intrinsic dimensionality,
the local expansion dimension (ED)~\citep{ged,KargerR02}. 


The model of intrinsic dimensionality considered here is a variant of ED, in which the notion of volume is generalized to a probability measure, and distances are modeled as a random variable~\citep{5d,LID1sisap17}.
Let $x$ be a reference point,
and $X$ be a random variable expressing
a distance measurement from $X$.
Let $P_X$ be the cumulative distribution function associated with $X$.
For any $r$ such that $P_X(r) > 0$ and $P_X$ is continuously differentiable
in an interval containing $r$, 
the LID of $X$ at distance $r$ is defined, if the limit exists, as
\begin{equation}\label{eq:LID}
\mathrm{LID}_{P_X}(r) 
	\triangleq \!\!\! \lim\limits_{\epsilon \to 0^+} \!\!
	\frac{\mathrm{ln}(P_X((1+\epsilon) r) / P_X(r))}
	{\mathrm{ln}(1+\epsilon)}
= \frac{r \cdot P'_X(r)}{P_X(r)} 
\end{equation}


\subsection{Indiscriminability}
%

A distance measure is `discriminative' 
when an expansion in distance results 
in a relatively small increase in the number of observations.
Discriminability of points improves the efficiency
of data mining and machine learning tasks,
minimizes error and ensures robustness.
Capturing a huge number of points by a small increase
in distance measure greatly increases the computational cost of applications,
and has a negative impact on effectiveness.
Moreover, the relative ranking of points with reference
to an indiscriminative distance measure 
can easily be affected by noise.
When selecting features, it is strongly desirable 
that they be chosen so as to ensure better discriminability.


Formally, 
let $x$ be a reference point
and let $X$ be a random distance variable
with respect to that reference point, as defined in Section~\ref{s:expdim}.
For any distance $r$ from $x$ such that $P_X(r) > 0$,
the indiscriminability of $X$ at $r$ is the following limit, if it exists~\citep{5d,LID1sisap17}:
\begin{equation}
\begin{split}
\mathrm{InDiscr}_{P_X}(r) 
	&\triangleq \lim\limits_{\epsilon \to 0^+} 
	\Bigg(\frac{P_X((1+\epsilon) r) - P_X(r)}{\epsilon \cdot F_R(r)} \Bigg) \\
&= \frac{r \cdot P'_X(r)}{P_X(r)} = \mathrm{LID}_{P_X}(r).
\end{split}
\end{equation}
This definition of indiscriminability is unitless,
does not require the parameters
of the underlying distance distribution,
and coincides precisely with local ID (Eq.~\ref{eq:LID}).



\subsection{Estimation of Local Intrinsic Dimensionality}

As the number of data elements increases,
the distance from a query to its $k$-th nearest neighbor tends to decrease.
For continuous distance distributions,
the radius of the local vicinity within which a query result is determined
tends to zero. Accordingly, the discriminability of
distances to the query should be assessed as the limit $\mathrm{LID}^{*}_{F_R}$
of $\mathrm{LID}_{F_R}(r)$ as $r$ tends to zero, where $F_R$ is the Cumulative Distribution Function (CDF) of the random distance variable $R$, representing distances from the query point to other data objects.
This limit can be estimated using
Extreme Value Theory (EVT), which states that
the tail of the distribution of the random distance variable $R$
can be modeled as a Weibull distribution.
The Weibull distribution is specified by two parameters:
the scale $\sigma$ and the shape $\xi$.
For a sufficiently small neighborhood distance threshold, 
the random distance variable representing the threshold excesses is
Weibull-distributed~\citep{kdd2015estimation}.

Concretely, the distances $(r_1,\ldots,r_k)$ to the $k$-nearest neighbors 
of a reference point $x$ are instances of the random variable $R$.
These instances can be used to estimate the Weibull shape parameter $\xi$.
A maximum likelihood estimator (MLE)
can be used to estimate $\xi$ and therefore
$\mathrm{LID}^{*}_{F_R}$~\citep{ kdd2015estimation}:
\begin{equation}\label{eq:LID_MLE}
\mathrm{\widehat{LID}}_{F_R} = -\frac{1}{\widehat{\xi}} 
= \Bigg( -\frac{1}{k} \sum_{i=1}^{k} \ln\frac{r_i}{r_k} \Bigg) ^{-1}
\: .
\end{equation}

\begin{figure}[!tb]
\centering
\includegraphics[width=0.9\textwidth]{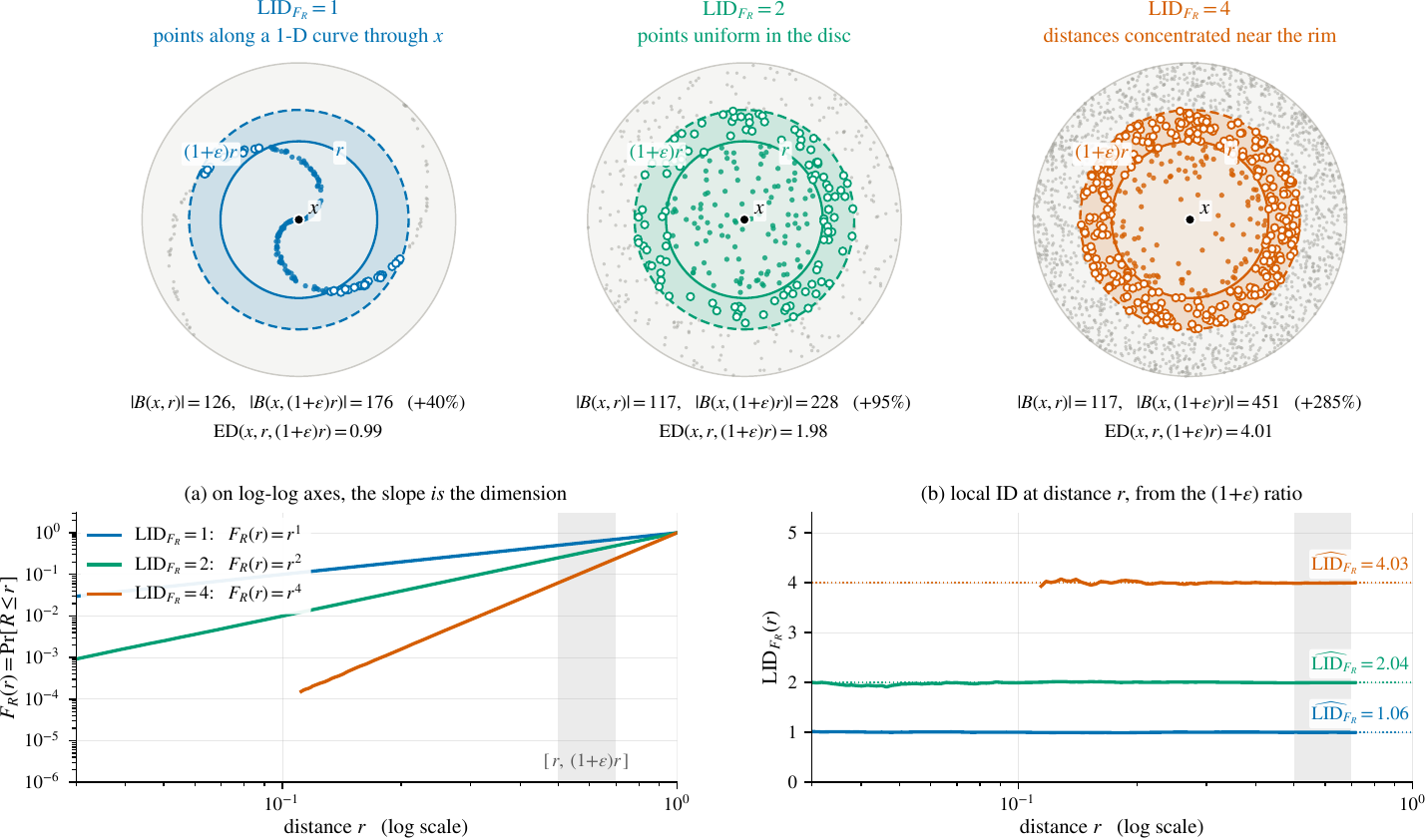}
\caption{LID as neighborhood growth. Top: three point sets sampled so that $F_R(r) = r^{m}$ holds exactly for $m = 1, 2, 4$, which is possible in the plane since LID depends on the distance distribution alone. Each disc starts from a neighborhood of the same size, $|B(x,r)| \approx 120$, and expands the radius by the same factor $1+\epsilon = 1.4$: the expansion dimension over the two concentric balls returns the LID, so it is the relative gain in points, not the absolute gain, that reveals the dimension. (a)~On log-log axes, the slope of $F_R$ is the dimension. (b)~The $(1+\epsilon)$ ratio across distances, with the MLE $\widehat{\mathrm{LID}}_{F_R}$ from $k = 500$ neighbors.}
\label{fg:lid}
\end{figure}

Figure~\ref{fg:lid} illustrates the model on three planar point sets whose distance distributions from the reference point satisfy $F_R(r) = r^{m}$ exactly, for $m \in \{1,2,4\}$. Starting from neighborhoods of equal size, an expansion of the radius by a common factor $1+\epsilon$ admits comparatively few additional points when the local ID is low, and a great many when it is high; the corresponding expansion dimension recovers $d$ in each case, as does the MLE. The rightmost distance measure is thus the least discriminative, even though every case is embedded in the plane.

\section{Method description}
\label{s:method}

For a given reference point, LID quantifies the local rate of neighborhood growth, or equivalently, the difficulty of discriminating among nearby objects, with lower LID indicating greater local discriminability. Because higher local discriminability benefits distance-based learning, our selection strategy applies this principle in the transposed feature space, treating each feature as a point and evaluating it by the discriminability of its distances to neighboring features. A feature's overall quality is then given by the inverse of the LID estimate computed from its $k$-nearest-neighbor distances.

\begin{algorithm}[t!]
\caption{LID Unsupervised Feature Selection}
\label{alg:fs}
\begin{algorithmic}[1]
\REQUIRE Dataset $X \in \mathbb{R}^{n \times d}$ and feature-neighborhood size $k \in [2, d)$.
\ENSURE Feature-importance score vector $S$.
\STATE $X^\top \gets \operatorname{transpose}(X)$
\STATE Remove duplicate rows of $X^\top$.
\STATE For each $i \in \{1,\ldots,d\}$, compute the $k$ nearest neighboring features of $f_i$ in $X^\top$, excluding $f_i$ itself. Store the neighbor distances in $D$, where $D_{i\ell}$ is the distance from $f_i$ to its $\ell$-th nearest neighbor, and store the corresponding feature indices in $I$, where $I_{i\ell}$ is the index of the $\ell$-th nearest neighbor of $f_i$.
\WHILE{$\exists\, i: D_{i1} = D_{ik}$}
\STATE $k \gets k + 1$
\STATE Extend $D$ and $I$ with the $k$-th nearest neighbor of each feature.
\ENDWHILE
\STATE Estimate the LID of each feature and compute its initial score:
$S(f_i) \gets \frac{1}{\widehat{\mathrm{LID}}(f_i;D_{i,:})}$,
$i=1,\ldots,d$.
\STATE Let $\pi=(\pi_1,\pi_2,\ldots,\pi_d)$ be a permutation of $\{1,\ldots,d\}$ such that
$S(f_{\pi_r}) \geq S(f_{\pi_s})$ for all $r<s$.
\STATE Initialize the selected-index set $F \gets \emptyset$ and the reserve-index set $R \gets \emptyset$.
\FOR{$r \gets 1$ to $d$}
    \STATE $i \gets \pi_r$
    \STATE $\mathit{is\_neighbor} \gets \mathrm{false}$
    \FOR{$j \in F$}
        \IF{$i \in \{I_{j1},\ldots,I_{jk}\}$}
            \STATE $\mathit{is\_neighbor} \gets \mathrm{true}$
            \STATE \textbf{break}
        \ENDIF
    \ENDFOR
    \IF{$\mathit{is\_neighbor}$}
        \STATE $R \gets R \cup \{i\}$
    \ELSE
        \STATE $F \gets F \cup \{i\}$
    \ENDIF
\ENDFOR
\IF{$R \neq \emptyset$}
    \STATE $S_{\max} \gets \max_{i \in R} S(f_i)$
    \FOR{$i \in R$}
        \STATE $S(f_i) \gets -(1-S(f_i)/S_{\max})$
    \ENDFOR
\ENDIF
\RETURN $S=\bigl(S(f_1),\ldots,S(f_d)\bigr)$.
\end{algorithmic}
\end{algorithm}

Alg.~\ref{alg:fs} sketches our algorithm.
Let $X=[f_1,\ldots,f_d]\in\mathbb{R}^{n\times d}$ denote the input data matrix, where $n$ is the number of samples, $d$ is the number of features, and each column vector $f_i\in\mathbb{R}^n$ contains the values of the $i$-th feature across all samples. To evaluate relationships among features, LIDFS first transposes the data matrix, yielding ${X^\top\in\mathbb{R}^{d\times n}}$. In this representation, each row corresponds to a feature described by its values across the dataset, so that features with similar value profiles across the samples are close to one another in the transposed space.

Given a user-defined neighborhood size $k$ and a chosen distance measure, the algorithm computes the $k$ nearest neighboring features of each feature $f_i$. The resulting neighbor distances are stored in a matrix $D$, where $D_{i\ell}$ denotes the distance from $f_i$ to its $\ell$-th nearest neighbor. The corresponding neighbor indices are stored in a matrix $I$, where $I_{i\ell}$ identifies the index of the $\ell$-th nearest neighbor of $f_i$. The distances in the $i$-th row of $D$ are then used to estimate the Local Intrinsic Dimensionality of $f_i$, denoted by $\widehat{\mathrm{LID}}(f_i;D_{i,:})$.    Before the neighbor search, duplicate features are removed, since they would result in zero neighbor distances. Moreover, the LID estimate of a feature is undefined if all of its $k$ neighbor distances are equal; in this case, $k$ is increased until every feature has at least two distinct neighbor distances. 

A lower LID estimate indicates a lower relative rate of neighborhood growth and, according to the connection between LID and indiscriminability, greater local discriminability among neighboring features. In the transposed space, this suggests that the feature has a comparatively distinctive value profile across the samples and is less locally redundant with nearby features. Such a feature may therefore have greater potential to preserve distinctions among the points of $X$, although this implication is heuristic rather than guaranteed by the LID estimate alone.
The initial importance score of each feature is therefore defined as the inverse of its estimated LID, $S(f_i)=1/\widehat{\mathrm{LID}}(f_i;D_{i,:})$.
Accordingly, features with lower LID estimates receive higher initial scores.

After computing the initial scores, the features are sorted in descending order of importance. The resulting ordered list is denoted by $\Omega^* = (f^*_1, f^*_2, \ldots, f^*_d)$, where $f^*_i$ is the feature at rank $i$, and $S(f^*_i) \geq S(f^*_j)$ for all $i<j$. A refinement step then reduces redundancy among highly ranked features. Two sets are initialized: the selected-feature set $F$, which stores representative features, and the reserve set $R$, which stores features considered locally redundant with respect to previously selected features.

The algorithm scans $\Omega^*$ from the highest to the lowest-scoring feature. For each candidate feature $f^*_i$, it determines whether $f^*_i$ belongs to the $k$-neighborhood of any feature already included in $F$. If it does not belong to the neighborhood of any selected feature, it is added to $F$ and treated as non-redundant under the neighborhood criterion. Otherwise, it is added to $R$, as it is locally similar to a previously selected feature.

Finally, the scores of the features stored in the reserve set $R$ are adjusted. If $R \neq \varnothing$, let $S_{\max} = \max_{f \in R} S(f)$ denote the largest initial importance score among the reserved features. For each feature $f_r \in R$, the score is updated to $S(f_r) = -\bigl(1 - S(f_r)/S_{\max}\bigr)$.
Since the initial scores are positive and $S(f_r) \leq S_{\max}$, this transformation assigns a score in the interval $(-1,0]$ to every reserved feature while preserving their relative ordering. In particular, the highest-scoring feature in $R$ receives a refined score of $0$, whereas the remaining reserved features receive negative scores. Features in $F$ retain their positive initial scores and therefore precede all features in $R$.

As a result of this adjustment, a complete ranking is produced in which locally non-redundant features are prioritized, while the relative ordering of the reserved features is retained. Consequently, a user may select any desired number of top-ranked features without rerunning the feature-selection procedure. If $R=\varnothing$, no score adjustment is required. The final output of LIDFS is the feature-importance vector $S = \bigl(S(f_1), S(f_2), \ldots, S(f_d)\bigr)$, which is used to rank the features.

LID-based feature-importance scores quantify the local discriminability of features in the transposed feature space. A feature with a lower estimated LID has a more distinctive value profile relative to its neighboring features and is therefore treated as a potentially informative representative feature. The refinement step further prioritizes features from different neighborhoods, reducing the selection of locally redundant features. The neighborhood size $k$ controls the locality of both the LID estimates and the redundancy criterion: smaller values of $k$ emphasize fine-scale neighborhood structure but yield less stable estimates, whereas larger values yield smoother estimates but may merge distinct local structures. The choice of $k$ thus involves a trade-off between locality and estimation stability.

\paragraph{Computational complexity}
The time complexity of LIDFS is analyzed for a dataset with $n$ samples and $d$ features, assuming a distance measure that is computed in $\mathcal{O}(n)$ time for a pair of features, such as the Euclidean distance. Since the removal of duplicate features can only reduce the number of features, $d$ also bounds the number of features that remain after this step. Transposing the data matrix requires $\mathcal{O}(nd)$ time, and duplicate features are removed in $\mathcal{O}(nd)$ expected time by hashing the rows of $X^\top$. The dominant cost is the $k$-nearest-neighbor search in the transposed space. An exact search computes the distances between all pairs of features in $\mathcal{O}(d^2 n)$ time, after which the $k$ nearest neighbors of each feature are obtained by partial selection in $\mathcal{O}(d)$ time and sorted in $\mathcal{O}(k \log k)$ time, resulting in $\mathcal{O}(d^2 n + dk\log k)$ time in total. The loop that increases $k$ is only entered if all $k$ neighbor distances of a feature are equal, in which case the neighbor lists are extended from the already computed distances; even if $k$ grows to $d-1$, this requires at most $\mathcal{O}(d^2 \log d)$ time for sorting all rows of the distance matrix. The maximum-likelihood estimate of LID (Eq.~\ref{eq:LID_MLE})
requires a single pass over the $k$ neighbor distances of a feature, and therefore the initial scores of all features are computed in $\mathcal{O}(dk)$ time. Sorting the features by their scores requires $\mathcal{O}(d \log d)$ time. In the refinement step, the membership test is implemented with a Boolean array that marks the neighbors of all selected features, such that each candidate is checked in $\mathcal{O}(1)$ time and each insertion into $F$ requires $\mathcal{O}(k)$ time, resulting in $\mathcal{O}(dk)$ time for the entire scan.\footnote{A direct implementation that scans $F$ for every candidate requires $\mathcal{O}(d^2 k)$ time in the worst case, which does not exceed $\mathcal{O}(d^2 n)$ for $k \leq n$.} Finally, the adjustment of the scores of the reserved features requires $\mathcal{O}(|R|) \subseteq \mathcal{O}(d)$ time. Consequently, the overall time complexity of LIDFS is
\begin{equation*}
\mathcal{O}\bigl(d^2 n + d^2 \log d + dk \log k\bigr) = \mathcal{O}(d^2 n),
\end{equation*}
where the simplification holds for $\log d \leq n$, which is satisfied by any practical dataset. The running time of LIDFS is therefore quadratic in the number of features and linear in the number of samples, while the neighborhood size $k$ only affects lower-order terms. The space complexity is $\mathcal{O}(nd + d^2)$ if the full distance matrix is stored, which reduces to $\mathcal{O}(nd + dk)$ if the distances are computed in blocks and only the nearest neighbors of each feature are retained. For datasets with a very large number of features, the exact neighbor search can be replaced by an approximate nearest-neighbor search to alleviate the quadratic dependence on $d$.
\section{Experimental framework}
\label{s:framework}
This section describes the setup used to compare LIDFS with other methods. Competitors include traditional methods (Variance-based \citep{laplacianscore}, Correlation-based \citep{cfs, mitra2002unsupervised}, Laplacian Score~\citep{laplacianscore}, and MCFS~\citep{mcfs}) and recent methods (SCFS \citep{DBLP:journals/eaai/ParsaZG20}, SOGFS \citep{DBLP:conf/aaai/NieZL16}, LLSRFS \citep{DBLP:journals/prl/WANG2024110183}, VCSDFS~\citep{DBLP:journals/nn/KaramiSTMV23}, and BLUFS~\citep{liu2027blufs}).

\subsection{Evaluation Metrics}
To 
evaluate the feature selection methods across both supervised and unsupervised downstream tasks, four metrics are used: ACC, AUC, CLSACC, and NMI.

\paragraph{Accuracy (ACC)}
\label{par:acc}
Accuracy is the proportion of correct predictions in classification. For a set of $n$ instances with true labels $l_i$ and predicted labels $\hat{l}_i$, it is calculated as $\text{ACC} = \frac{1}{n} \sum_{i=1}^{n} \mathbb{I}(l_i = \hat{l}_i)$, where $\mathbb{I}(\cdot)$ is the indicator function.

\paragraph{Area Under the Curve (AUC)}
\label{par:auc}
AUC measures the ability of a classifier to distinguish between classes. It is the area under the Receiver Operating Characteristic (ROC) curve, which plots the True Positive Rate (Sensitivity) against the False Positive Rate (1 - Specificity) at various threshold settings. AUC ranges from 0 to 1, with $0.5$ indicating random performance and a higher value indicating better performance. For a given threshold, with $\text{TP}$, $\text{FP}$, $\text{TN}$, and $\text{FN}$ denoting the numbers of true positives, false positives, true negatives, and false negatives, the True Positive Rate ($\text{TPR}$) and False Positive Rate ($\text{FPR}$) are defined as $\text{TPR} = \text{TP}/(\text{TP} + \text{FN})$ and $\text{FPR} = \text{FP}/(\text{FP} + \text{TN})$.

\paragraph{Clustering Accuracy (CLSACC)}
\label{par:clsacc}
Clustering Accuracy assesses how well the clusters found by the algorithm match the true labels of the data. It requires finding the best one-to-one mapping (permutation $\pi$) between the cluster assignments ($c_i$) and the true labels ($l_i$), and is given by $\text{CLSACC} = \frac{1}{n} \sum_{i=1}^{n} \mathbb{I}(l_i = \pi(c_i))$, where $n$ is the number of data points. The optimal mapping $\pi$ is found by the Hungarian algorithm~\citep{kuhn1955hungarian}.

\paragraph{Normalized Mutual Information (NMI)}
\label{par:nmi}
NMI quantifies the degree of dependence between the clustering result ($C$) and the true labels ($L$). It is a normalization of the Mutual Information ($\text{MI}$), making it comparable across cluster numbers: $\text{NMI}(C, L) = 2\,\text{MI}(C, L)/(H(C) + H(L))$, where $H(C)$ and $H(L)$ are the entropies of $C$ and $L$, and
\begin{equation}
\text{MI}(C, L) = \sum_{c \in C} \sum_{l \in L} P(c, l) \log \left( \frac{P(c, l)}{P(c)P(l)} \right)
\label{eq:mi}
\end{equation}
\paragraph{Feature Selection Dynamic Evaluation Metric (FSDEM)}
\label{par:fsdem}
To evaluate the overall feature selection process, we combine FSDEM with the above metrics: 
\begin{equation}
    \label{eq:fsdem}
    \text{FSDEM} = \frac{\int_{a}^{b} g(x) \, dx}{b - a},
\end{equation}
where $g(x)$ approximates an evaluation metric as a function of the percentage $x$ of selected features, and $[a,b]$ is the evaluated range.

\subsection{Datasets}
We conduct 
experiments on 23 high-dimensional datasets from the scikit-feature repository \citep{fs_datasets}. Table~\ref{tbl:hdds} presents an overview of the characteristics of the datasets. The datasets vary in the number of instances, features, and domains. This repository is the standard source for feature selection benchmarks.

\subsection{Evaluation Configuration}
We perform our experiments using the \texttt{FSEVAL} benchmarking suite \citep{rajabinasab2026fsevalfeatureselectionevaluation}, a specialized framework designed for the extensive and comprehensive evaluation of feature selection algorithms. The supervised evaluation is conducted with five downstream classifiers, namely KNN \citep{cover1967nearest}, Decision Tree \citep{DBLP:books/wa/BreimanFOS84}, SVM \citep{svm}, Random Forest \citep{breiman2001random}, and XGBoost \citep{chen2016xgboost} with a 5-fold stratified cross-validation to provide a robust evaluation. For brevity, we present the detailed results for the Random Forest classifier in Section~\ref{s:res}. The results for the other classifiers (available in the Appendix).
lead to the same conclusions. Multiclass AUC is macro-averaged. For CLSACC and NMI, we use the average of 10 runs of the $k$-means algorithm. $k$-means is always given the true number of classes. For the random baseline, the evaluation is repeated 100 times to clearly reflect the standard deviation. The average value is used for further analysis. Methods that require the number of clusters (MCFS, SCFS, SOGFS, LLSRFS, and BLUFS) are given the true number of classes, whereas LIDFS uses no information about the labels. For LIDFS, we use $ k = 5$ as a default setting to calculate LID values for each feature. We favor smaller values for $k$, as fine-grained neighborhoods are deemed more important than the approximation accuracy of the LID. We use Euclidean distance as the chosen distance measure and include random feature selection~\citep{DBLP:journals/corr/abs-2605-22973} as a baseline. Runtime tests ran on an AMD Ryzen 9 9950X (128 GB RAM).

We conduct two different experiments to evaluate the performance of the feature selection process. First, we evaluate the overall feature selection process by selecting from 5\% to 100\% of the features with a step size of 5\%. As the datasets are very high-dimensional, we also conduct a different experiment by selecting from 0.5\% to 10\% of features with a step size of 0.5\%. This allows us to have a more realistic view of the performance of feature selection algorithms to make sure the high number of selected features relative to the number of instances does not obscure the results. The figures of the full-range experiment are presented in \ref{app:full}.
\section{Results}
\label{s:res}
In this section, we present the experimental results based on the described evaluation framework. We aim to evaluate different aspects of unsupervised feature selection algorithms, including efficiency and performance, to comprehensively assess the characteristics of LIDFS against the competitors.

\subsection{Runtime Analysis}

\begin{figure}[!tb]
    \centering
    \includegraphics[width=0.8\textwidth]{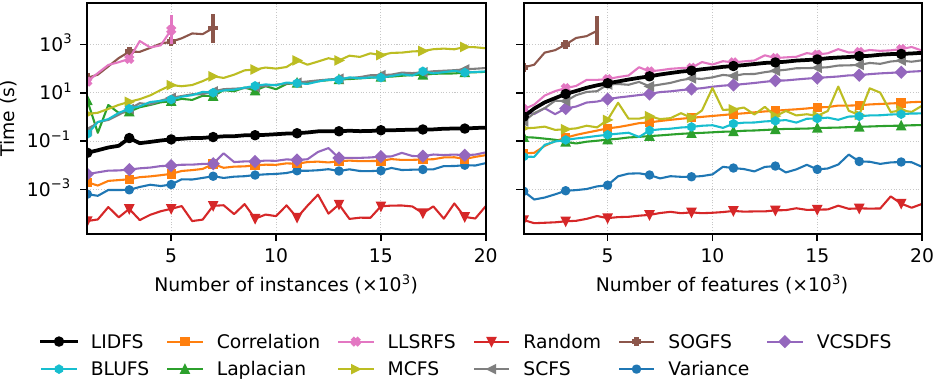} 
    \caption{Runtime analysis of the feature selection methods. The $y$-axis is presented in logarithmic scale.}
    \label{fg:runtime}
\end{figure}

Our first experiment evaluates the runtime of the algorithms, which is essential to understanding the practical significance of their performance. We report two settings, in each of which either the number of instances or the number of features is fixed at 100 while the other varies from 1000 to 20000 in steps of 500. Runtime is capped at one hour per method; the first time a method exceeds this cap, the elapsed time is recorded and no further settings are evaluated for that method. Figure~\ref{fg:runtime} summarizes the results, on a logarithmic $y$-axis.

As expected, the random baseline is by far the most efficient method, as it involves no significant computation. It is followed by traditional methods such as variance-based and correlation-based feature selection. Recent state-of-the-art methods are computationally more expensive. When the number of instances is varied, LIDFS is highly efficient, requiring less than 0.4 seconds for 20000 instances, and the only recent method that is faster is VCSDFS. For the same setting, Laplacian Score, BLUFS, SCFS, and MCFS require between one and twelve minutes. When the number of features is varied, the runtime of LIDFS grows quadratically, in accordance with its $\mathcal{O}(d^2 n)$ time complexity, and only SOGFS and LLSRFS are slower. SOGFS and LLSRFS are the most expensive methods. SOGFS exceeds the time limit in both settings, at 7000 instances and 4500 features, while LLSRFS exceeds it at 5000 instances but completes the setting with varying numbers of features. Nevertheless, both methods are included in the remaining experiments.

\subsection{Performance Analysis}
We compare LIDFS with the competitors on the 23 benchmark datasets and the four evaluation metrics.

\begin{figure}[!tb]
    \centering
    \includegraphics[width=0.75\textwidth]{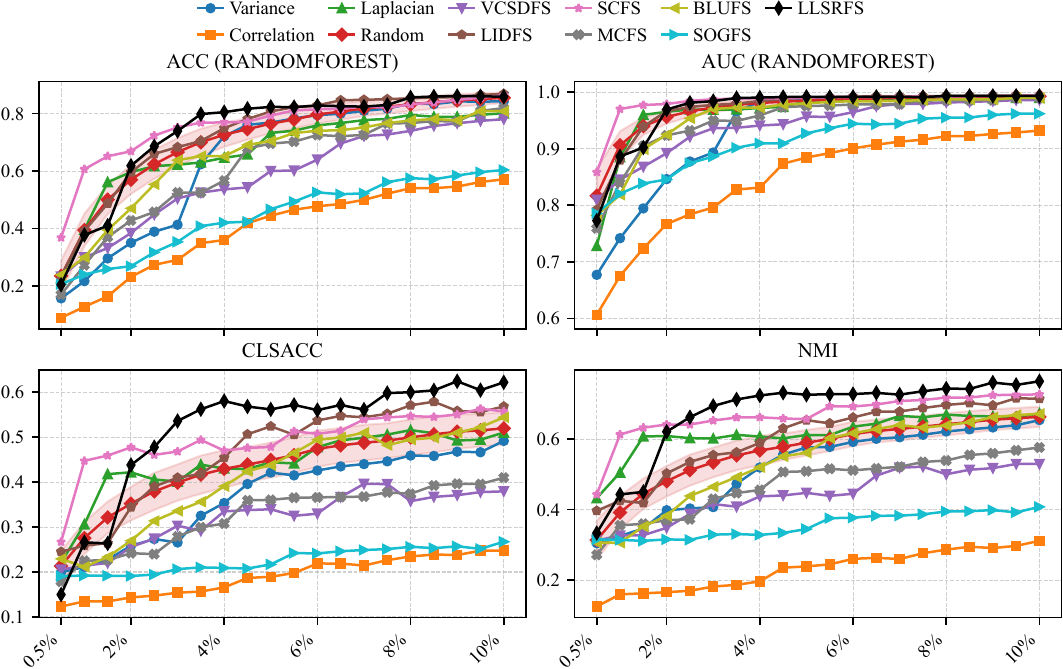} 
    \caption{Comparison of the performance of the unsupervised feature selection methods on the Isolet dataset with the random baseline over the first 10\% of the features.}
    \label{fg:hdline2}
\end{figure}

Figure~\ref{fg:hdline2} presents an example of the four evaluation metrics on the Isolet dataset for the aggressive selection experiment, in which the number of selected features is limited to 10\%; the corresponding results for the full range are presented in Fig.~\ref{fg:hdline1} in \ref{app:full}. For the supervised metrics, LIDFS is among the top-performing methods together with SCFS, while LLSRFS achieves the best clustering performance on this dataset. LIDFS generally outperforms the random baseline. Many feature selection methods, including recent state-of-the-art algorithms, fail to outperform the random baseline. This might seem surprising, but is in fact consistent with previous observations~\citep{DBLP:journals/corr/abs-2605-22973}. The same pattern holds for the full range.

An interesting observation in Fig.~\ref{fg:hdline2} is that, for the top-performing methods, the performance curve mostly flattens after roughly the top $10\%$ of features have been selected. Na\"ive, low-performing methods such as correlation-based feature selection continue to improve up to around the $80\%$ selection mark, or beyond. This is consistent with our assumption in LIDFS: within each feature neighborhood, one feature is designated as the representative and the rest are moved to the reserve set, reflecting their possible redundancy. The results suggest that, with an appropriate $k$, LIDFS identifies this redundancy, assigning positive importance values only to features whose selection is expected to increase the informational value of the lower-dimensional representation. 

\begin{figure}[!tb]
    \centering
    \includegraphics[width=\textwidth]{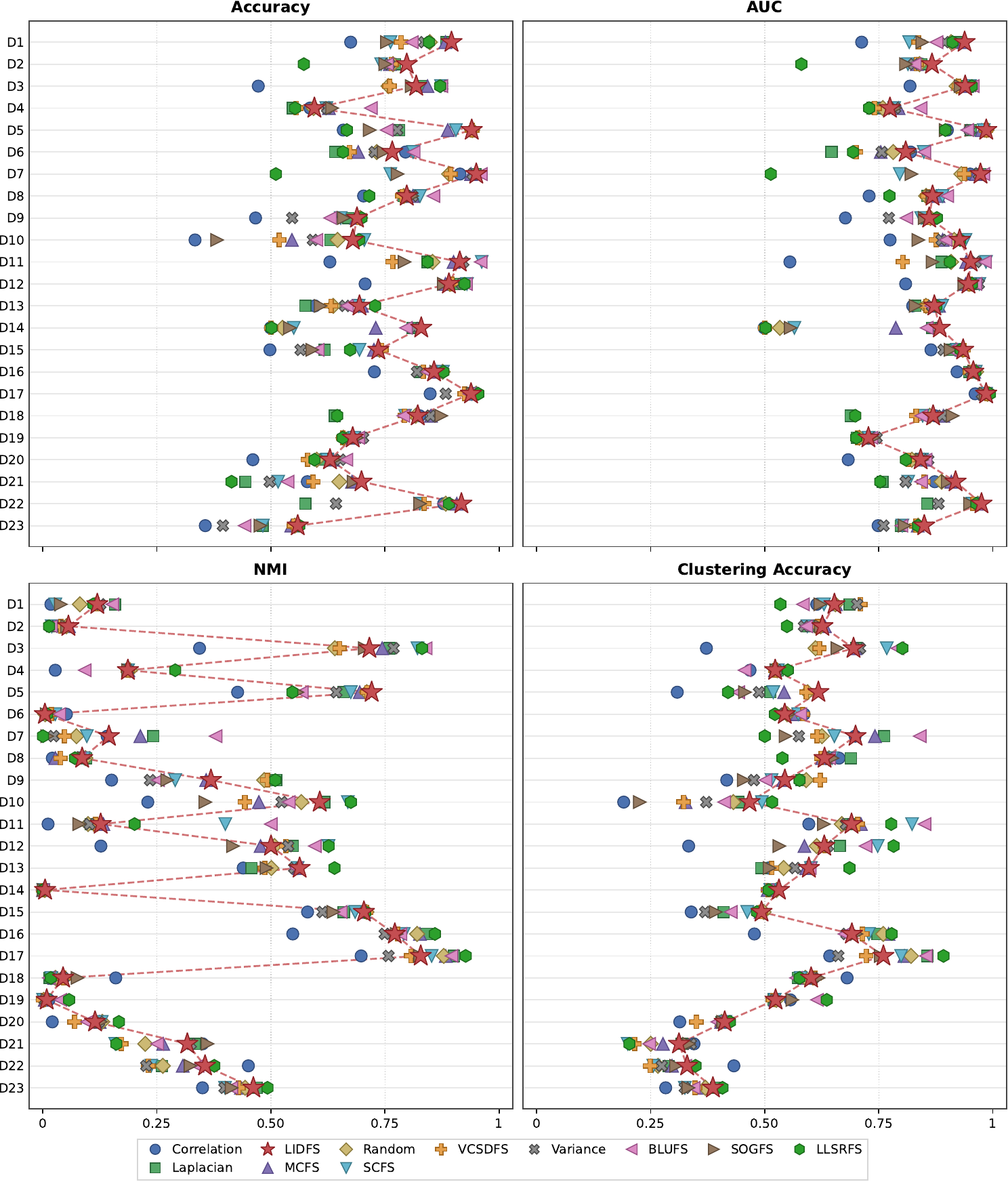}
    \caption{Mean AUC, ACC, NMI, and CLSACC (FSDEM) for the aggressive (10\%) selection scenarios. LIDFS constitutes or lies close to the frontier on most datasets for ACC and AUC.}
    \label{fig:full10}
\end{figure}

\begin{figure}[!tb]
    \centering
    \newcommand{\imgwidth}{0.47\textwidth}
    
    \begin{tabular}{@{}cc@{}}
        \includegraphics[width=\imgwidth]{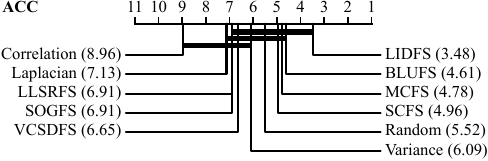} &
        \includegraphics[width=\imgwidth]{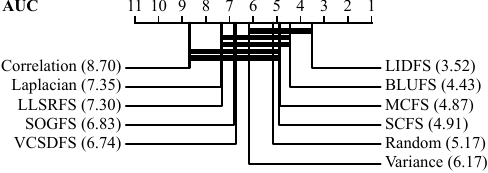} \\
        \rule{0pt}{0.05em} \\
        \includegraphics[width=\imgwidth]{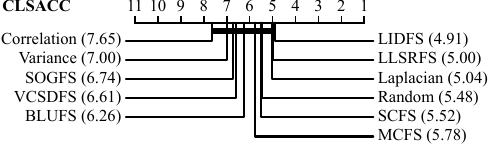} &
        \includegraphics[width=\imgwidth]{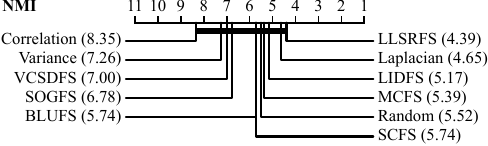} \\
    \end{tabular}
    \caption{Critical difference diagram over the first 10\% of number of features for the four different metrics based on the average performance measured by the FSDEM score.}
    \label{fg:cdd2}
\end{figure}

Feature selection performance varies across datasets; we selected one exhibiting a commonly observed pattern for illustration. Individual results for all datasets are available in the Appendix.
The FSDEM score, which shows the overall performance over a range, is presented in Fig.~\ref{fig:full10} for the aggressive selection scenario and in Fig.~\ref{fig:full100} in \ref{app:full} for the full range. We observe that LIDFS constitutes or lies close to the frontier on most datasets for ACC and AUC, providing a reliable and efficient solution for unsupervised feature selection, while for NMI and CLSACC, LLSRFS forms the frontier most often. To analyze the overall performance across all datasets, we use critical difference diagrams~\citep{demsar2006statistical}, which rank methods by performance from right to left.
The critical difference diagrams for the different metrics, over the full range of feature numbers and the $10\%$ range, are shown in Fig.~\ref{fg:cdd1} in \ref{app:full} and Fig.~\ref{fg:cdd2}, respectively. This analysis uses the FSDEM score~\citep{rajabinasab2024} of each metric.

LIDFS is in most cases the highest-performing method for unsupervised feature selection. For supervised downstream tasks, LIDFS, BLUFS, SCFS, and MCFS form the top statistical group, with LIDFS ranked first in three of the four cases and second, by a negligible margin behind BLUFS, for AUC over the full range. LIDFS is also the only method that significantly outperforms the random baseline, for ACC in both scenarios and for AUC in the aggressive scenario. Other state-of-the-art methods, such as VCSDFS, SOGFS, and LLSRFS, perform even worse than the random baseline. For the unsupervised evaluation, LIDFS is ranked first for CLSACC in the aggressive scenario, second for CLSACC over the full range, and third and fourth for NMI in the aggressive and full scenarios, respectively. LLSRFS ranks first in the other three cases, although it performs worse than the random baseline in the supervised evaluation. None of the differences between LIDFS and the other methods is statistically significant in the unsupervised evaluation. On the efficiency frontier, LIDFS shows a lower runtime compared to the closest competitors, BLUFS and SCFS, when the number of instances grows, while its quadratic dependence on the number of features makes it slower when the number of features grows. It is noteworthy that LIDFS works in a truly unsupervised setting, while most of its closest competitors, including BLUFS, SCFS, MCFS, and LLSRFS, use information extracted from the labels, such as the number of unique classes. The results for the other classifiers are almost identical to those for the Random Forest classifier and lead to the same conclusions; across the five classifiers, the two supervised metrics, and the two scenarios, LIDFS ranks first in 16 of the 20 cases and second in the remaining four.


\subsection{Sensitivity Analysis}
\label{sec:sensitivity}
The neighborhood size $k$ is the central hyperparameter of LIDFS. Theoretically, larger values of $k$ are expected to produce more accurate LID estimates, since the estimator averages over a larger sample. At the same time, a larger $k$ enlarges the neighborhood over which LID is computed possibly beyond local characteristics, reducing the granularity of the discrimination between features. As these two effects act in opposite directions, the net effect of $k$ must be assessed empirically. We evaluate $k \in \{5, 10, 25, 50, 100\}$ on all 23 benchmark datasets under the two feature-selection scenarios, retaining up to 10\% of the original features, and full selection up to 100\%.

\begin{figure}[!tb]
    \centering
    \includegraphics[width=0.85\textwidth]{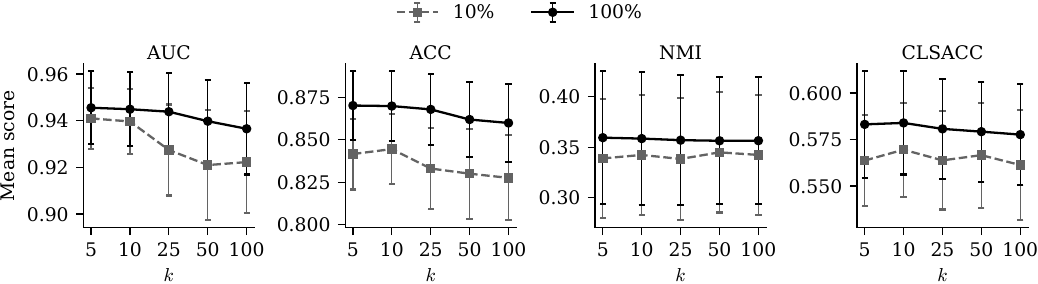}
    \caption{Mean AUC, ACC, NMI, and CLSACC as a function of the neighborhood size $k$, averaged over datasets and over the feature-selection fraction, for the aggressive (10\%) and full (100\%) selection scenarios. Error bars show $\pm 1$ standard error of the mean over the 23 datasets. For all metrics higher values are better.}
    \label{fig:perf-vs-k-colored}
\end{figure}

\paragraph{Downstream performance}
Figure~\ref{fig:perf-vs-k-colored} summarizes different evaluation metrics as a function of $k$, averaged over datasets and over the feature-selection fraction within each scenario. The metrics generally decrease as $k$ increases, which illustrates that the granularity of neighborhoods in the feature selection process matters more than a smooth LID estimate.

\paragraph{Per-dataset heterogeneity and practical guidance}
The per-dataset results are consistent with the aggregate trend in Figure~\ref{fig:perf-vs-k-colored}: smaller values of $k$, which yield finer neighborhoods, generally outperform larger values, which yield more stable LID estimates, and this preference is strongest for AUC and ACC, while NMI and CLSACC are comparatively insensitive to $k$. Nevertheless, no single value of $k$ is optimal on every dataset, and a considerable fraction of the datasets favors larger neighborhoods, so dataset-specific tuning of $k$ remains worthwhile when the computational budget allows. The per-dataset FSDEM scores and the corresponding rank heatmap are provided in \ref{app:full}. In the absence of tuning, our results support using a small $k$ ($k=5$, as in the main experiments) rather than a large neighborhood.

\paragraph{Summary}
The sensitivity analysis was conducted with all classifiers and shows the same trend as for the Random Forest classifier, where small values of $k$, particularly $k=5$ and $k=10$, perform comparably, and larger values lead to a slight degradation in performance. This confirms the theoretical trade-off motivating this analysis: a larger $k$ improves the fidelity of the LID estimate, yet this improvement does not transfer into better downstream performance, as the coarser local granularity mildly degrades all four metrics on average. A small-to-moderate neighborhood size is therefore the recommended default, with per-dataset tuning of $k$ available as an option when the observed heterogeneity across datasets justifies the additional cost.


\section{Conclusion}
\label{s:conclusion}
We proposed a novel unsupervised feature selection algorithm based on LID to compute feature importance values and select informative features without requiring labeled data. We theoretically motivated LID and its suitability for the unsupervised feature selection task, and provided a brief but comprehensive review of progress in the field. We evaluated the proposed method on a large number of datasets against classic and state-of-the-art algorithms and a random baseline, and conducted an in-depth runtime analysis to compare the computational efficiency of the algorithms, and a sensitivity analysis of the neighborhood size $k$.

The experimental results show that the proposed unsupervised feature selection method, LIDFS, offers the best overall observed performance in the supervised evaluation, where it ranks first or second for all five classifiers, and remains competitive in the clustering-based evaluation, where no competitor is significantly better. Its advantage is most pronounced in the aggressive selection scenario, which is the most relevant one for high-dimensional data in practice. The superiority of LIDFS, while not statistically significant with respect to its closest competitors, is remarkable in particular as LIDFS scales considerably better than the closest competitors (e.g., BLUFS and SCFS) with the number of instances. These observations highlight the effectiveness and efficiency of LIDFS as a truly unsupervised method that, unlike some competitors (e.g., SCFS, MCFS), uses no information from the labels.

\subsection{Discussion}

LIDFS differs from most other methods in that it uses neither information-theoretic nor spectral measures. Instead, it directly assesses the discriminability of features via the estimation of LID. One major advantage of LID is that, aside from continuity, it requires no knowledge or assumptions concerning the distribution of the data. The main drawback is that LIDFS requires a relatively smooth distribution of distance values, which is more likely to occur when the data are dense, high-dimensional, or both. Nevertheless, such 
settings are those in which classical feature selection methods often fail, due to high computational complexity or unstable performance, sometimes falling below the random baseline.

The underlying logic of LIDFS is to assign greater importance, via an inverse relationship with the LID estimate, to features that reflect a lower-dimensional local structure relative to others, and therefore exhibit higher discriminability. It also includes a pruning step that prevents the selection of too many mutually close features, maintaining low redundancy in the selection process. As the experiments show, this discrimination provides a systematic approach to identifying redundancy, an ability tied to the neighborhood size $k$, and is reflected in the strong performance of LIDFS.

The experiments also reveal the limitations of LIDFS. First, its runtime grows quadratically with the number of features, since the $k$-nearest-neighbor search is performed in the transposed space. Methods whose cost is dominated by the number of instances, such as BLUFS and MCFS, are therefore faster on data with many features and few instances, whereas LIDFS is considerably faster when the number of instances grows. Second, the advantage of LIDFS over its closest competitors is not statistically significant, and in the clustering-based evaluation LLSRFS achieves better average ranks, at the price of a much higher computational cost and a performance below the random baseline in the supervised evaluation. Third, the best neighborhood size $k$ varies across datasets, although a small default value performs well on average. Finally, the evaluation is restricted to continuous features, the Euclidean distance, and the benchmark collection of Li et al.~\citep{fs_datasets}; the behavior of LIDFS on categorical or mixed data, with other distance measures, or under heavy noise remains to be studied.

Several aspects of this work can benefit researchers and practitioners in pattern recognition. LIDFS produces a complete feature ranking in a single run, so that any number of features can be selected without rerunning the procedure, which is convenient when the target dimensionality is unknown or must be tuned for a downstream model. As it requires neither labels nor the number of classes, it can be used as a preprocessing step for clustering, similarity search, and anomaly detection, or as a pre-filter that reduces the dimensionality of data with many instances before more expensive embedded methods are applied. The LID-based criterion also provides an interpretable notion of feature quality, as it relates the importance of a feature directly to the discriminability of distances in its neighborhood. Moreover, the results show that random feature selection is a strong reference point that several recent methods fail to outperform; we therefore recommend including such a baseline in future evaluations of unsupervised feature selection. The implementation of LIDFS and the benchmark code are publicly available to support reproducibility and further comparisons.

\subsection{Future work}

As data volume increases, the cost of building an exact index becomes prohibitive, motivating the use of approximate nearest-neighbor algorithms such as $\epsilon$-NNS~\citep{enns} or NN-Descent~\citep{NNdescent} in place of exact search. Approximate distances may affect LID estimation and the feature ranking, but trading off feature quality for computation time can be advantageous for large or high-dimensional data.

One possible extension of this approach is to consider other models of intrinsic dimensionality as an alternative to LID for assessing feature quality. Although such models may lack the theoretical guarantees regarding the discriminability of distance measures, or may have higher computational complexity or greater sensitivity to noise~\citep{kdd2015estimation}, they may still merit consideration, particularly if they prove more reliable for non-smooth distance distributions or for smaller data samples.

Several other directions are possible. First, the parameters of ID-based feature selection could be tuned in an unsupervised manner. Also, our approach to unsupervised feature ranking could be extended to supervised applications by restricting the ID estimation to training examples with common labels. Finally, the algorithmic design of LIDFS could be revisited to yield a more granular feature-importance calculation, either through a refinement step that avoids assigning similar importance values within the same neighborhood, or through other approaches independent of $k$. This must be done while maintaining the low computational overhead of the current procedure.

The quadratic dependence on the number of features could be reduced by estimating the distances between features from a subsample of the instances, by random projections, or by restricting the neighborhood search to candidate features obtained from an inexpensive pre-screening. Another direction is to integrate the LID-based criterion into embedded frameworks that jointly learn the clustering structure and the feature weights, such as the multi-subspace graph clustering approach of Cai et al.~\citep{cai2026multisubspace}, where LID-based scores could serve as feature weights or priors. LIDFS could also be extended to multi-view data, where LID estimates from different views may be combined to select features consistently across views, and to streaming settings, in which the feature ranking is updated incrementally as new instances arrive. Finally, the stability of the selected subsets under data perturbations, and the behavior of LIDFS on data with categorical or mixed features under suitable distance measures, deserve further investigation.
\section*{CRediT authorship contribution statement}
\textbf{Oussama Chelly:} Conceptualization, Investigation, Methodology. \textbf{Michael E.\ Houle:} Conceptualization, Formal analysis, Methodology, Supervision, Validation, Writing \textendash{} original draft, Writing \textendash{} review \& editing. \textbf{Ken-ichi Kawarabayashi:} Conceptualization, Investigation, Writing \textendash{} review \& editing. \textbf{Muhammad Rajabinasab:} Conceptualization, Data curation, Formal analysis, Investigation, Methodology, Software, Validation, Visualization, Writing \textendash{} original draft, Writing \textendash{} review \& editing. \textbf{Arthur Zimek:} Conceptualization, Formal analysis, Investigation, Methodology, Project administration, Supervision, Validation, Writing \textendash{} review \& editing.

\section*{Acknowledgments}
This study was funded by the Innovation Fund Denmark project ``PREPARE: Personalized Risk Estimation and Prevention of Cardiovascular Disease''.

\section*{Declaration of generative AI and AI-assisted technologies}
During the preparation of this work, the authors used Generative AI in order to check grammar and spelling, adjust the formatting of the manuscript, and clean up the accompanying code. The tool was not used to generate scientific content, analyses, or results. After using this tool, the authors reviewed and edited the content as needed and take full responsibility for the content of the publication.
\bibliographystyle{elsarticle-num}
\bibliography{ref,pref,gref}

\begin{thebibliography}{10}
\expandafter\ifx\csname url\endcsname\relax
  \def\url#1{\texttt{#1}}\fi
\expandafter\ifx\csname urlprefix\endcsname\relax\def\urlprefix{URL }\fi
\expandafter\ifx\csname href\endcsname\relax
  \def\href#1#2{#2} \def\path#1{#1}\fi

\bibitem{guyon}
I.~Guyon, A.~Elisseeff, An introduction to variable and feature selection, JMLR 3 (2003).

\bibitem{enns}
P.~Indyk, R.~Motwani, Approximate nearest neighbors: towards removing the curse of dimensionality, in: STOC, ACM, 1998.

\bibitem{rajabinasab2024}
M.~Rajabinasab, A.~D. Lautrup, T.~Hyrup, A.~Zimek, A dynamic evaluation metric for feature selection, in: Similarity Search and Applications, Springer Nature Switzerland, Cham, 2025.

\bibitem{Cam03}
F.~Camastra, Data dimensionality estimation methods: a survey, Pattern Recognition 36~(12) (2003).

\bibitem{fsinbioinformatics}
Y.~Saeys, I.~Inza, P.~Larra{\~n}aga, A review of feature selection techniques in bioinformatics, Bioinformatics 23~(19) (2007).

\bibitem{robustPCA}
J.~Feng, H.~Xu, S.~Yan, Robust {PCA} in high-dimension: A deterministic approach, in: ICML, 2012.

\bibitem{LapAOFS}
X.~He, M.~Ji, C.~Zhang, H.~Bao, A variance minimization criterion to feature selection using laplacian regularization, PAMI 33~(10) (2011).

\bibitem{spectral}
Z.~Zhao, H.~Liu, Spectral feature selection for supervised and unsupervised learning, in: ICML, ACM, 2007.

\bibitem{laplacianscore}
X.~He, D.~Cai, P.~Niyogi, Laplacian score for feature selection, in: NIPS, 2005.

\bibitem{mcfs}
D.~Cai, C.~Zhang, X.~He, Unsupervised feature selection for multi-cluster data, in: KDD, ACM, 2010.

\bibitem{wang2026dualgraph}
X.~Wang, R.~Shang, C.~Liu, Y.~Li, S.~Xu, Unsupervised feature selection based on dual-graph clustering learning and adaptive weighting, Pattern Recognition 178 (2026) 113401.
\newblock \href {https://doi.org/10.1016/j.patcog.2026.113401} {\path{doi:10.1016/j.patcog.2026.113401}}.

\bibitem{miao2026redundancy}
J.~Miao, J.~Zhao, T.~Yang, C.~Fan, Y.~Tian, M.~Xu, Sparse subspace learning based redundancy-aware unsupervised feature selection, Pattern Recognition 172 (2026) 112611.
\newblock \href {https://doi.org/10.1016/j.patcog.2025.112611} {\path{doi:10.1016/j.patcog.2025.112611}}.

\bibitem{5d}
M.~E. Houle, {D}imensionality, {D}iscriminability, {D}ensity \& {D}istance {D}istributions, in: ICDMW, IEEE, 2013.

\bibitem{LID1sisap17}
M.~E. Houle, Local intrinsic dimensionality {I}: An extreme-value-theoretic foundation for similarity applications, in: SISAP, 2017.

\bibitem{kdd2015estimation}
L.~Amsaleg, O.~Chelly, T.~Furon, S.~Girard, M.~E. Houle, K.~Kawarabayashi, M.~Nett, Estimating local intrinsic dimensionality, in: KDD, ACM, 2015.

\bibitem{mrmr}
H.~Peng, F.~Long, C.~Ding, Feature selection based on mutual information criteria of max-dependency, max-relevance, and min-redundancy, PAMI 27~(8) (2005).

\bibitem{infogain}
T.~M. Cover, J.~A. Thomas, Elements of Information Theory, 1991.

\bibitem{cfs}
M.~A. Hall, L.~A. Smith, Feature selection for machine learning: Comparing a correlation-based filter approach to the wrapper, in: FLAIRS Conference, 1999.

\bibitem{chi2}
H.~Liu, R.~Setiono, Chi2: Feature selection and discretization of numeric attributes, in: {TAI}, IEEE, 1995.

\bibitem{fisher}
R.~O. Duda, P.~E. Hart, D.~G. Stork, Pattern Classification, 2001.

\bibitem{DBLP:conf/aaai/NieZL16}
F.~Nie, W.~Zhu, X.~Li, Unsupervised feature selection with structured graph optimization, in: {AAAI}, {AAAI} Press, 2016, pp. 1302--1308.

\bibitem{DBLP:journals/prl/WANG2024110183}
C.~Wang, J.~Wang, Z.~Gu, J.-M. Wei, J.~Liu, Unsupervised feature selection by learning exponential weights, Pattern Recognition 148 (2024).
\newblock \href {https://doi.org/https://doi.org/10.1016/j.patcog.2023.110183} {\path{doi:https://doi.org/10.1016/j.patcog.2023.110183}}.

\bibitem{DBLP:journals/eaai/ParsaZG20}
M.~G. Parsa, H.~Zare, M.~Ghatee, Unsupervised feature selection based on adaptive similarity learning and subspace clustering, Eng. Appl. Artif. Intell. 95 (2020).
\newblock \href {https://doi.org/10.1016/J.ENGAPPAI.2020.103855} {\path{doi:10.1016/J.ENGAPPAI.2020.103855}}.

\bibitem{liu2027blufs}
J.~Liu, X.~Ju, X.~Xiu, W.~Liu, Bi-level unsupervised feature selection with $\ell_{2,0}$-norm, Pattern Recognition 181 (2027) 114687.
\newblock \href {https://doi.org/10.1016/j.patcog.2026.114687} {\path{doi:10.1016/j.patcog.2026.114687}}.

\bibitem{DBLP:journals/nn/KaramiSTMV23}
S.~Karami, F.~Saberi{-}Movahed, P.~Tiwari, P.~Marttinen, S.~Vahdati, Unsupervised feature selection based on variance-covariance subspace distance, Neural Networks 166 (2023).
\newblock \href {https://doi.org/10.1016/J.NEUNET.2023.06.018} {\path{doi:10.1016/J.NEUNET.2023.06.018}}.

\bibitem{zhang2025scalableunsupervisedfeatureselection}
X.~Zhang, R.~C. de~Amorim, Scalable unsupervised feature selection via weight stability (2025).
\newblock \href {http://arxiv.org/abs/2506.06114} {\path{arXiv:2506.06114}}.

\bibitem{Jiang2024}
K.~Jiang, T.~Cao, L.~Zhu, Q.~Sun, {Adaptive and flexible $\ell _1$-norm graph embedding for unsupervised feature selection}, Applied Intelligence 54~(22) (2024).
\newblock \href {https://doi.org/10.1007/s10489-024-05760-z} {\path{doi:10.1007/s10489-024-05760-z}}.

\bibitem{cao2024multi}
Z.~Cao, X.~Xie, Y.~Li, Multi-view unsupervised feature selection with consensus partition and diverse graph, Information Sciences 661 (2024).

\bibitem{wu2024collaborative}
J.-S. Wu, Y.~Li, J.-X. Gong, W.~Min, Collaborative and discriminative subspace learning for unsupervised multi-view feature selection, Engineering Applications of Artificial Intelligence 133 (2024).

\bibitem{Man67}
B.~B. Mandelbrot, How long is the coast of britain{?}, Science 156~(3775) (1967).

\bibitem{CG99}
T.~W. Gamelin, R.~E. Greene, Introduction to Topology: Second Edition, Courier Dover Publications, 2013.

\bibitem{ged}
M.~E. Houle, H.~Kashima, M.~Nett, Generalized expansion dimension, in: ICDMW, IEEE, 2012.

\bibitem{KargerR02}
D.~R. Karger, M.~Ruhl, Finding nearest neighbors in growth-restricted metrics, in: STOC, ACM, 2002.

\bibitem{xiguo}
M.~E. Houle, X.~Ma, M.~Nett, V.~Oria, Dimensional testing for multi-step similarity search, in: ICDM, IEEE, 2012.

\bibitem{DBLP:journals/pvldb/CasanovaEHKNSZ17}
G.~Casanova, E.~Englmeier, M.~E. Houle, P.~Kr{\"{o}}ger, M.~Nett, E.~Schubert, A.~Zimek, Dimensional testing for reverse k-nearest neighbor search, Proc. {VLDB} Endow. 10~(7) (2017).

\bibitem{Anderberg}
A.~Anderberg, J.~Bailey, R.~J. G.~B. Campello, M.~E. Houle, H.~O. Marques, M.~Radovanovic, A.~Zimek, Dimensionality-aware outlier detection, in: {SIAM International Conference on Data Mining}, 2024.

\bibitem{mitra2002unsupervised}
P.~Mitra, C.~A. Murty, S.~K. Pal, Unsupervised feature selection using feature similarity, IEEE Transactions on Pattern Analysis and Machine Intelligence 24~(3) (2002).
\newblock \href {https://doi.org/10.1109/34.990132} {\path{doi:10.1109/34.990132}}.

\bibitem{kuhn1955hungarian}
H.~W. Kuhn, {The Hungarian Method for the assignment problem}, Naval Research Logistics Quarterly 2~(1-2) (1955).

\bibitem{fs_datasets}
J.~Li, K.~Cheng, S.~Wang, F.~Morstatter, R.~P. Trevino, J.~Tang, H.~Liu, Feature selection: A data perspective, ACM computing surveys (CSUR) 50~(6) (2017).

\bibitem{rajabinasab2026fsevalfeatureselectionevaluation}
M.~Rajabinasab, A.~Zimek, {FSEVAL:} feature selection evaluation toolbox and dashboard, CoRR abs/2604.18227 (2026).
\newblock \href {http://arxiv.org/abs/2604.18227} {\path{arXiv:2604.18227}}, \href {https://doi.org/10.48550/ARXIV.2604.18227} {\path{doi:10.48550/ARXIV.2604.18227}}.

\bibitem{cover1967nearest}
T.~M. Cover, P.~E. Hart, Nearest neighbor pattern classification, IEEE Transactions on Information Theory 13~(1) (1967).

\bibitem{DBLP:books/wa/BreimanFOS84}
L.~Breiman, J.~H. Friedman, R.~A. Olshen, C.~J. Stone, Classification and Regression Trees, Wadsworth, 1984.

\bibitem{svm}
C.~Cortes, V.~Vapnik, Support-vector networks, Machine learning 20~(3) (1995).

\bibitem{breiman2001random}
L.~Breiman, {Random Forests}, Machine Learning 45~(1) (2001).

\bibitem{chen2016xgboost}
T.~Chen, C.~Guestrin, {XGBoost}: A scalable tree boosting system, in: Proceedings of the 22nd ACM SIGKDD International Conference on Knowledge Discovery and Data Mining, ACM, 2016.

\bibitem{DBLP:journals/corr/abs-2605-22973}
M.~Rajabinasab, M.~E. Houle, O.~Chelly, A.~Zimek, Worse than random: The importance of a baseline for unsupervised feature selection -- extended version, CoRR abs/2605.22973 (2026).
\newblock \href {http://arxiv.org/abs/2605.22973} {\path{arXiv:2605.22973}}, \href {https://doi.org/10.48550/ARXIV.2605.22973} {\path{doi:10.48550/ARXIV.2605.22973}}.

\bibitem{demsar2006statistical}
J.~Demšar, Statistical comparisons of classifiers over multiple data sets, Journal of Machine Learning Research 7 (2006).

\bibitem{NNdescent}
W.~Dong, C.~Moses, K.~Li, Efficient k-nearest neighbor graph construction for generic similarity measures, in: WWW, ACM, 2011.

\bibitem{cai2026multisubspace}
Y.~Cai, H.~Yang, J.~Zhu, F.~Nie, Multi-subspace graph clustering joint dimensionality reduction and feature selection, Pattern Recognition 172 (2026) 112557.
\newblock \href {https://doi.org/10.1016/j.patcog.2025.112557} {\path{doi:10.1016/j.patcog.2025.112557}}.

\end{thebibliography}

\newcommand{\supMainFig}{fg:hdline2}
\newcommand{\supMainFigii}{fg:hdline1}
\newcommand{\supMainCddTen}{fg:cdd2}
\newcommand{\supMainCddHund}{fg:cdd1}
\newcommand{\supdir}{supfigz}
\newcommand{\supcdddir}{supcddz}
\newcommand{\supcddwidth}{0.495\textwidth}
\newcommand{\supfigheight}{0.25\textheight}
\newcommand{\supback}{%
  {\hfill\hyperref[tab:supplot-index]{$\uparrow$\ Back to Table~\ref*{tab:supplot-index}}\par}}
\newcommand{\supbackcdd}{%
  {\hfill\hyperref[tab:supcdd-index]{$\uparrow$\ Back to Table~\ref*{tab:supcdd-index}}\par}}

\clearpage
\appendix 

\makeatletter
\setcounter{topnumber}{100}
\setcounter{bottomnumber}{100}
\setcounter{totalnumber}{100}
\setcounter{dbltopnumber}{100}
\renewcommand{\topfraction}{1}
\renewcommand{\bottomfraction}{1}
\renewcommand{\dbltopfraction}{1}
\renewcommand{\textfraction}{0}
\renewcommand{\floatpagefraction}{0.8}
\renewcommand{\dblfloatpagefraction}{0.8}
\setlength{\@fptop}{0pt}
\setlength{\@fpsep}{12pt}
\setlength{\@fpbot}{0pt plus 1fil}
\renewenvironment{table}{\@float{table}}{\end@float}
\renewenvironment{table*}{\@dblfloat{table}}{\end@dblfloat}
\def\fps@table{tbp}
\makeatother

\section{Results over the full selection range}\label{app:full}

\setcounter{figure}{0}
\setcounter{table}{0}
\renewcommand{\thefigure}{\Alph{section}\arabic{figure}}
\renewcommand{\thetable}{\Alph{section}\arabic{table}}
\renewcommand{\theHfigure}{full.\Alph{section}\arabic{figure}}
\renewcommand{\theHtable}{full.\Alph{section}\arabic{table}}

\begin{figure}[!htbp]
    \centering
    \includegraphics[width=\textwidth]{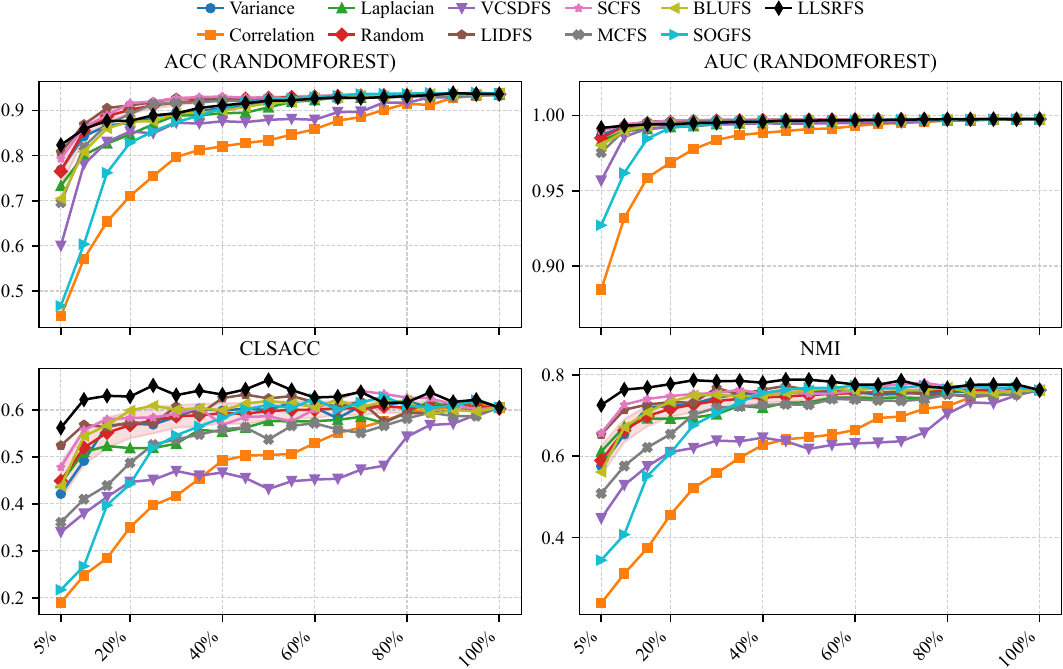} 
    \caption{Comparison of the feature selection performance of the unsupervised feature selection methods on the Isolet dataset with the random baseline over the full range of the number of selected features.}
    \label{fg:hdline1}
\end{figure}

\begin{figure}[!htbp]
    \centering
    \includegraphics[width=\textwidth]{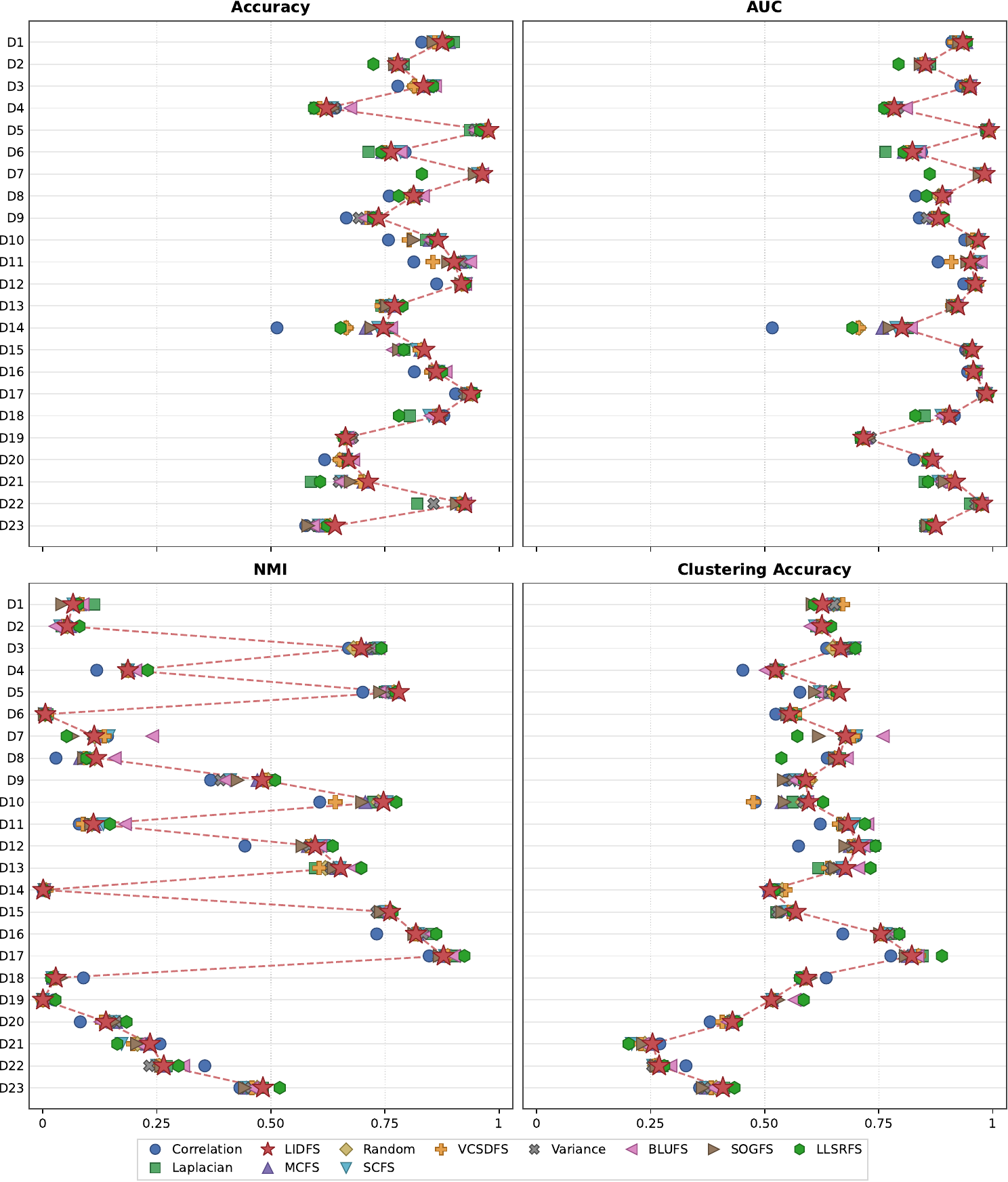}
    \caption{Mean AUC, ACC, NMI, and CLSACC (FSDEM) for the full (100\%) selection scenarios. LIDFS constitutes or lies close to the frontier on most datasets for ACC and AUC.}
    \label{fig:full100}
\end{figure}

\begin{figure}[!htbp]
    \centering
    \newcommand{\imgwidth}{0.47\textwidth}
    
    \begin{tabular}{@{}cc@{}}
        \includegraphics[width=\imgwidth]{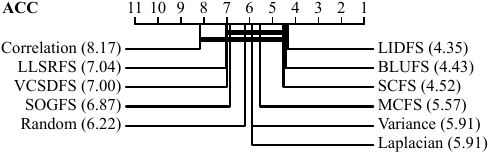} &
        \includegraphics[width=\imgwidth]{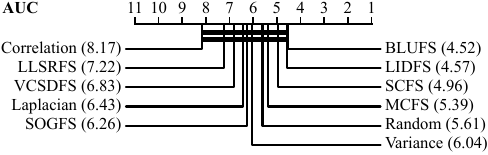} \\
        \rule{0pt}{0.05em} \\
        \includegraphics[width=\imgwidth]{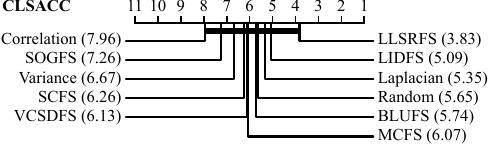} &
        \includegraphics[width=\imgwidth]{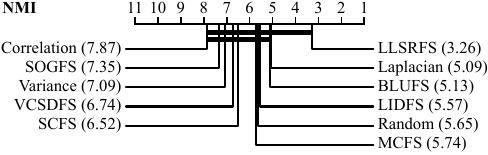} \\
    \end{tabular}
    \caption{Critical difference diagram over the full range of number of features for the four different metrics based on the average performance measured by the FSDEM score.}
    \label{fg:cdd1}
\end{figure}

\begin{figure}[!htbp]
    \centering
    \includegraphics[width=0.9\textwidth]{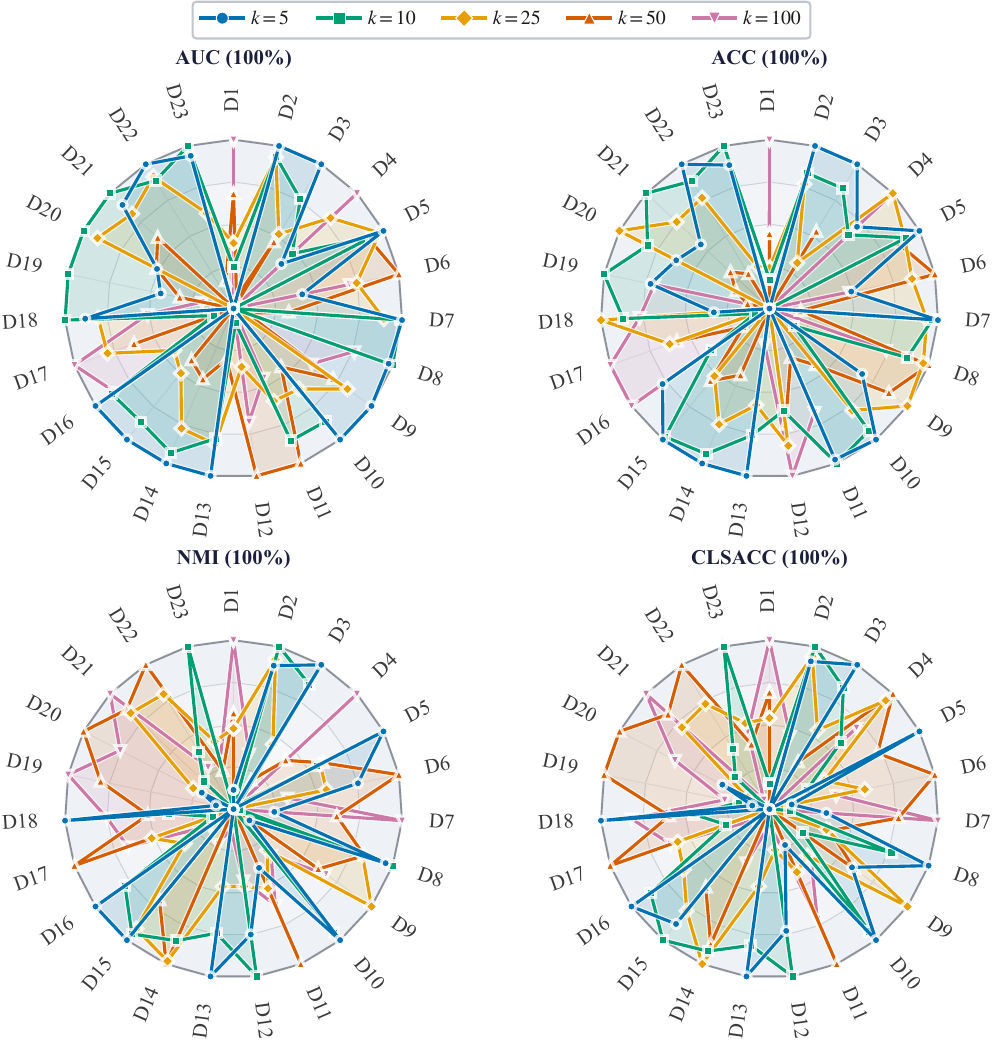}
    \caption{The performance of LIDFS with different choices of $k$ on different datasets and metrics for the full (100\%) selection scenario. Smaller values of $k$ show a better overall performance.}
    \label{fig:radar100}
\end{figure}

\begin{figure}[!htbp]
    \centering
    \includegraphics[width=0.9\textwidth]{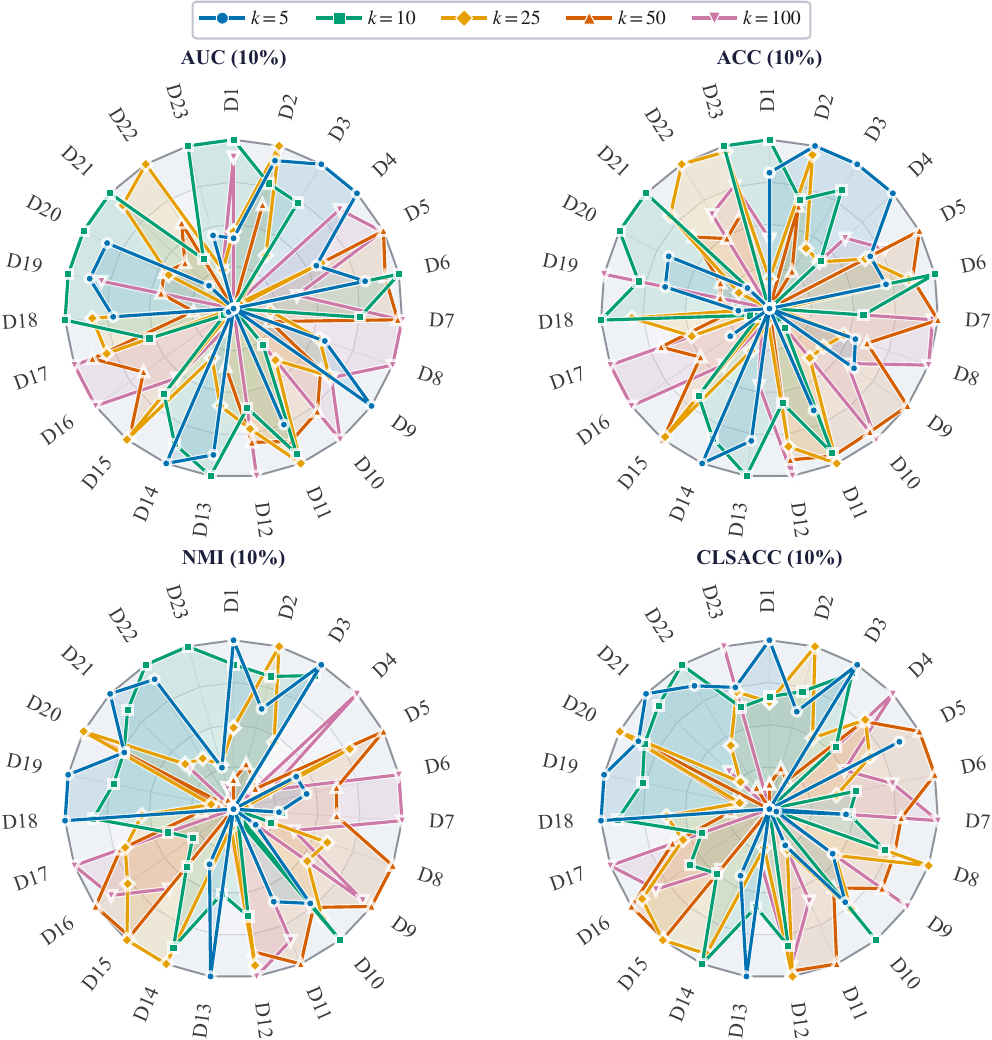}
    \caption{The performance of LIDFS with different choices of $k$ on different datasets and metrics for the aggressive (10\%) selection scenario. Smaller values of $k$ show a better overall performance.}
    \label{fig:radar10}
\end{figure}

\begin{figure}[!htbp]
    \centering
    \includegraphics[width=0.9\textwidth]{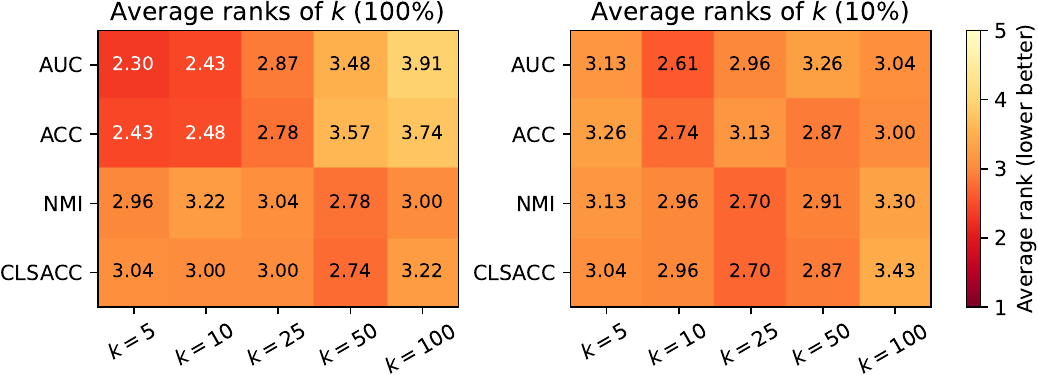}\\
    \caption{Average ranks across the metrics for both scenarios. The preference for small neighborhoods is strongest for AUC and ACC; NMI and CLSACC are comparatively flat.}
    \label{fig:rank-heatmap}
\end{figure}

This appendix presents the figures of the experiment in which 5\% to 100\% of the features are selected, namely the results on the Isolet dataset (Fig.~\ref{fg:hdline1}), the FSDEM scores on all datasets (Fig.~\ref{fig:full100}), the critical difference diagrams (Fig.~\ref{fg:cdd1}), and the sensitivity of LIDFS to the neighborhood size $k$ for both scenarios (Figs.~\ref{fig:radar100} and~\ref{fig:radar10}), as well as the corresponding average ranks (Fig.~\ref{fig:rank-heatmap}). These results are discussed in Section~\ref{s:res}.


\begin{table}[!htbp]
    \centering
    \caption{Characteristics of the high-dimensional benchmark datasets, with $n$ instances and $d$ features.}
    \label{tbl:hdds}
    \setlength{\tabcolsep}{4pt}
    \begin{tabular}{|l|l|r|r||l|l|r|r|}
        \hline
        \textbf{ID} & \textbf{Dataset} & \textbf{$n$} & \textbf{$d$} & \textbf{ID} & \textbf{Dataset} & \textbf{$n$} & \textbf{$d$} \\
        \hline
        D1 & ALLAML & 72 & 7,129 & D13 & lung\_discrete & 73 & 325 \\
        D2 & arcene & 200 & 10,000 & D14 & madelon & 2,600 & 500 \\
        D3 & Carcinom & 174 & 9,182 & D15 & ORL & 400 & 1,024 \\
        D4 & CLL\_SUB\_111 & 111 & 11,340 & D16 & orlraws10P & 100 & 10,304 \\
        D5 & COIL20 & 1,440 & 1,024 & D17 & pixraw10P & 100 & 10,000 \\
        D6 & colon & 62 & 2,000 & D18 & Prostate\_GE & 102 & 5,966 \\
        D7 & gisette & 7,000 & 5,000 & D19 & SMK\_CAN\_187 & 187 & 19,993 \\
        D8 & GLI\_85 & 85 & 22,283 & D20 & TOX\_171 & 171 & 5,748 \\
        D9 & GLIOMA & 50 & 4,434 & D21 & warpAR10P & 130 & 2,400 \\
        D10 & Isolet & 1,560 & 617 & D22 & warpPIE10P & 210 & 2,420 \\
        D11 & leukemia & 72 & 7,070 & D23 & Yale & 165 & 1,024 \\
        D12 & lung & 203 & 3,312 &  & & &  \\
        \hline
    \end{tabular}
\end{table}
\clearpage
\section{Supplementary results}\label{app:supplots}

\setcounter{figure}{0}
\setcounter{table}{0}
\renewcommand{\thefigure}{\Alph{section}\arabic{figure}}
\renewcommand{\thetable}{\Alph{section}\arabic{table}}
\renewcommand{\theHfigure}{supplot.\Alph{section}\arabic{figure}}
\renewcommand{\theHtable}{supplot.\Alph{section}\arabic{table}}

This appendix presents the results that are not included in the main text or in \ref{app:full}. It consists of two parts. \ref{app:supcdd} presents the critical difference diagrams of the supervised metrics (ACC and AUC) for the KNN, decision tree, SVM, and XGBoost classifiers, and \ref{app:supperf} presents the performance plots on all 23 benchmark datasets. All supervised results are reported for each of the five downstream classifiers separately, whereas the unsupervised metrics, namely clustering accuracy (CLSACC) and NMI, do not depend on the classifier and are therefore reported once. The results for the random forest classifier and for the unsupervised metrics are presented in the main text and in \ref{app:full}, and they are not repeated here.

\subsection{Critical difference diagrams}\label{app:supcdd}

The critical difference diagrams are computed in the same way as in the main text, i.e., over all 23 benchmark datasets based on the average performance measured by the FSDEM score, using Wilcoxon-Holm with a significance level of $0.2$. The diagrams of the random forest classifier and of the unsupervised metrics for the first 10\% and for the entire range of the number of selected features are presented in Fig.~\ref{\supMainCddTen} of the main text and Fig.~\ref{\supMainCddHund} in \ref{app:full}, respectively. Each figure below contains the diagrams of ACC (left) and AUC (right) for one classifier and one range, and Table~\ref{tab:supcdd-index} provides an index of all figures, where each entry links to the corresponding figure.

\begin{table}[h!]
\centering
\caption{Index of the supplementary critical difference diagrams. Each entry gives the number of the figure with the diagrams of ACC and AUC for the corresponding classifier and range of the number of selected features (up to 10\% or 100\% of the features), and links to the figure. Entries marked with $^{\dagger}$ refer to the main text (0.5\% to 10\%) or to \ref{app:full} (5\% to 100\%).}
\label{tab:supcdd-index}
\begin{tabular}{lcc}
\toprule
 & \multicolumn{2}{c}{Range of the number of selected features} \\
\cmidrule(lr){2-3}
Classifier / metrics & 0.5\% to 10\% & 5\% to 100\% \\
\midrule
KNN (ACC, AUC) & \ref{fig:supcdd-knn-10} & \ref{fig:supcdd-knn-100} \\
Decision tree (ACC, AUC) & \ref{fig:supcdd-dt-10} & \ref{fig:supcdd-dt-100} \\
SVM (ACC, AUC) & \ref{fig:supcdd-svm-10} & \ref{fig:supcdd-svm-100} \\
Random forest (ACC, AUC) & \ref{\supMainCddTen}$^{\dagger}$ & \ref{\supMainCddHund}$^{\dagger}$ \\
XGBoost (ACC, AUC) & \ref{fig:supcdd-xgb-10} & \ref{fig:supcdd-xgb-100} \\
CLSACC, NMI (classifier independent) & \ref{\supMainCddTen}$^{\dagger}$ & \ref{\supMainCddHund}$^{\dagger}$ \\
\bottomrule
\end{tabular}
\end{table}


\begin{figure}[p]
\centering
\pdfbookmark[3]{KNN}{supcdd-bm-knn}
\includegraphics[width=\supcddwidth]{\supcdddir/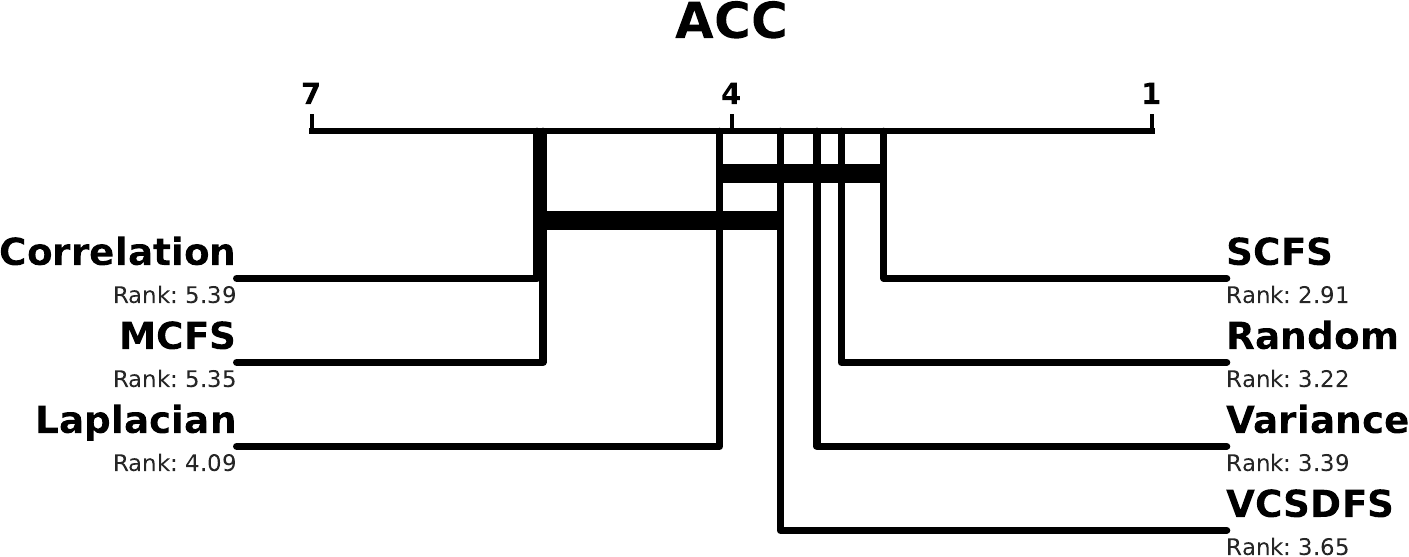}\hfill
\includegraphics[width=\supcddwidth]{\supcdddir/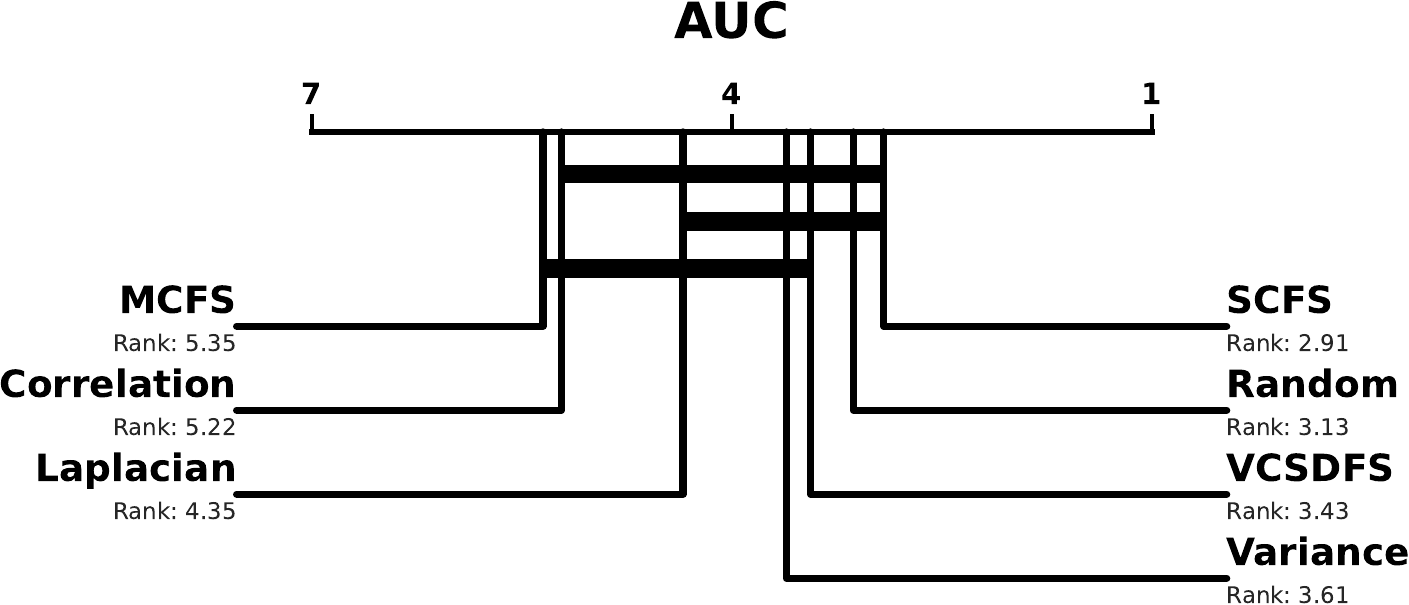}
\caption{Critical difference diagrams of ACC (left) and AUC (right) for the KNN classifier (0.5\% to 10\% of the features).}\label{fig:supcdd-knn-10}
\supbackcdd
\end{figure}
\begin{figure}[p]
\centering
\includegraphics[width=\supcddwidth]{\supcdddir/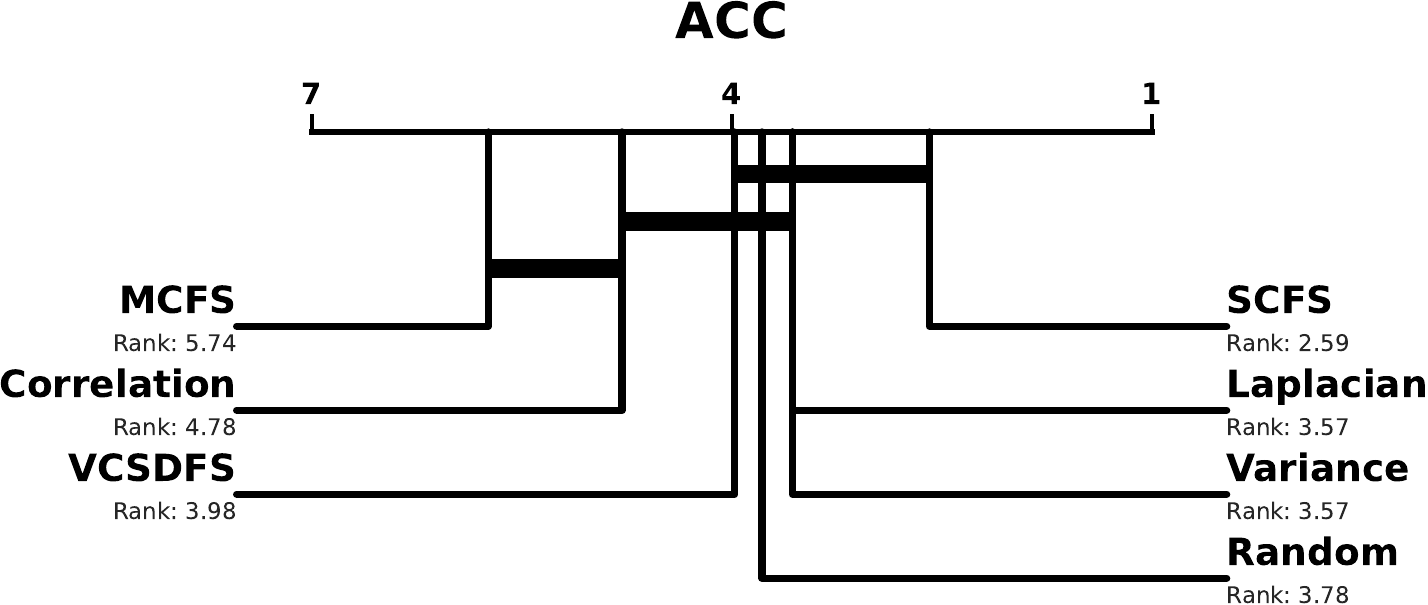}\hfill
\includegraphics[width=\supcddwidth]{\supcdddir/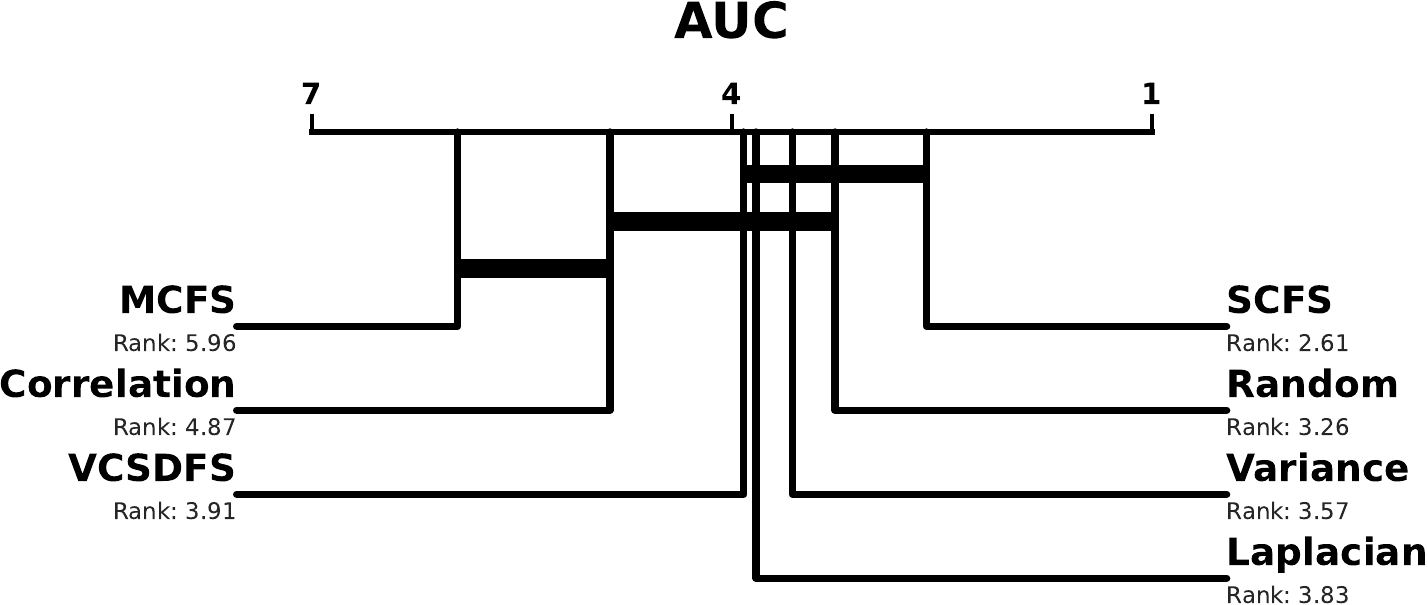}
\caption{Critical difference diagrams of ACC (left) and AUC (right) for the KNN classifier (5\% to 100\% of the features).}\label{fig:supcdd-knn-100}
\supbackcdd
\end{figure}
\begin{figure}[p]
\centering
\pdfbookmark[3]{Decision tree}{supcdd-bm-dt}
\includegraphics[width=\supcddwidth]{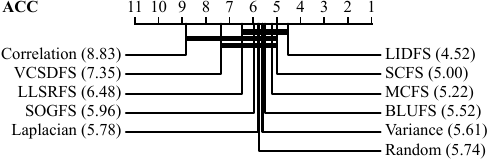}\hfill
\includegraphics[width=\supcddwidth]{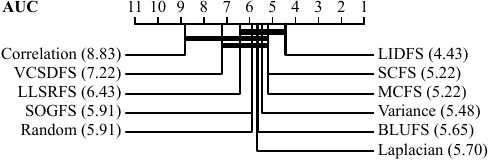}
\caption{Critical difference diagrams of ACC (left) and AUC (right) for the decision tree classifier (0.5\% to 10\% of the features).}\label{fig:supcdd-dt-10}
\supbackcdd
\end{figure}
\begin{figure}[p]
\centering
\includegraphics[width=\supcddwidth]{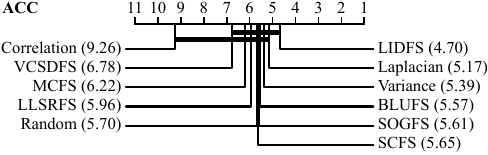}\hfill
\includegraphics[width=\supcddwidth]{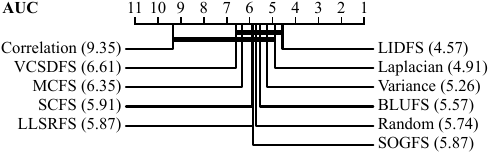}
\caption{Critical difference diagrams of ACC (left) and AUC (right) for the decision tree classifier (5\% to 100\% of the features).}\label{fig:supcdd-dt-100}
\supbackcdd
\end{figure}
\begin{figure}[p]
\centering
\pdfbookmark[3]{SVM}{supcdd-bm-svm}
\includegraphics[width=\supcddwidth]{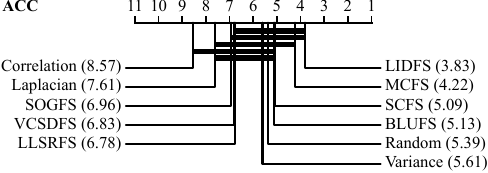}\hfill
\includegraphics[width=\supcddwidth]{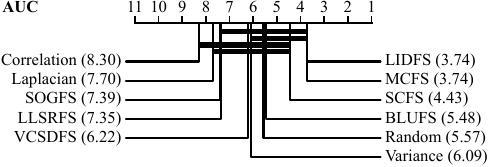}
\caption{Critical difference diagrams of ACC (left) and AUC (right) for the SVM classifier (0.5\% to 10\% of the features).}\label{fig:supcdd-svm-10}
\supbackcdd
\end{figure}
\begin{figure}[p]
\centering
\includegraphics[width=\supcddwidth]{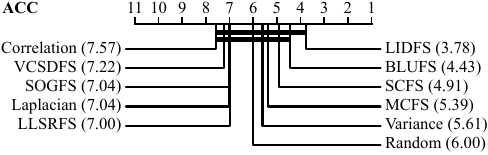}\hfill
\includegraphics[width=\supcddwidth]{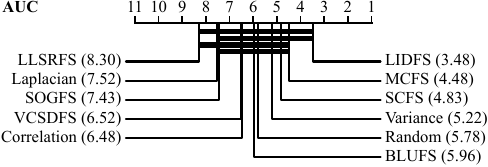}
\caption{Critical difference diagrams of ACC (left) and AUC (right) for the SVM classifier (5\% to 100\% of the features).}\label{fig:supcdd-svm-100}
\supbackcdd
\end{figure}
\begin{figure}[p]
\centering
\pdfbookmark[3]{XGBoost}{supcdd-bm-xgb}
\includegraphics[width=\supcddwidth]{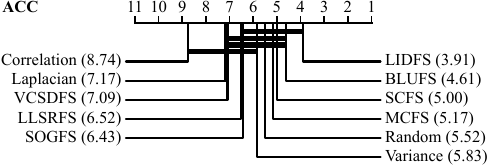}\hfill
\includegraphics[width=\supcddwidth]{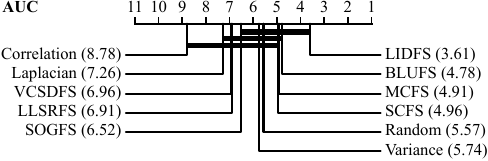}
\caption{Critical difference diagrams of ACC (left) and AUC (right) for the XGBoost classifier (0.5\% to 10\% of the features).}\label{fig:supcdd-xgb-10}
\supbackcdd
\end{figure}
\begin{figure}[p]
\centering
\includegraphics[width=\supcddwidth]{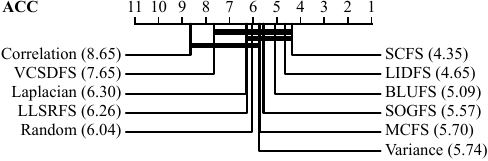}\hfill
\includegraphics[width=\supcddwidth]{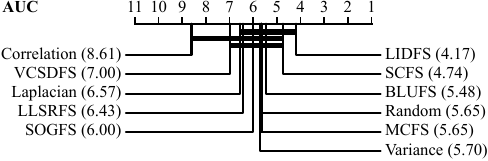}
\caption{Critical difference diagrams of ACC (left) and AUC (right) for the XGBoost classifier (5\% to 100\% of the features).}\label{fig:supcdd-xgb-100}
\supbackcdd
\end{figure}

\clearpage

\subsection{Performance plots}\label{app:supperf}

This part presents the performance plots on all 23 benchmark datasets and for both experimental settings, in which the number of selected features ranges from 0.5\% to 10\% and from 5\% to 100\% of the features, respectively. The unsupervised metrics (CLSACC and NMI) are reported once for each dataset, whereas the supervised metrics (ACC and AUC) are reported separately for each of the five classifiers, i.e., KNN, decision tree, SVM, random forest, and XGBoost. In all plots, the random baseline is the average of 100 independent runs, and the shaded area denotes one standard deviation. The results on the Isolet dataset for the random forest classifier and for the unsupervised metrics are presented in Fig.~\ref{\supMainFig} of the main text and Fig.~\ref{\supMainFigii} in \ref{app:full} for the first 10\% and for the entire range of the number of selected features, respectively, and are not repeated here. Table~\ref{tab:supplot-index} provides an index of all plots, where each entry links to the corresponding figure.

\begin{table}[b!]
\small
\centering
\caption{Index of the supplementary performance plots. Each entry gives the number of the figure for the corresponding dataset, metric group or classifier, and range of the number of selected features (up to 10\% or 100\% of the features), and links to the figure. Entries marked with $^{\dagger}$ refer to the main text (0.5\% to 10\%) or to \ref{app:full} (5\% to 100\%). DT, RF, and XGB denote decision tree, random forest, and XGBoost. Column headings 10 and 100 refer to the selection ranges up to 10\% and 100\% of the features.}
\label{tab:supplot-index}
\setlength{\tabcolsep}{0pt}
\begin{tabular*}{\textwidth}{@{\extracolsep{\fill}}l*{12}{c}@{}}
\toprule
 & \multicolumn{2}{c}{Unsupervised} & \multicolumn{10}{c}{Supervised (ACC, AUC)} \\
\cmidrule(lr){2-3}\cmidrule(lr){4-13}
 & \multicolumn{2}{c}{CLSACC, NMI} & \multicolumn{2}{c}{KNN} & \multicolumn{2}{c}{DT} & \multicolumn{2}{c}{SVM} & \multicolumn{2}{c}{RF} & \multicolumn{2}{c}{XGB} \\
\cmidrule(lr){2-3}\cmidrule(lr){4-5}\cmidrule(lr){6-7}\cmidrule(lr){8-9}\cmidrule(lr){10-11}\cmidrule(lr){12-13}
Dataset & 10 & 100 & 10 & 100 & 10 & 100 & 10 & 100 & 10 & 100 & 10 & 100 \\
\midrule
ALLAML & \ref{fig:supplot-unsup-10-ALLAML} & \ref{fig:supplot-unsup-100-ALLAML} & \ref{fig:supplot-knn-10-ALLAML} & \ref{fig:supplot-knn-100-ALLAML} & \ref{fig:supplot-dt-10-ALLAML} & \ref{fig:supplot-dt-100-ALLAML} & \ref{fig:supplot-svm-10-ALLAML} & \ref{fig:supplot-svm-100-ALLAML} & \ref{fig:supplot-rf-10-ALLAML} & \ref{fig:supplot-rf-100-ALLAML} & \ref{fig:supplot-xgb-10-ALLAML} & \ref{fig:supplot-xgb-100-ALLAML} \\
arcene & \ref{fig:supplot-unsup-10-arcene} & \ref{fig:supplot-unsup-100-arcene} & \ref{fig:supplot-knn-10-arcene} & \ref{fig:supplot-knn-100-arcene} & \ref{fig:supplot-dt-10-arcene} & \ref{fig:supplot-dt-100-arcene} & \ref{fig:supplot-svm-10-arcene} & \ref{fig:supplot-svm-100-arcene} & \ref{fig:supplot-rf-10-arcene} & \ref{fig:supplot-rf-100-arcene} & \ref{fig:supplot-xgb-10-arcene} & \ref{fig:supplot-xgb-100-arcene} \\
Carcinom & \ref{fig:supplot-unsup-10-Carcinom} & \ref{fig:supplot-unsup-100-Carcinom} & \ref{fig:supplot-knn-10-Carcinom} & \ref{fig:supplot-knn-100-Carcinom} & \ref{fig:supplot-dt-10-Carcinom} & \ref{fig:supplot-dt-100-Carcinom} & \ref{fig:supplot-svm-10-Carcinom} & \ref{fig:supplot-svm-100-Carcinom} & \ref{fig:supplot-rf-10-Carcinom} & \ref{fig:supplot-rf-100-Carcinom} & \ref{fig:supplot-xgb-10-Carcinom} & \ref{fig:supplot-xgb-100-Carcinom} \\
CLL\_SUB\_111 & \ref{fig:supplot-unsup-10-CLL_SUB_111} & \ref{fig:supplot-unsup-100-CLL_SUB_111} & \ref{fig:supplot-knn-10-CLL_SUB_111} & \ref{fig:supplot-knn-100-CLL_SUB_111} & \ref{fig:supplot-dt-10-CLL_SUB_111} & \ref{fig:supplot-dt-100-CLL_SUB_111} & \ref{fig:supplot-svm-10-CLL_SUB_111} & \ref{fig:supplot-svm-100-CLL_SUB_111} & \ref{fig:supplot-rf-10-CLL_SUB_111} & \ref{fig:supplot-rf-100-CLL_SUB_111} & \ref{fig:supplot-xgb-10-CLL_SUB_111} & \ref{fig:supplot-xgb-100-CLL_SUB_111} \\
COIL20 & \ref{fig:supplot-unsup-10-COIL20} & \ref{fig:supplot-unsup-100-COIL20} & \ref{fig:supplot-knn-10-COIL20} & \ref{fig:supplot-knn-100-COIL20} & \ref{fig:supplot-dt-10-COIL20} & \ref{fig:supplot-dt-100-COIL20} & \ref{fig:supplot-svm-10-COIL20} & \ref{fig:supplot-svm-100-COIL20} & \ref{fig:supplot-rf-10-COIL20} & \ref{fig:supplot-rf-100-COIL20} & \ref{fig:supplot-xgb-10-COIL20} & \ref{fig:supplot-xgb-100-COIL20} \\
colon & \ref{fig:supplot-unsup-10-colon} & \ref{fig:supplot-unsup-100-colon} & \ref{fig:supplot-knn-10-colon} & \ref{fig:supplot-knn-100-colon} & \ref{fig:supplot-dt-10-colon} & \ref{fig:supplot-dt-100-colon} & \ref{fig:supplot-svm-10-colon} & \ref{fig:supplot-svm-100-colon} & \ref{fig:supplot-rf-10-colon} & \ref{fig:supplot-rf-100-colon} & \ref{fig:supplot-xgb-10-colon} & \ref{fig:supplot-xgb-100-colon} \\
gisette & \ref{fig:supplot-unsup-10-gisette} & \ref{fig:supplot-unsup-100-gisette} & \ref{fig:supplot-knn-10-gisette} & \ref{fig:supplot-knn-100-gisette} & \ref{fig:supplot-dt-10-gisette} & \ref{fig:supplot-dt-100-gisette} & \ref{fig:supplot-svm-10-gisette} & \ref{fig:supplot-svm-100-gisette} & \ref{fig:supplot-rf-10-gisette} & \ref{fig:supplot-rf-100-gisette} & \ref{fig:supplot-xgb-10-gisette} & \ref{fig:supplot-xgb-100-gisette} \\
GLI\_85 & \ref{fig:supplot-unsup-10-GLI_85} & \ref{fig:supplot-unsup-100-GLI_85} & \ref{fig:supplot-knn-10-GLI_85} & \ref{fig:supplot-knn-100-GLI_85} & \ref{fig:supplot-dt-10-GLI_85} & \ref{fig:supplot-dt-100-GLI_85} & \ref{fig:supplot-svm-10-GLI_85} & \ref{fig:supplot-svm-100-GLI_85} & \ref{fig:supplot-rf-10-GLI_85} & \ref{fig:supplot-rf-100-GLI_85} & \ref{fig:supplot-xgb-10-GLI_85} & \ref{fig:supplot-xgb-100-GLI_85} \\
GLIOMA & \ref{fig:supplot-unsup-10-GLIOMA} & \ref{fig:supplot-unsup-100-GLIOMA} & \ref{fig:supplot-knn-10-GLIOMA} & \ref{fig:supplot-knn-100-GLIOMA} & \ref{fig:supplot-dt-10-GLIOMA} & \ref{fig:supplot-dt-100-GLIOMA} & \ref{fig:supplot-svm-10-GLIOMA} & \ref{fig:supplot-svm-100-GLIOMA} & \ref{fig:supplot-rf-10-GLIOMA} & \ref{fig:supplot-rf-100-GLIOMA} & \ref{fig:supplot-xgb-10-GLIOMA} & \ref{fig:supplot-xgb-100-GLIOMA} \\
Isolet & \ref{\supMainFig}$^{\dagger}$ & \ref{\supMainFigii}$^{\dagger}$ & \ref{fig:supplot-knn-10-Isolet} & \ref{fig:supplot-knn-100-Isolet} & \ref{fig:supplot-dt-10-Isolet} & \ref{fig:supplot-dt-100-Isolet} & \ref{fig:supplot-svm-10-Isolet} & \ref{fig:supplot-svm-100-Isolet} & \ref{\supMainFig}$^{\dagger}$ & \ref{\supMainFigii}$^{\dagger}$ & \ref{fig:supplot-xgb-10-Isolet} & \ref{fig:supplot-xgb-100-Isolet} \\
leukemia & \ref{fig:supplot-unsup-10-leukemia} & \ref{fig:supplot-unsup-100-leukemia} & \ref{fig:supplot-knn-10-leukemia} & \ref{fig:supplot-knn-100-leukemia} & \ref{fig:supplot-dt-10-leukemia} & \ref{fig:supplot-dt-100-leukemia} & \ref{fig:supplot-svm-10-leukemia} & \ref{fig:supplot-svm-100-leukemia} & \ref{fig:supplot-rf-10-leukemia} & \ref{fig:supplot-rf-100-leukemia} & \ref{fig:supplot-xgb-10-leukemia} & \ref{fig:supplot-xgb-100-leukemia} \\
lung & \ref{fig:supplot-unsup-10-lung} & \ref{fig:supplot-unsup-100-lung} & \ref{fig:supplot-knn-10-lung} & \ref{fig:supplot-knn-100-lung} & \ref{fig:supplot-dt-10-lung} & \ref{fig:supplot-dt-100-lung} & \ref{fig:supplot-svm-10-lung} & \ref{fig:supplot-svm-100-lung} & \ref{fig:supplot-rf-10-lung} & \ref{fig:supplot-rf-100-lung} & \ref{fig:supplot-xgb-10-lung} & \ref{fig:supplot-xgb-100-lung} \\
lung\_discrete & \ref{fig:supplot-unsup-10-lung_discrete} & \ref{fig:supplot-unsup-100-lung_discrete} & \ref{fig:supplot-knn-10-lung_discrete} & \ref{fig:supplot-knn-100-lung_discrete} & \ref{fig:supplot-dt-10-lung_discrete} & \ref{fig:supplot-dt-100-lung_discrete} & \ref{fig:supplot-svm-10-lung_discrete} & \ref{fig:supplot-svm-100-lung_discrete} & \ref{fig:supplot-rf-10-lung_discrete} & \ref{fig:supplot-rf-100-lung_discrete} & \ref{fig:supplot-xgb-10-lung_discrete} & \ref{fig:supplot-xgb-100-lung_discrete} \\
madelon & \ref{fig:supplot-unsup-10-madelon} & \ref{fig:supplot-unsup-100-madelon} & \ref{fig:supplot-knn-10-madelon} & \ref{fig:supplot-knn-100-madelon} & \ref{fig:supplot-dt-10-madelon} & \ref{fig:supplot-dt-100-madelon} & \ref{fig:supplot-svm-10-madelon} & \ref{fig:supplot-svm-100-madelon} & \ref{fig:supplot-rf-10-madelon} & \ref{fig:supplot-rf-100-madelon} & \ref{fig:supplot-xgb-10-madelon} & \ref{fig:supplot-xgb-100-madelon} \\
ORL & \ref{fig:supplot-unsup-10-ORL} & \ref{fig:supplot-unsup-100-ORL} & \ref{fig:supplot-knn-10-ORL} & \ref{fig:supplot-knn-100-ORL} & \ref{fig:supplot-dt-10-ORL} & \ref{fig:supplot-dt-100-ORL} & \ref{fig:supplot-svm-10-ORL} & \ref{fig:supplot-svm-100-ORL} & \ref{fig:supplot-rf-10-ORL} & \ref{fig:supplot-rf-100-ORL} & \ref{fig:supplot-xgb-10-ORL} & \ref{fig:supplot-xgb-100-ORL} \\
orlraws10P & \ref{fig:supplot-unsup-10-orlraws10P} & \ref{fig:supplot-unsup-100-orlraws10P} & \ref{fig:supplot-knn-10-orlraws10P} & \ref{fig:supplot-knn-100-orlraws10P} & \ref{fig:supplot-dt-10-orlraws10P} & \ref{fig:supplot-dt-100-orlraws10P} & \ref{fig:supplot-svm-10-orlraws10P} & \ref{fig:supplot-svm-100-orlraws10P} & \ref{fig:supplot-rf-10-orlraws10P} & \ref{fig:supplot-rf-100-orlraws10P} & \ref{fig:supplot-xgb-10-orlraws10P} & \ref{fig:supplot-xgb-100-orlraws10P} \\
pixraw10P & \ref{fig:supplot-unsup-10-pixraw10P} & \ref{fig:supplot-unsup-100-pixraw10P} & \ref{fig:supplot-knn-10-pixraw10P} & \ref{fig:supplot-knn-100-pixraw10P} & \ref{fig:supplot-dt-10-pixraw10P} & \ref{fig:supplot-dt-100-pixraw10P} & \ref{fig:supplot-svm-10-pixraw10P} & \ref{fig:supplot-svm-100-pixraw10P} & \ref{fig:supplot-rf-10-pixraw10P} & \ref{fig:supplot-rf-100-pixraw10P} & \ref{fig:supplot-xgb-10-pixraw10P} & \ref{fig:supplot-xgb-100-pixraw10P} \\
Prostate\_GE & \ref{fig:supplot-unsup-10-Prostate_GE} & \ref{fig:supplot-unsup-100-Prostate_GE} & \ref{fig:supplot-knn-10-Prostate_GE} & \ref{fig:supplot-knn-100-Prostate_GE} & \ref{fig:supplot-dt-10-Prostate_GE} & \ref{fig:supplot-dt-100-Prostate_GE} & \ref{fig:supplot-svm-10-Prostate_GE} & \ref{fig:supplot-svm-100-Prostate_GE} & \ref{fig:supplot-rf-10-Prostate_GE} & \ref{fig:supplot-rf-100-Prostate_GE} & \ref{fig:supplot-xgb-10-Prostate_GE} & \ref{fig:supplot-xgb-100-Prostate_GE} \\
SMK\_CAN\_187 & \ref{fig:supplot-unsup-10-SMK_CAN_187} & \ref{fig:supplot-unsup-100-SMK_CAN_187} & \ref{fig:supplot-knn-10-SMK_CAN_187} & \ref{fig:supplot-knn-100-SMK_CAN_187} & \ref{fig:supplot-dt-10-SMK_CAN_187} & \ref{fig:supplot-dt-100-SMK_CAN_187} & \ref{fig:supplot-svm-10-SMK_CAN_187} & \ref{fig:supplot-svm-100-SMK_CAN_187} & \ref{fig:supplot-rf-10-SMK_CAN_187} & \ref{fig:supplot-rf-100-SMK_CAN_187} & \ref{fig:supplot-xgb-10-SMK_CAN_187} & \ref{fig:supplot-xgb-100-SMK_CAN_187} \\
TOX\_171 & \ref{fig:supplot-unsup-10-TOX_171} & \ref{fig:supplot-unsup-100-TOX_171} & \ref{fig:supplot-knn-10-TOX_171} & \ref{fig:supplot-knn-100-TOX_171} & \ref{fig:supplot-dt-10-TOX_171} & \ref{fig:supplot-dt-100-TOX_171} & \ref{fig:supplot-svm-10-TOX_171} & \ref{fig:supplot-svm-100-TOX_171} & \ref{fig:supplot-rf-10-TOX_171} & \ref{fig:supplot-rf-100-TOX_171} & \ref{fig:supplot-xgb-10-TOX_171} & \ref{fig:supplot-xgb-100-TOX_171} \\
warpAR10P & \ref{fig:supplot-unsup-10-warpAR10P} & \ref{fig:supplot-unsup-100-warpAR10P} & \ref{fig:supplot-knn-10-warpAR10P} & \ref{fig:supplot-knn-100-warpAR10P} & \ref{fig:supplot-dt-10-warpAR10P} & \ref{fig:supplot-dt-100-warpAR10P} & \ref{fig:supplot-svm-10-warpAR10P} & \ref{fig:supplot-svm-100-warpAR10P} & \ref{fig:supplot-rf-10-warpAR10P} & \ref{fig:supplot-rf-100-warpAR10P} & \ref{fig:supplot-xgb-10-warpAR10P} & \ref{fig:supplot-xgb-100-warpAR10P} \\
warpPIE10P & \ref{fig:supplot-unsup-10-warpPIE10P} & \ref{fig:supplot-unsup-100-warpPIE10P} & \ref{fig:supplot-knn-10-warpPIE10P} & \ref{fig:supplot-knn-100-warpPIE10P} & \ref{fig:supplot-dt-10-warpPIE10P} & \ref{fig:supplot-dt-100-warpPIE10P} & \ref{fig:supplot-svm-10-warpPIE10P} & \ref{fig:supplot-svm-100-warpPIE10P} & \ref{fig:supplot-rf-10-warpPIE10P} & \ref{fig:supplot-rf-100-warpPIE10P} & \ref{fig:supplot-xgb-10-warpPIE10P} & \ref{fig:supplot-xgb-100-warpPIE10P} \\
Yale & \ref{fig:supplot-unsup-10-Yale} & \ref{fig:supplot-unsup-100-Yale} & \ref{fig:supplot-knn-10-Yale} & \ref{fig:supplot-knn-100-Yale} & \ref{fig:supplot-dt-10-Yale} & \ref{fig:supplot-dt-100-Yale} & \ref{fig:supplot-svm-10-Yale} & \ref{fig:supplot-svm-100-Yale} & \ref{fig:supplot-rf-10-Yale} & \ref{fig:supplot-rf-100-Yale} & \ref{fig:supplot-xgb-10-Yale} & \ref{fig:supplot-xgb-100-Yale} \\
\bottomrule
\end{tabular*}
\end{table}

\begin{figure}[p]
\centering
\pdfbookmark[3]{Unsupervised metrics}{supplot-bm-unsup}
\pdfbookmark[4]{0.5\% to 10\%}{supplot-bm-unsup-10}
\includegraphics[width=\textwidth,height=\supfigheight,keepaspectratio]{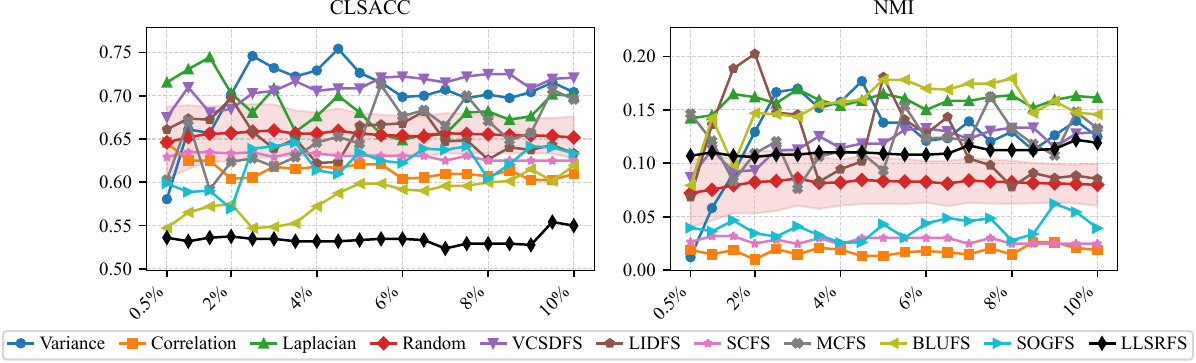}
\caption{CLSACC and NMI on the ALLAML dataset (0.5\% to 10\% of the features).}\label{fig:supplot-unsup-10-ALLAML}
\supback
\end{figure}
\begin{figure}[p]
\centering
\includegraphics[width=\textwidth,height=\supfigheight,keepaspectratio]{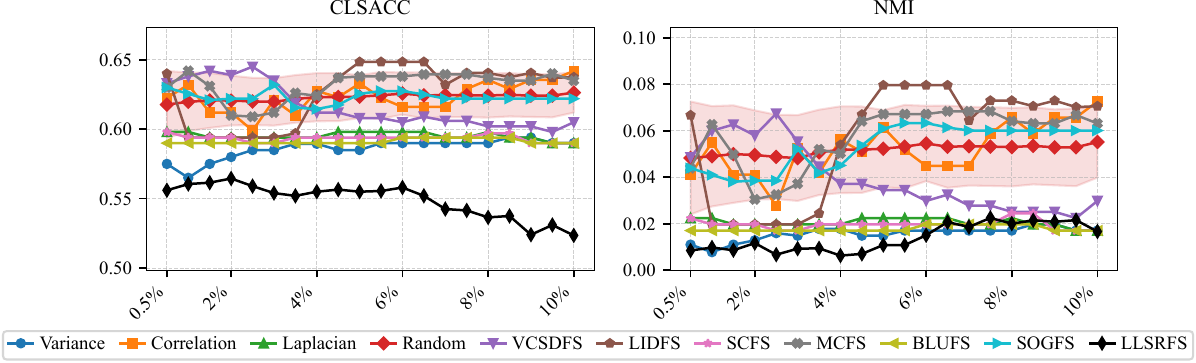}
\caption{CLSACC and NMI on the arcene dataset (0.5\% to 10\% of the features).}\label{fig:supplot-unsup-10-arcene}
\supback
\end{figure}
\begin{figure}[p]
\centering
\includegraphics[width=\textwidth,height=\supfigheight,keepaspectratio]{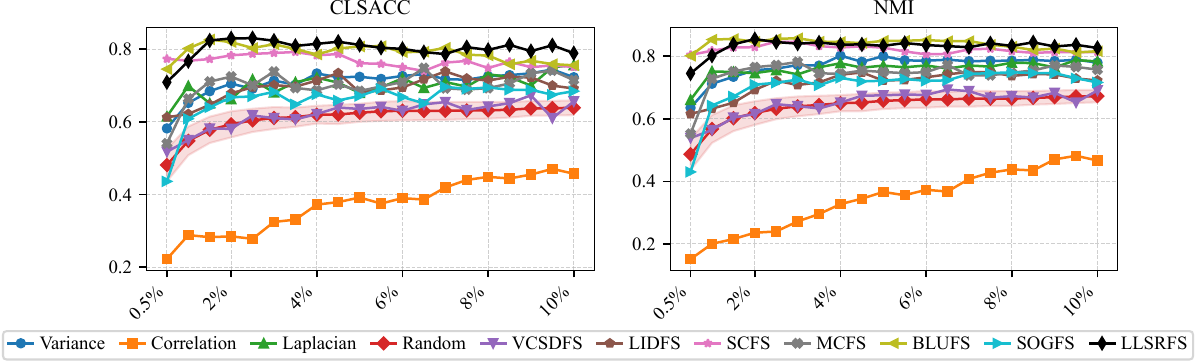}
\caption{CLSACC and NMI on the Carcinom dataset (0.5\% to 10\% of the features).}\label{fig:supplot-unsup-10-Carcinom}
\supback
\end{figure}
\begin{figure}[p]
\centering
\includegraphics[width=\textwidth,height=\supfigheight,keepaspectratio]{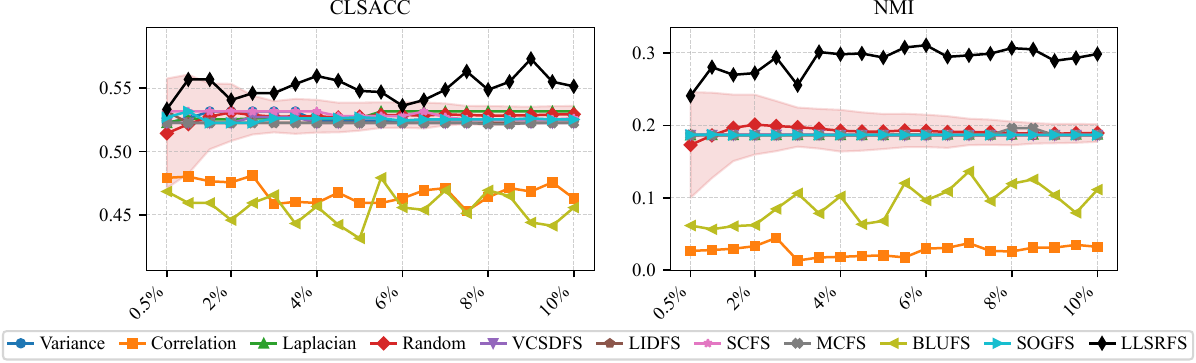}
\caption{CLSACC and NMI on the CLL\_SUB\_111 dataset (0.5\% to 10\% of the features).}\label{fig:supplot-unsup-10-CLL_SUB_111}
\supback
\end{figure}
\begin{figure}[p]
\centering
\includegraphics[width=\textwidth,height=\supfigheight,keepaspectratio]{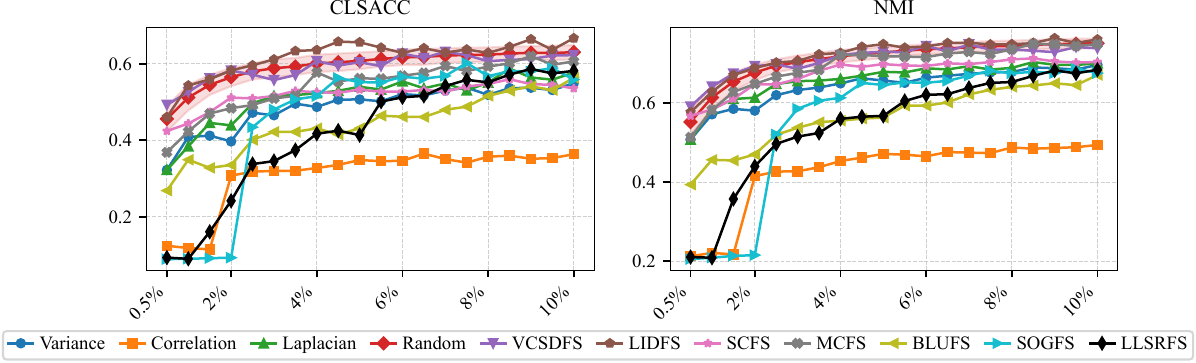}
\caption{CLSACC and NMI on the COIL20 dataset (0.5\% to 10\% of the features).}\label{fig:supplot-unsup-10-COIL20}
\supback
\end{figure}
\begin{figure}[p]
\centering
\includegraphics[width=\textwidth,height=\supfigheight,keepaspectratio]{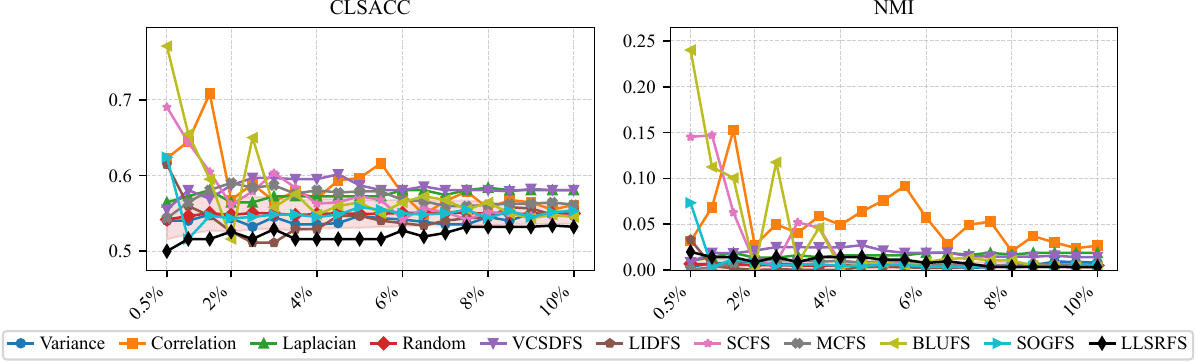}
\caption{CLSACC and NMI on the colon dataset (0.5\% to 10\% of the features).}\label{fig:supplot-unsup-10-colon}
\supback
\end{figure}
\begin{figure}[p]
\centering
\includegraphics[width=\textwidth,height=\supfigheight,keepaspectratio]{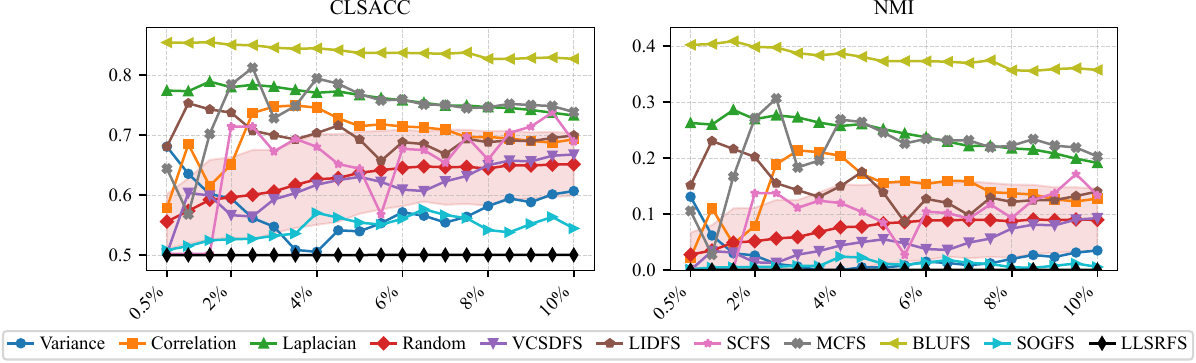}
\caption{CLSACC and NMI on the gisette dataset (0.5\% to 10\% of the features).}\label{fig:supplot-unsup-10-gisette}
\supback
\end{figure}
\begin{figure}[p]
\centering
\includegraphics[width=\textwidth,height=\supfigheight,keepaspectratio]{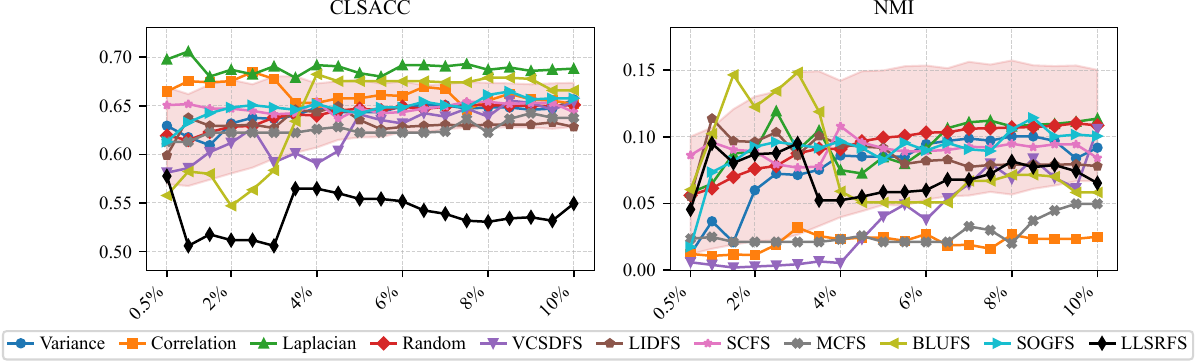}
\caption{CLSACC and NMI on the GLI\_85 dataset (0.5\% to 10\% of the features).}\label{fig:supplot-unsup-10-GLI_85}
\supback
\end{figure}
\begin{figure}[p]
\centering
\includegraphics[width=\textwidth,height=\supfigheight,keepaspectratio]{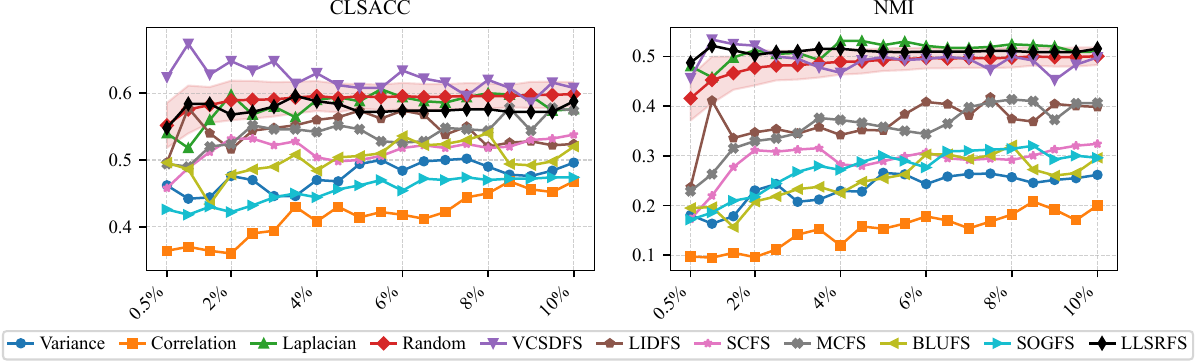}
\caption{CLSACC and NMI on the GLIOMA dataset (0.5\% to 10\% of the features).}\label{fig:supplot-unsup-10-GLIOMA}
\supback
\end{figure}
\begin{figure}[p]
\centering
\includegraphics[width=\textwidth,height=\supfigheight,keepaspectratio]{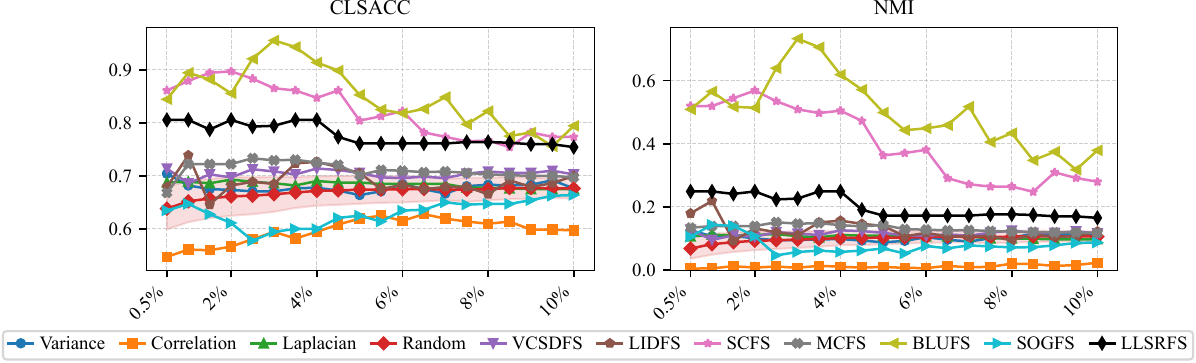}
\caption{CLSACC and NMI on the leukemia dataset (0.5\% to 10\% of the features).}\label{fig:supplot-unsup-10-leukemia}
\supback
\end{figure}
\begin{figure}[p]
\centering
\includegraphics[width=\textwidth,height=\supfigheight,keepaspectratio]{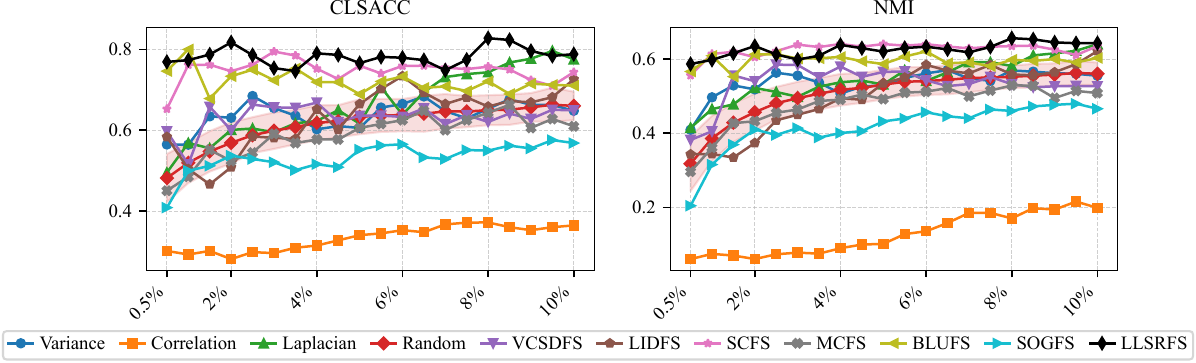}
\caption{CLSACC and NMI on the lung dataset (0.5\% to 10\% of the features).}\label{fig:supplot-unsup-10-lung}
\supback
\end{figure}
\begin{figure}[p]
\centering
\includegraphics[width=\textwidth,height=\supfigheight,keepaspectratio]{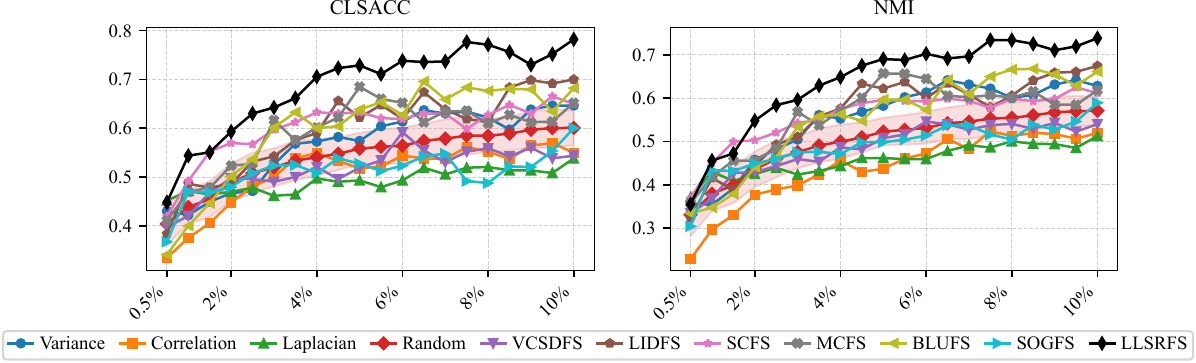}
\caption{CLSACC and NMI on the lung\_discrete dataset (0.5\% to 10\% of the features).}\label{fig:supplot-unsup-10-lung_discrete}
\supback
\end{figure}
\begin{figure}[p]
\centering
\includegraphics[width=\textwidth,height=\supfigheight,keepaspectratio]{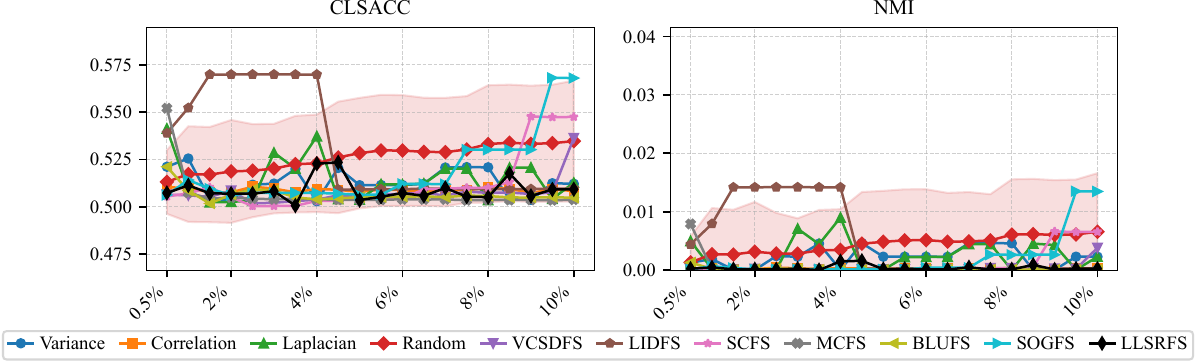}
\caption{CLSACC and NMI on the madelon dataset (0.5\% to 10\% of the features).}\label{fig:supplot-unsup-10-madelon}
\supback
\end{figure}
\begin{figure}[p]
\centering
\includegraphics[width=\textwidth,height=\supfigheight,keepaspectratio]{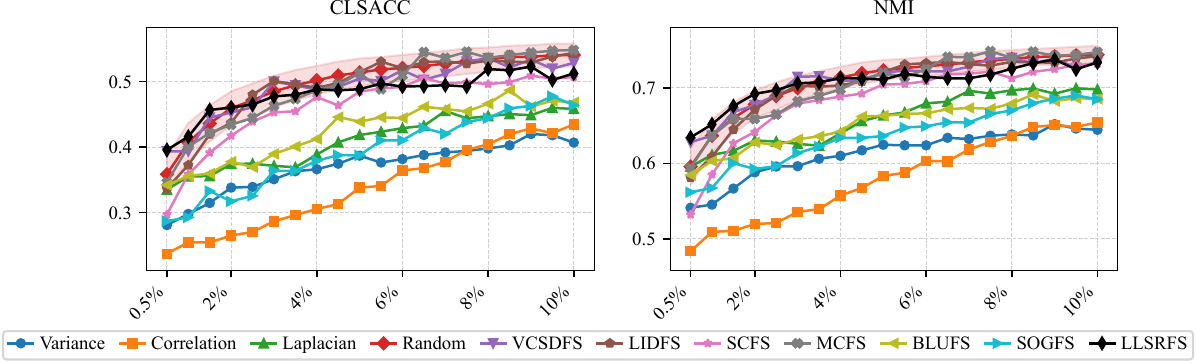}
\caption{CLSACC and NMI on the ORL dataset (0.5\% to 10\% of the features).}\label{fig:supplot-unsup-10-ORL}
\supback
\end{figure}
\begin{figure}[p]
\centering
\includegraphics[width=\textwidth,height=\supfigheight,keepaspectratio]{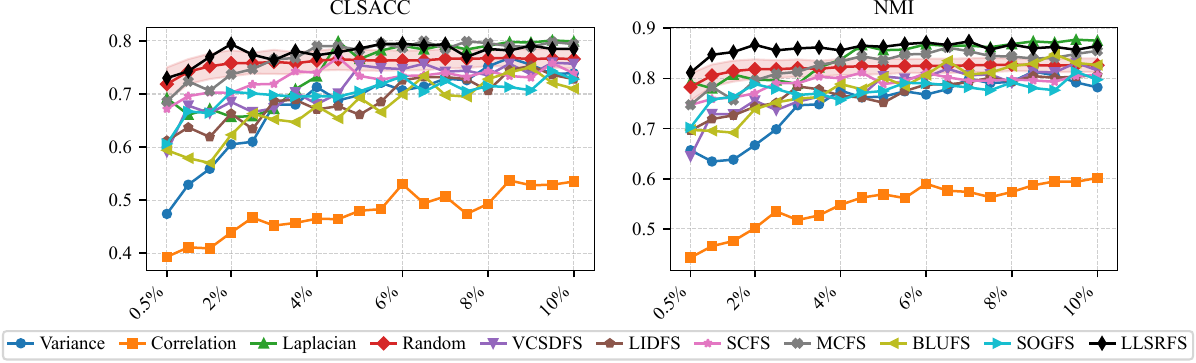}
\caption{CLSACC and NMI on the orlraws10P dataset (0.5\% to 10\% of the features).}\label{fig:supplot-unsup-10-orlraws10P}
\supback
\end{figure}
\begin{figure}[p]
\centering
\includegraphics[width=\textwidth,height=\supfigheight,keepaspectratio]{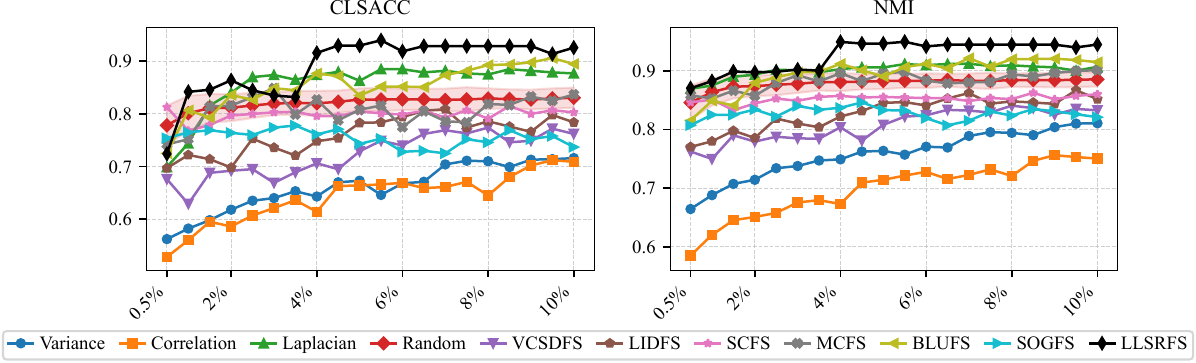}
\caption{CLSACC and NMI on the pixraw10P dataset (0.5\% to 10\% of the features).}\label{fig:supplot-unsup-10-pixraw10P}
\supback
\end{figure}
\begin{figure}[p]
\centering
\includegraphics[width=\textwidth,height=\supfigheight,keepaspectratio]{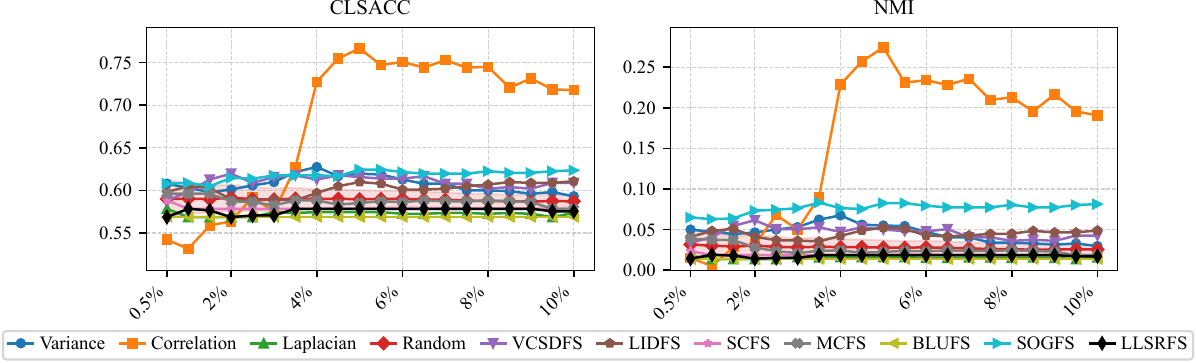}
\caption{CLSACC and NMI on the Prostate\_GE dataset (0.5\% to 10\% of the features).}\label{fig:supplot-unsup-10-Prostate_GE}
\supback
\end{figure}
\begin{figure}[p]
\centering
\includegraphics[width=\textwidth,height=\supfigheight,keepaspectratio]{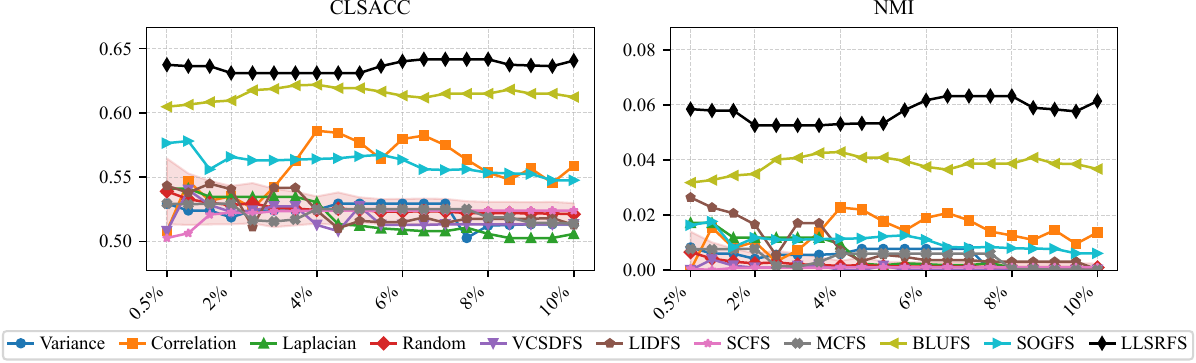}
\caption{CLSACC and NMI on the SMK\_CAN\_187 dataset (0.5\% to 10\% of the features).}\label{fig:supplot-unsup-10-SMK_CAN_187}
\supback
\end{figure}
\begin{figure}[p]
\centering
\includegraphics[width=\textwidth,height=\supfigheight,keepaspectratio]{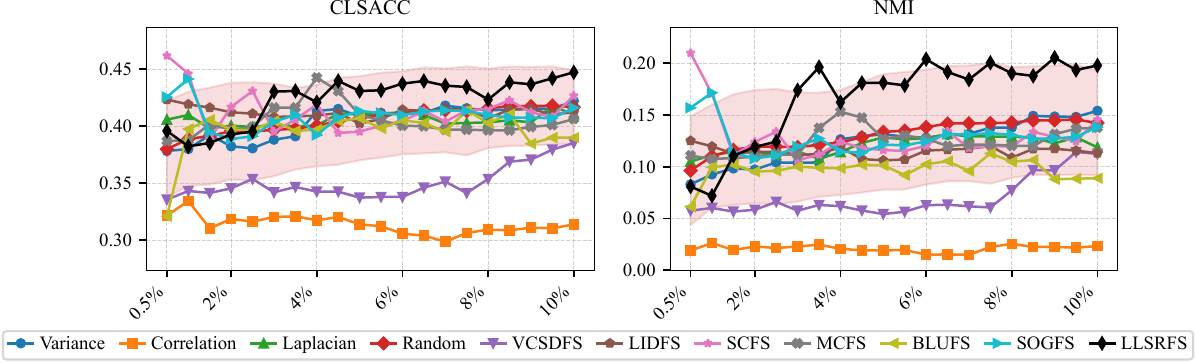}
\caption{CLSACC and NMI on the TOX\_171 dataset (0.5\% to 10\% of the features).}\label{fig:supplot-unsup-10-TOX_171}
\supback
\end{figure}
\begin{figure}[p]
\centering
\includegraphics[width=\textwidth,height=\supfigheight,keepaspectratio]{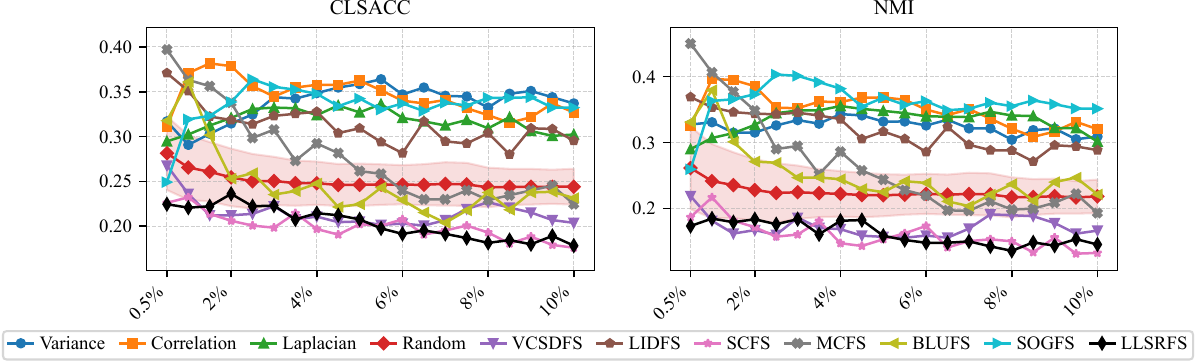}
\caption{CLSACC and NMI on the warpAR10P dataset (0.5\% to 10\% of the features).}\label{fig:supplot-unsup-10-warpAR10P}
\supback
\end{figure}
\begin{figure}[p]
\centering
\includegraphics[width=\textwidth,height=\supfigheight,keepaspectratio]{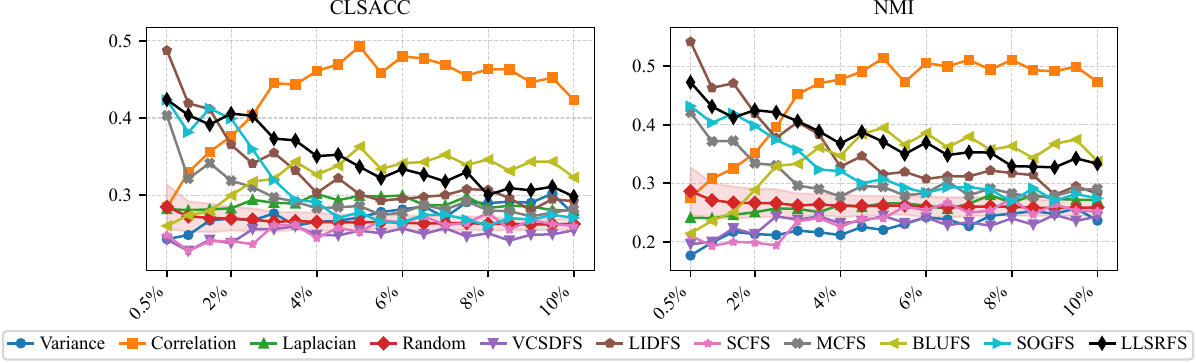}
\caption{CLSACC and NMI on the warpPIE10P dataset (0.5\% to 10\% of the features).}\label{fig:supplot-unsup-10-warpPIE10P}
\supback
\end{figure}
\begin{figure}[p]
\centering
\includegraphics[width=\textwidth,height=\supfigheight,keepaspectratio]{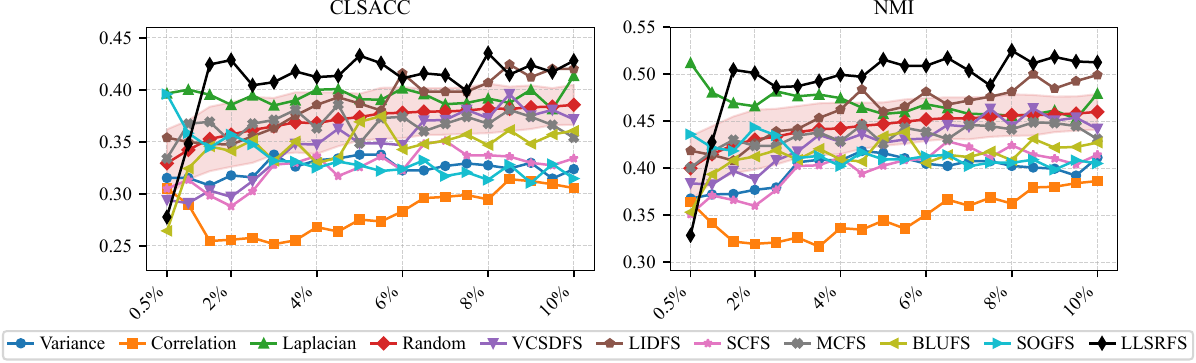}
\caption{CLSACC and NMI on the Yale dataset (0.5\% to 10\% of the features).}\label{fig:supplot-unsup-10-Yale}
\supback
\end{figure}
\begin{figure}[p]
\centering
\pdfbookmark[4]{5\% to 100\%}{supplot-bm-unsup-100}
\includegraphics[width=\textwidth,height=\supfigheight,keepaspectratio]{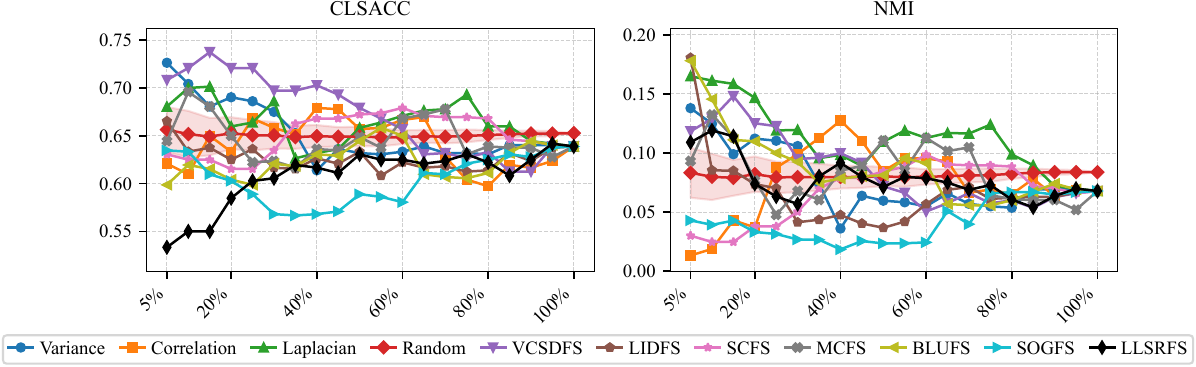}
\caption{CLSACC and NMI on the ALLAML dataset (5\% to 100\% of the features).}\label{fig:supplot-unsup-100-ALLAML}
\supback
\end{figure}
\begin{figure}[p]
\centering
\includegraphics[width=\textwidth,height=\supfigheight,keepaspectratio]{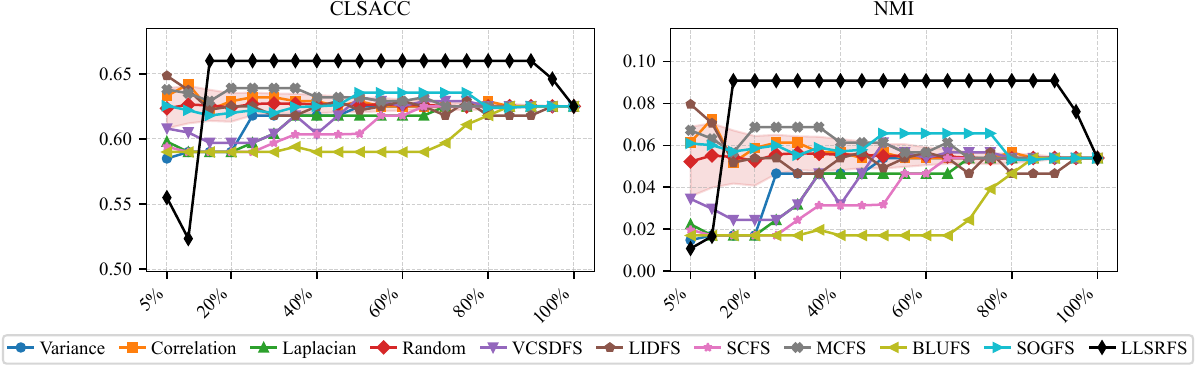}
\caption{CLSACC and NMI on the arcene dataset (5\% to 100\% of the features).}\label{fig:supplot-unsup-100-arcene}
\supback
\end{figure}
\begin{figure}[p]
\centering
\includegraphics[width=\textwidth,height=\supfigheight,keepaspectratio]{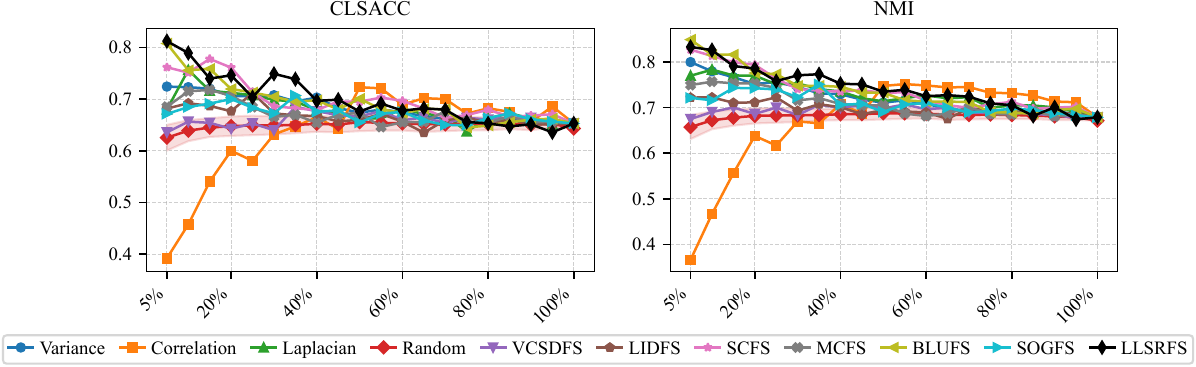}
\caption{CLSACC and NMI on the Carcinom dataset (5\% to 100\% of the features).}\label{fig:supplot-unsup-100-Carcinom}
\supback
\end{figure}
\begin{figure}[p]
\centering
\includegraphics[width=\textwidth,height=\supfigheight,keepaspectratio]{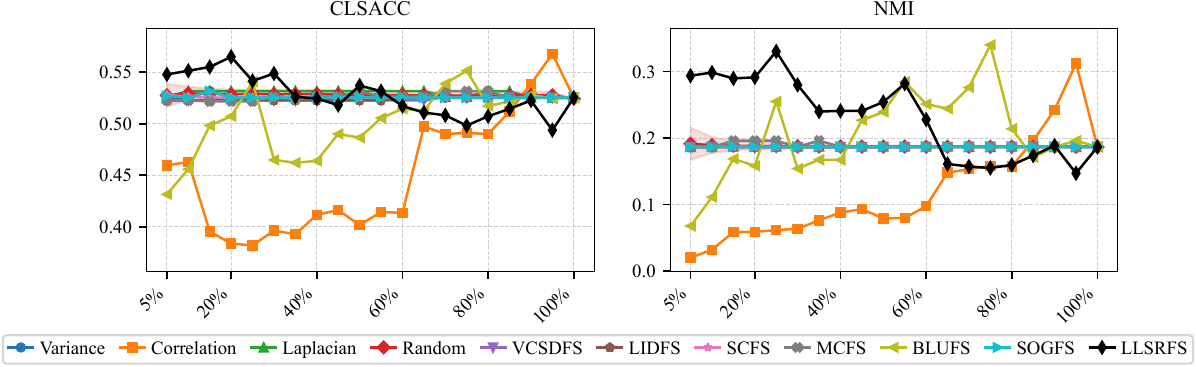}
\caption{CLSACC and NMI on the CLL\_SUB\_111 dataset (5\% to 100\% of the features).}\label{fig:supplot-unsup-100-CLL_SUB_111}
\supback
\end{figure}
\begin{figure}[p]
\centering
\includegraphics[width=\textwidth,height=\supfigheight,keepaspectratio]{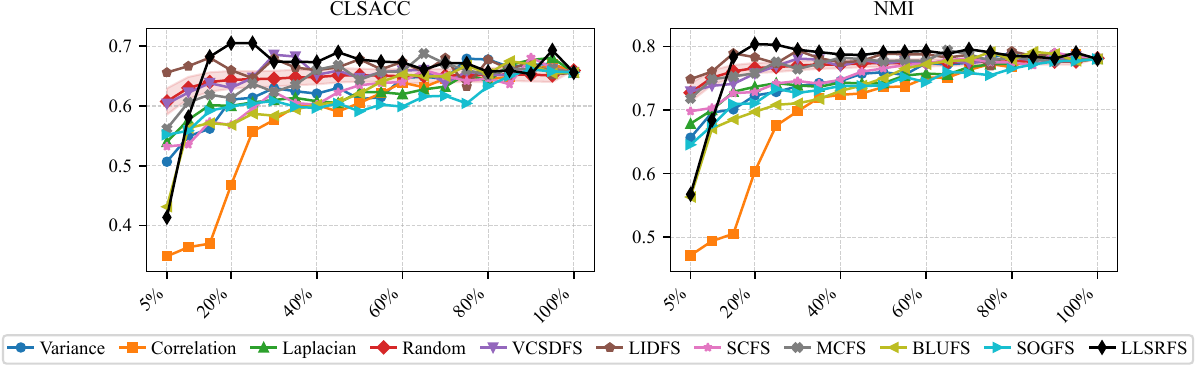}
\caption{CLSACC and NMI on the COIL20 dataset (5\% to 100\% of the features).}\label{fig:supplot-unsup-100-COIL20}
\supback
\end{figure}
\begin{figure}[p]
\centering
\includegraphics[width=\textwidth,height=\supfigheight,keepaspectratio]{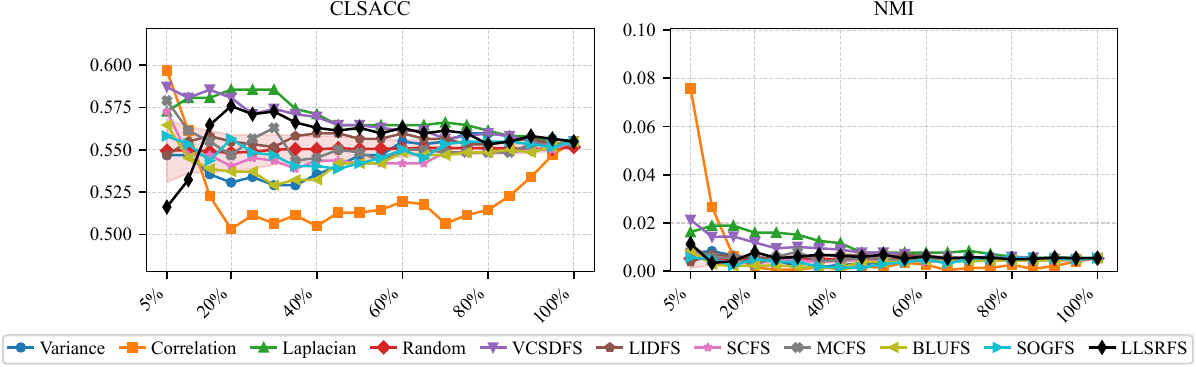}
\caption{CLSACC and NMI on the colon dataset (5\% to 100\% of the features).}\label{fig:supplot-unsup-100-colon}
\supback
\end{figure}
\begin{figure}[p]
\centering
\includegraphics[width=\textwidth,height=\supfigheight,keepaspectratio]{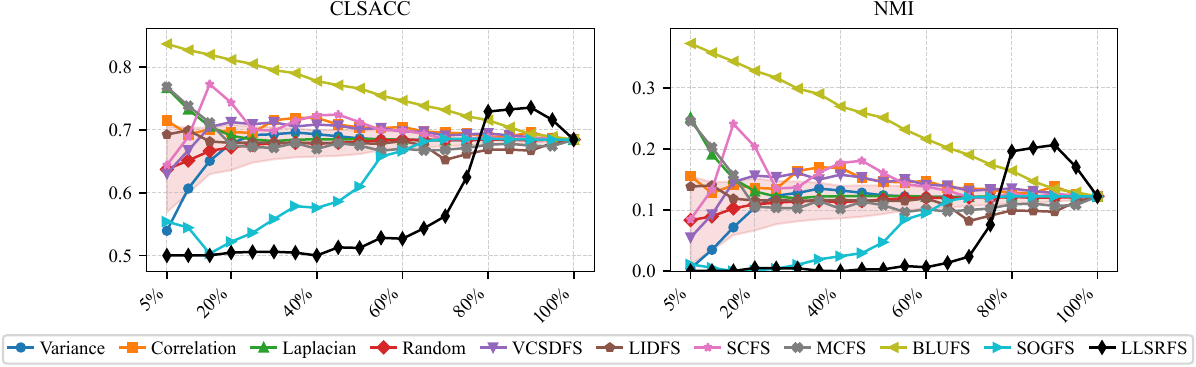}
\caption{CLSACC and NMI on the gisette dataset (5\% to 100\% of the features).}\label{fig:supplot-unsup-100-gisette}
\supback
\end{figure}
\begin{figure}[p]
\centering
\includegraphics[width=\textwidth,height=\supfigheight,keepaspectratio]{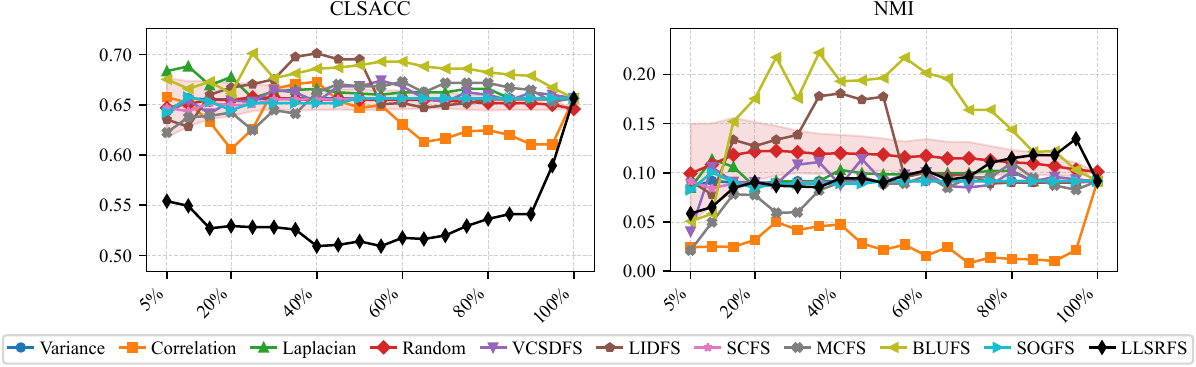}
\caption{CLSACC and NMI on the GLI\_85 dataset (5\% to 100\% of the features).}\label{fig:supplot-unsup-100-GLI_85}
\supback
\end{figure}
\begin{figure}[p]
\centering
\includegraphics[width=\textwidth,height=\supfigheight,keepaspectratio]{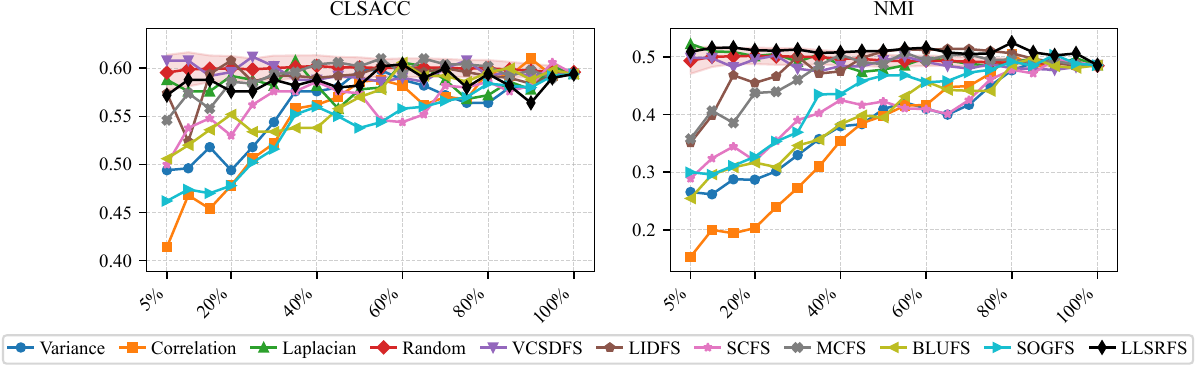}
\caption{CLSACC and NMI on the GLIOMA dataset (5\% to 100\% of the features).}\label{fig:supplot-unsup-100-GLIOMA}
\supback
\end{figure}
\begin{figure}[p]
\centering
\includegraphics[width=\textwidth,height=\supfigheight,keepaspectratio]{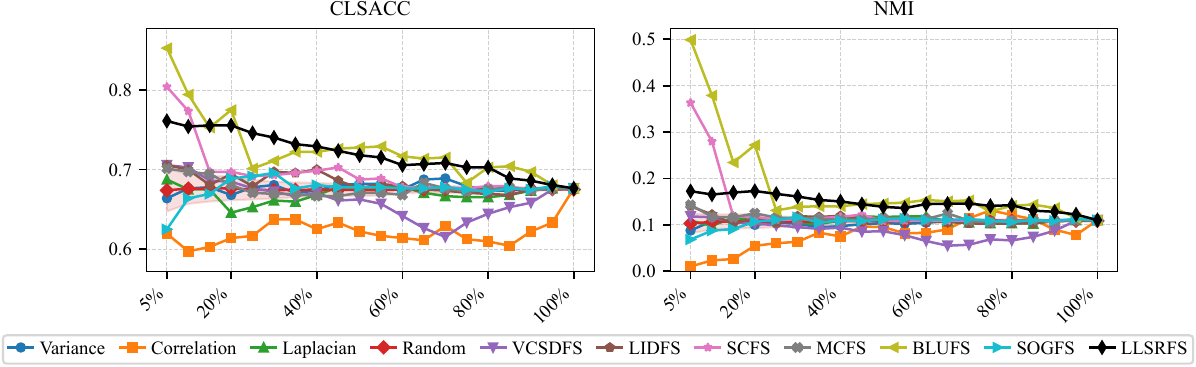}
\caption{CLSACC and NMI on the leukemia dataset (5\% to 100\% of the features).}\label{fig:supplot-unsup-100-leukemia}
\supback
\end{figure}
\begin{figure}[p]
\centering
\includegraphics[width=\textwidth,height=\supfigheight,keepaspectratio]{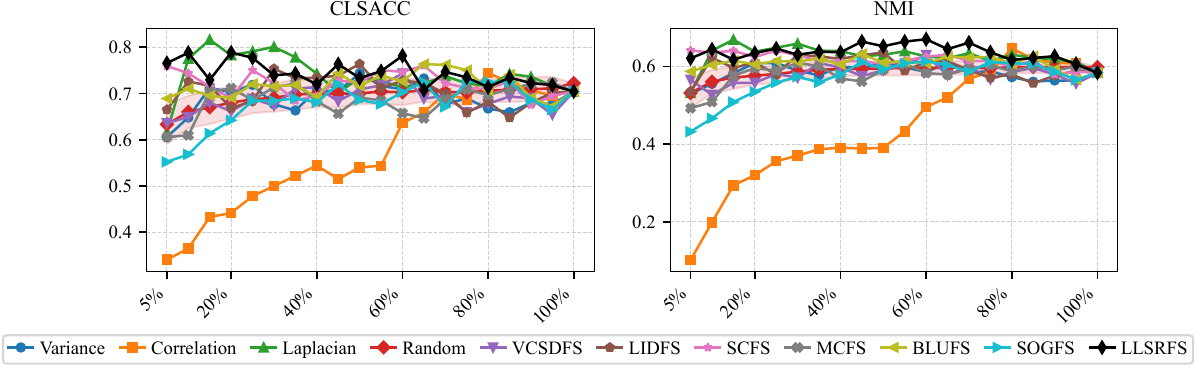}
\caption{CLSACC and NMI on the lung dataset (5\% to 100\% of the features).}\label{fig:supplot-unsup-100-lung}
\supback
\end{figure}
\begin{figure}[p]
\centering
\includegraphics[width=\textwidth,height=\supfigheight,keepaspectratio]{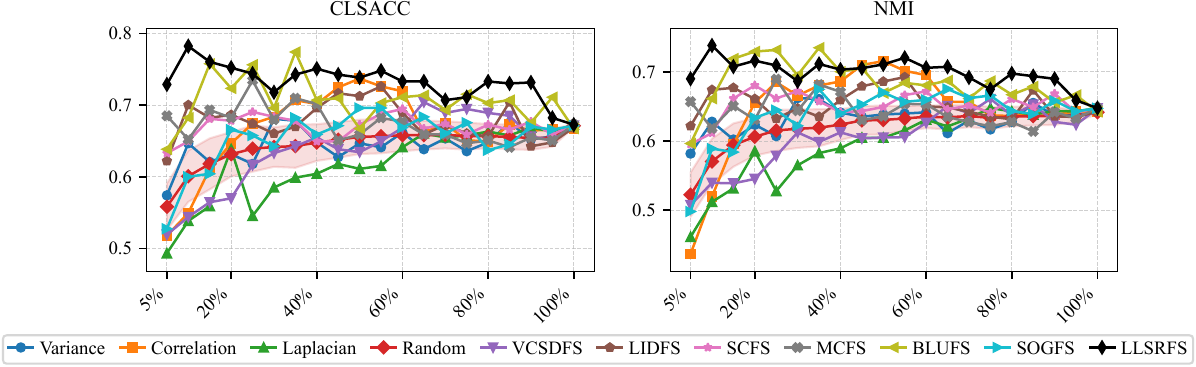}
\caption{CLSACC and NMI on the lung\_discrete dataset (5\% to 100\% of the features).}\label{fig:supplot-unsup-100-lung_discrete}
\supback
\end{figure}
\begin{figure}[p]
\centering
\includegraphics[width=\textwidth,height=\supfigheight,keepaspectratio]{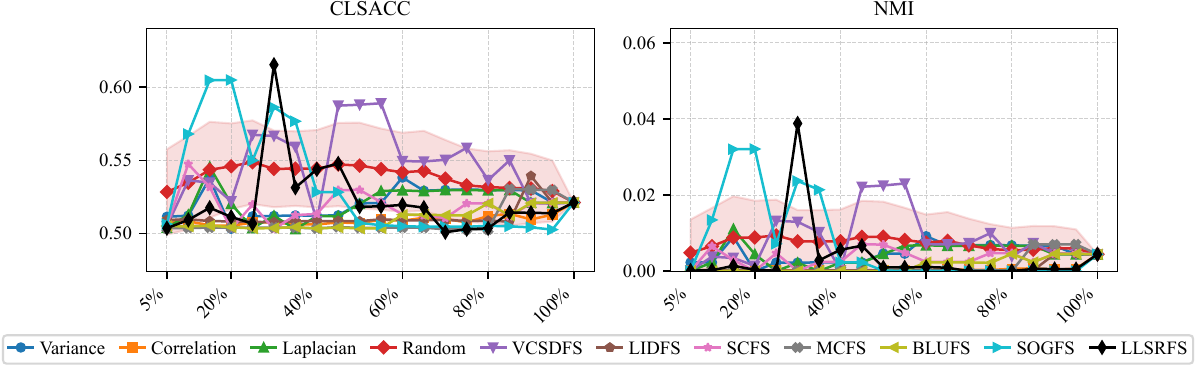}
\caption{CLSACC and NMI on the madelon dataset (5\% to 100\% of the features).}\label{fig:supplot-unsup-100-madelon}
\supback
\end{figure}
\begin{figure}[p]
\centering
\includegraphics[width=\textwidth,height=\supfigheight,keepaspectratio]{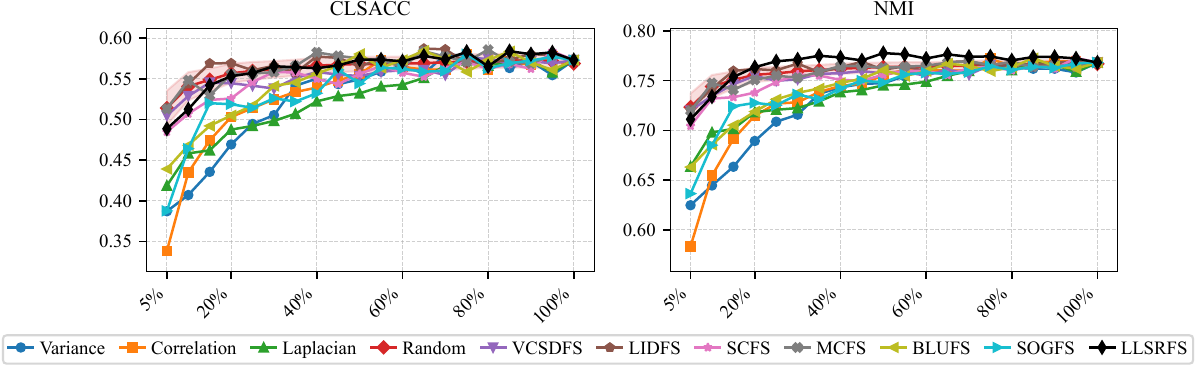}
\caption{CLSACC and NMI on the ORL dataset (5\% to 100\% of the features).}\label{fig:supplot-unsup-100-ORL}
\supback
\end{figure}
\begin{figure}[p]
\centering
\includegraphics[width=\textwidth,height=\supfigheight,keepaspectratio]{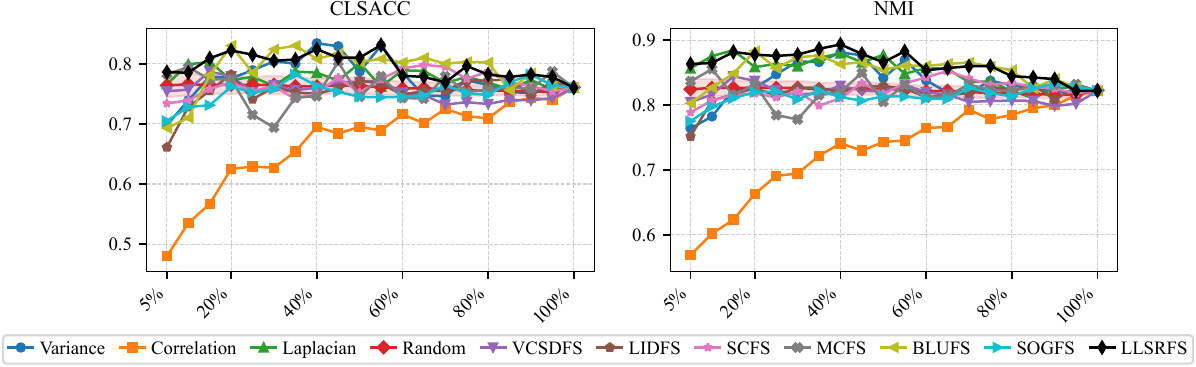}
\caption{CLSACC and NMI on the orlraws10P dataset (5\% to 100\% of the features).}\label{fig:supplot-unsup-100-orlraws10P}
\supback
\end{figure}
\begin{figure}[p]
\centering
\includegraphics[width=\textwidth,height=\supfigheight,keepaspectratio]{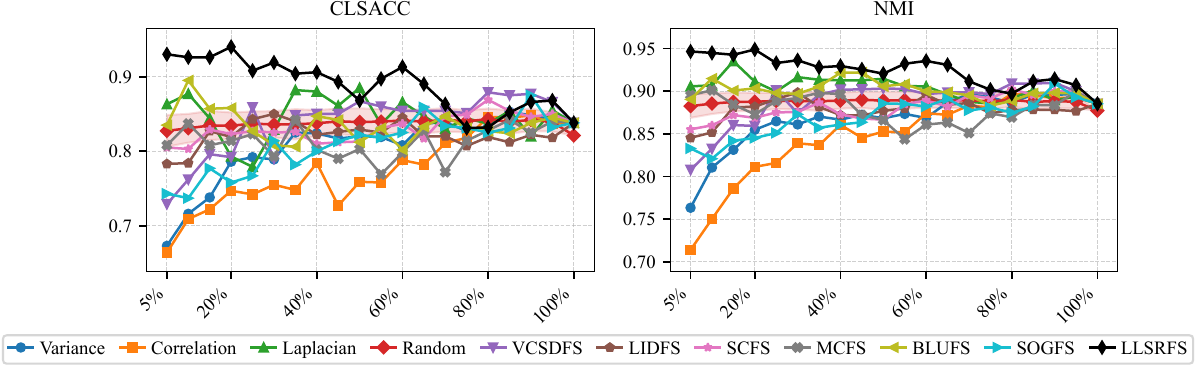}
\caption{CLSACC and NMI on the pixraw10P dataset (5\% to 100\% of the features).}\label{fig:supplot-unsup-100-pixraw10P}
\supback
\end{figure}
\begin{figure}[p]
\centering
\includegraphics[width=\textwidth,height=\supfigheight,keepaspectratio]{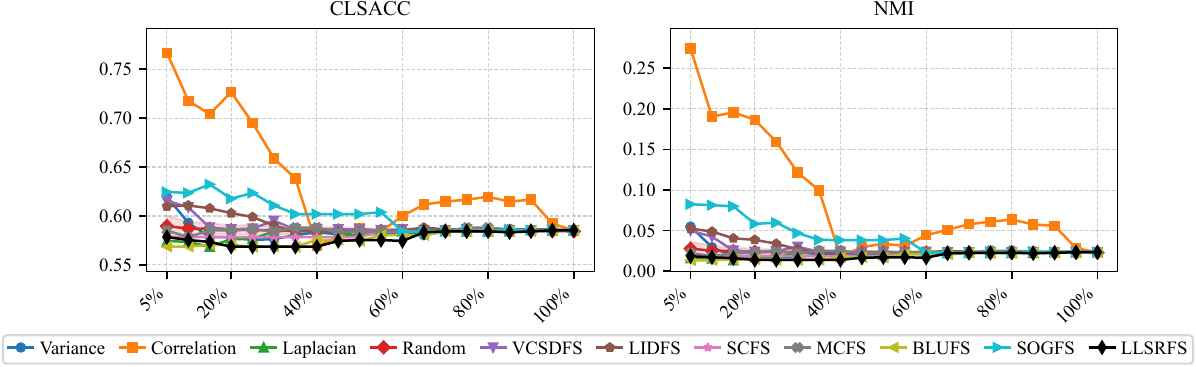}
\caption{CLSACC and NMI on the Prostate\_GE dataset (5\% to 100\% of the features).}\label{fig:supplot-unsup-100-Prostate_GE}
\supback
\end{figure}
\begin{figure}[p]
\centering
\includegraphics[width=\textwidth,height=\supfigheight,keepaspectratio]{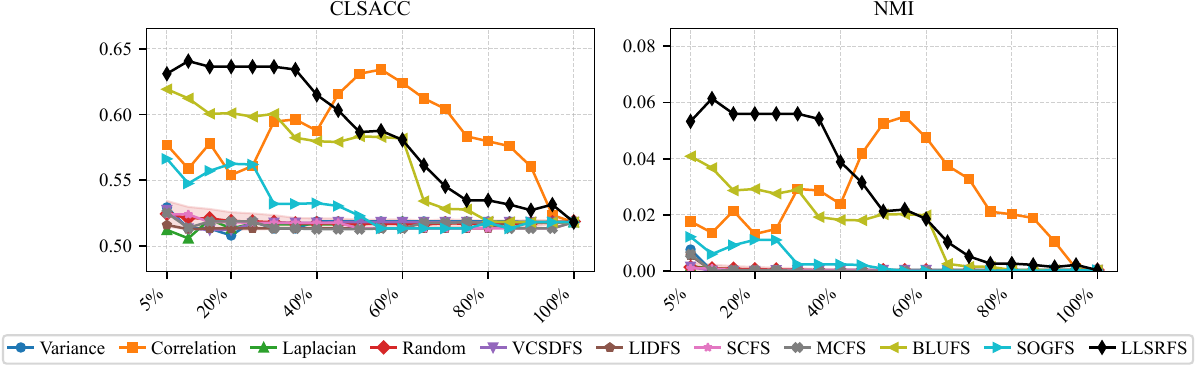}
\caption{CLSACC and NMI on the SMK\_CAN\_187 dataset (5\% to 100\% of the features).}\label{fig:supplot-unsup-100-SMK_CAN_187}
\supback
\end{figure}
\begin{figure}[p]
\centering
\includegraphics[width=\textwidth,height=\supfigheight,keepaspectratio]{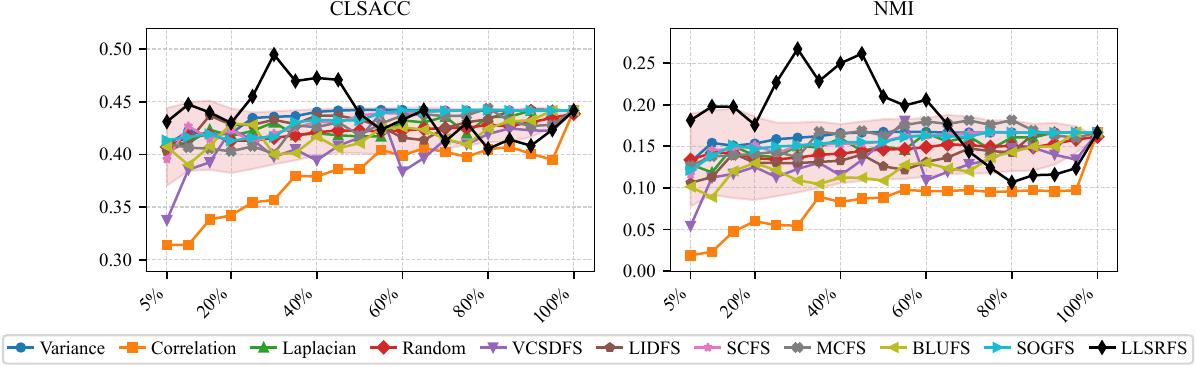}
\caption{CLSACC and NMI on the TOX\_171 dataset (5\% to 100\% of the features).}\label{fig:supplot-unsup-100-TOX_171}
\supback
\end{figure}
\begin{figure}[p]
\centering
\includegraphics[width=\textwidth,height=\supfigheight,keepaspectratio]{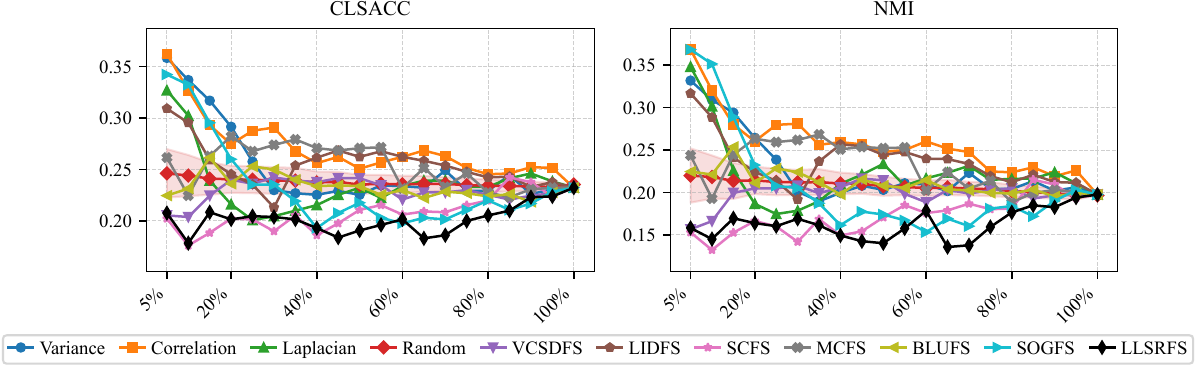}
\caption{CLSACC and NMI on the warpAR10P dataset (5\% to 100\% of the features).}\label{fig:supplot-unsup-100-warpAR10P}
\supback
\end{figure}
\begin{figure}[p]
\centering
\includegraphics[width=\textwidth,height=\supfigheight,keepaspectratio]{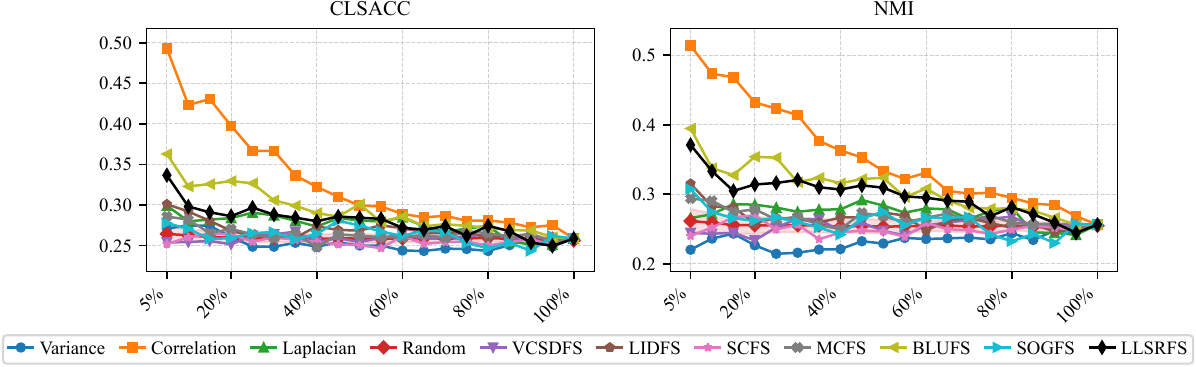}
\caption{CLSACC and NMI on the warpPIE10P dataset (5\% to 100\% of the features).}\label{fig:supplot-unsup-100-warpPIE10P}
\supback
\end{figure}
\begin{figure}[p]
\centering
\includegraphics[width=\textwidth,height=\supfigheight,keepaspectratio]{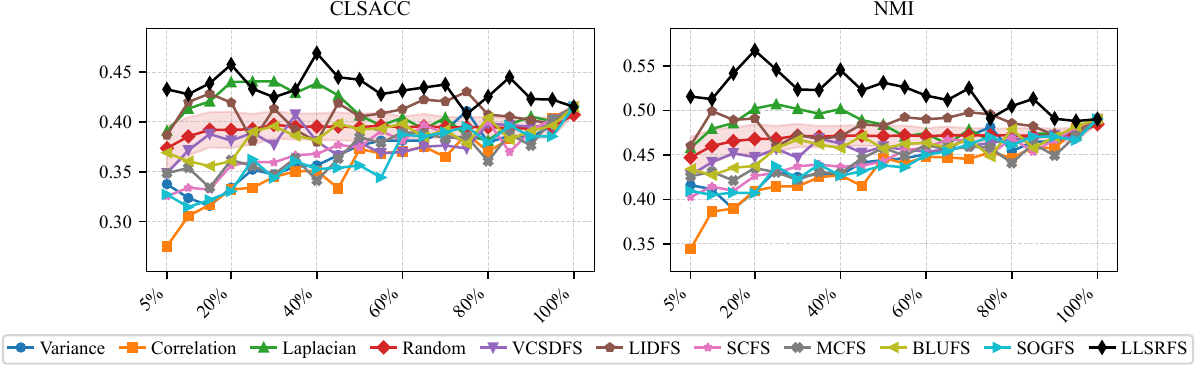}
\caption{CLSACC and NMI on the Yale dataset (5\% to 100\% of the features).}\label{fig:supplot-unsup-100-Yale}
\supback
\end{figure}
\clearpage

\begin{figure}[p]
\centering
\pdfbookmark[3]{KNN}{supplot-bm-knn}
\pdfbookmark[4]{0.5\% to 10\%}{supplot-bm-knn-10}
\includegraphics[width=\textwidth,height=\supfigheight,keepaspectratio]{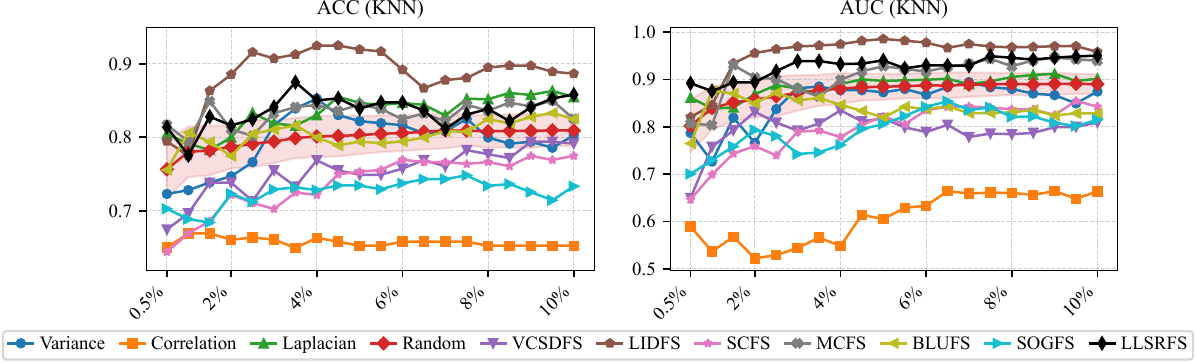}
\caption{ACC and AUC of the KNN classifier on the ALLAML dataset (0.5\% to 10\% of the features).}\label{fig:supplot-knn-10-ALLAML}
\supback
\end{figure}
\begin{figure}[p]
\centering
\includegraphics[width=\textwidth,height=\supfigheight,keepaspectratio]{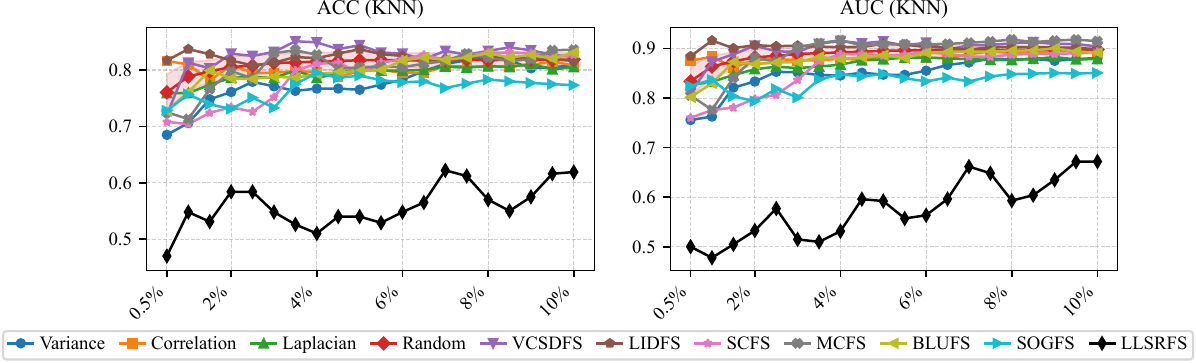}
\caption{ACC and AUC of the KNN classifier on the arcene dataset (0.5\% to 10\% of the features).}\label{fig:supplot-knn-10-arcene}
\supback
\end{figure}
\begin{figure}[p]
\centering
\includegraphics[width=\textwidth,height=\supfigheight,keepaspectratio]{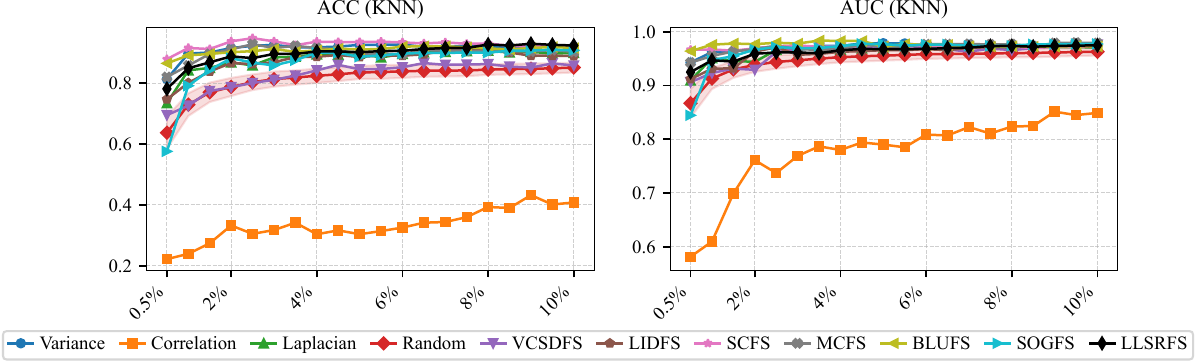}
\caption{ACC and AUC of the KNN classifier on the Carcinom dataset (0.5\% to 10\% of the features).}\label{fig:supplot-knn-10-Carcinom}
\supback
\end{figure}
\begin{figure}[p]
\centering
\includegraphics[width=\textwidth,height=\supfigheight,keepaspectratio]{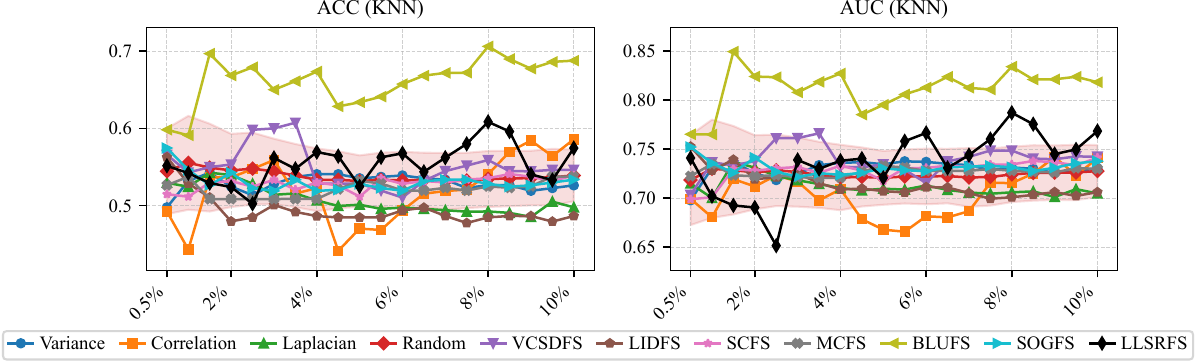}
\caption{ACC and AUC of the KNN classifier on the CLL\_SUB\_111 dataset (0.5\% to 10\% of the features).}\label{fig:supplot-knn-10-CLL_SUB_111}
\supback
\end{figure}
\begin{figure}[p]
\centering
\includegraphics[width=\textwidth,height=\supfigheight,keepaspectratio]{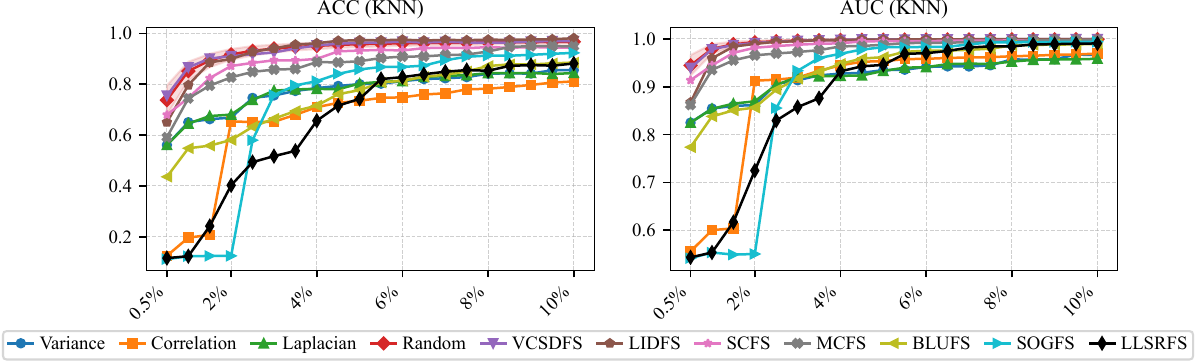}
\caption{ACC and AUC of the KNN classifier on the COIL20 dataset (0.5\% to 10\% of the features).}\label{fig:supplot-knn-10-COIL20}
\supback
\end{figure}
\begin{figure}[p]
\centering
\includegraphics[width=\textwidth,height=\supfigheight,keepaspectratio]{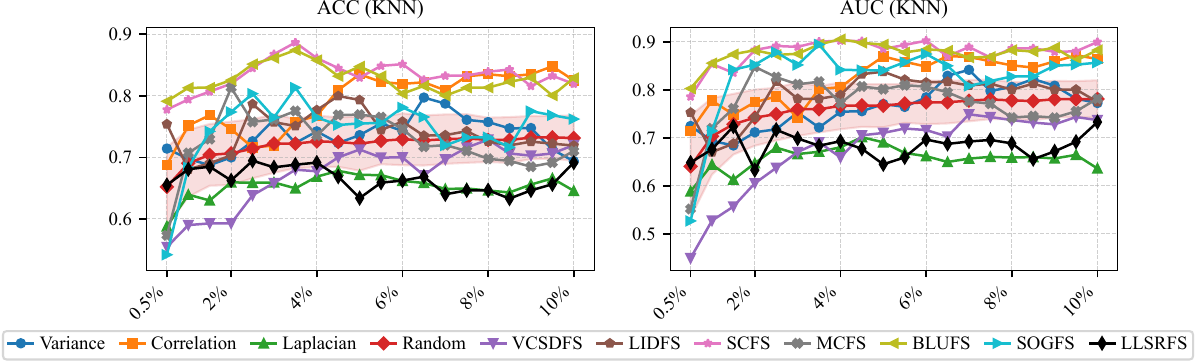}
\caption{ACC and AUC of the KNN classifier on the colon dataset (0.5\% to 10\% of the features).}\label{fig:supplot-knn-10-colon}
\supback
\end{figure}
\begin{figure}[p]
\centering
\includegraphics[width=\textwidth,height=\supfigheight,keepaspectratio]{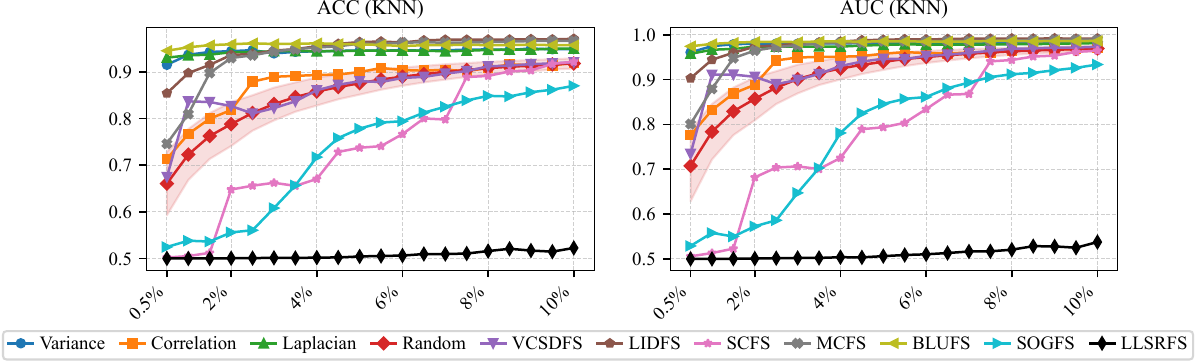}
\caption{ACC and AUC of the KNN classifier on the gisette dataset (0.5\% to 10\% of the features).}\label{fig:supplot-knn-10-gisette}
\supback
\end{figure}
\begin{figure}[p]
\centering
\includegraphics[width=\textwidth,height=\supfigheight,keepaspectratio]{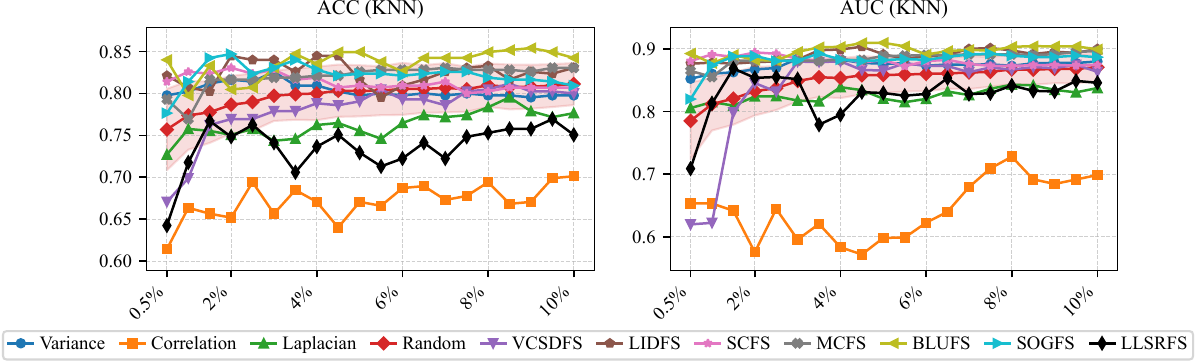}
\caption{ACC and AUC of the KNN classifier on the GLI\_85 dataset (0.5\% to 10\% of the features).}\label{fig:supplot-knn-10-GLI_85}
\supback
\end{figure}
\begin{figure}[p]
\centering
\includegraphics[width=\textwidth,height=\supfigheight,keepaspectratio]{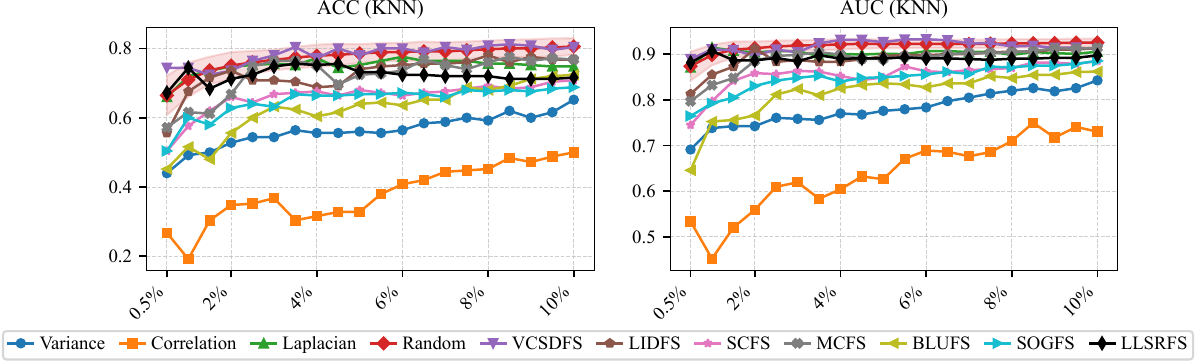}
\caption{ACC and AUC of the KNN classifier on the GLIOMA dataset (0.5\% to 10\% of the features).}\label{fig:supplot-knn-10-GLIOMA}
\supback
\end{figure}
\begin{figure}[p]
\centering
\includegraphics[width=\textwidth,height=\supfigheight,keepaspectratio]{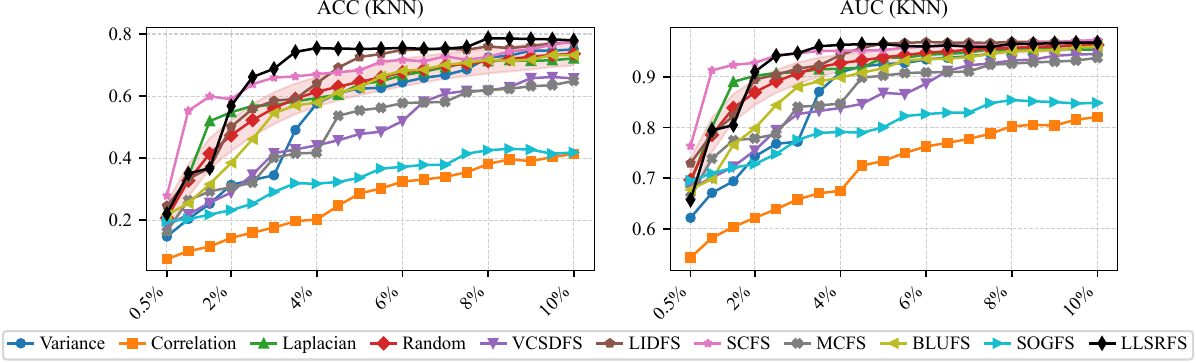}
\caption{ACC and AUC of the KNN classifier on the Isolet dataset (0.5\% to 10\% of the features).}\label{fig:supplot-knn-10-Isolet}
\supback
\end{figure}
\begin{figure}[p]
\centering
\includegraphics[width=\textwidth,height=\supfigheight,keepaspectratio]{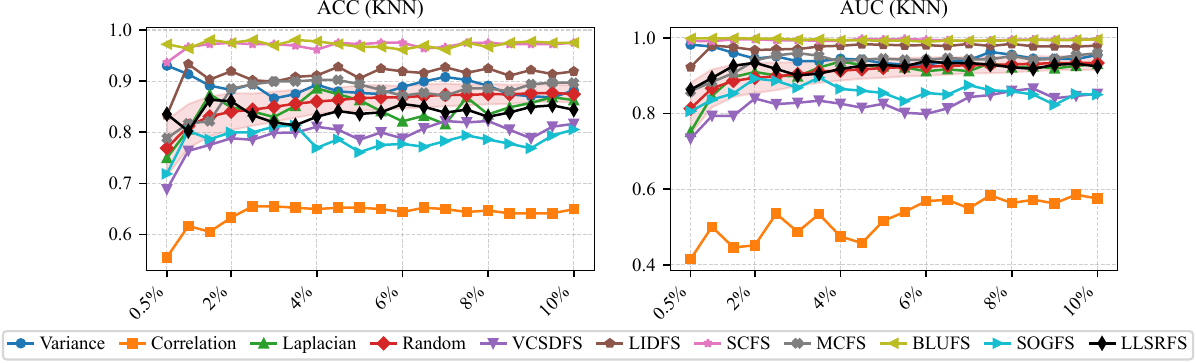}
\caption{ACC and AUC of the KNN classifier on the leukemia dataset (0.5\% to 10\% of the features).}\label{fig:supplot-knn-10-leukemia}
\supback
\end{figure}
\begin{figure}[p]
\centering
\includegraphics[width=\textwidth,height=\supfigheight,keepaspectratio]{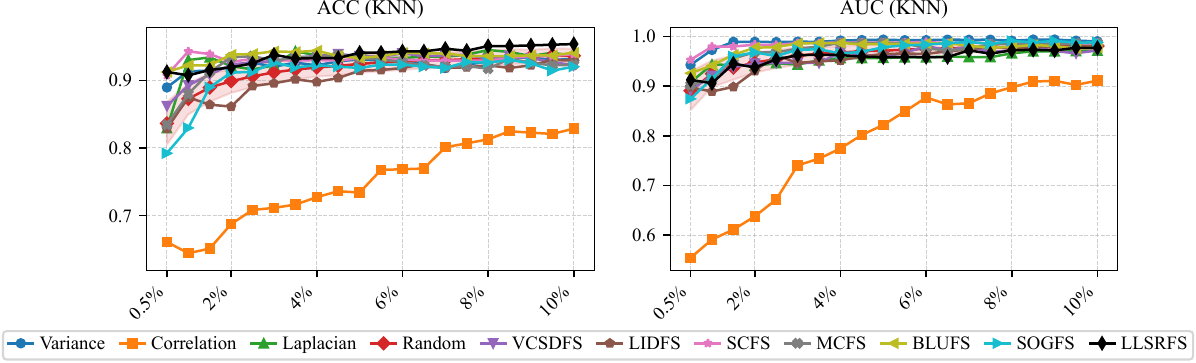}
\caption{ACC and AUC of the KNN classifier on the lung dataset (0.5\% to 10\% of the features).}\label{fig:supplot-knn-10-lung}
\supback
\end{figure}
\begin{figure}[p]
\centering
\includegraphics[width=\textwidth,height=\supfigheight,keepaspectratio]{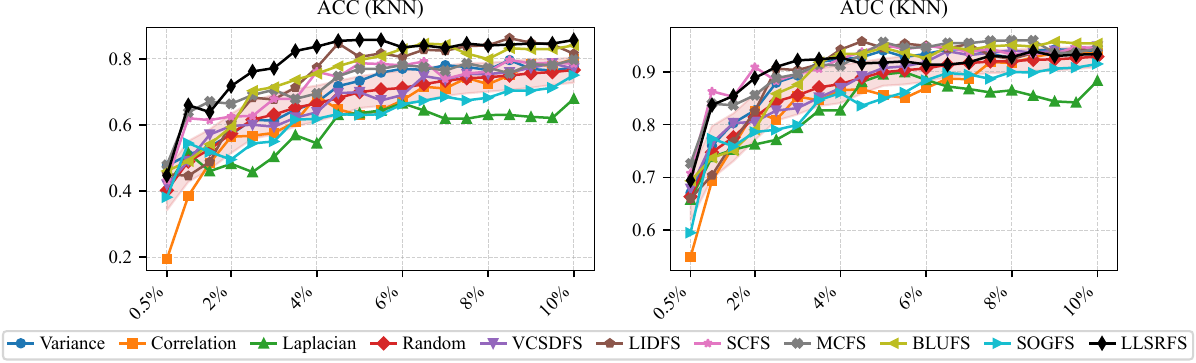}
\caption{ACC and AUC of the KNN classifier on the lung\_discrete dataset (0.5\% to 10\% of the features).}\label{fig:supplot-knn-10-lung_discrete}
\supback
\end{figure}
\begin{figure}[p]
\centering
\includegraphics[width=\textwidth,height=\supfigheight,keepaspectratio]{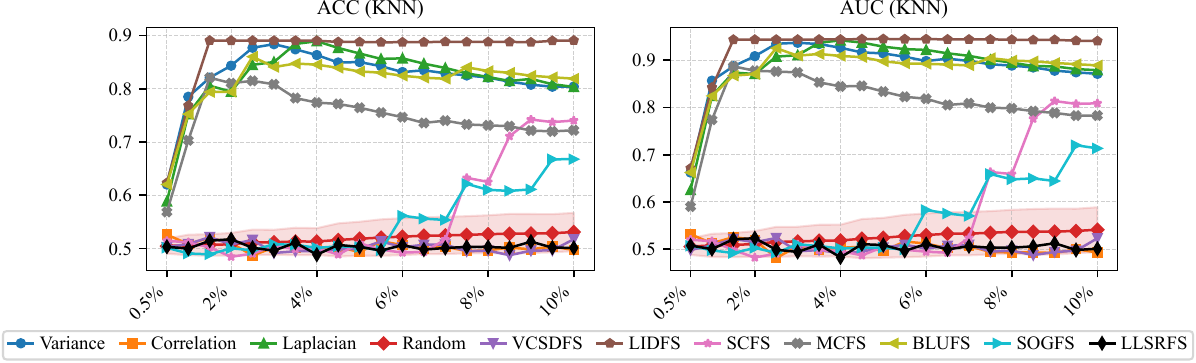}
\caption{ACC and AUC of the KNN classifier on the madelon dataset (0.5\% to 10\% of the features).}\label{fig:supplot-knn-10-madelon}
\supback
\end{figure}
\begin{figure}[p]
\centering
\includegraphics[width=\textwidth,height=\supfigheight,keepaspectratio]{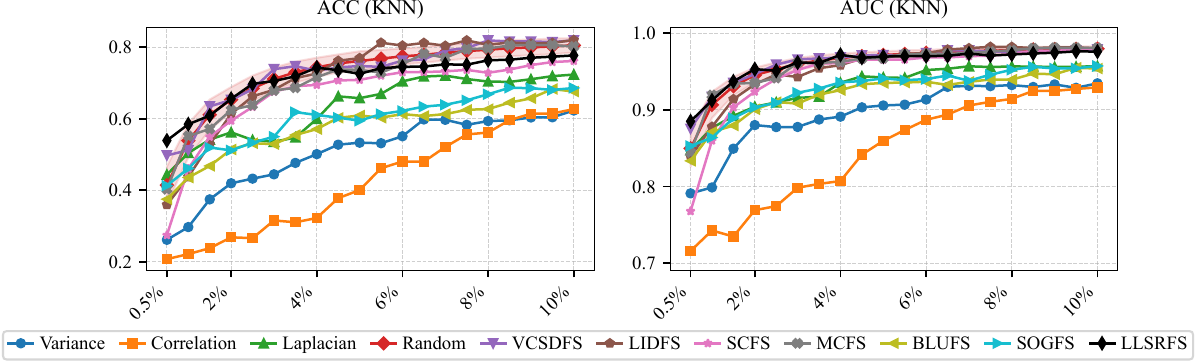}
\caption{ACC and AUC of the KNN classifier on the ORL dataset (0.5\% to 10\% of the features).}\label{fig:supplot-knn-10-ORL}
\supback
\end{figure}
\begin{figure}[p]
\centering
\includegraphics[width=\textwidth,height=\supfigheight,keepaspectratio]{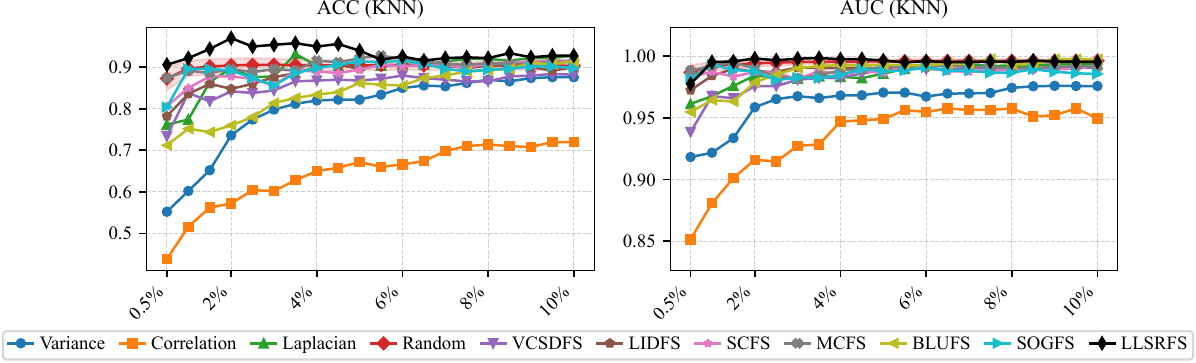}
\caption{ACC and AUC of the KNN classifier on the orlraws10P dataset (0.5\% to 10\% of the features).}\label{fig:supplot-knn-10-orlraws10P}
\supback
\end{figure}
\begin{figure}[p]
\centering
\includegraphics[width=\textwidth,height=\supfigheight,keepaspectratio]{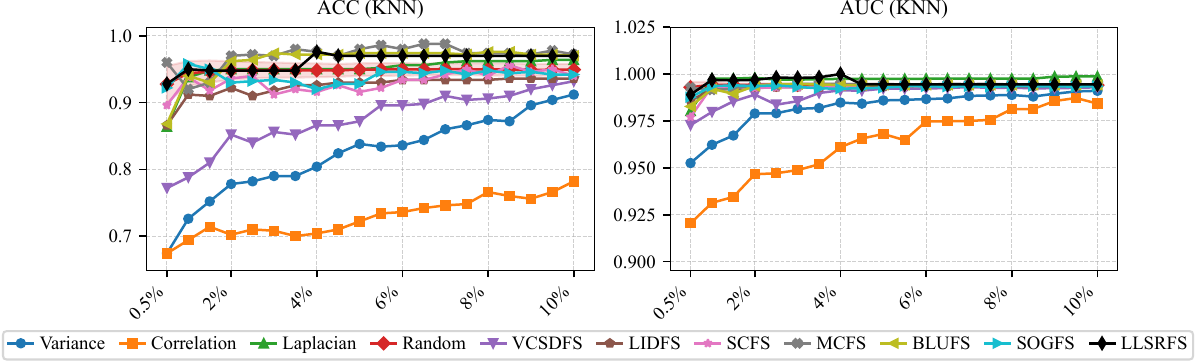}
\caption{ACC and AUC of the KNN classifier on the pixraw10P dataset (0.5\% to 10\% of the features).}\label{fig:supplot-knn-10-pixraw10P}
\supback
\end{figure}
\begin{figure}[p]
\centering
\includegraphics[width=\textwidth,height=\supfigheight,keepaspectratio]{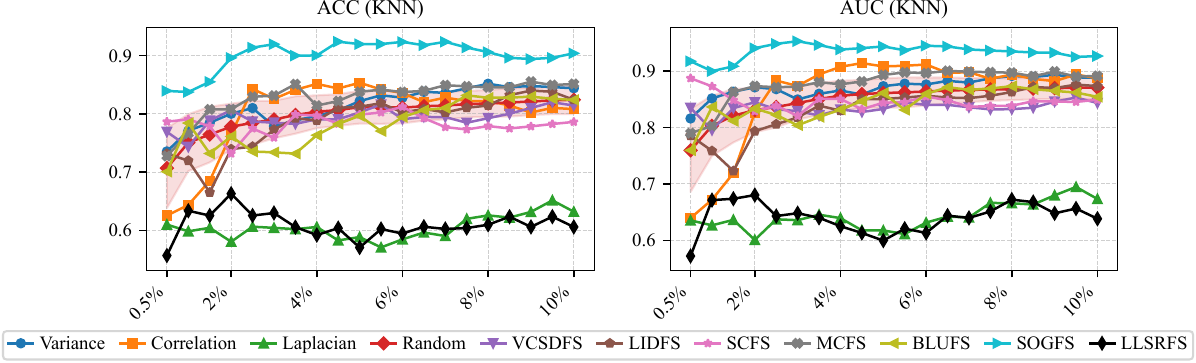}
\caption{ACC and AUC of the KNN classifier on the Prostate\_GE dataset (0.5\% to 10\% of the features).}\label{fig:supplot-knn-10-Prostate_GE}
\supback
\end{figure}
\begin{figure}[p]
\centering
\includegraphics[width=\textwidth,height=\supfigheight,keepaspectratio]{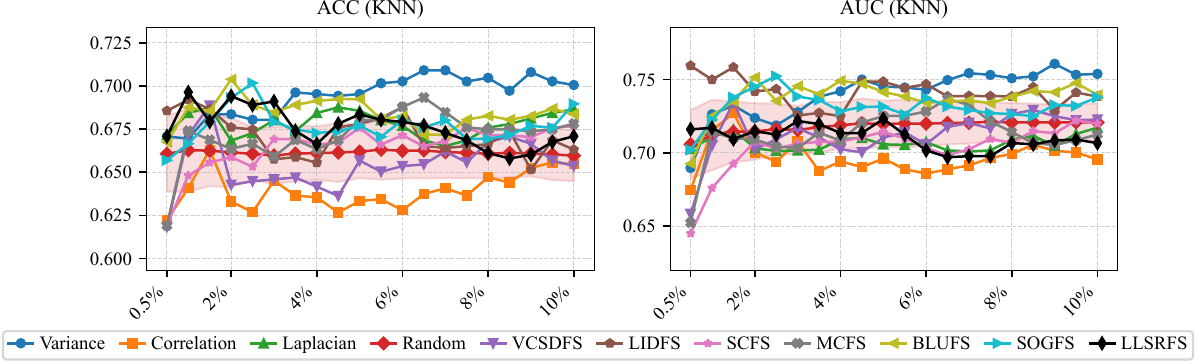}
\caption{ACC and AUC of the KNN classifier on the SMK\_CAN\_187 dataset (0.5\% to 10\% of the features).}\label{fig:supplot-knn-10-SMK_CAN_187}
\supback
\end{figure}
\begin{figure}[p]
\centering
\includegraphics[width=\textwidth,height=\supfigheight,keepaspectratio]{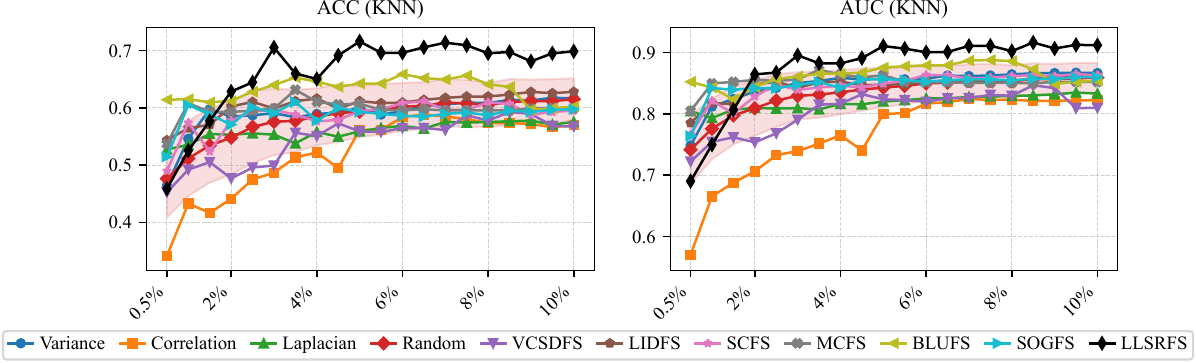}
\caption{ACC and AUC of the KNN classifier on the TOX\_171 dataset (0.5\% to 10\% of the features).}\label{fig:supplot-knn-10-TOX_171}
\supback
\end{figure}
\begin{figure}[p]
\centering
\includegraphics[width=\textwidth,height=\supfigheight,keepaspectratio]{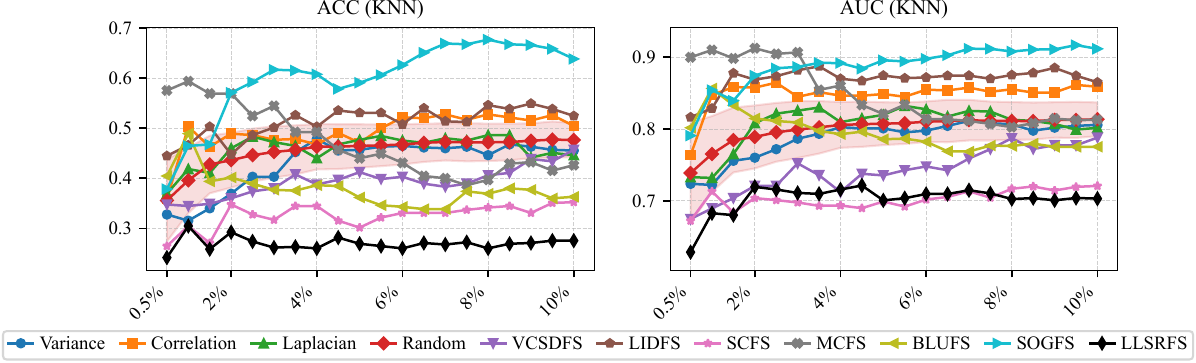}
\caption{ACC and AUC of the KNN classifier on the warpAR10P dataset (0.5\% to 10\% of the features).}\label{fig:supplot-knn-10-warpAR10P}
\supback
\end{figure}
\begin{figure}[p]
\centering
\includegraphics[width=\textwidth,height=\supfigheight,keepaspectratio]{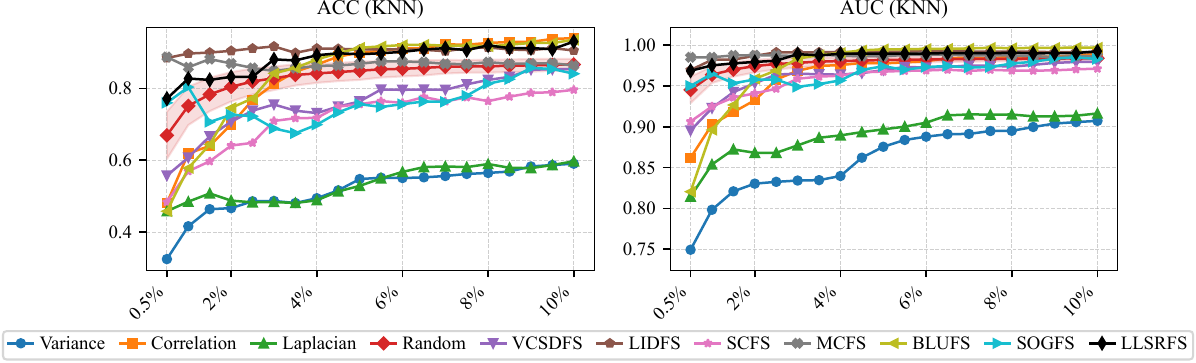}
\caption{ACC and AUC of the KNN classifier on the warpPIE10P dataset (0.5\% to 10\% of the features).}\label{fig:supplot-knn-10-warpPIE10P}
\supback
\end{figure}
\begin{figure}[p]
\centering
\includegraphics[width=\textwidth,height=\supfigheight,keepaspectratio]{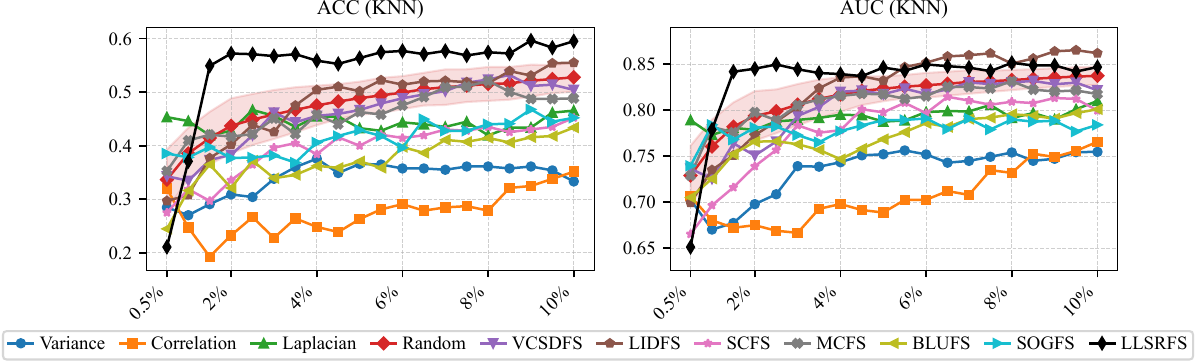}
\caption{ACC and AUC of the KNN classifier on the Yale dataset (0.5\% to 10\% of the features).}\label{fig:supplot-knn-10-Yale}
\supback
\end{figure}
\begin{figure}[p]
\centering
\pdfbookmark[4]{5\% to 100\%}{supplot-bm-knn-100}
\includegraphics[width=\textwidth,height=\supfigheight,keepaspectratio]{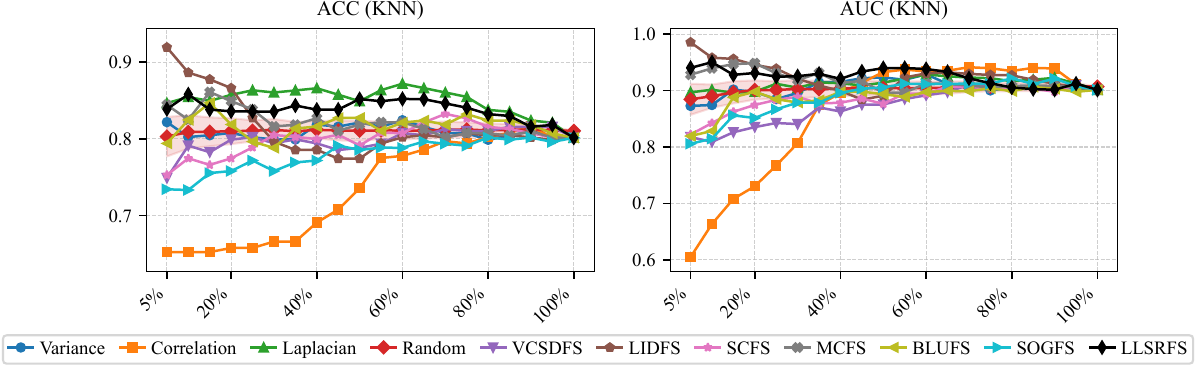}
\caption{ACC and AUC of the KNN classifier on the ALLAML dataset (5\% to 100\% of the features).}\label{fig:supplot-knn-100-ALLAML}
\supback
\end{figure}
\begin{figure}[p]
\centering
\includegraphics[width=\textwidth,height=\supfigheight,keepaspectratio]{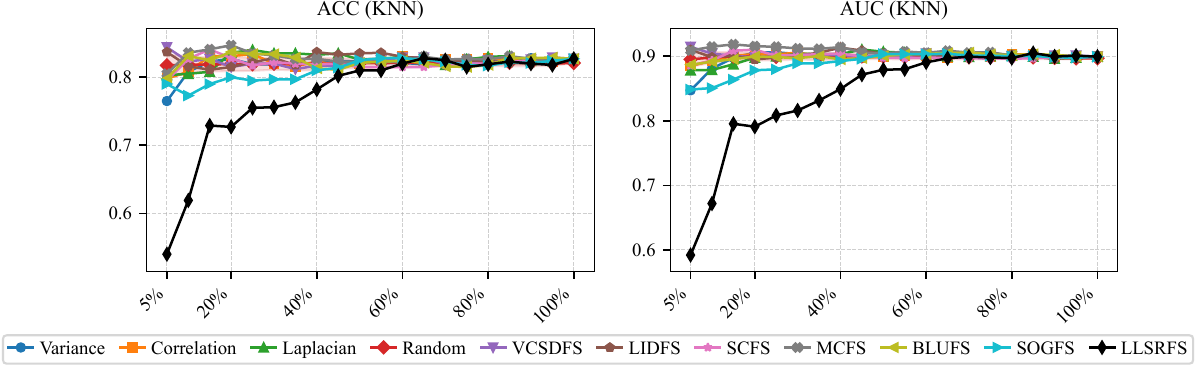}
\caption{ACC and AUC of the KNN classifier on the arcene dataset (5\% to 100\% of the features).}\label{fig:supplot-knn-100-arcene}
\supback
\end{figure}
\begin{figure}[p]
\centering
\includegraphics[width=\textwidth,height=\supfigheight,keepaspectratio]{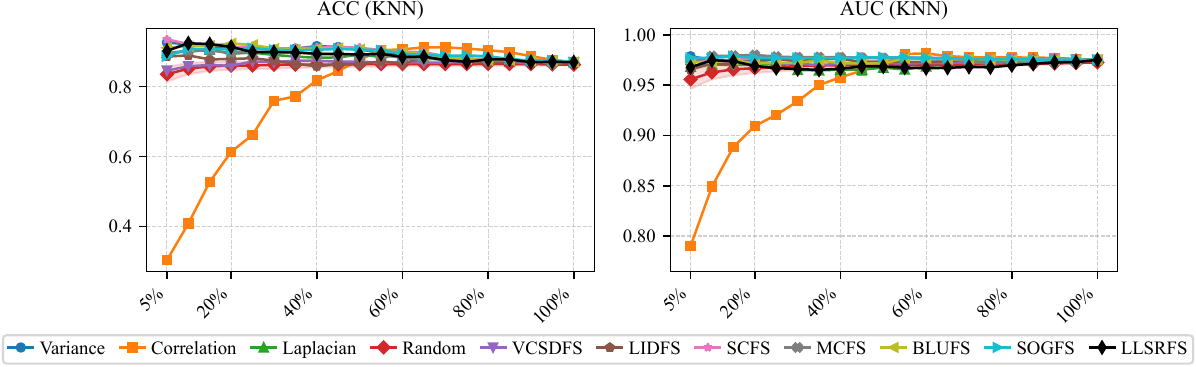}
\caption{ACC and AUC of the KNN classifier on the Carcinom dataset (5\% to 100\% of the features).}\label{fig:supplot-knn-100-Carcinom}
\supback
\end{figure}
\begin{figure}[p]
\centering
\includegraphics[width=\textwidth,height=\supfigheight,keepaspectratio]{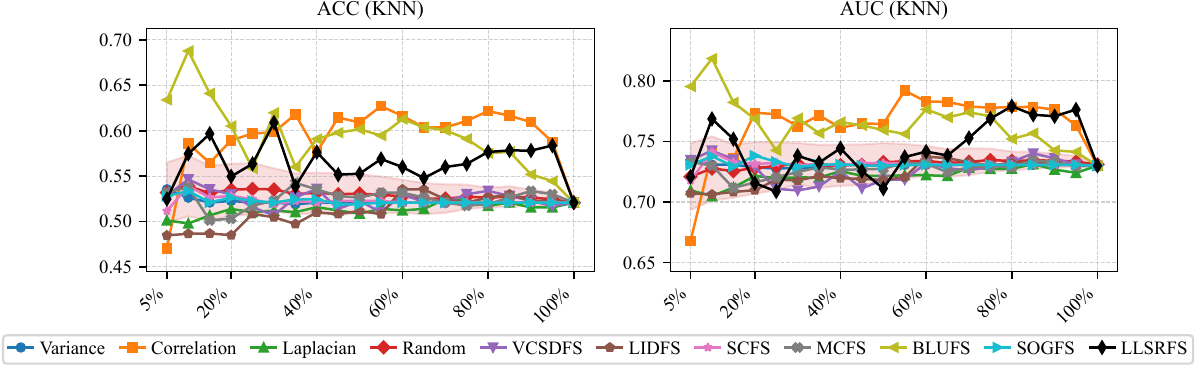}
\caption{ACC and AUC of the KNN classifier on the CLL\_SUB\_111 dataset (5\% to 100\% of the features).}\label{fig:supplot-knn-100-CLL_SUB_111}
\supback
\end{figure}
\begin{figure}[p]
\centering
\includegraphics[width=\textwidth,height=\supfigheight,keepaspectratio]{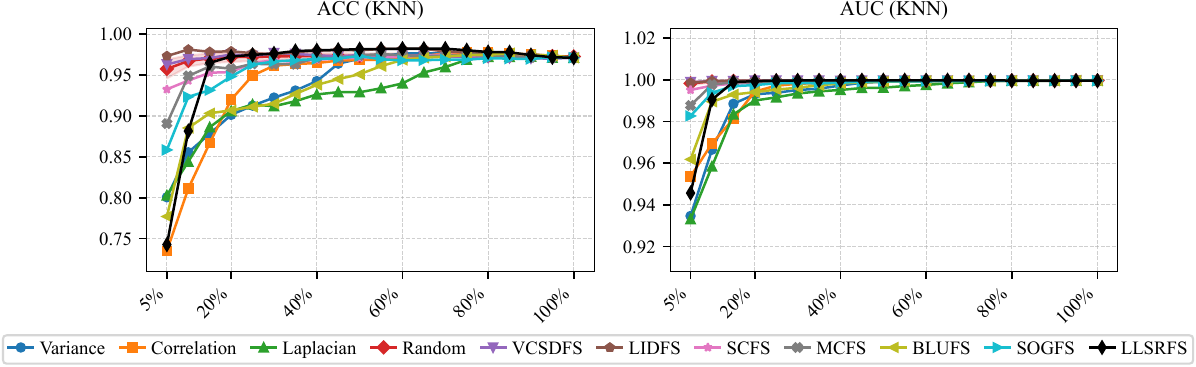}
\caption{ACC and AUC of the KNN classifier on the COIL20 dataset (5\% to 100\% of the features).}\label{fig:supplot-knn-100-COIL20}
\supback
\end{figure}
\begin{figure}[p]
\centering
\includegraphics[width=\textwidth,height=\supfigheight,keepaspectratio]{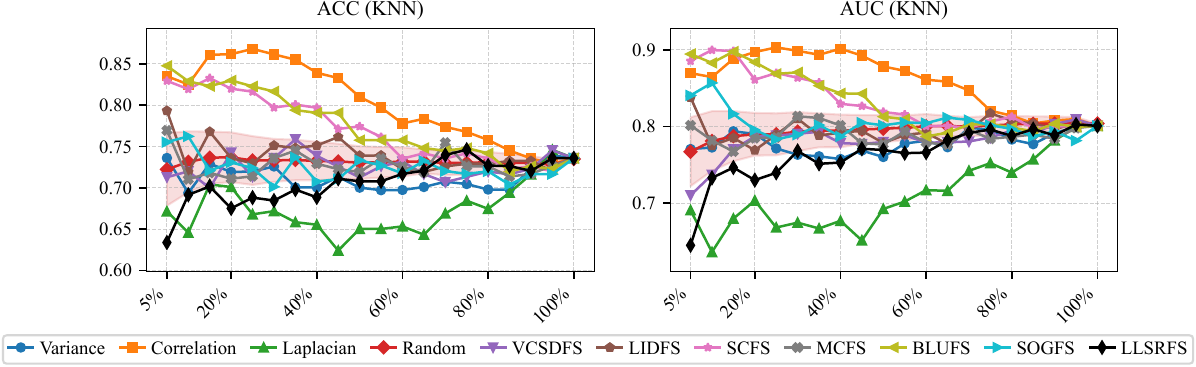}
\caption{ACC and AUC of the KNN classifier on the colon dataset (5\% to 100\% of the features).}\label{fig:supplot-knn-100-colon}
\supback
\end{figure}
\begin{figure}[p]
\centering
\includegraphics[width=\textwidth,height=\supfigheight,keepaspectratio]{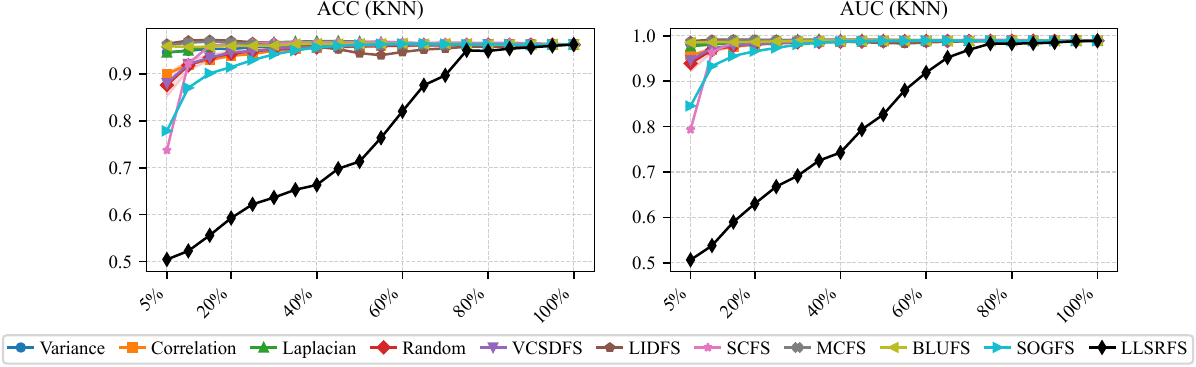}
\caption{ACC and AUC of the KNN classifier on the gisette dataset (5\% to 100\% of the features).}\label{fig:supplot-knn-100-gisette}
\supback
\end{figure}
\begin{figure}[p]
\centering
\includegraphics[width=\textwidth,height=\supfigheight,keepaspectratio]{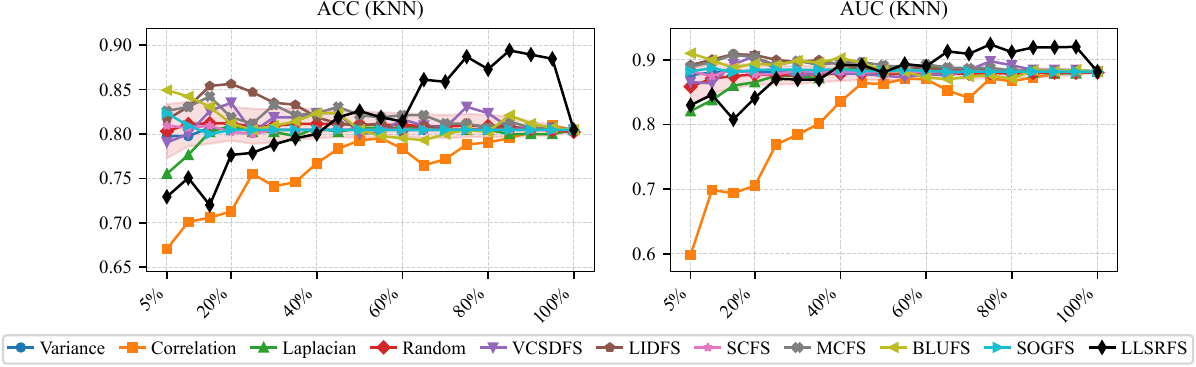}
\caption{ACC and AUC of the KNN classifier on the GLI\_85 dataset (5\% to 100\% of the features).}\label{fig:supplot-knn-100-GLI_85}
\supback
\end{figure}
\begin{figure}[p]
\centering
\includegraphics[width=\textwidth,height=\supfigheight,keepaspectratio]{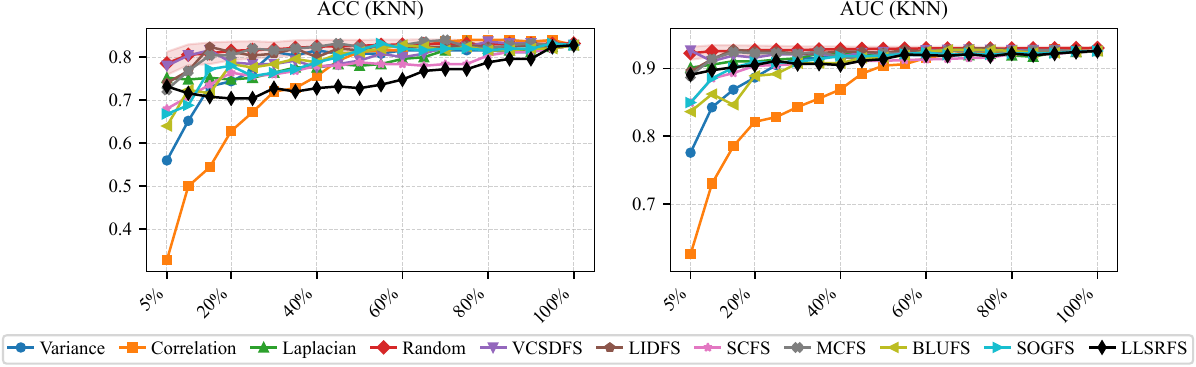}
\caption{ACC and AUC of the KNN classifier on the GLIOMA dataset (5\% to 100\% of the features).}\label{fig:supplot-knn-100-GLIOMA}
\supback
\end{figure}
\begin{figure}[p]
\centering
\includegraphics[width=\textwidth,height=\supfigheight,keepaspectratio]{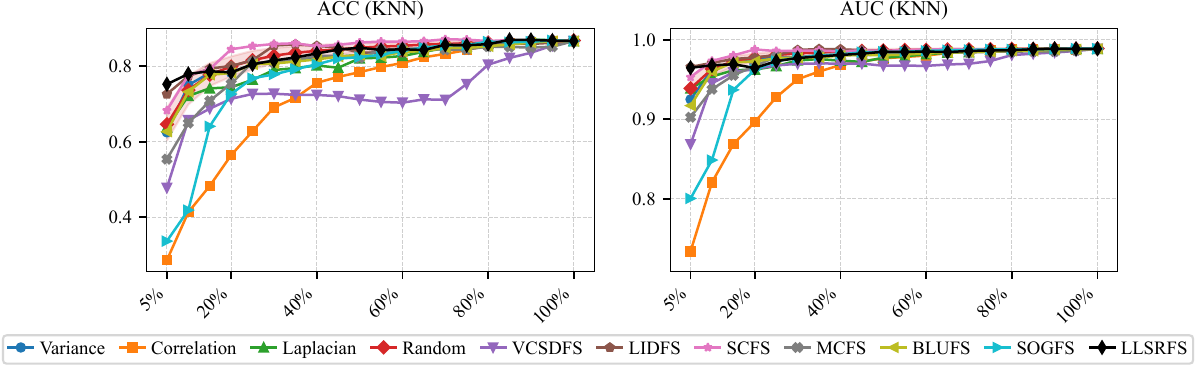}
\caption{ACC and AUC of the KNN classifier on the Isolet dataset (5\% to 100\% of the features).}\label{fig:supplot-knn-100-Isolet}
\supback
\end{figure}
\begin{figure}[p]
\centering
\includegraphics[width=\textwidth,height=\supfigheight,keepaspectratio]{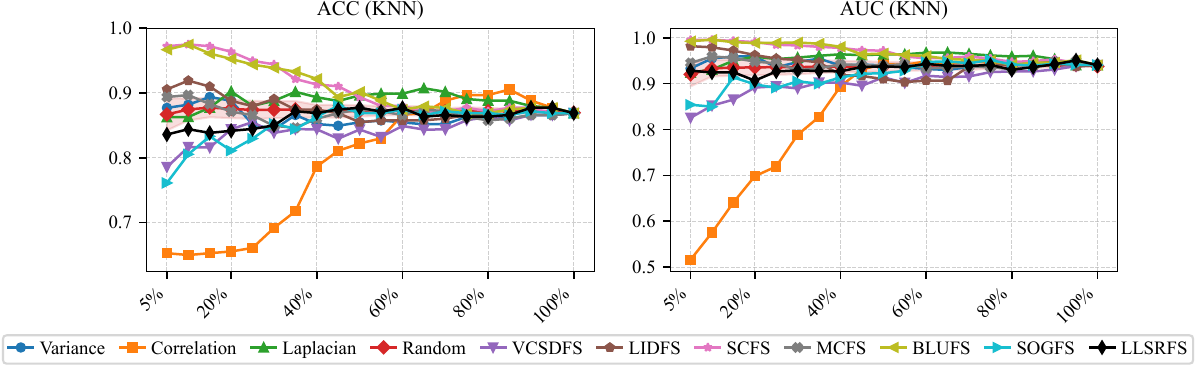}
\caption{ACC and AUC of the KNN classifier on the leukemia dataset (5\% to 100\% of the features).}\label{fig:supplot-knn-100-leukemia}
\supback
\end{figure}
\begin{figure}[p]
\centering
\includegraphics[width=\textwidth,height=\supfigheight,keepaspectratio]{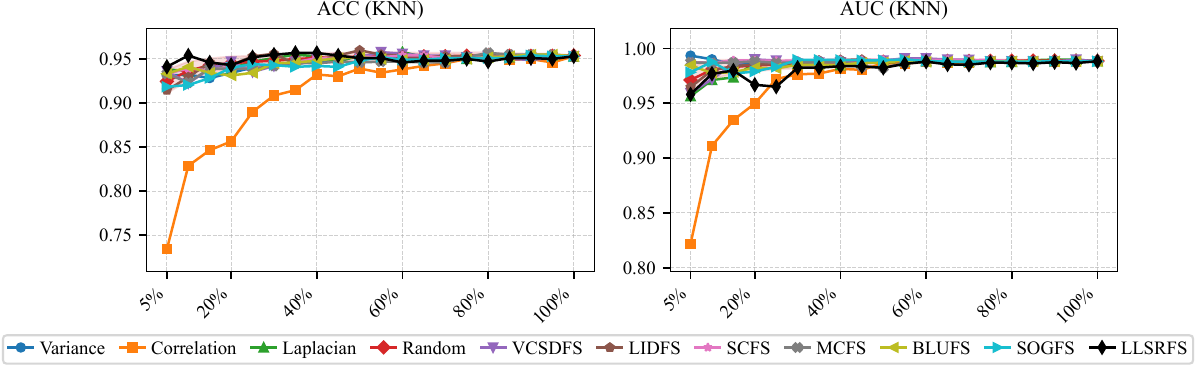}
\caption{ACC and AUC of the KNN classifier on the lung dataset (5\% to 100\% of the features).}\label{fig:supplot-knn-100-lung}
\supback
\end{figure}
\begin{figure}[p]
\centering
\includegraphics[width=\textwidth,height=\supfigheight,keepaspectratio]{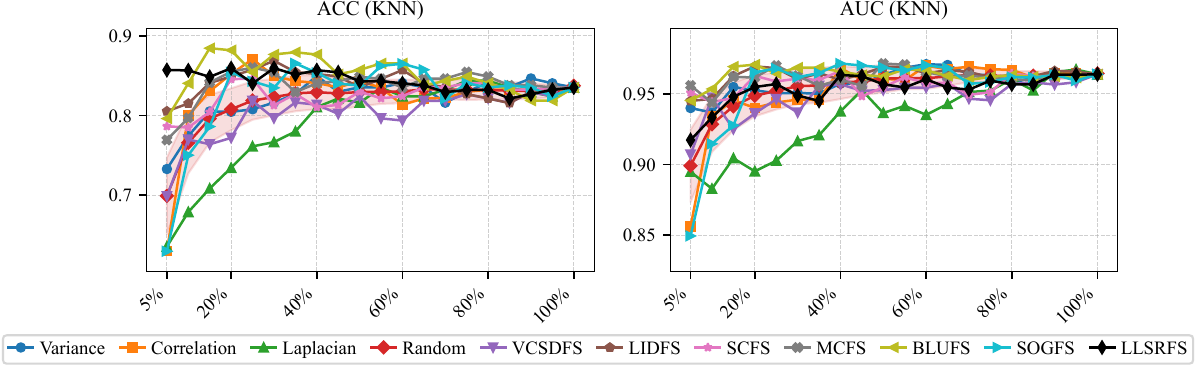}
\caption{ACC and AUC of the KNN classifier on the lung\_discrete dataset (5\% to 100\% of the features).}\label{fig:supplot-knn-100-lung_discrete}
\supback
\end{figure}
\begin{figure}[p]
\centering
\includegraphics[width=\textwidth,height=\supfigheight,keepaspectratio]{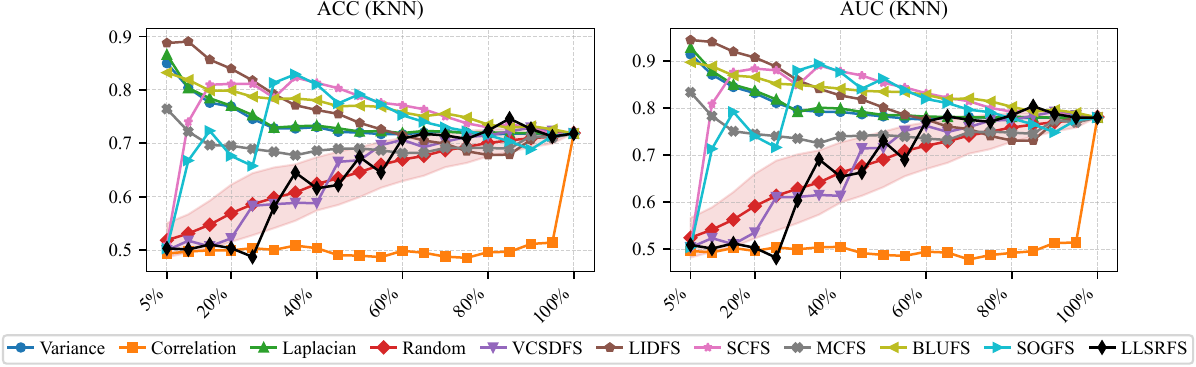}
\caption{ACC and AUC of the KNN classifier on the madelon dataset (5\% to 100\% of the features).}\label{fig:supplot-knn-100-madelon}
\supback
\end{figure}
\begin{figure}[p]
\centering
\includegraphics[width=\textwidth,height=\supfigheight,keepaspectratio]{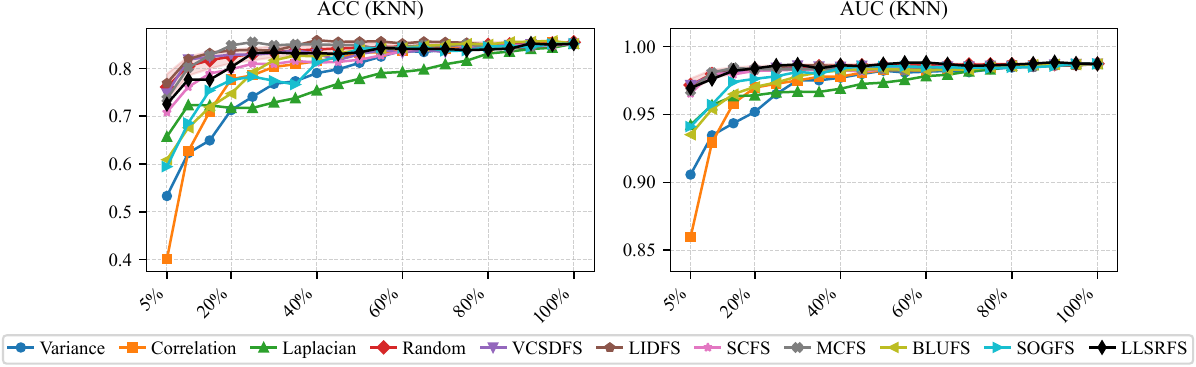}
\caption{ACC and AUC of the KNN classifier on the ORL dataset (5\% to 100\% of the features).}\label{fig:supplot-knn-100-ORL}
\supback
\end{figure}
\begin{figure}[p]
\centering
\includegraphics[width=\textwidth,height=\supfigheight,keepaspectratio]{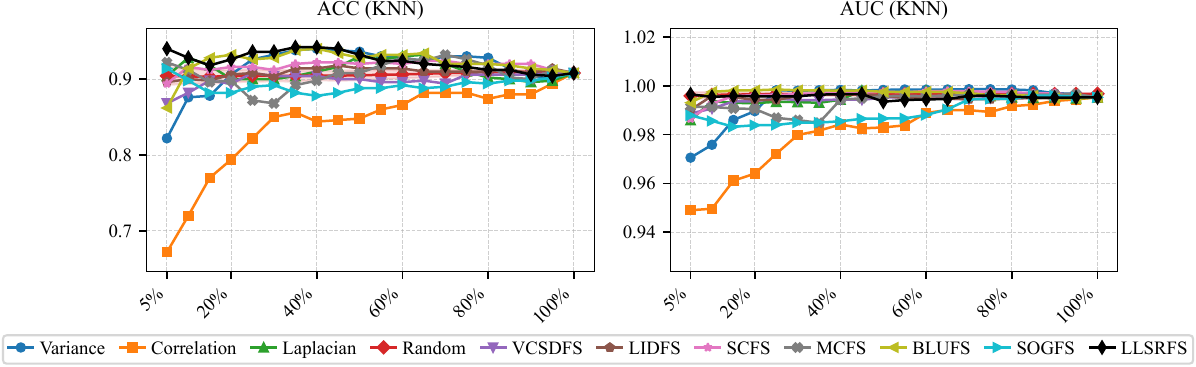}
\caption{ACC and AUC of the KNN classifier on the orlraws10P dataset (5\% to 100\% of the features).}\label{fig:supplot-knn-100-orlraws10P}
\supback
\end{figure}
\begin{figure}[p]
\centering
\includegraphics[width=\textwidth,height=\supfigheight,keepaspectratio]{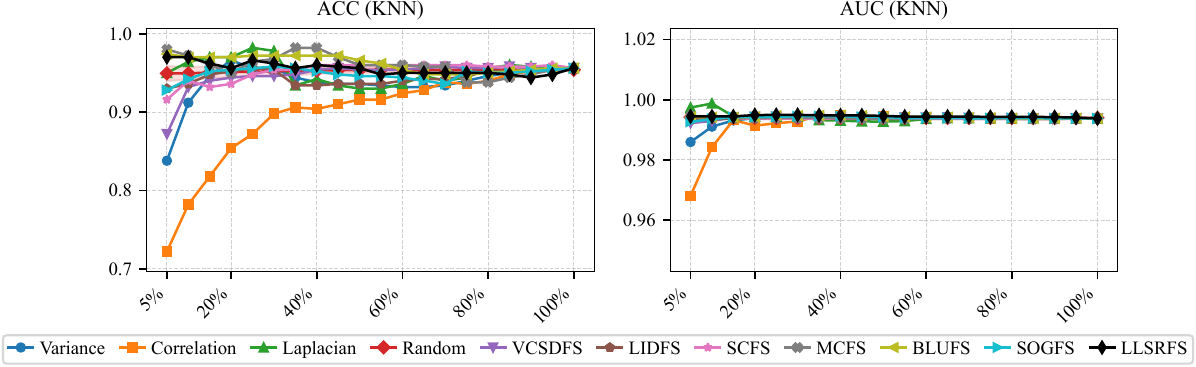}
\caption{ACC and AUC of the KNN classifier on the pixraw10P dataset (5\% to 100\% of the features).}\label{fig:supplot-knn-100-pixraw10P}
\supback
\end{figure}
\begin{figure}[p]
\centering
\includegraphics[width=\textwidth,height=\supfigheight,keepaspectratio]{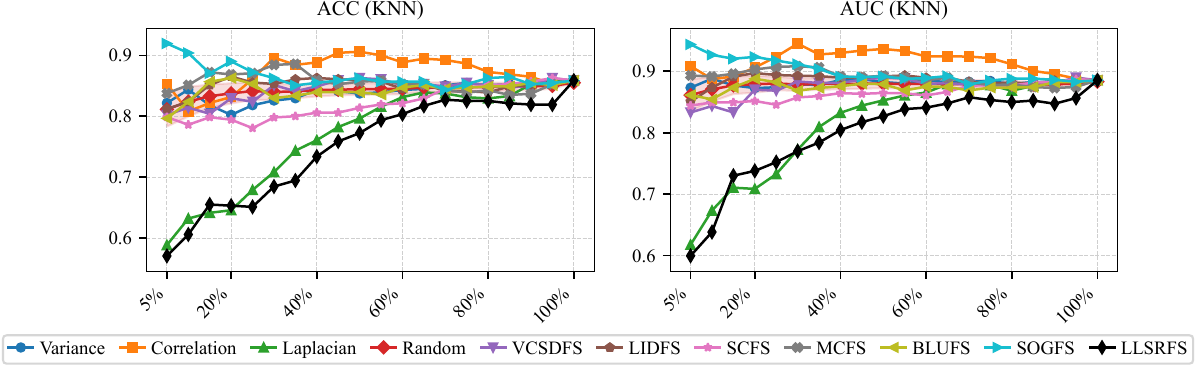}
\caption{ACC and AUC of the KNN classifier on the Prostate\_GE dataset (5\% to 100\% of the features).}\label{fig:supplot-knn-100-Prostate_GE}
\supback
\end{figure}
\begin{figure}[p]
\centering
\includegraphics[width=\textwidth,height=\supfigheight,keepaspectratio]{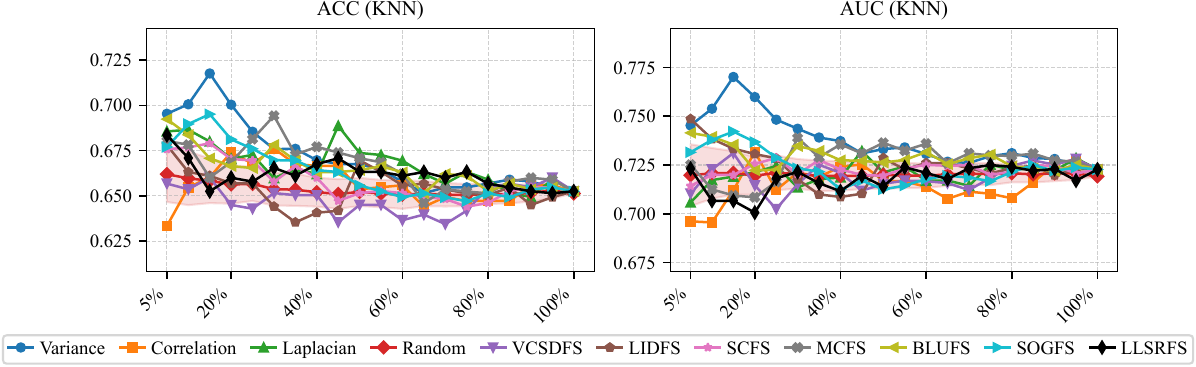}
\caption{ACC and AUC of the KNN classifier on the SMK\_CAN\_187 dataset (5\% to 100\% of the features).}\label{fig:supplot-knn-100-SMK_CAN_187}
\supback
\end{figure}
\begin{figure}[p]
\centering
\includegraphics[width=\textwidth,height=\supfigheight,keepaspectratio]{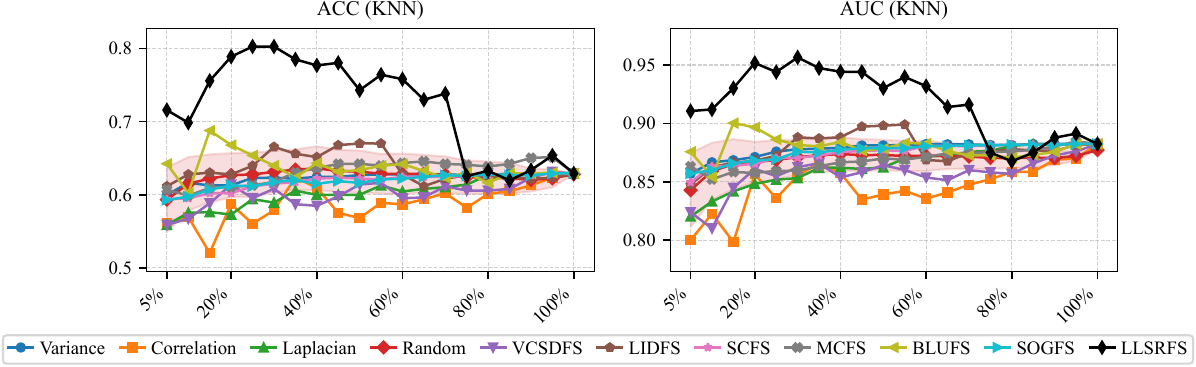}
\caption{ACC and AUC of the KNN classifier on the TOX\_171 dataset (5\% to 100\% of the features).}\label{fig:supplot-knn-100-TOX_171}
\supback
\end{figure}
\begin{figure}[p]
\centering
\includegraphics[width=\textwidth,height=\supfigheight,keepaspectratio]{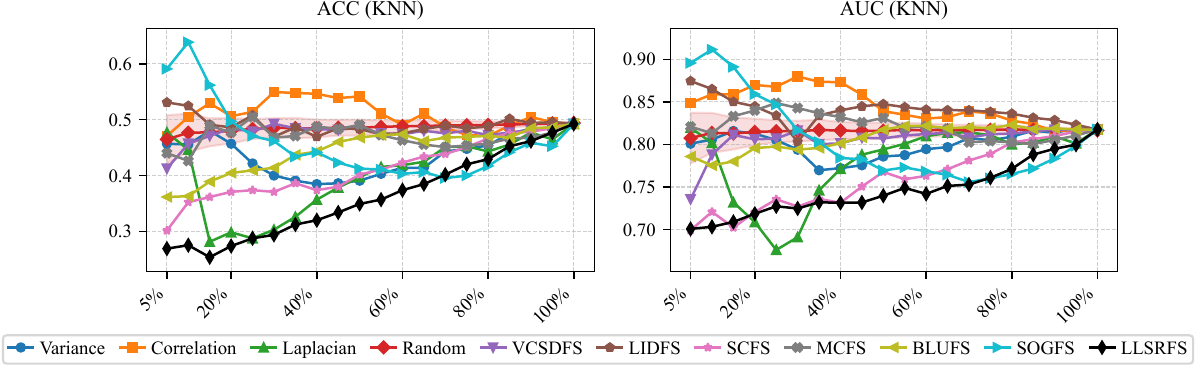}
\caption{ACC and AUC of the KNN classifier on the warpAR10P dataset (5\% to 100\% of the features).}\label{fig:supplot-knn-100-warpAR10P}
\supback
\end{figure}
\begin{figure}[p]
\centering
\includegraphics[width=\textwidth,height=\supfigheight,keepaspectratio]{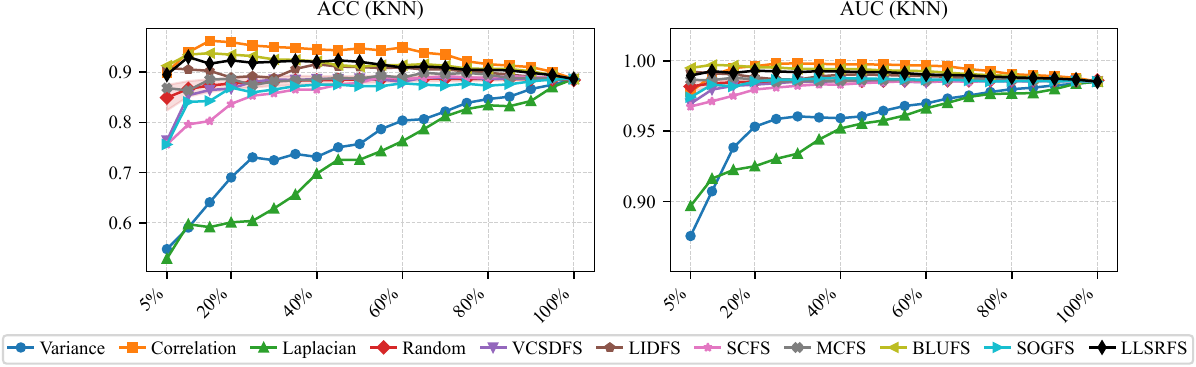}
\caption{ACC and AUC of the KNN classifier on the warpPIE10P dataset (5\% to 100\% of the features).}\label{fig:supplot-knn-100-warpPIE10P}
\supback
\end{figure}
\begin{figure}[p]
\centering
\includegraphics[width=\textwidth,height=\supfigheight,keepaspectratio]{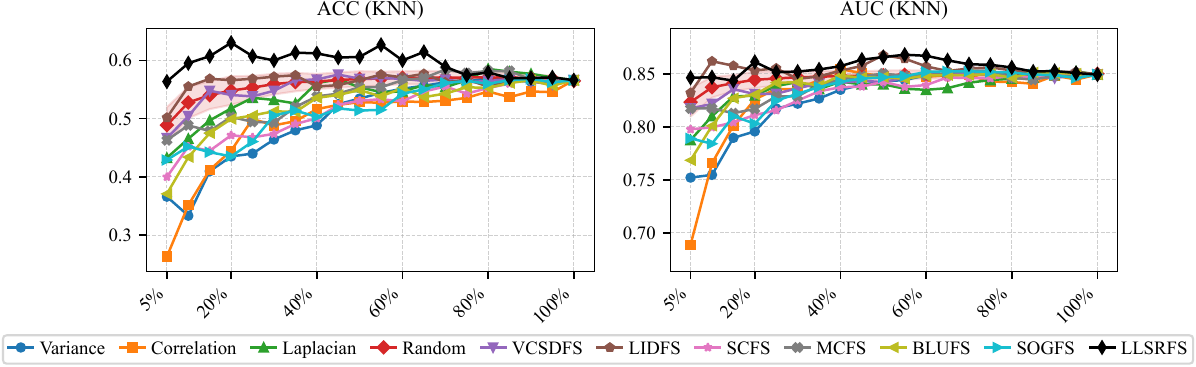}
\caption{ACC and AUC of the KNN classifier on the Yale dataset (5\% to 100\% of the features).}\label{fig:supplot-knn-100-Yale}
\supback
\end{figure}
\clearpage

\begin{figure}[p]
\centering
\pdfbookmark[3]{Decision tree}{supplot-bm-dt}
\pdfbookmark[4]{0.5\% to 10\%}{supplot-bm-dt-10}
\includegraphics[width=\textwidth,height=\supfigheight,keepaspectratio]{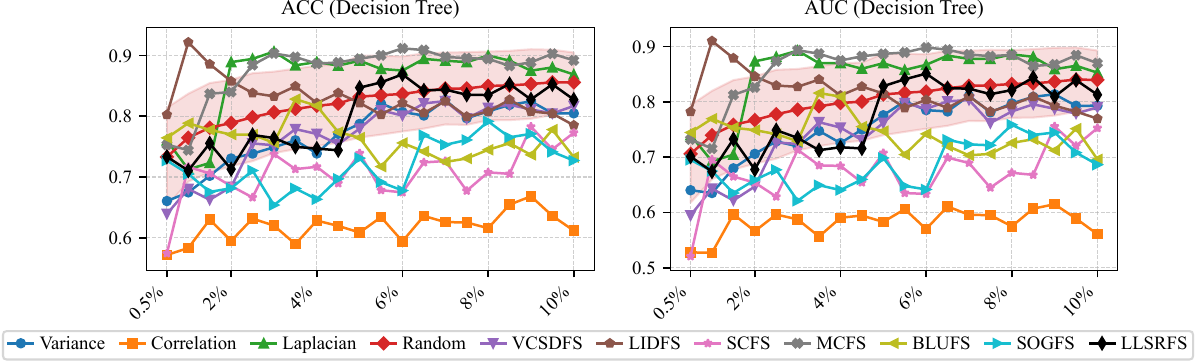}
\caption{ACC and AUC of the decision tree classifier on the ALLAML dataset (0.5\% to 10\% of the features).}\label{fig:supplot-dt-10-ALLAML}
\supback
\end{figure}
\begin{figure}[p]
\centering
\includegraphics[width=\textwidth,height=\supfigheight,keepaspectratio]{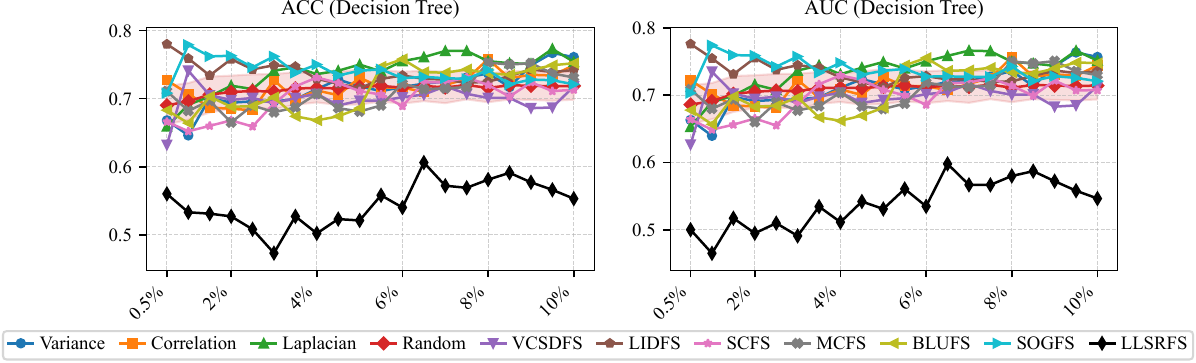}
\caption{ACC and AUC of the decision tree classifier on the arcene dataset (0.5\% to 10\% of the features).}\label{fig:supplot-dt-10-arcene}
\supback
\end{figure}
\begin{figure}[p]
\centering
\includegraphics[width=\textwidth,height=\supfigheight,keepaspectratio]{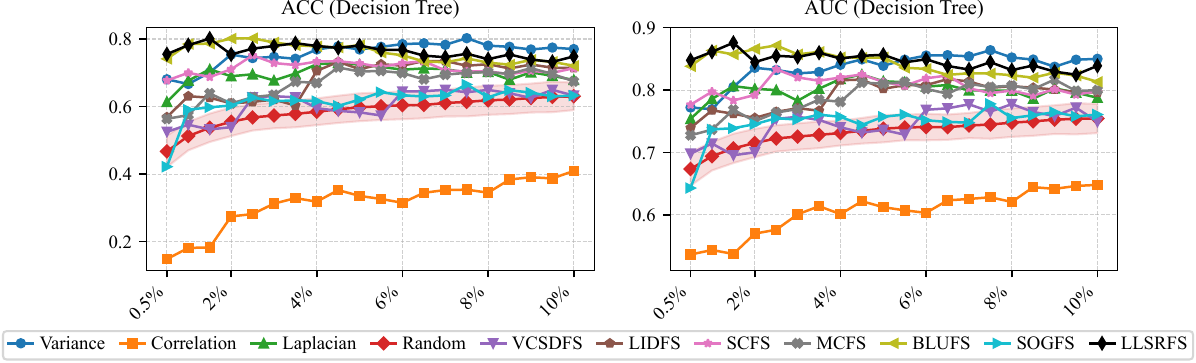}
\caption{ACC and AUC of the decision tree classifier on the Carcinom dataset (0.5\% to 10\% of the features).}\label{fig:supplot-dt-10-Carcinom}
\supback
\end{figure}
\begin{figure}[p]
\centering
\includegraphics[width=\textwidth,height=\supfigheight,keepaspectratio]{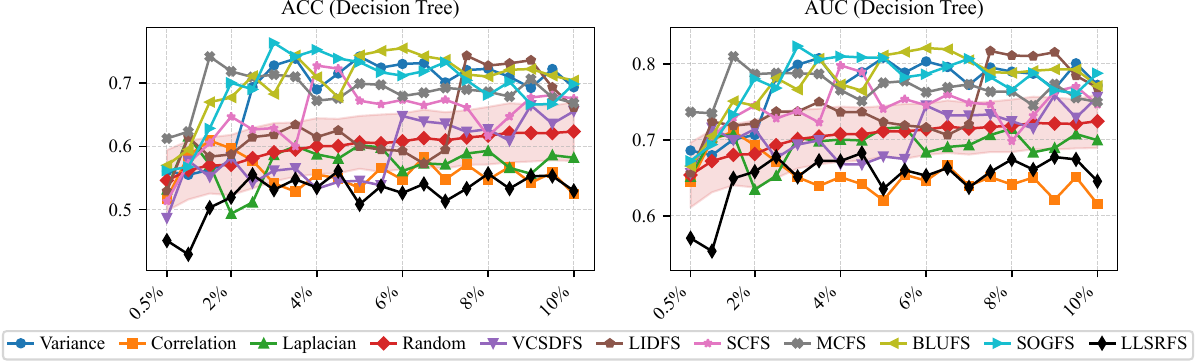}
\caption{ACC and AUC of the decision tree classifier on the CLL\_SUB\_111 dataset (0.5\% to 10\% of the features).}\label{fig:supplot-dt-10-CLL_SUB_111}
\supback
\end{figure}
\begin{figure}[p]
\centering
\includegraphics[width=\textwidth,height=\supfigheight,keepaspectratio]{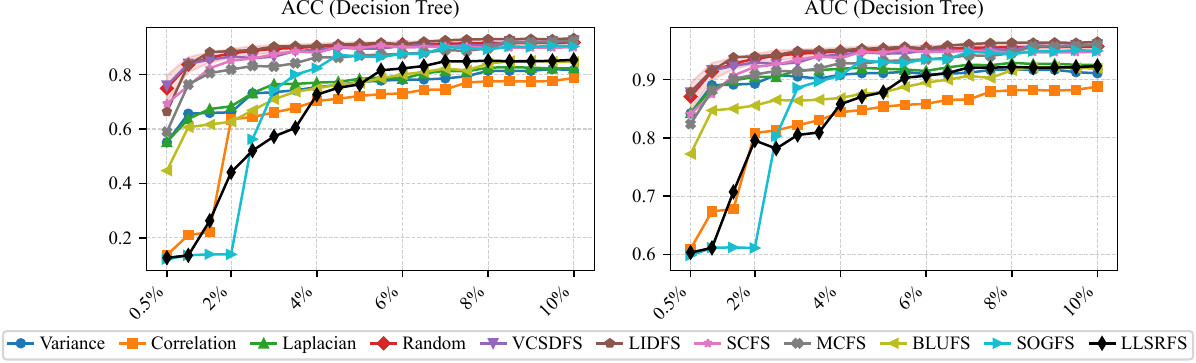}
\caption{ACC and AUC of the decision tree classifier on the COIL20 dataset (0.5\% to 10\% of the features).}\label{fig:supplot-dt-10-COIL20}
\supback
\end{figure}
\begin{figure}[p]
\centering
\includegraphics[width=\textwidth,height=\supfigheight,keepaspectratio]{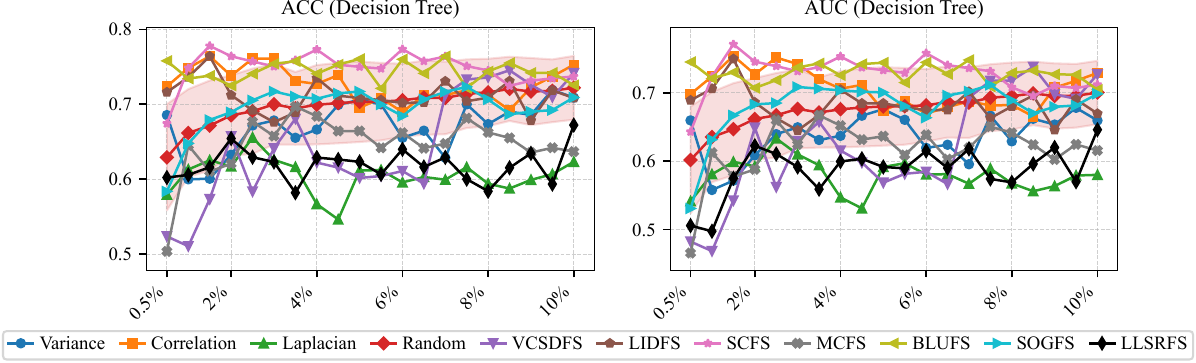}
\caption{ACC and AUC of the decision tree classifier on the colon dataset (0.5\% to 10\% of the features).}\label{fig:supplot-dt-10-colon}
\supback
\end{figure}
\begin{figure}[p]
\centering
\includegraphics[width=\textwidth,height=\supfigheight,keepaspectratio]{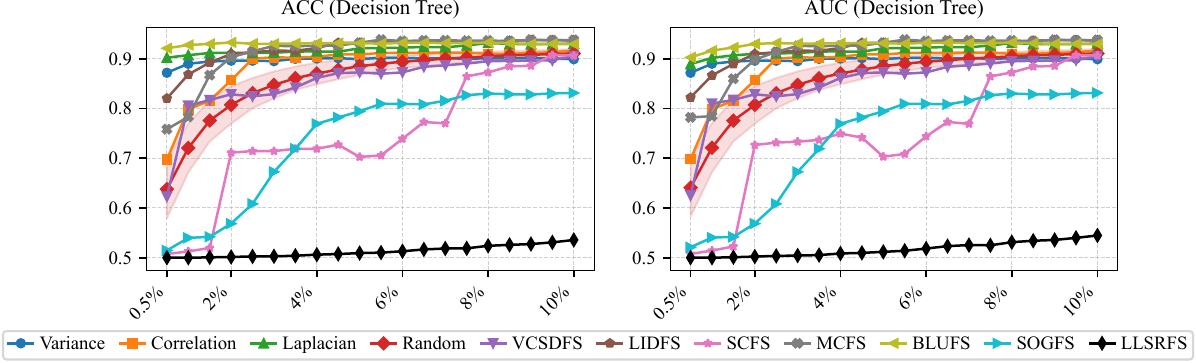}
\caption{ACC and AUC of the decision tree classifier on the gisette dataset (0.5\% to 10\% of the features).}\label{fig:supplot-dt-10-gisette}
\supback
\end{figure}
\begin{figure}[p]
\centering
\includegraphics[width=\textwidth,height=\supfigheight,keepaspectratio]{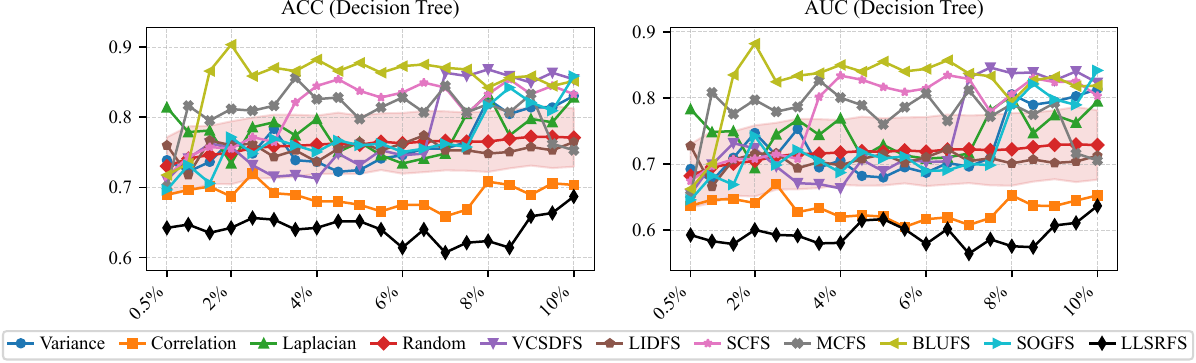}
\caption{ACC and AUC of the decision tree classifier on the GLI\_85 dataset (0.5\% to 10\% of the features).}\label{fig:supplot-dt-10-GLI_85}
\supback
\end{figure}
\begin{figure}[p]
\centering
\includegraphics[width=\textwidth,height=\supfigheight,keepaspectratio]{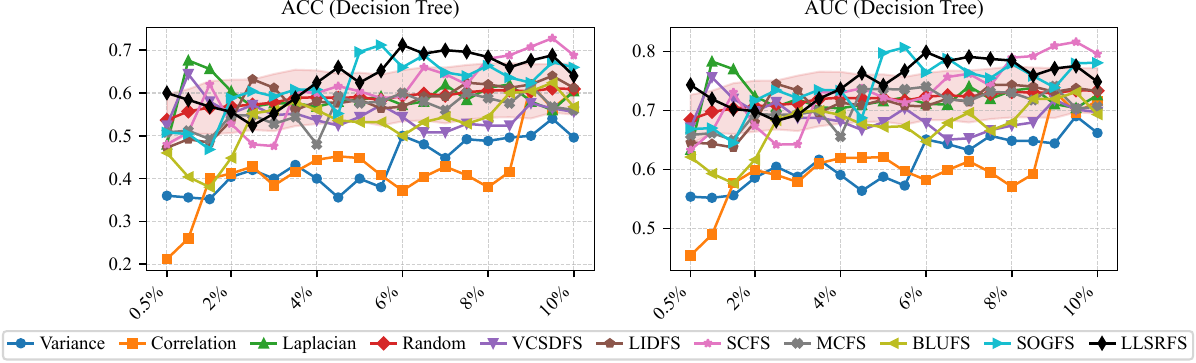}
\caption{ACC and AUC of the decision tree classifier on the GLIOMA dataset (0.5\% to 10\% of the features).}\label{fig:supplot-dt-10-GLIOMA}
\supback
\end{figure}
\begin{figure}[p]
\centering
\includegraphics[width=\textwidth,height=\supfigheight,keepaspectratio]{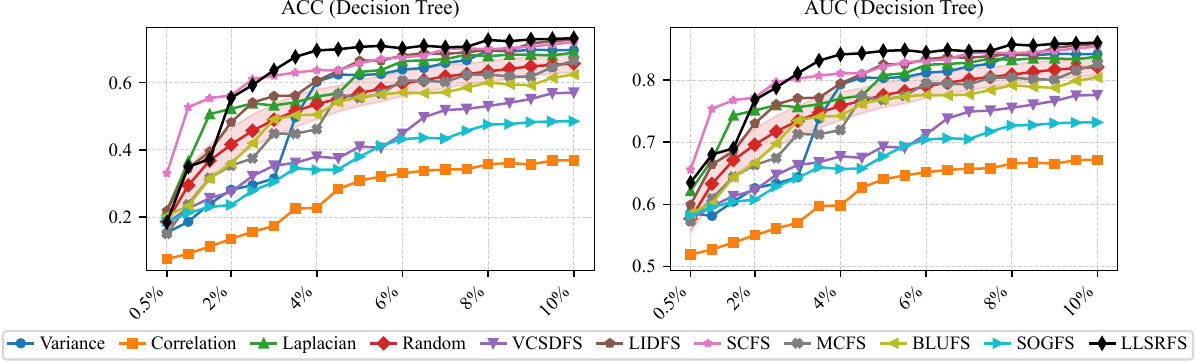}
\caption{ACC and AUC of the decision tree classifier on the Isolet dataset (0.5\% to 10\% of the features).}\label{fig:supplot-dt-10-Isolet}
\supback
\end{figure}
\begin{figure}[p]
\centering
\includegraphics[width=\textwidth,height=\supfigheight,keepaspectratio]{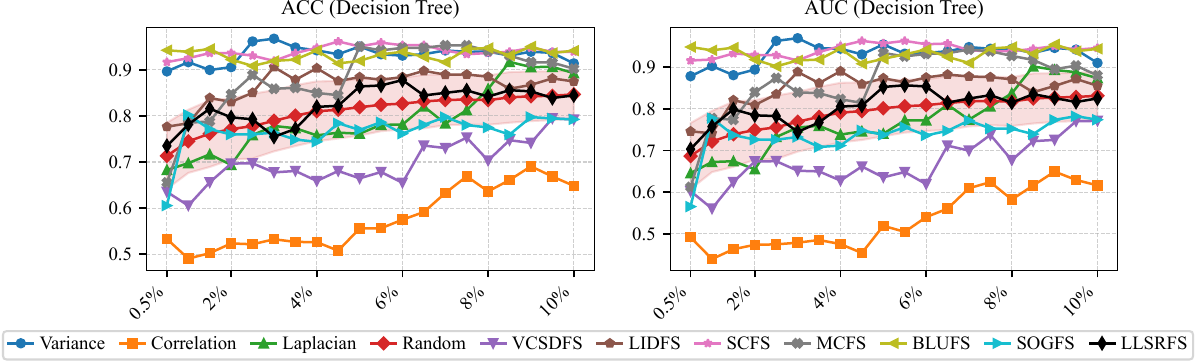}
\caption{ACC and AUC of the decision tree classifier on the leukemia dataset (0.5\% to 10\% of the features).}\label{fig:supplot-dt-10-leukemia}
\supback
\end{figure}
\begin{figure}[p]
\centering
\includegraphics[width=\textwidth,height=\supfigheight,keepaspectratio]{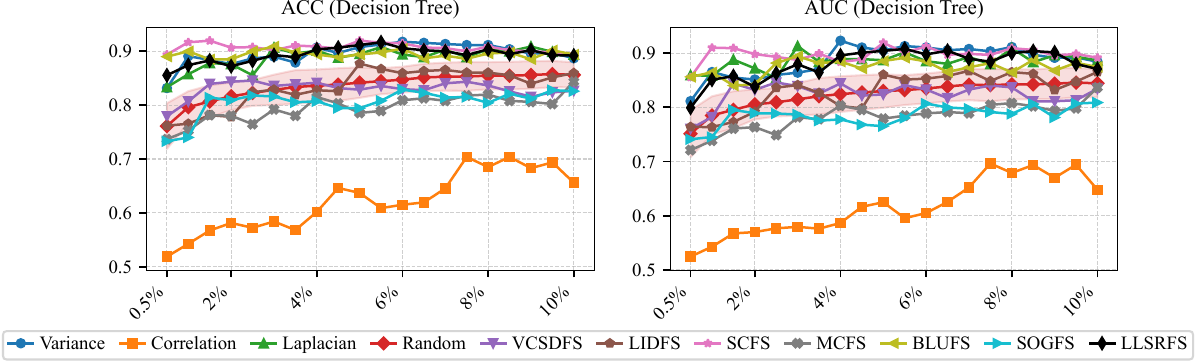}
\caption{ACC and AUC of the decision tree classifier on the lung dataset (0.5\% to 10\% of the features).}\label{fig:supplot-dt-10-lung}
\supback
\end{figure}
\begin{figure}[p]
\centering
\includegraphics[width=\textwidth,height=\supfigheight,keepaspectratio]{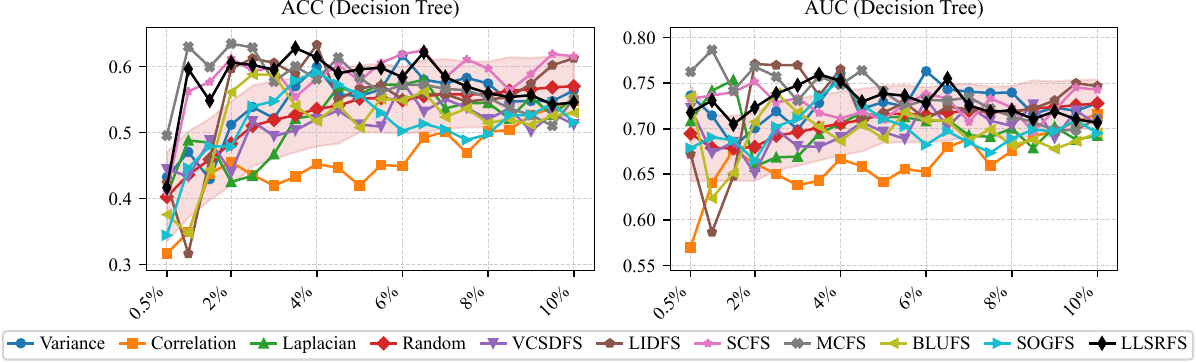}
\caption{ACC and AUC of the decision tree classifier on the lung\_discrete dataset (0.5\% to 10\% of the features).}\label{fig:supplot-dt-10-lung_discrete}
\supback
\end{figure}
\begin{figure}[p]
\centering
\includegraphics[width=\textwidth,height=\supfigheight,keepaspectratio]{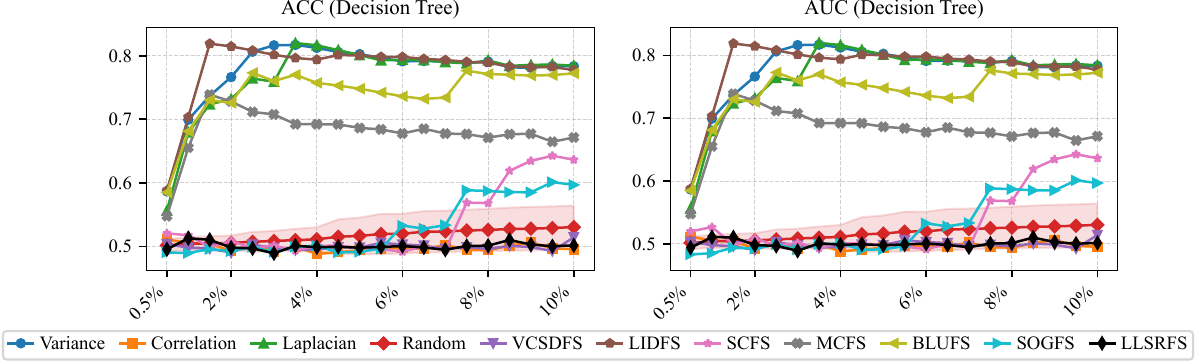}
\caption{ACC and AUC of the decision tree classifier on the madelon dataset (0.5\% to 10\% of the features).}\label{fig:supplot-dt-10-madelon}
\supback
\end{figure}
\begin{figure}[p]
\centering
\includegraphics[width=\textwidth,height=\supfigheight,keepaspectratio]{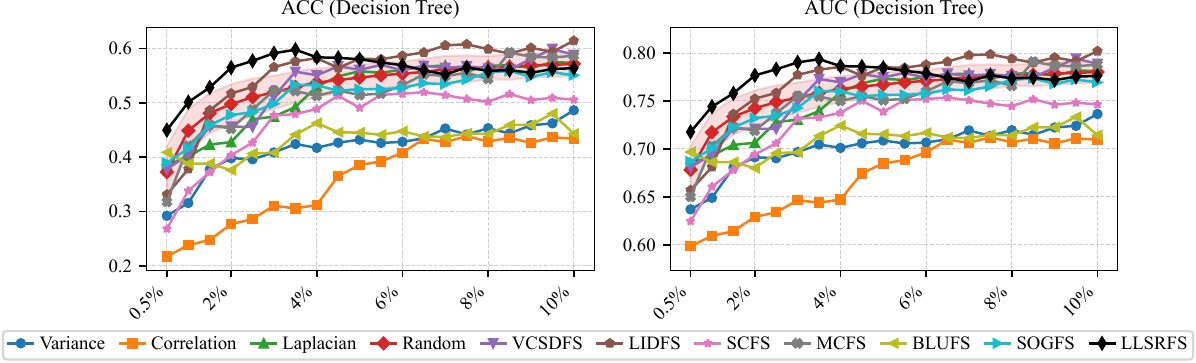}
\caption{ACC and AUC of the decision tree classifier on the ORL dataset (0.5\% to 10\% of the features).}\label{fig:supplot-dt-10-ORL}
\supback
\end{figure}
\begin{figure}[p]
\centering
\includegraphics[width=\textwidth,height=\supfigheight,keepaspectratio]{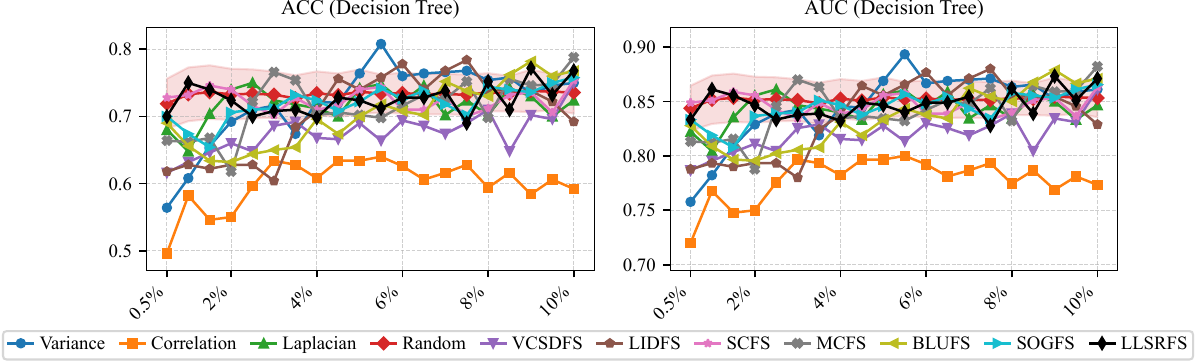}
\caption{ACC and AUC of the decision tree classifier on the orlraws10P dataset (0.5\% to 10\% of the features).}\label{fig:supplot-dt-10-orlraws10P}
\supback
\end{figure}
\begin{figure}[p]
\centering
\includegraphics[width=\textwidth,height=\supfigheight,keepaspectratio]{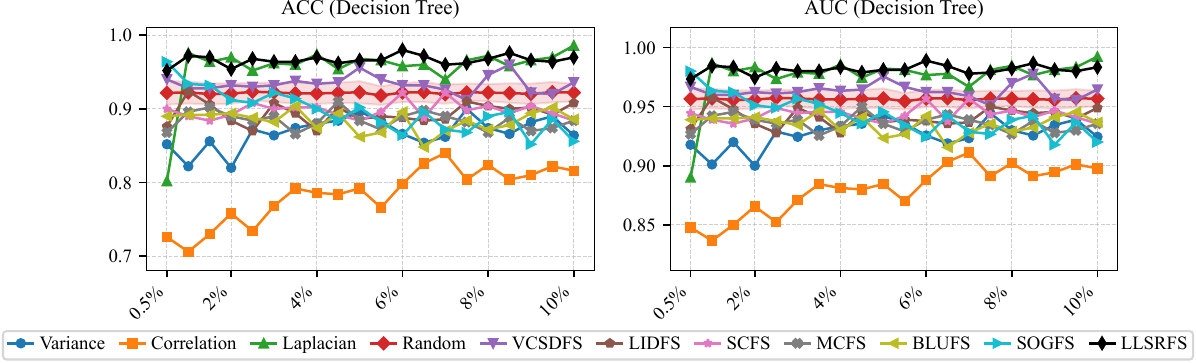}
\caption{ACC and AUC of the decision tree classifier on the pixraw10P dataset (0.5\% to 10\% of the features).}\label{fig:supplot-dt-10-pixraw10P}
\supback
\end{figure}
\begin{figure}[p]
\centering
\includegraphics[width=\textwidth,height=\supfigheight,keepaspectratio]{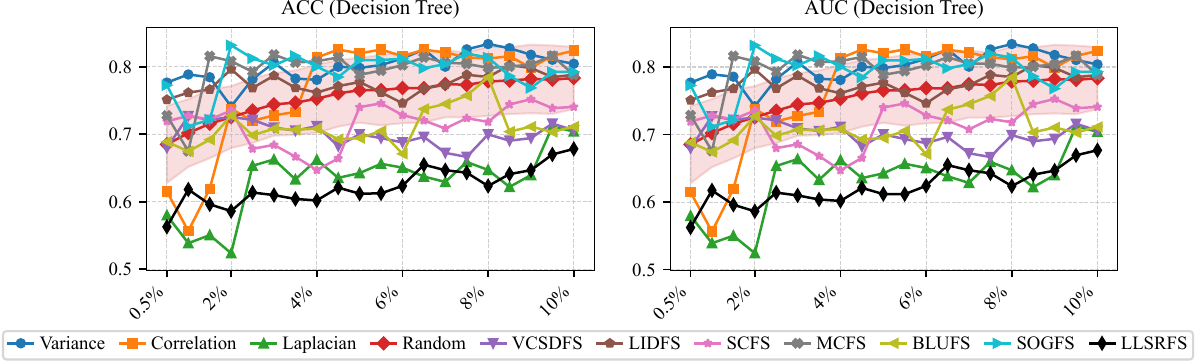}
\caption{ACC and AUC of the decision tree classifier on the Prostate\_GE dataset (0.5\% to 10\% of the features).}\label{fig:supplot-dt-10-Prostate_GE}
\supback
\end{figure}
\begin{figure}[p]
\centering
\includegraphics[width=\textwidth,height=\supfigheight,keepaspectratio]{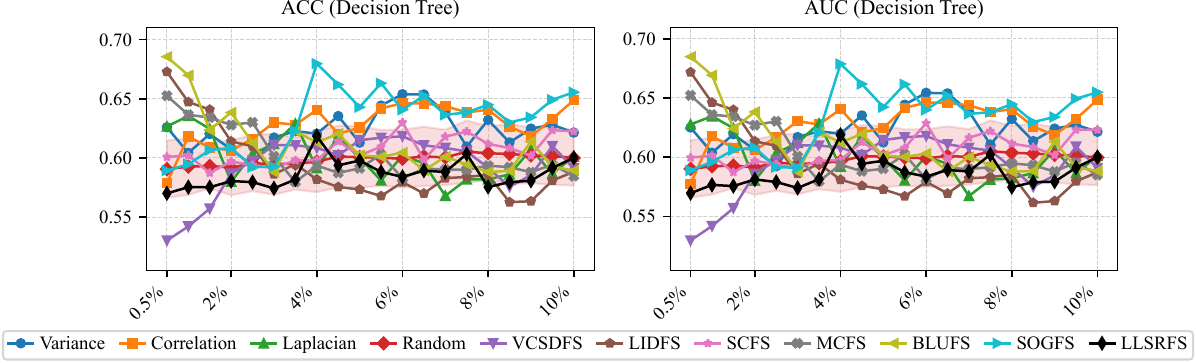}
\caption{ACC and AUC of the decision tree classifier on the SMK\_CAN\_187 dataset (0.5\% to 10\% of the features).}\label{fig:supplot-dt-10-SMK_CAN_187}
\supback
\end{figure}
\begin{figure}[p]
\centering
\includegraphics[width=\textwidth,height=\supfigheight,keepaspectratio]{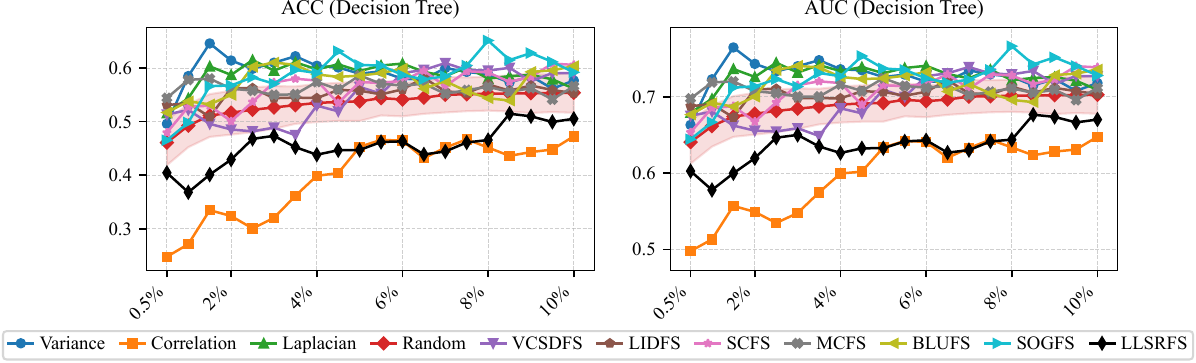}
\caption{ACC and AUC of the decision tree classifier on the TOX\_171 dataset (0.5\% to 10\% of the features).}\label{fig:supplot-dt-10-TOX_171}
\supback
\end{figure}
\begin{figure}[p]
\centering
\includegraphics[width=\textwidth,height=\supfigheight,keepaspectratio]{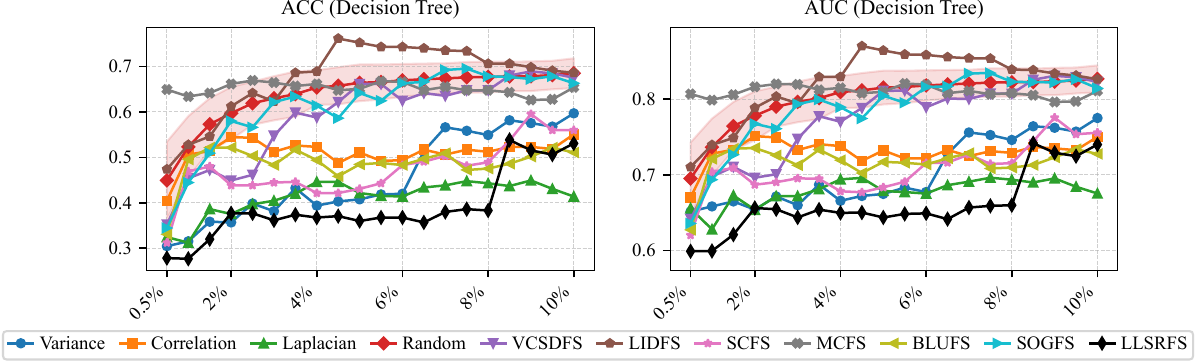}
\caption{ACC and AUC of the decision tree classifier on the warpAR10P dataset (0.5\% to 10\% of the features).}\label{fig:supplot-dt-10-warpAR10P}
\supback
\end{figure}
\begin{figure}[p]
\centering
\includegraphics[width=\textwidth,height=\supfigheight,keepaspectratio]{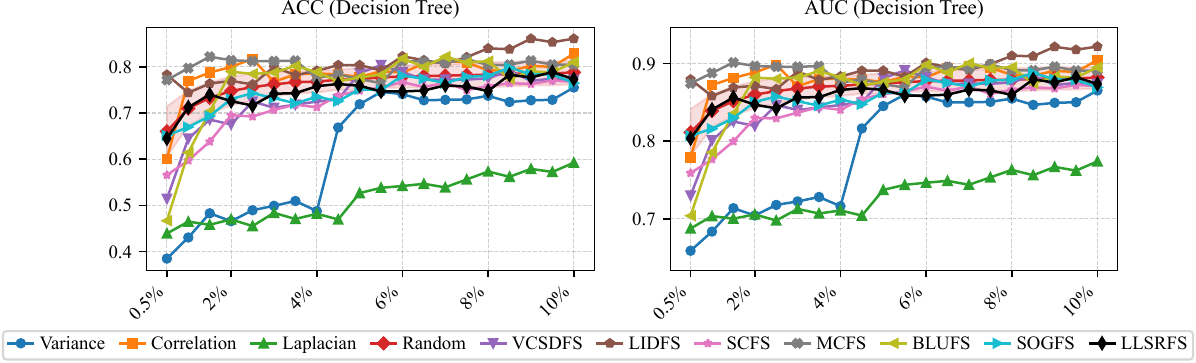}
\caption{ACC and AUC of the decision tree classifier on the warpPIE10P dataset (0.5\% to 10\% of the features).}\label{fig:supplot-dt-10-warpPIE10P}
\supback
\end{figure}
\begin{figure}[p]
\centering
\includegraphics[width=\textwidth,height=\supfigheight,keepaspectratio]{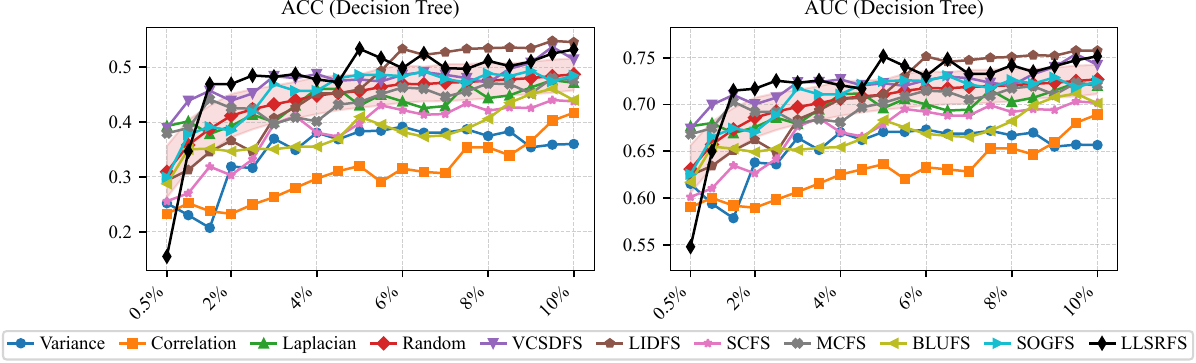}
\caption{ACC and AUC of the decision tree classifier on the Yale dataset (0.5\% to 10\% of the features).}\label{fig:supplot-dt-10-Yale}
\supback
\end{figure}
\begin{figure}[p]
\centering
\pdfbookmark[4]{5\% to 100\%}{supplot-bm-dt-100}
\includegraphics[width=\textwidth,height=\supfigheight,keepaspectratio]{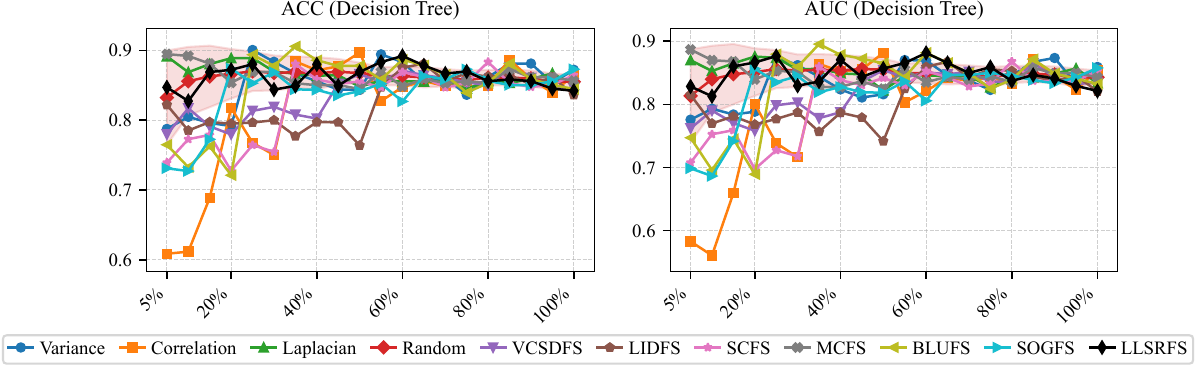}
\caption{ACC and AUC of the decision tree classifier on the ALLAML dataset (5\% to 100\% of the features).}\label{fig:supplot-dt-100-ALLAML}
\supback
\end{figure}
\begin{figure}[p]
\centering
\includegraphics[width=\textwidth,height=\supfigheight,keepaspectratio]{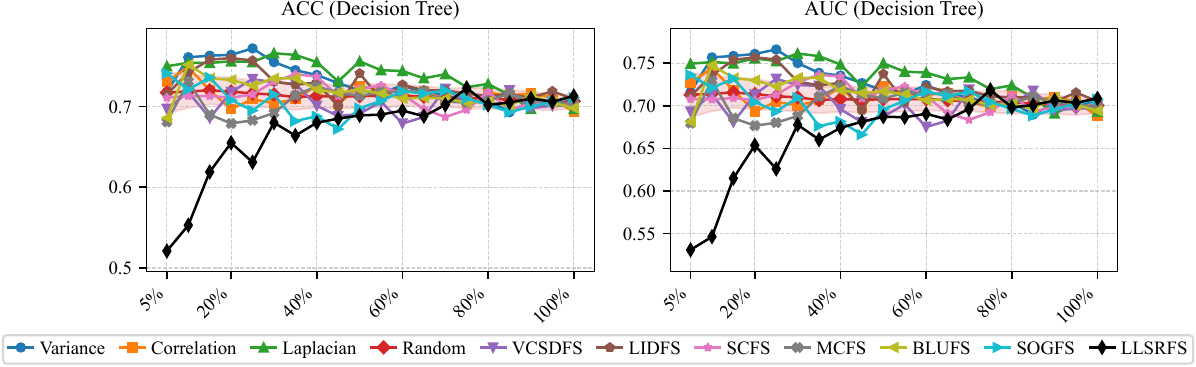}
\caption{ACC and AUC of the decision tree classifier on the arcene dataset (5\% to 100\% of the features).}\label{fig:supplot-dt-100-arcene}
\supback
\end{figure}
\begin{figure}[p]
\centering
\includegraphics[width=\textwidth,height=\supfigheight,keepaspectratio]{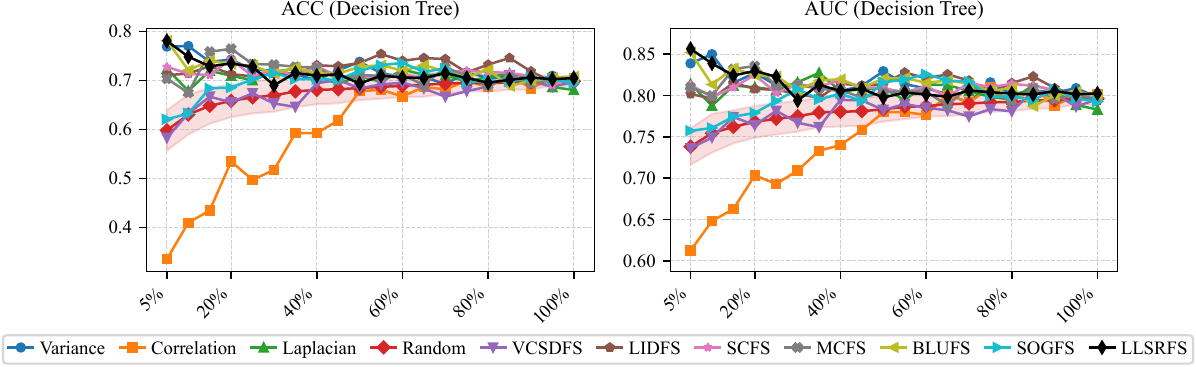}
\caption{ACC and AUC of the decision tree classifier on the Carcinom dataset (5\% to 100\% of the features).}\label{fig:supplot-dt-100-Carcinom}
\supback
\end{figure}
\begin{figure}[p]
\centering
\includegraphics[width=\textwidth,height=\supfigheight,keepaspectratio]{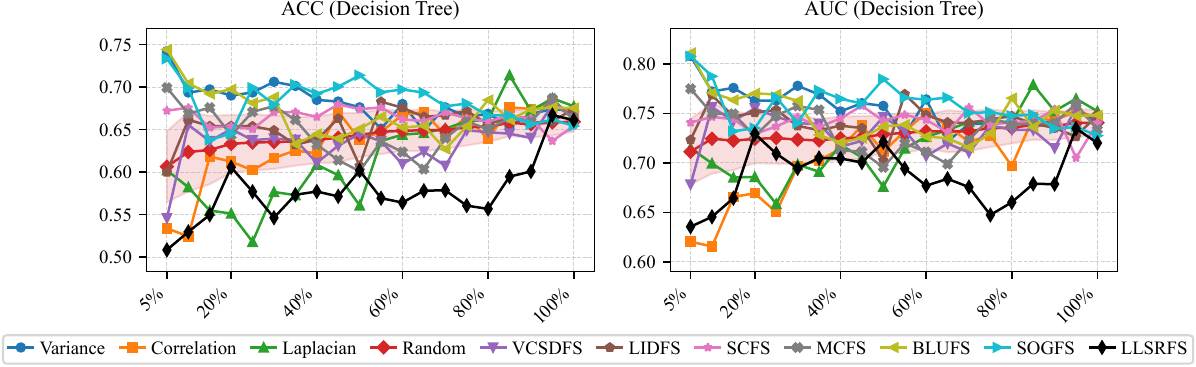}
\caption{ACC and AUC of the decision tree classifier on the CLL\_SUB\_111 dataset (5\% to 100\% of the features).}\label{fig:supplot-dt-100-CLL_SUB_111}
\supback
\end{figure}
\begin{figure}[p]
\centering
\includegraphics[width=\textwidth,height=\supfigheight,keepaspectratio]{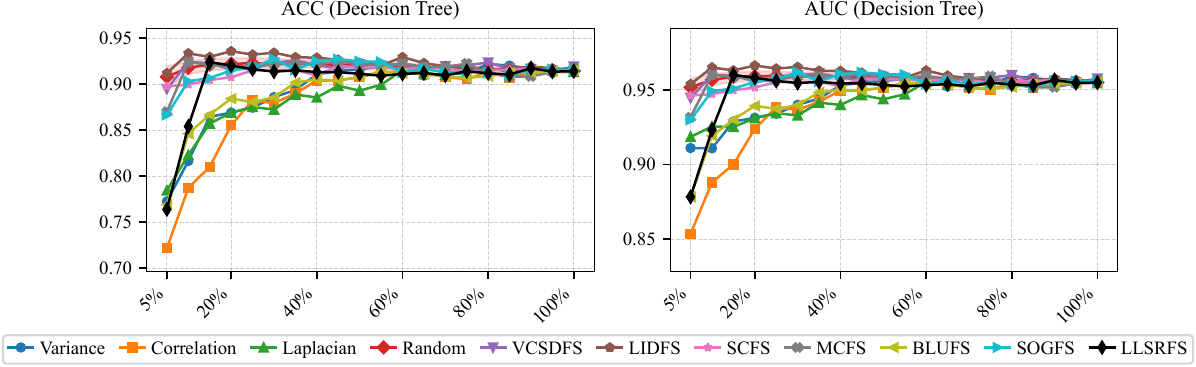}
\caption{ACC and AUC of the decision tree classifier on the COIL20 dataset (5\% to 100\% of the features).}\label{fig:supplot-dt-100-COIL20}
\supback
\end{figure}
\begin{figure}[p]
\centering
\includegraphics[width=\textwidth,height=\supfigheight,keepaspectratio]{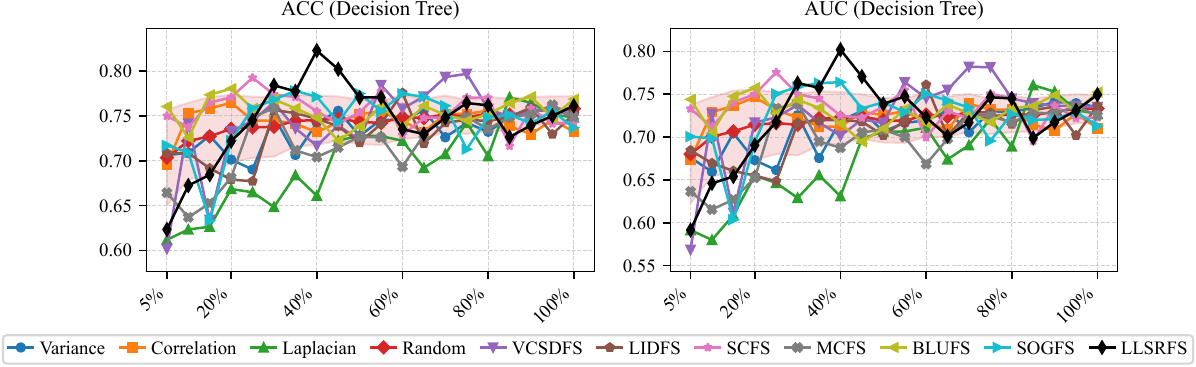}
\caption{ACC and AUC of the decision tree classifier on the colon dataset (5\% to 100\% of the features).}\label{fig:supplot-dt-100-colon}
\supback
\end{figure}
\begin{figure}[p]
\centering
\includegraphics[width=\textwidth,height=\supfigheight,keepaspectratio]{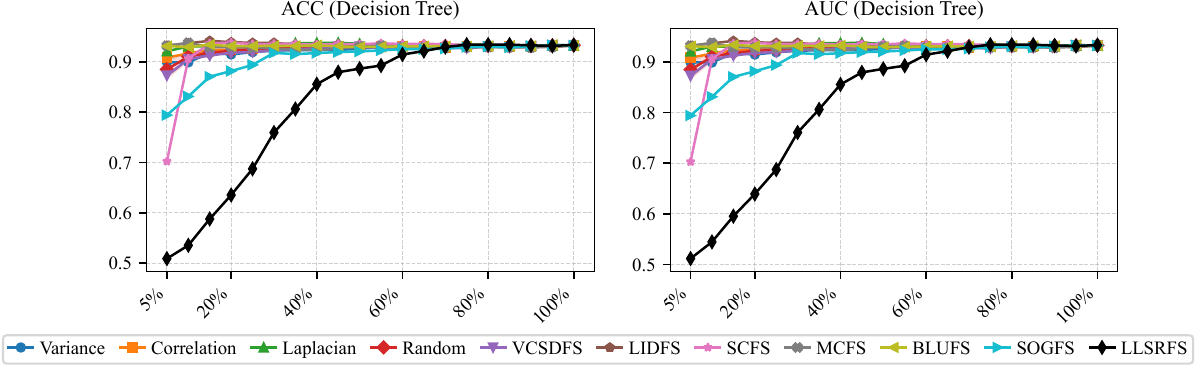}
\caption{ACC and AUC of the decision tree classifier on the gisette dataset (5\% to 100\% of the features).}\label{fig:supplot-dt-100-gisette}
\supback
\end{figure}
\begin{figure}[p]
\centering
\includegraphics[width=\textwidth,height=\supfigheight,keepaspectratio]{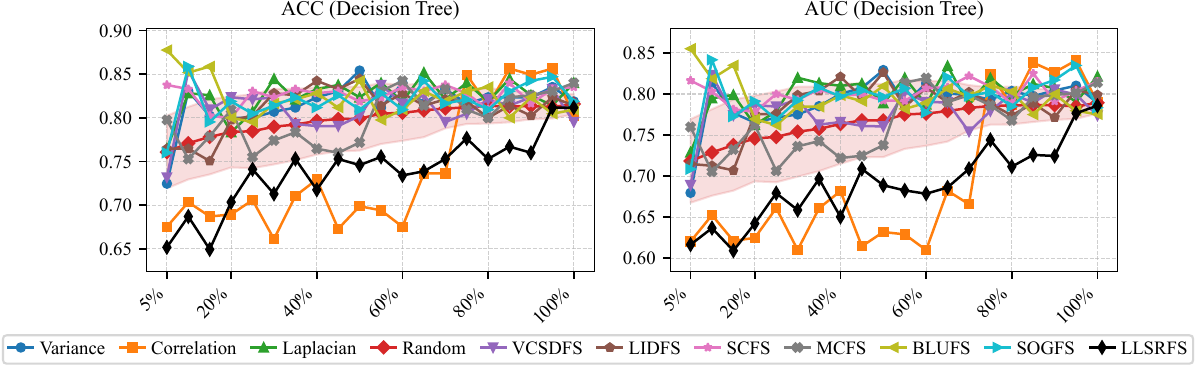}
\caption{ACC and AUC of the decision tree classifier on the GLI\_85 dataset (5\% to 100\% of the features).}\label{fig:supplot-dt-100-GLI_85}
\supback
\end{figure}
\begin{figure}[p]
\centering
\includegraphics[width=\textwidth,height=\supfigheight,keepaspectratio]{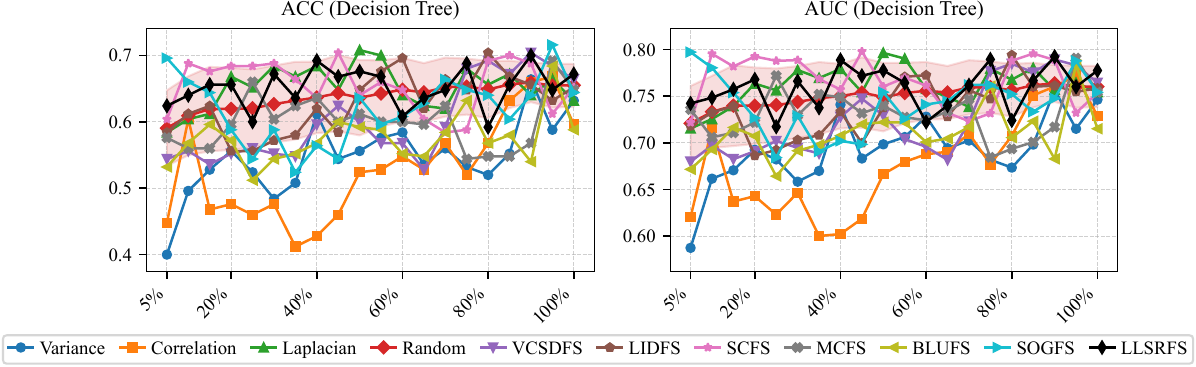}
\caption{ACC and AUC of the decision tree classifier on the GLIOMA dataset (5\% to 100\% of the features).}\label{fig:supplot-dt-100-GLIOMA}
\supback
\end{figure}
\begin{figure}[p]
\centering
\includegraphics[width=\textwidth,height=\supfigheight,keepaspectratio]{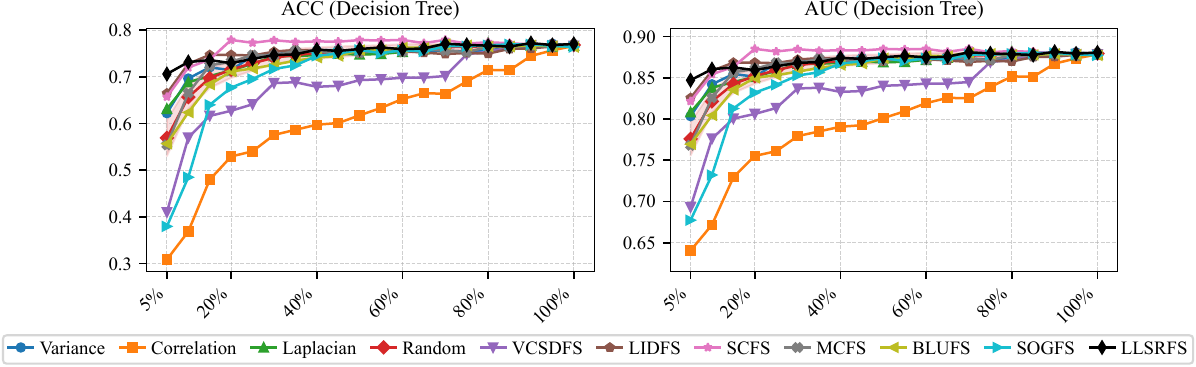}
\caption{ACC and AUC of the decision tree classifier on the Isolet dataset (5\% to 100\% of the features).}\label{fig:supplot-dt-100-Isolet}
\supback
\end{figure}
\begin{figure}[p]
\centering
\includegraphics[width=\textwidth,height=\supfigheight,keepaspectratio]{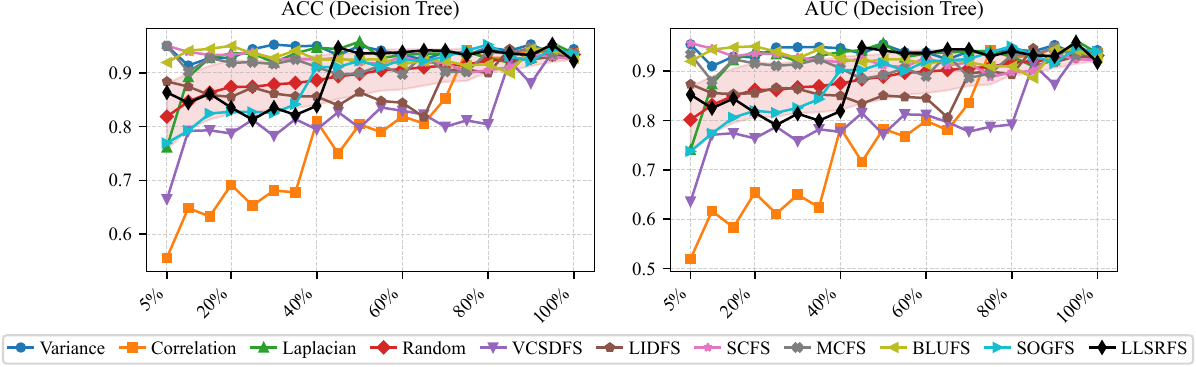}
\caption{ACC and AUC of the decision tree classifier on the leukemia dataset (5\% to 100\% of the features).}\label{fig:supplot-dt-100-leukemia}
\supback
\end{figure}
\begin{figure}[p]
\centering
\includegraphics[width=\textwidth,height=\supfigheight,keepaspectratio]{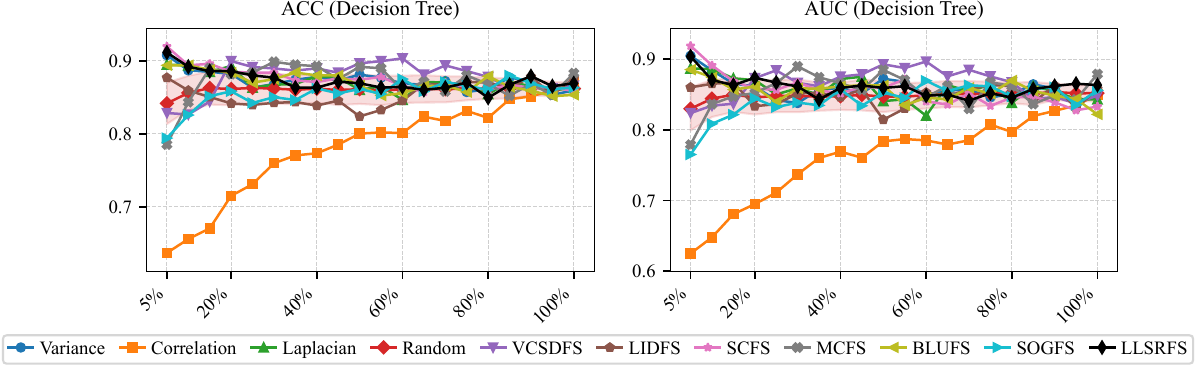}
\caption{ACC and AUC of the decision tree classifier on the lung dataset (5\% to 100\% of the features).}\label{fig:supplot-dt-100-lung}
\supback
\end{figure}
\begin{figure}[p]
\centering
\includegraphics[width=\textwidth,height=\supfigheight,keepaspectratio]{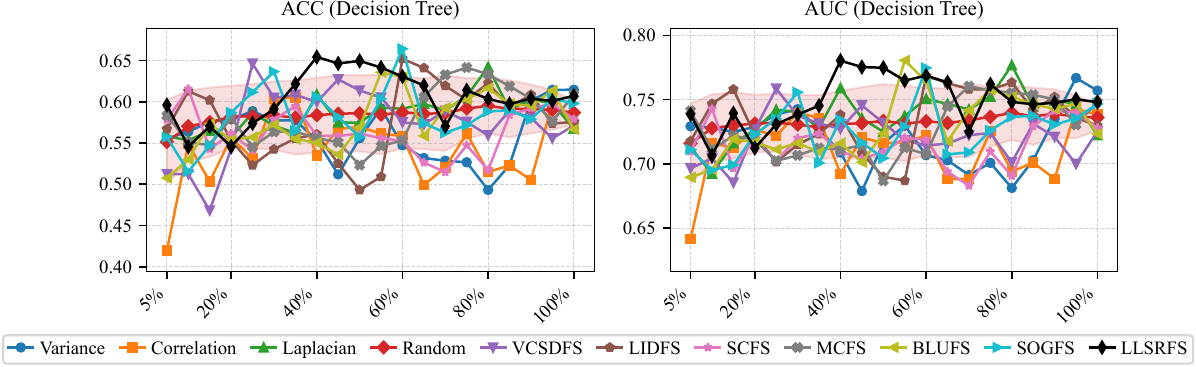}
\caption{ACC and AUC of the decision tree classifier on the lung\_discrete dataset (5\% to 100\% of the features).}\label{fig:supplot-dt-100-lung_discrete}
\supback
\end{figure}
\begin{figure}[p]
\centering
\includegraphics[width=\textwidth,height=\supfigheight,keepaspectratio]{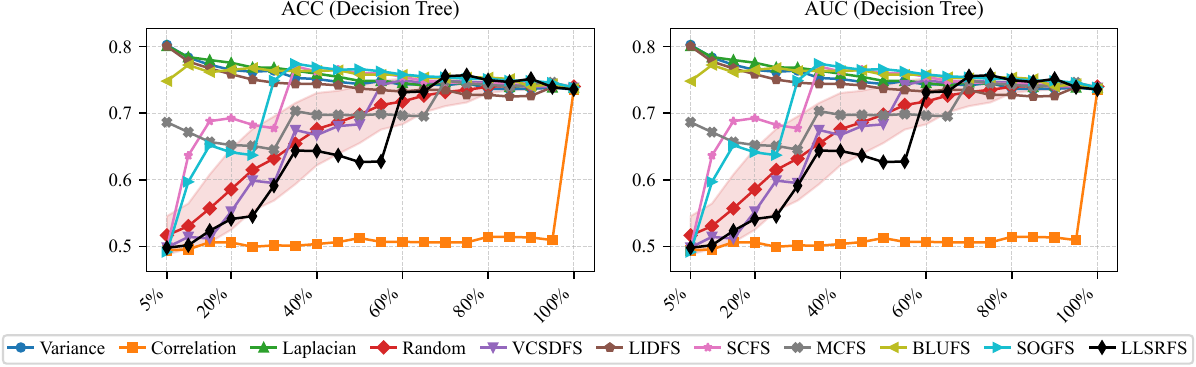}
\caption{ACC and AUC of the decision tree classifier on the madelon dataset (5\% to 100\% of the features).}\label{fig:supplot-dt-100-madelon}
\supback
\end{figure}
\begin{figure}[p]
\centering
\includegraphics[width=\textwidth,height=\supfigheight,keepaspectratio]{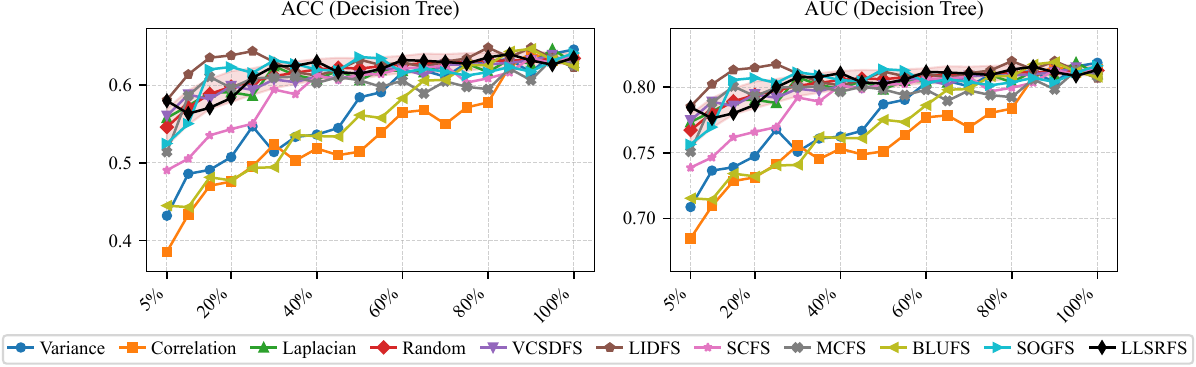}
\caption{ACC and AUC of the decision tree classifier on the ORL dataset (5\% to 100\% of the features).}\label{fig:supplot-dt-100-ORL}
\supback
\end{figure}
\begin{figure}[p]
\centering
\includegraphics[width=\textwidth,height=\supfigheight,keepaspectratio]{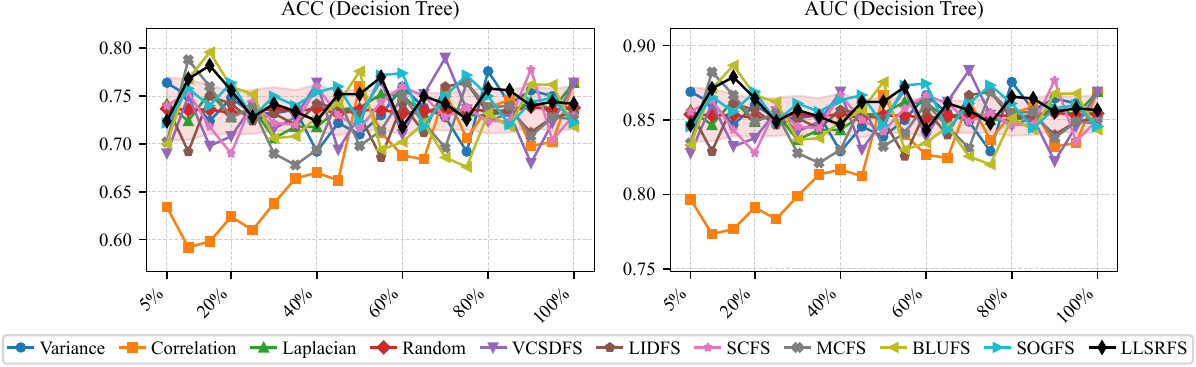}
\caption{ACC and AUC of the decision tree classifier on the orlraws10P dataset (5\% to 100\% of the features).}\label{fig:supplot-dt-100-orlraws10P}
\supback
\end{figure}
\begin{figure}[p]
\centering
\includegraphics[width=\textwidth,height=\supfigheight,keepaspectratio]{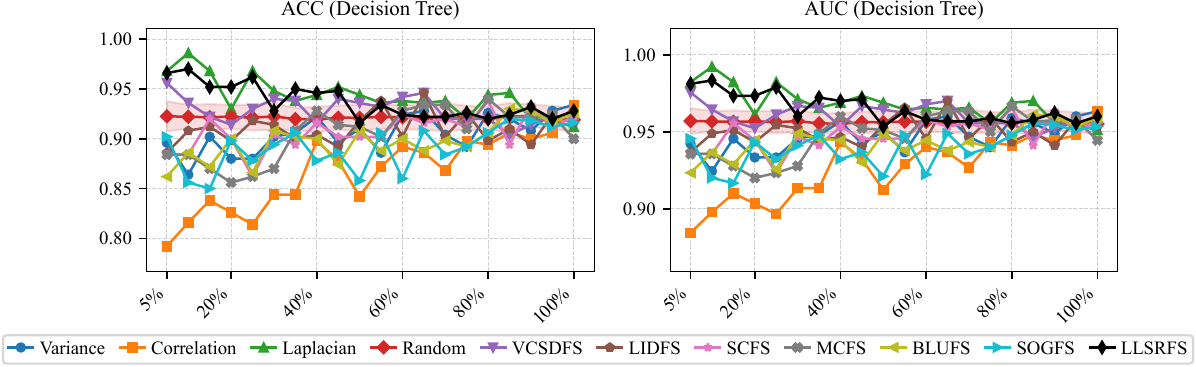}
\caption{ACC and AUC of the decision tree classifier on the pixraw10P dataset (5\% to 100\% of the features).}\label{fig:supplot-dt-100-pixraw10P}
\supback
\end{figure}
\begin{figure}[p]
\centering
\includegraphics[width=\textwidth,height=\supfigheight,keepaspectratio]{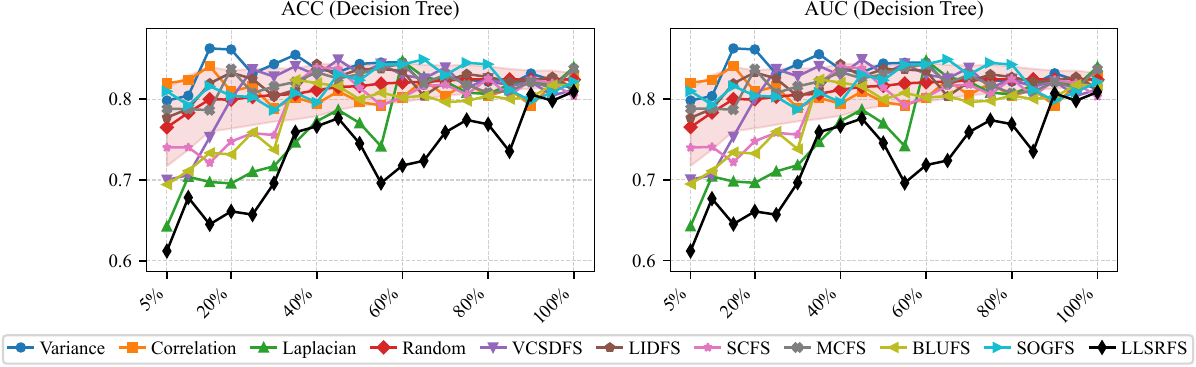}
\caption{ACC and AUC of the decision tree classifier on the Prostate\_GE dataset (5\% to 100\% of the features).}\label{fig:supplot-dt-100-Prostate_GE}
\supback
\end{figure}
\begin{figure}[p]
\centering
\includegraphics[width=\textwidth,height=\supfigheight,keepaspectratio]{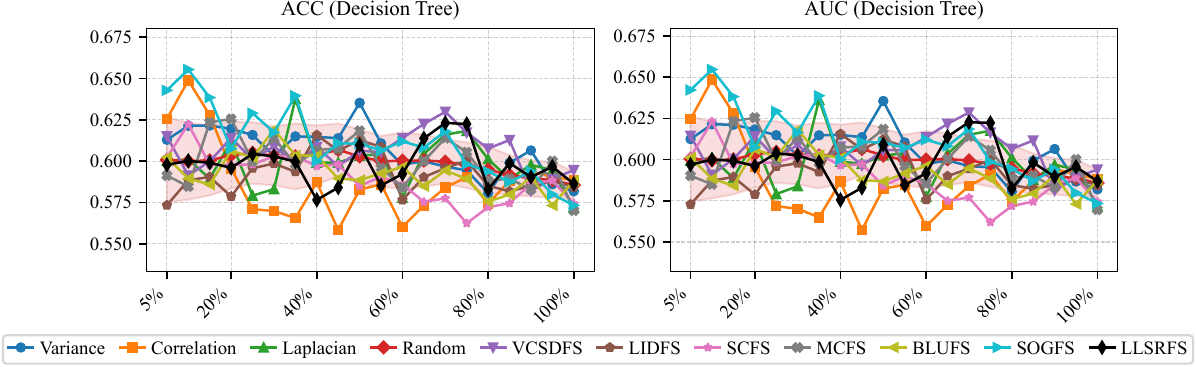}
\caption{ACC and AUC of the decision tree classifier on the SMK\_CAN\_187 dataset (5\% to 100\% of the features).}\label{fig:supplot-dt-100-SMK_CAN_187}
\supback
\end{figure}
\begin{figure}[p]
\centering
\includegraphics[width=\textwidth,height=\supfigheight,keepaspectratio]{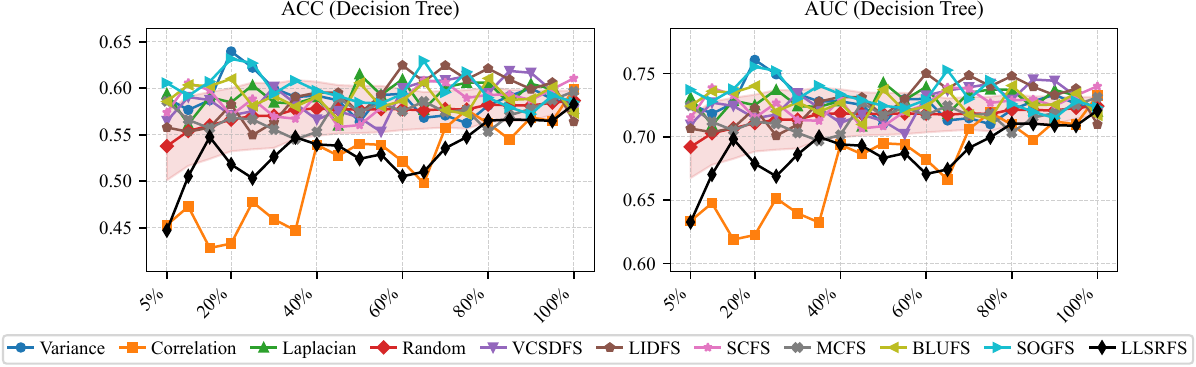}
\caption{ACC and AUC of the decision tree classifier on the TOX\_171 dataset (5\% to 100\% of the features).}\label{fig:supplot-dt-100-TOX_171}
\supback
\end{figure}
\begin{figure}[p]
\centering
\includegraphics[width=\textwidth,height=\supfigheight,keepaspectratio]{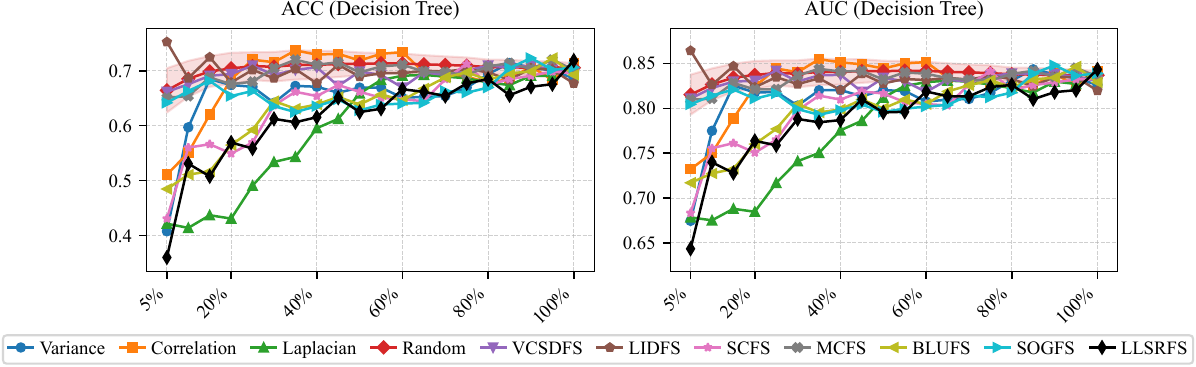}
\caption{ACC and AUC of the decision tree classifier on the warpAR10P dataset (5\% to 100\% of the features).}\label{fig:supplot-dt-100-warpAR10P}
\supback
\end{figure}
\begin{figure}[p]
\centering
\includegraphics[width=\textwidth,height=\supfigheight,keepaspectratio]{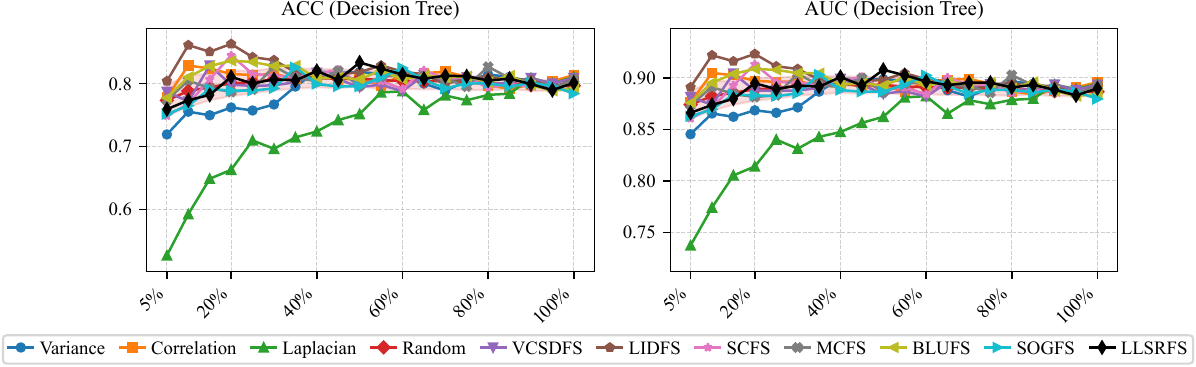}
\caption{ACC and AUC of the decision tree classifier on the warpPIE10P dataset (5\% to 100\% of the features).}\label{fig:supplot-dt-100-warpPIE10P}
\supback
\end{figure}
\begin{figure}[p]
\centering
\includegraphics[width=\textwidth,height=\supfigheight,keepaspectratio]{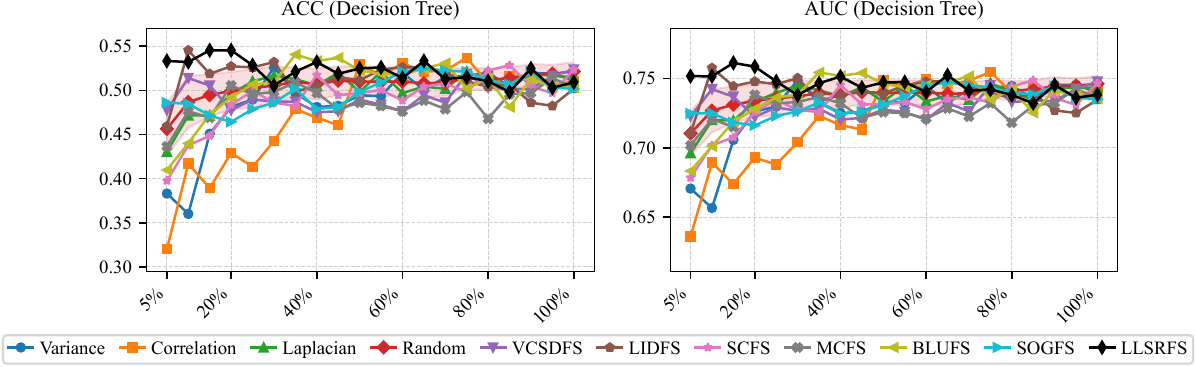}
\caption{ACC and AUC of the decision tree classifier on the Yale dataset (5\% to 100\% of the features).}\label{fig:supplot-dt-100-Yale}
\supback
\end{figure}
\clearpage

\begin{figure}[p]
\centering
\pdfbookmark[3]{SVM}{supplot-bm-svm}
\pdfbookmark[4]{0.5\% to 10\%}{supplot-bm-svm-10}
\includegraphics[width=\textwidth,height=\supfigheight,keepaspectratio]{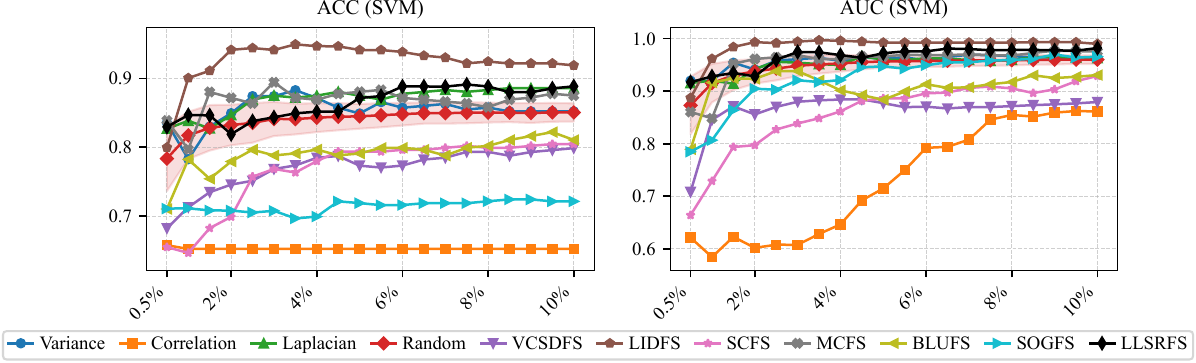}
\caption{ACC and AUC of the SVM classifier on the ALLAML dataset (0.5\% to 10\% of the features).}\label{fig:supplot-svm-10-ALLAML}
\supback
\end{figure}
\begin{figure}[p]
\centering
\includegraphics[width=\textwidth,height=\supfigheight,keepaspectratio]{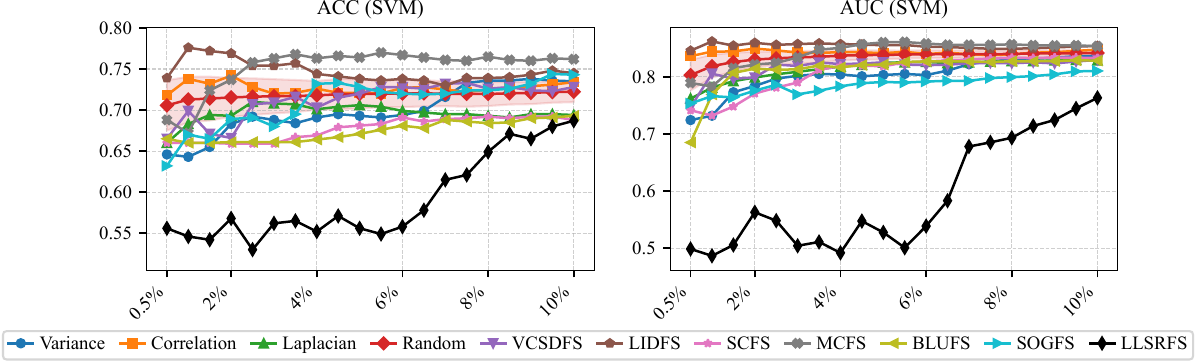}
\caption{ACC and AUC of the SVM classifier on the arcene dataset (0.5\% to 10\% of the features).}\label{fig:supplot-svm-10-arcene}
\supback
\end{figure}
\begin{figure}[p]
\centering
\includegraphics[width=\textwidth,height=\supfigheight,keepaspectratio]{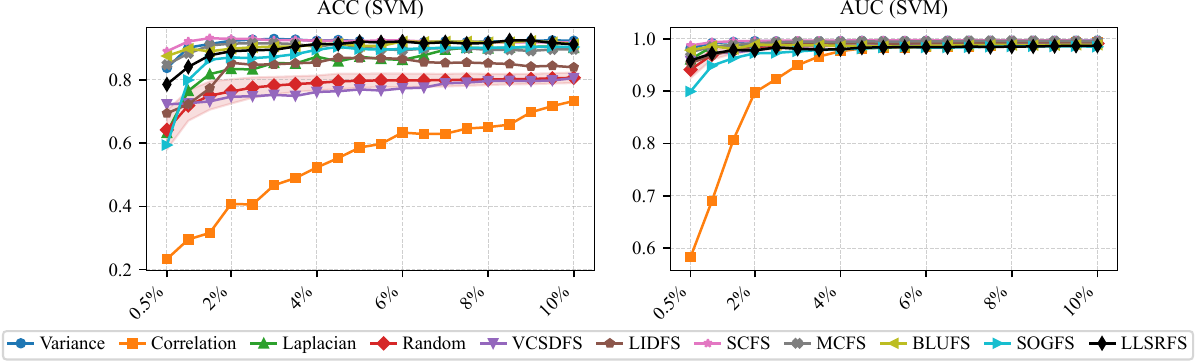}
\caption{ACC and AUC of the SVM classifier on the Carcinom dataset (0.5\% to 10\% of the features).}\label{fig:supplot-svm-10-Carcinom}
\supback
\end{figure}
\begin{figure}[p]
\centering
\includegraphics[width=\textwidth,height=\supfigheight,keepaspectratio]{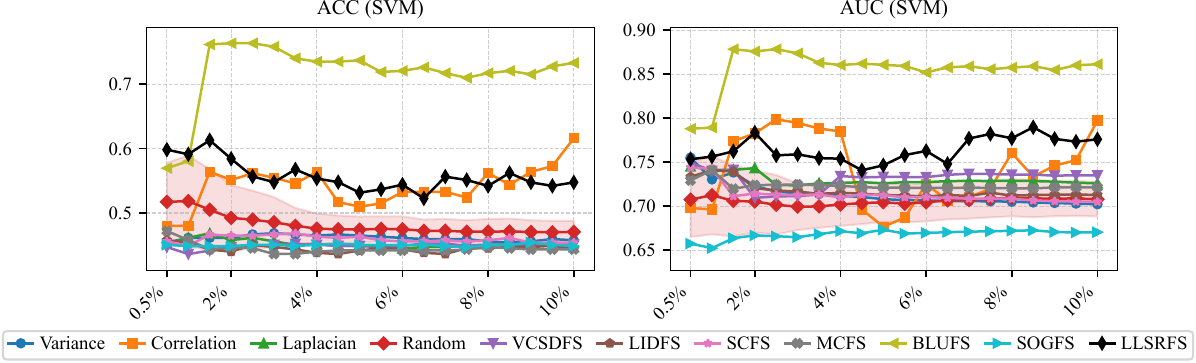}
\caption{ACC and AUC of the SVM classifier on the CLL\_SUB\_111 dataset (0.5\% to 10\% of the features).}\label{fig:supplot-svm-10-CLL_SUB_111}
\supback
\end{figure}
\begin{figure}[p]
\centering
\includegraphics[width=\textwidth,height=\supfigheight,keepaspectratio]{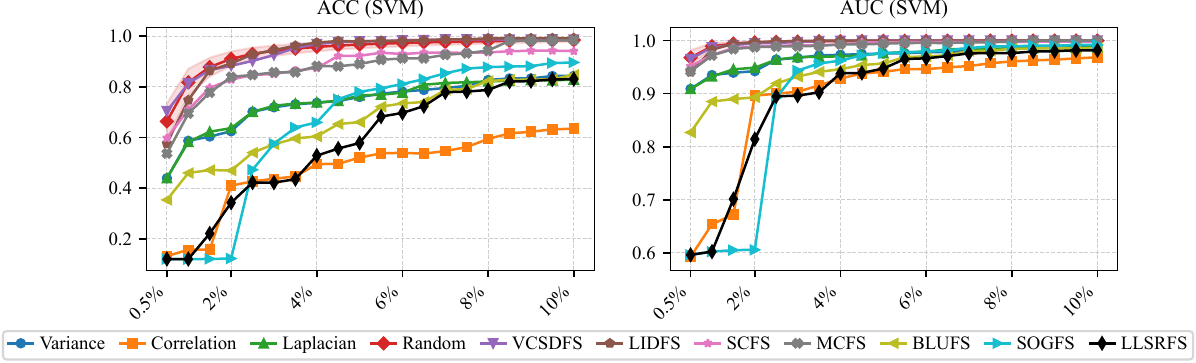}
\caption{ACC and AUC of the SVM classifier on the COIL20 dataset (0.5\% to 10\% of the features).}\label{fig:supplot-svm-10-COIL20}
\supback
\end{figure}
\begin{figure}[p]
\centering
\includegraphics[width=\textwidth,height=\supfigheight,keepaspectratio]{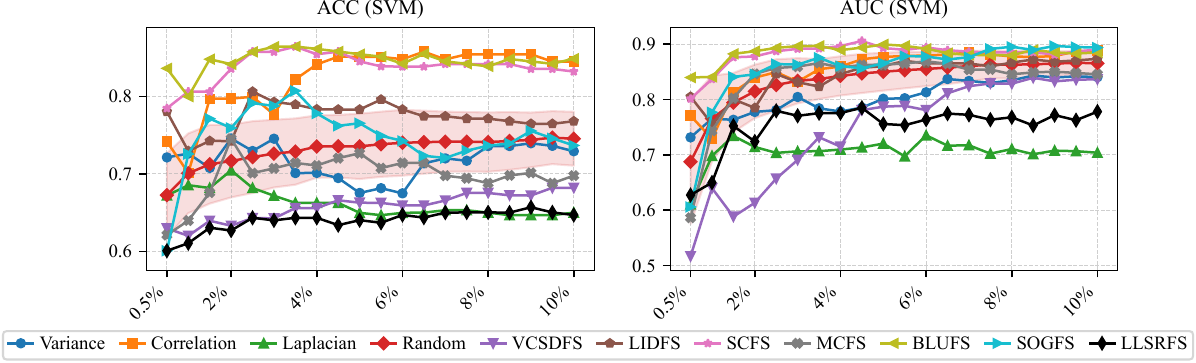}
\caption{ACC and AUC of the SVM classifier on the colon dataset (0.5\% to 10\% of the features).}\label{fig:supplot-svm-10-colon}
\supback
\end{figure}
\begin{figure}[p]
\centering
\includegraphics[width=\textwidth,height=\supfigheight,keepaspectratio]{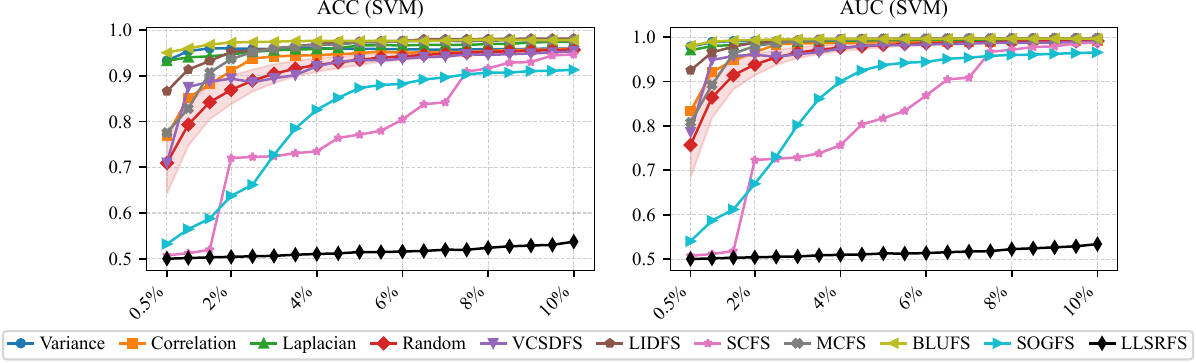}
\caption{ACC and AUC of the SVM classifier on the gisette dataset (0.5\% to 10\% of the features).}\label{fig:supplot-svm-10-gisette}
\supback
\end{figure}
\begin{figure}[p]
\centering
\includegraphics[width=\textwidth,height=\supfigheight,keepaspectratio]{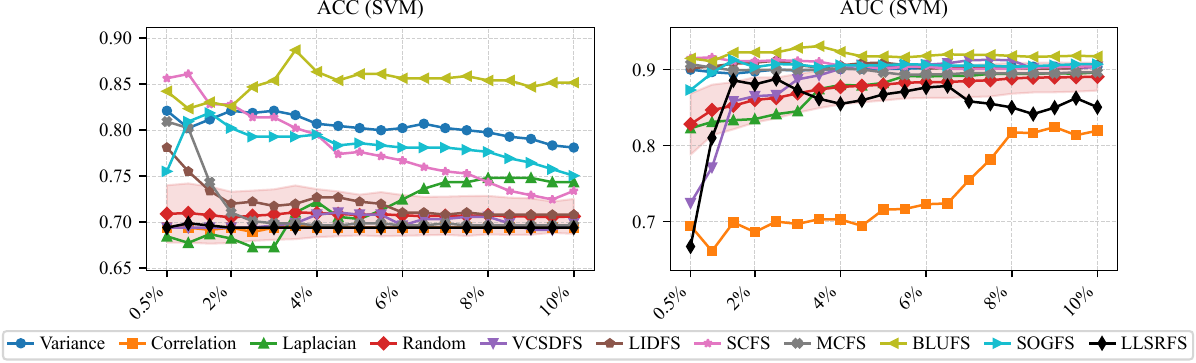}
\caption{ACC and AUC of the SVM classifier on the GLI\_85 dataset (0.5\% to 10\% of the features).}\label{fig:supplot-svm-10-GLI_85}
\supback
\end{figure}
\begin{figure}[p]
\centering
\includegraphics[width=\textwidth,height=\supfigheight,keepaspectratio]{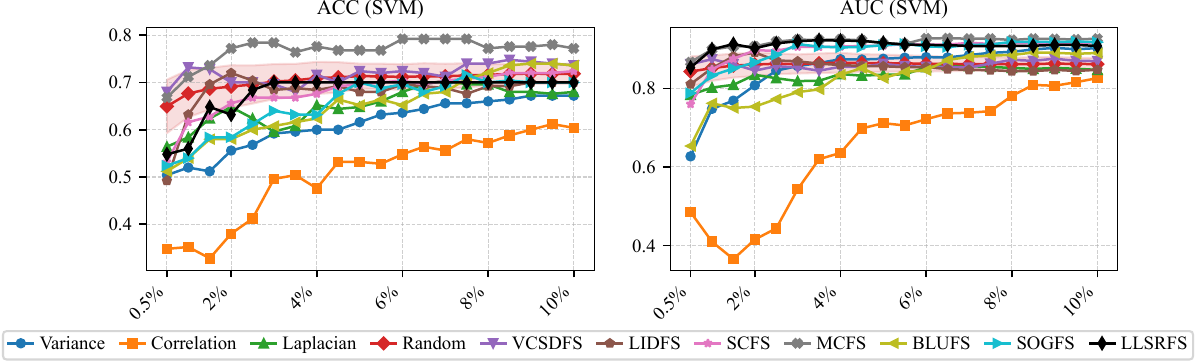}
\caption{ACC and AUC of the SVM classifier on the GLIOMA dataset (0.5\% to 10\% of the features).}\label{fig:supplot-svm-10-GLIOMA}
\supback
\end{figure}
\begin{figure}[p]
\centering
\includegraphics[width=\textwidth,height=\supfigheight,keepaspectratio]{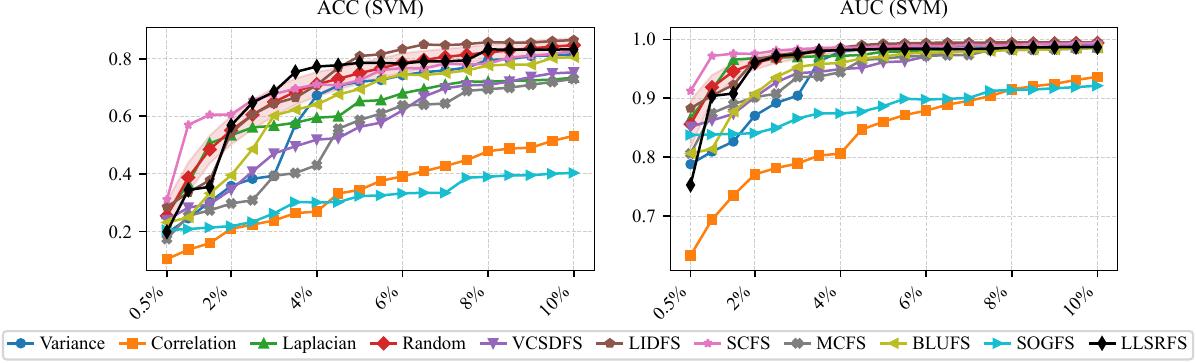}
\caption{ACC and AUC of the SVM classifier on the Isolet dataset (0.5\% to 10\% of the features).}\label{fig:supplot-svm-10-Isolet}
\supback
\end{figure}
\begin{figure}[p]
\centering
\includegraphics[width=\textwidth,height=\supfigheight,keepaspectratio]{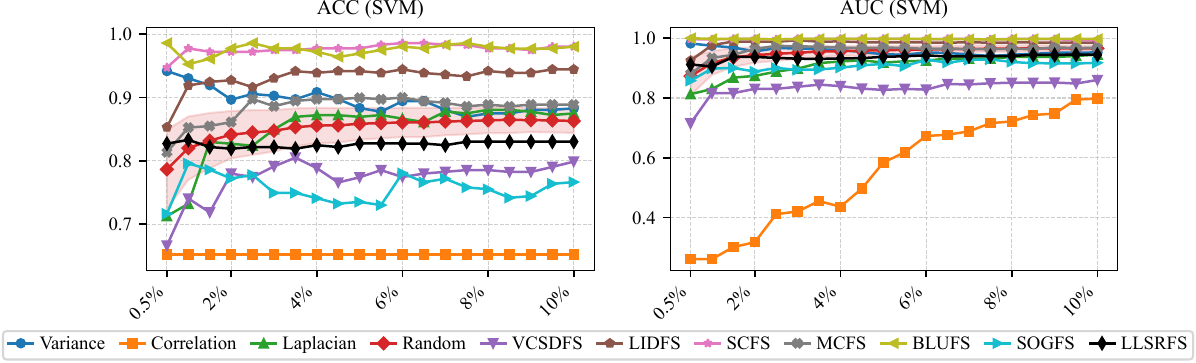}
\caption{ACC and AUC of the SVM classifier on the leukemia dataset (0.5\% to 10\% of the features).}\label{fig:supplot-svm-10-leukemia}
\supback
\end{figure}
\begin{figure}[p]
\centering
\includegraphics[width=\textwidth,height=\supfigheight,keepaspectratio]{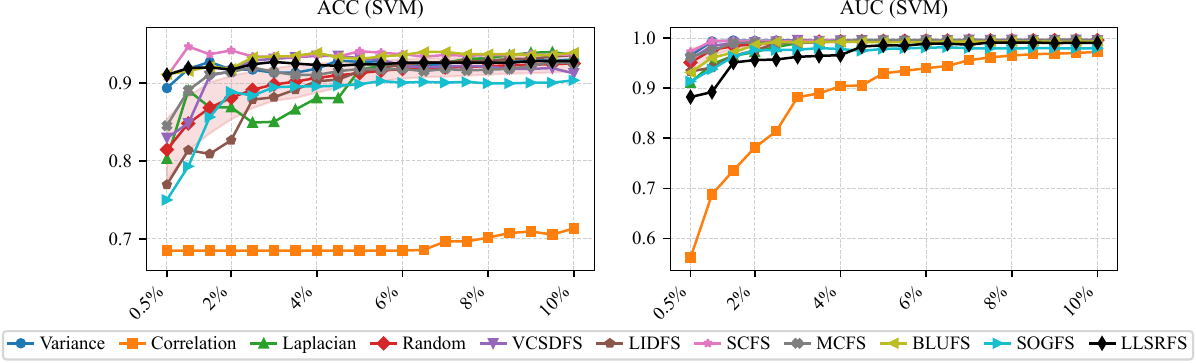}
\caption{ACC and AUC of the SVM classifier on the lung dataset (0.5\% to 10\% of the features).}\label{fig:supplot-svm-10-lung}
\supback
\end{figure}
\begin{figure}[p]
\centering
\includegraphics[width=\textwidth,height=\supfigheight,keepaspectratio]{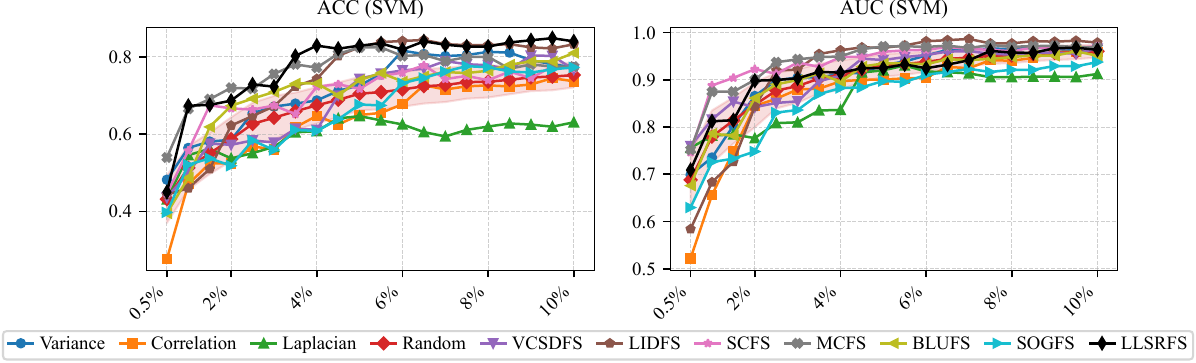}
\caption{ACC and AUC of the SVM classifier on the lung\_discrete dataset (0.5\% to 10\% of the features).}\label{fig:supplot-svm-10-lung_discrete}
\supback
\end{figure}
\begin{figure}[p]
\centering
\includegraphics[width=\textwidth,height=\supfigheight,keepaspectratio]{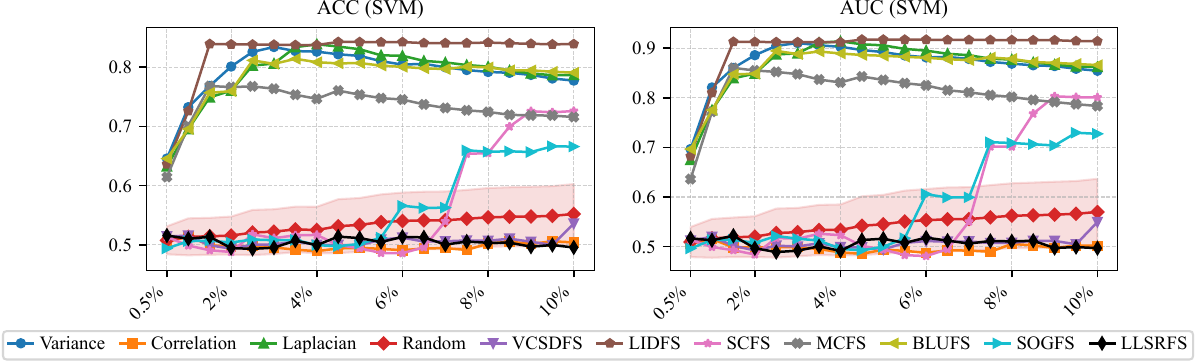}
\caption{ACC and AUC of the SVM classifier on the madelon dataset (0.5\% to 10\% of the features).}\label{fig:supplot-svm-10-madelon}
\supback
\end{figure}
\begin{figure}[p]
\centering
\includegraphics[width=\textwidth,height=\supfigheight,keepaspectratio]{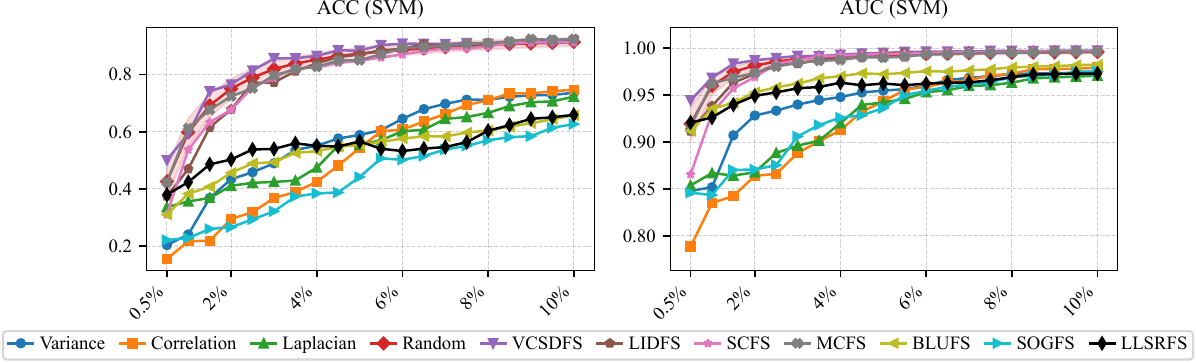}
\caption{ACC and AUC of the SVM classifier on the ORL dataset (0.5\% to 10\% of the features).}\label{fig:supplot-svm-10-ORL}
\supback
\end{figure}
\begin{figure}[p]
\centering
\includegraphics[width=\textwidth,height=\supfigheight,keepaspectratio]{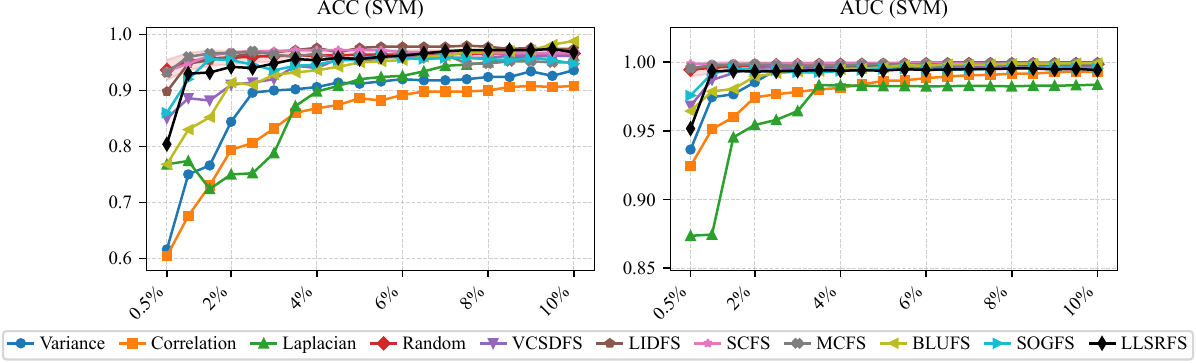}
\caption{ACC and AUC of the SVM classifier on the orlraws10P dataset (0.5\% to 10\% of the features).}\label{fig:supplot-svm-10-orlraws10P}
\supback
\end{figure}
\begin{figure}[p]
\centering
\includegraphics[width=\textwidth,height=\supfigheight,keepaspectratio]{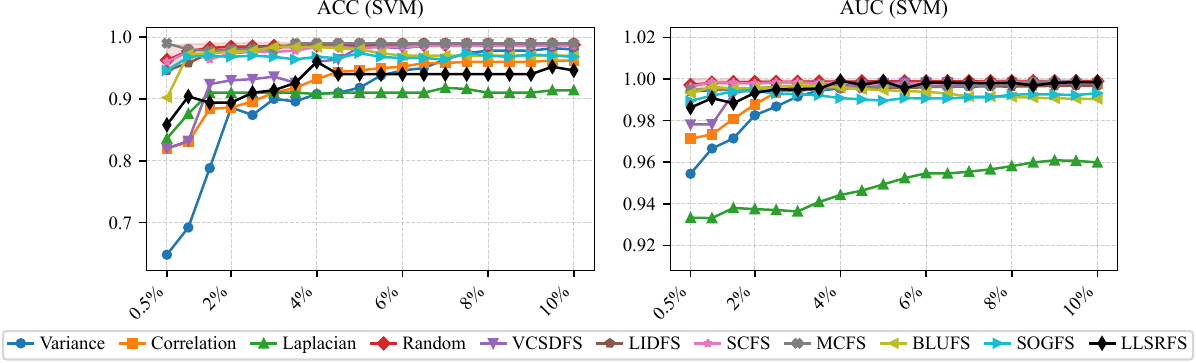}
\caption{ACC and AUC of the SVM classifier on the pixraw10P dataset (0.5\% to 10\% of the features).}\label{fig:supplot-svm-10-pixraw10P}
\supback
\end{figure}
\begin{figure}[p]
\centering
\includegraphics[width=\textwidth,height=\supfigheight,keepaspectratio]{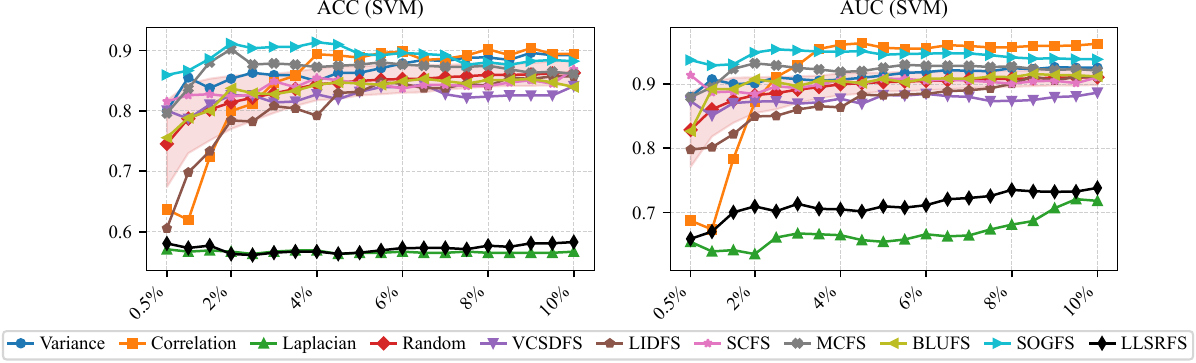}
\caption{ACC and AUC of the SVM classifier on the Prostate\_GE dataset (0.5\% to 10\% of the features).}\label{fig:supplot-svm-10-Prostate_GE}
\supback
\end{figure}
\begin{figure}[p]
\centering
\includegraphics[width=\textwidth,height=\supfigheight,keepaspectratio]{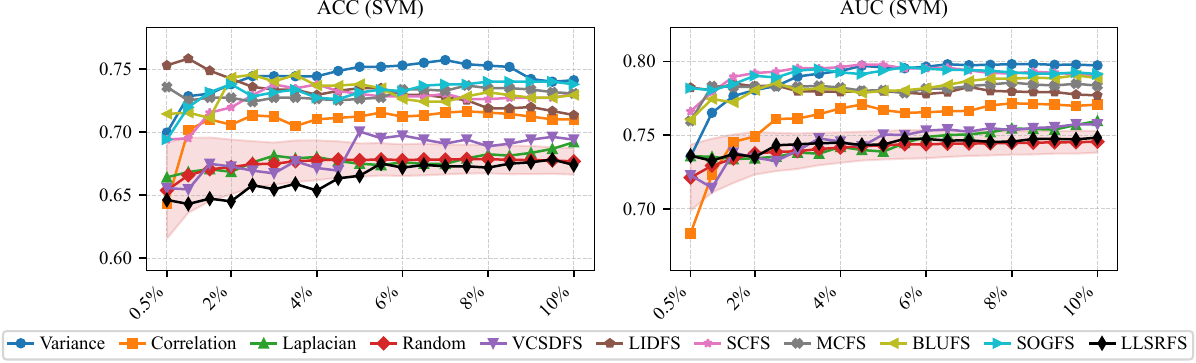}
\caption{ACC and AUC of the SVM classifier on the SMK\_CAN\_187 dataset (0.5\% to 10\% of the features).}\label{fig:supplot-svm-10-SMK_CAN_187}
\supback
\end{figure}
\begin{figure}[p]
\centering
\includegraphics[width=\textwidth,height=\supfigheight,keepaspectratio]{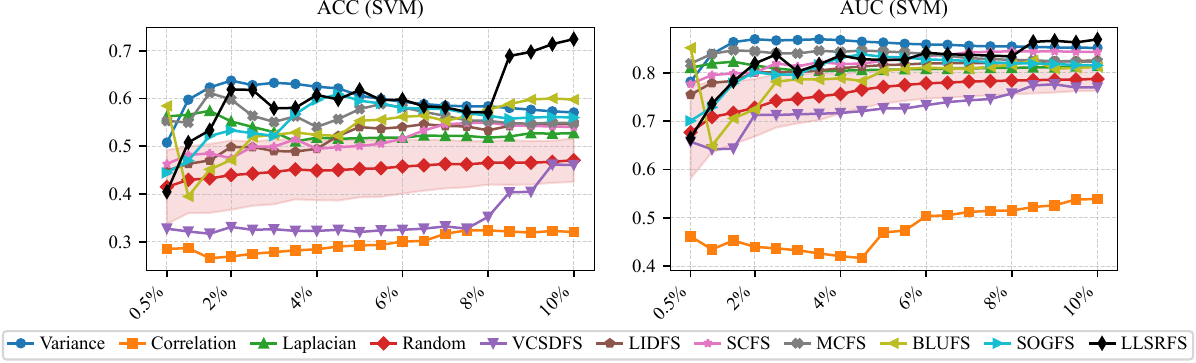}
\caption{ACC and AUC of the SVM classifier on the TOX\_171 dataset (0.5\% to 10\% of the features).}\label{fig:supplot-svm-10-TOX_171}
\supback
\end{figure}
\begin{figure}[p]
\centering
\includegraphics[width=\textwidth,height=\supfigheight,keepaspectratio]{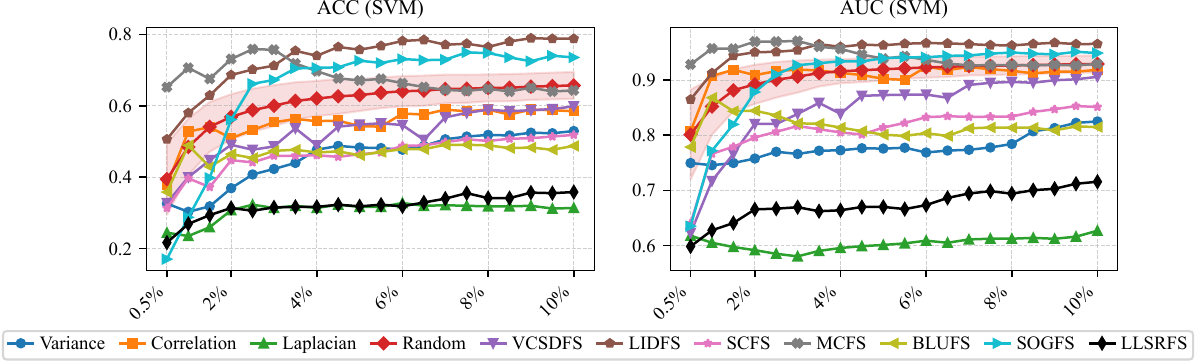}
\caption{ACC and AUC of the SVM classifier on the warpAR10P dataset (0.5\% to 10\% of the features).}\label{fig:supplot-svm-10-warpAR10P}
\supback
\end{figure}
\begin{figure}[p]
\centering
\includegraphics[width=\textwidth,height=\supfigheight,keepaspectratio]{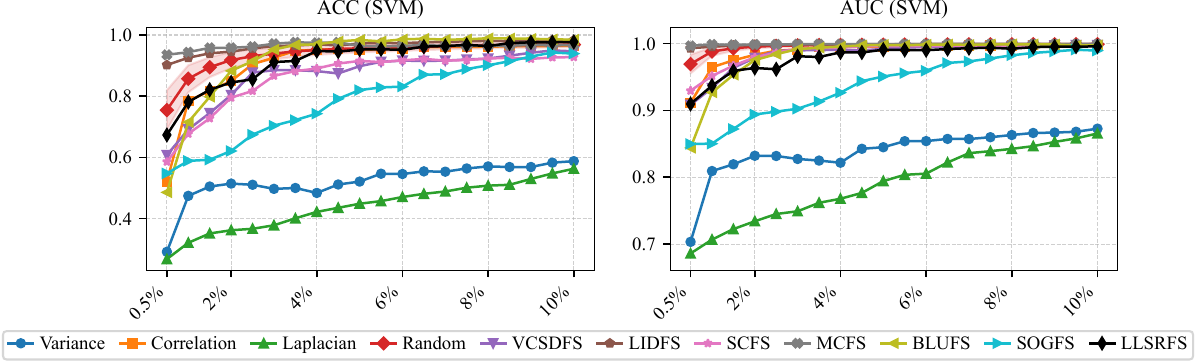}
\caption{ACC and AUC of the SVM classifier on the warpPIE10P dataset (0.5\% to 10\% of the features).}\label{fig:supplot-svm-10-warpPIE10P}
\supback
\end{figure}
\begin{figure}[p]
\centering
\includegraphics[width=\textwidth,height=\supfigheight,keepaspectratio]{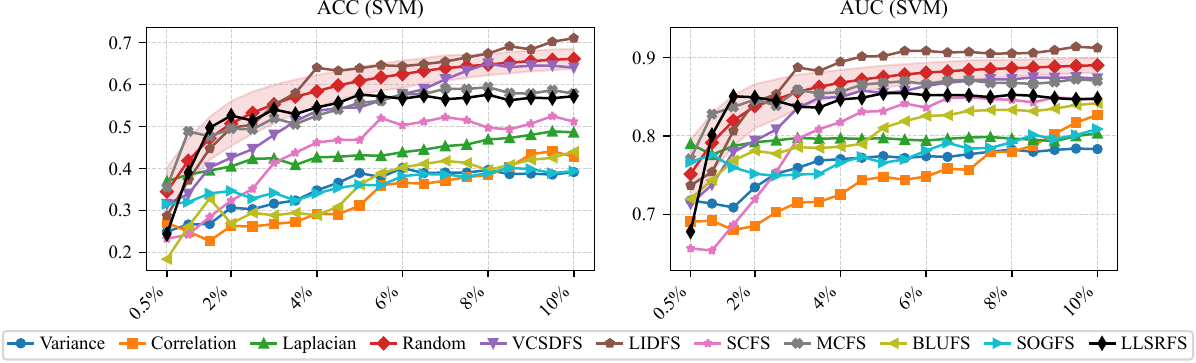}
\caption{ACC and AUC of the SVM classifier on the Yale dataset (0.5\% to 10\% of the features).}\label{fig:supplot-svm-10-Yale}
\supback
\end{figure}
\begin{figure}[p]
\centering
\pdfbookmark[4]{5\% to 100\%}{supplot-bm-svm-100}
\includegraphics[width=\textwidth,height=\supfigheight,keepaspectratio]{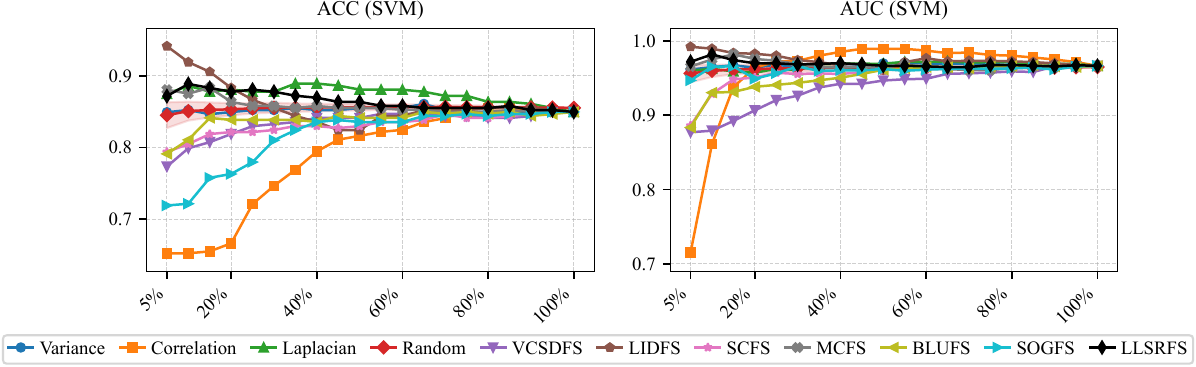}
\caption{ACC and AUC of the SVM classifier on the ALLAML dataset (5\% to 100\% of the features).}\label{fig:supplot-svm-100-ALLAML}
\supback
\end{figure}
\begin{figure}[p]
\centering
\includegraphics[width=\textwidth,height=\supfigheight,keepaspectratio]{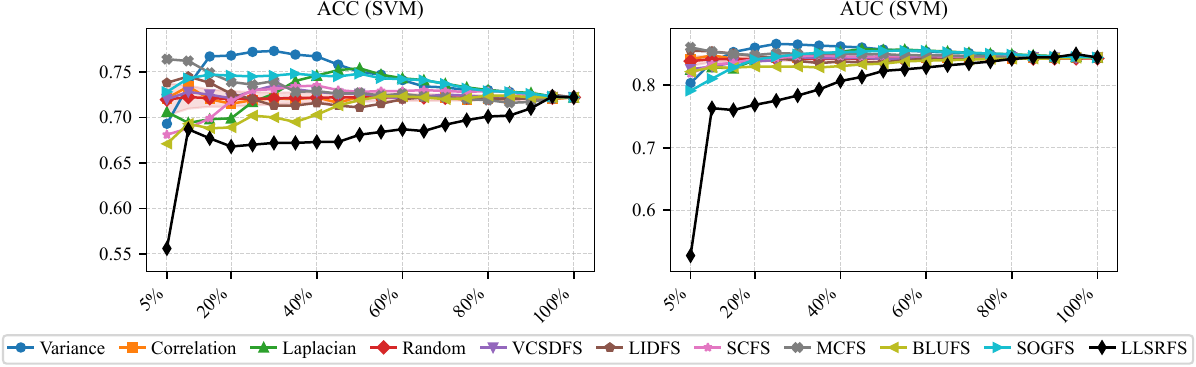}
\caption{ACC and AUC of the SVM classifier on the arcene dataset (5\% to 100\% of the features).}\label{fig:supplot-svm-100-arcene}
\supback
\end{figure}
\begin{figure}[p]
\centering
\includegraphics[width=\textwidth,height=\supfigheight,keepaspectratio]{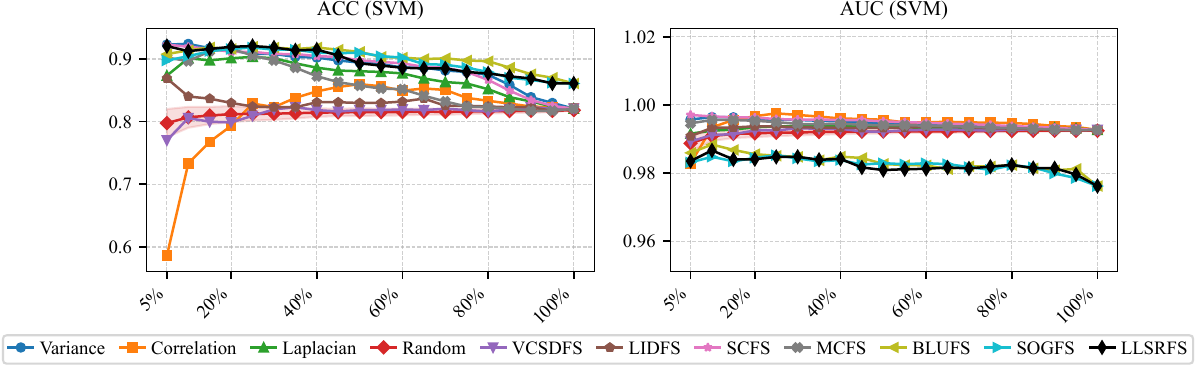}
\caption{ACC and AUC of the SVM classifier on the Carcinom dataset (5\% to 100\% of the features).}\label{fig:supplot-svm-100-Carcinom}
\supback
\end{figure}
\begin{figure}[p]
\centering
\includegraphics[width=\textwidth,height=\supfigheight,keepaspectratio]{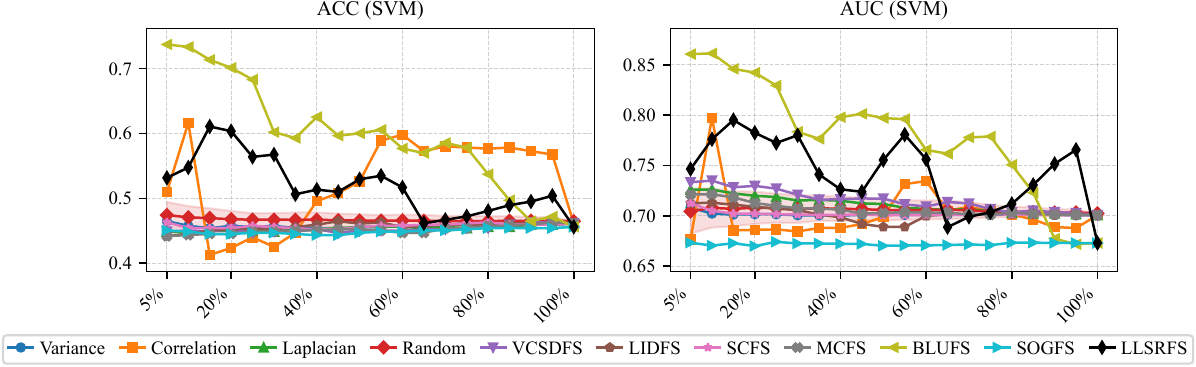}
\caption{ACC and AUC of the SVM classifier on the CLL\_SUB\_111 dataset (5\% to 100\% of the features).}\label{fig:supplot-svm-100-CLL_SUB_111}
\supback
\end{figure}
\begin{figure}[p]
\centering
\includegraphics[width=\textwidth,height=\supfigheight,keepaspectratio]{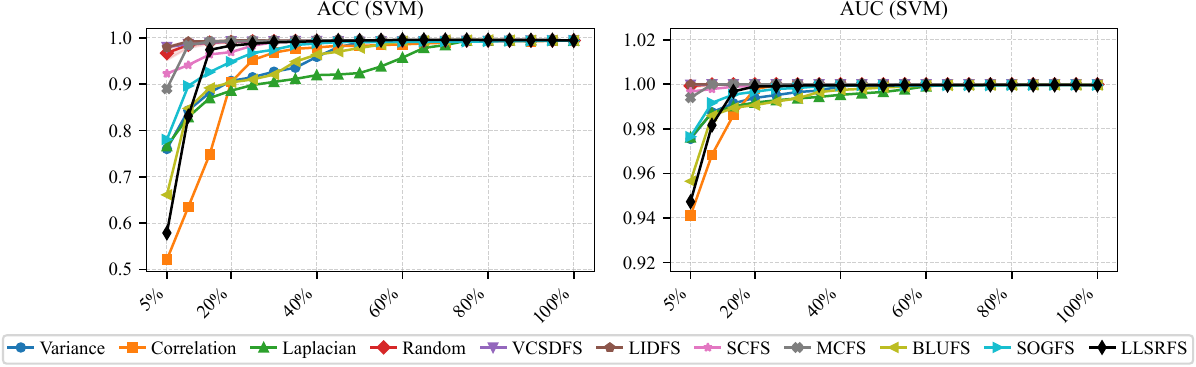}
\caption{ACC and AUC of the SVM classifier on the COIL20 dataset (5\% to 100\% of the features).}\label{fig:supplot-svm-100-COIL20}
\supback
\end{figure}
\begin{figure}[p]
\centering
\includegraphics[width=\textwidth,height=\supfigheight,keepaspectratio]{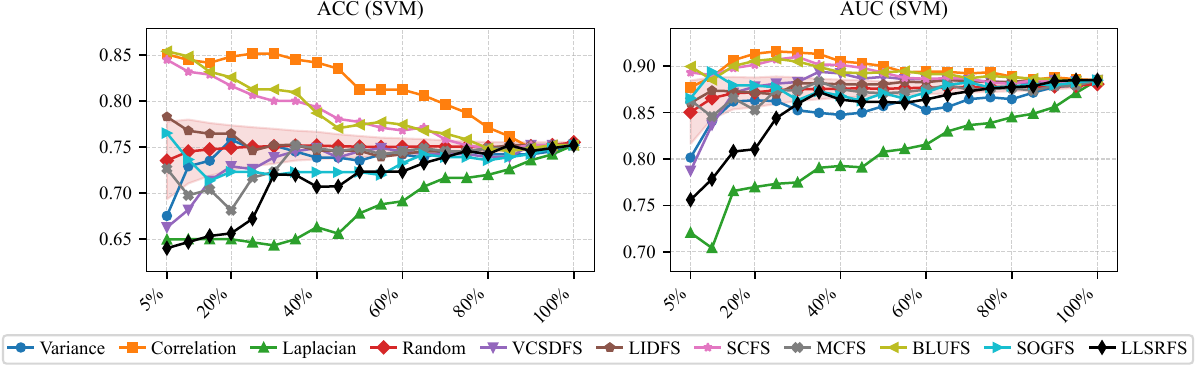}
\caption{ACC and AUC of the SVM classifier on the colon dataset (5\% to 100\% of the features).}\label{fig:supplot-svm-100-colon}
\supback
\end{figure}
\begin{figure}[p]
\centering
\includegraphics[width=\textwidth,height=\supfigheight,keepaspectratio]{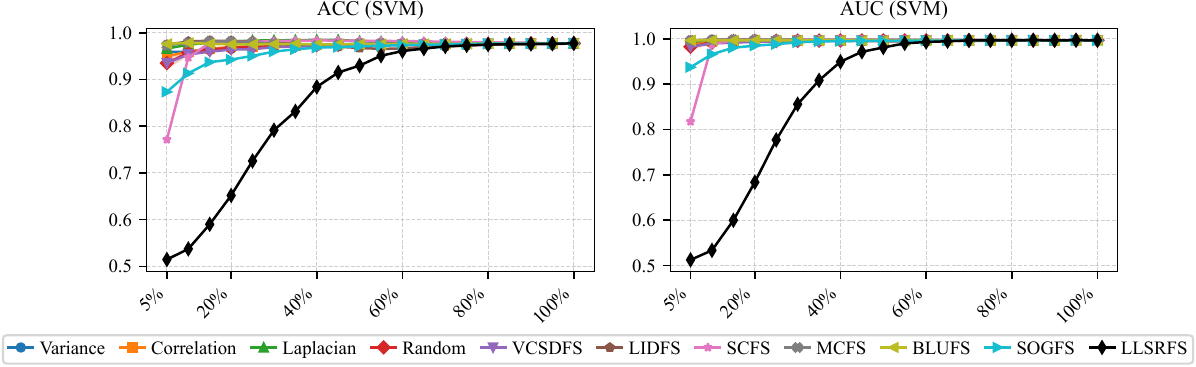}
\caption{ACC and AUC of the SVM classifier on the gisette dataset (5\% to 100\% of the features).}\label{fig:supplot-svm-100-gisette}
\supback
\end{figure}
\begin{figure}[p]
\centering
\includegraphics[width=\textwidth,height=\supfigheight,keepaspectratio]{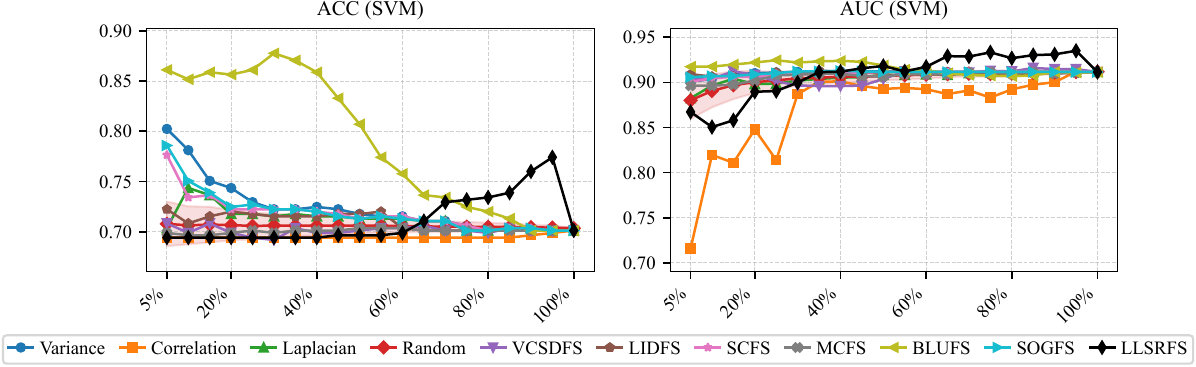}
\caption{ACC and AUC of the SVM classifier on the GLI\_85 dataset (5\% to 100\% of the features).}\label{fig:supplot-svm-100-GLI_85}
\supback
\end{figure}
\begin{figure}[p]
\centering
\includegraphics[width=\textwidth,height=\supfigheight,keepaspectratio]{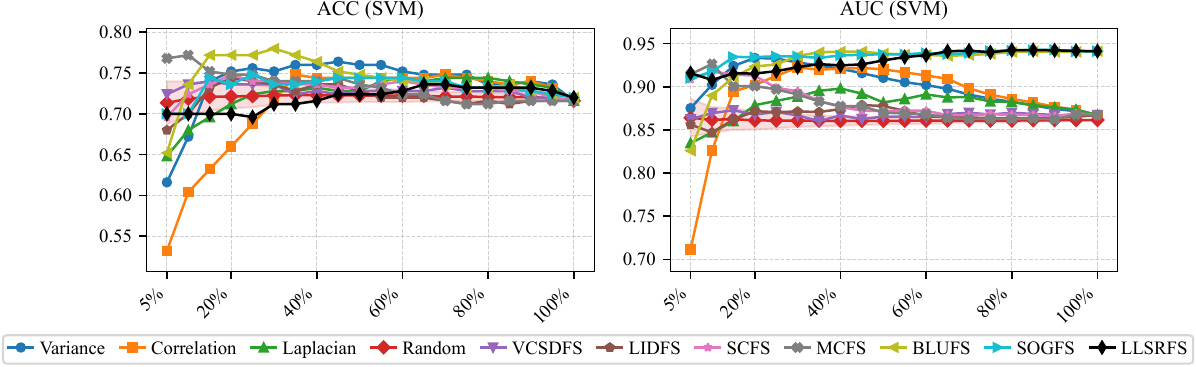}
\caption{ACC and AUC of the SVM classifier on the GLIOMA dataset (5\% to 100\% of the features).}\label{fig:supplot-svm-100-GLIOMA}
\supback
\end{figure}
\begin{figure}[p]
\centering
\includegraphics[width=\textwidth,height=\supfigheight,keepaspectratio]{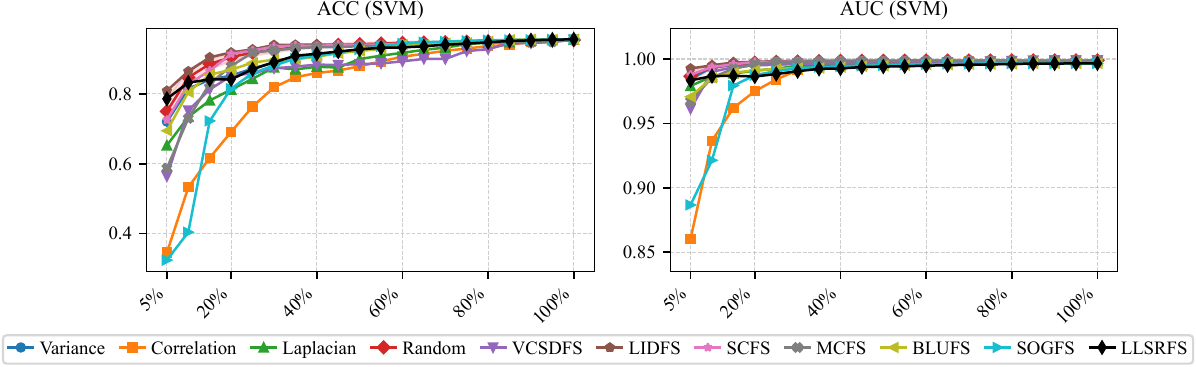}
\caption{ACC and AUC of the SVM classifier on the Isolet dataset (5\% to 100\% of the features).}\label{fig:supplot-svm-100-Isolet}
\supback
\end{figure}
\begin{figure}[p]
\centering
\includegraphics[width=\textwidth,height=\supfigheight,keepaspectratio]{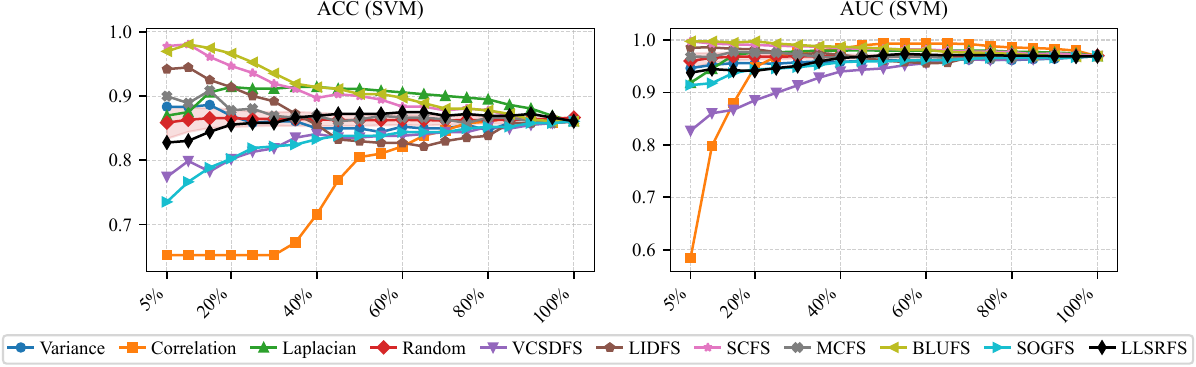}
\caption{ACC and AUC of the SVM classifier on the leukemia dataset (5\% to 100\% of the features).}\label{fig:supplot-svm-100-leukemia}
\supback
\end{figure}
\begin{figure}[p]
\centering
\includegraphics[width=\textwidth,height=\supfigheight,keepaspectratio]{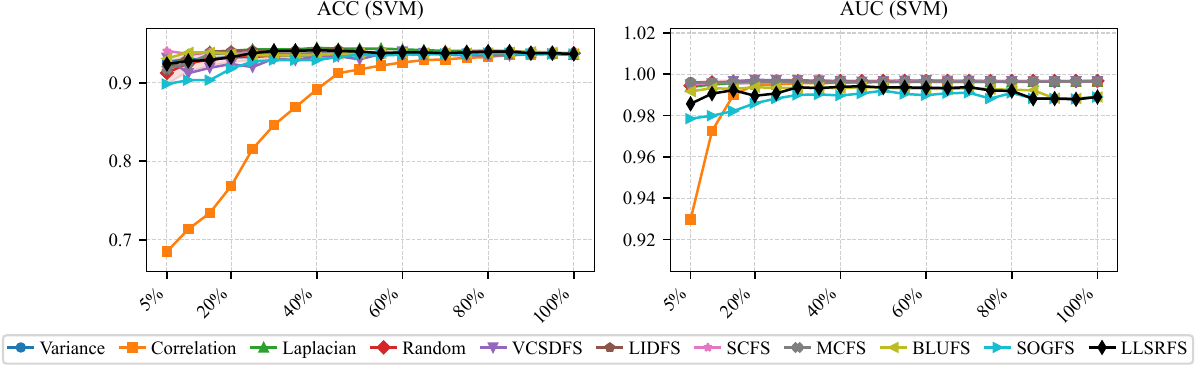}
\caption{ACC and AUC of the SVM classifier on the lung dataset (5\% to 100\% of the features).}\label{fig:supplot-svm-100-lung}
\supback
\end{figure}
\begin{figure}[p]
\centering
\includegraphics[width=\textwidth,height=\supfigheight,keepaspectratio]{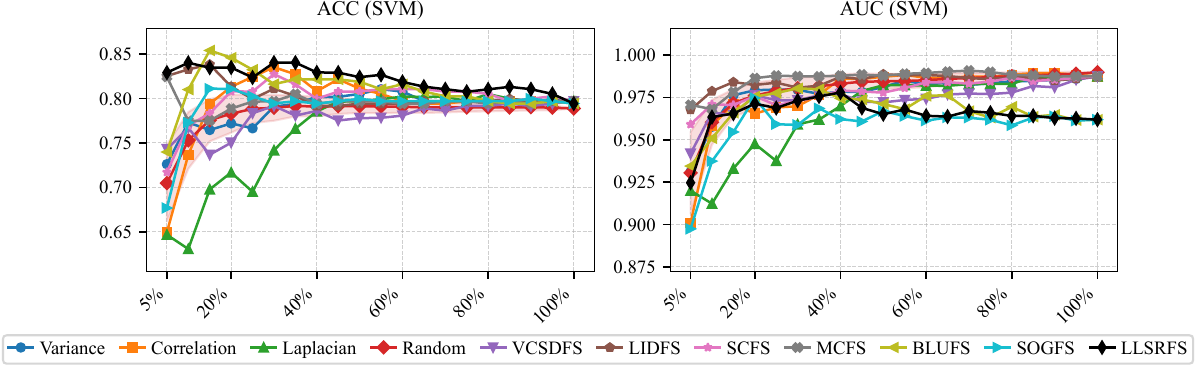}
\caption{ACC and AUC of the SVM classifier on the lung\_discrete dataset (5\% to 100\% of the features).}\label{fig:supplot-svm-100-lung_discrete}
\supback
\end{figure}
\begin{figure}[p]
\centering
\includegraphics[width=\textwidth,height=\supfigheight,keepaspectratio]{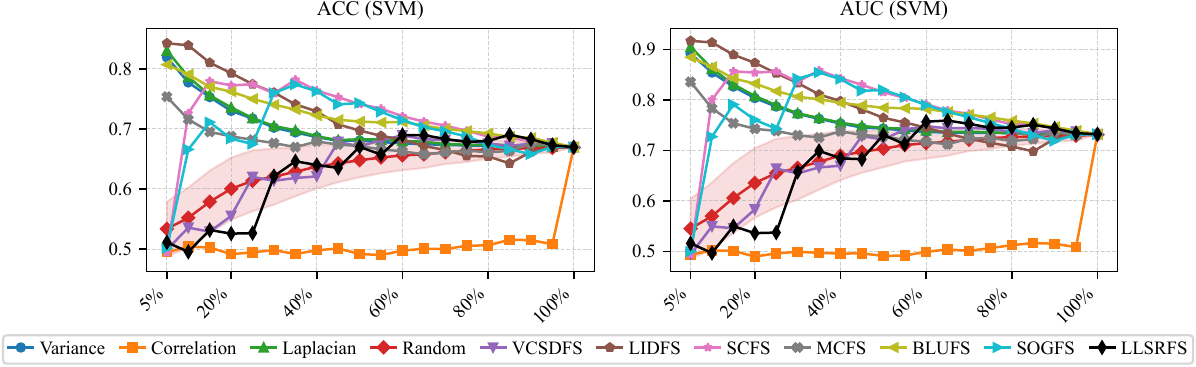}
\caption{ACC and AUC of the SVM classifier on the madelon dataset (5\% to 100\% of the features).}\label{fig:supplot-svm-100-madelon}
\supback
\end{figure}
\begin{figure}[p]
\centering
\includegraphics[width=\textwidth,height=\supfigheight,keepaspectratio]{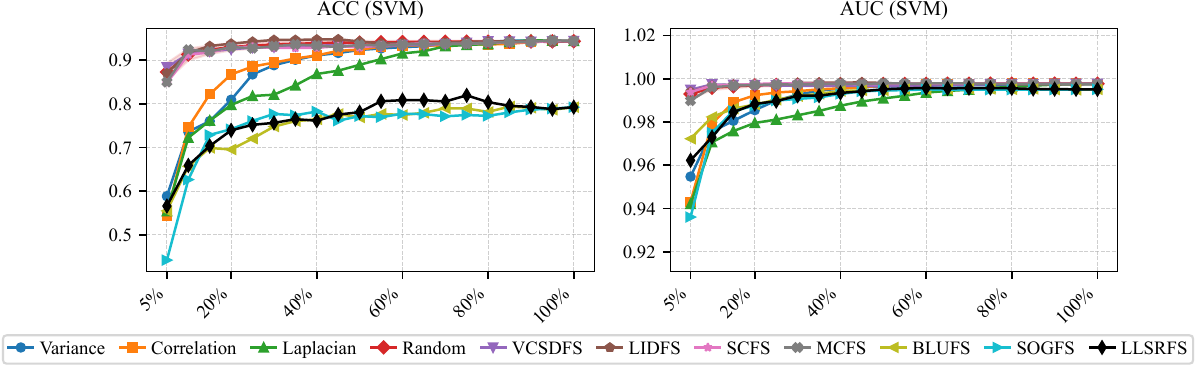}
\caption{ACC and AUC of the SVM classifier on the ORL dataset (5\% to 100\% of the features).}\label{fig:supplot-svm-100-ORL}
\supback
\end{figure}
\begin{figure}[p]
\centering
\includegraphics[width=\textwidth,height=\supfigheight,keepaspectratio]{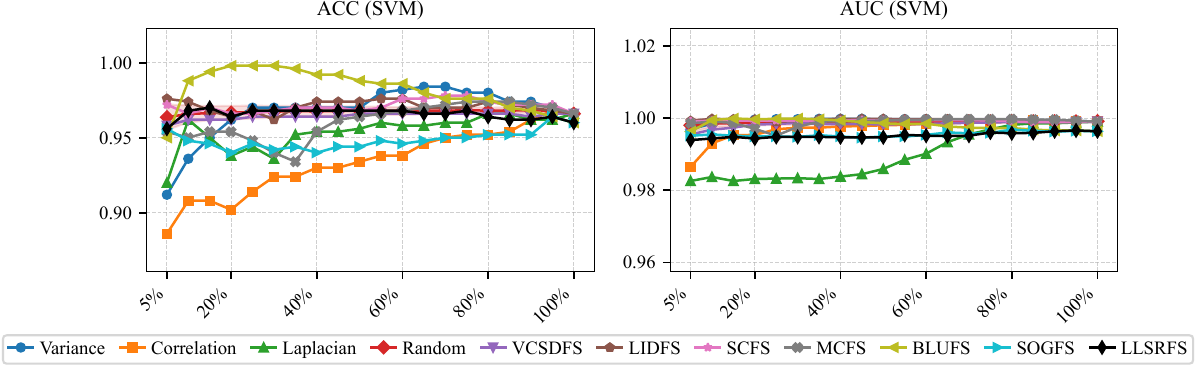}
\caption{ACC and AUC of the SVM classifier on the orlraws10P dataset (5\% to 100\% of the features).}\label{fig:supplot-svm-100-orlraws10P}
\supback
\end{figure}
\begin{figure}[p]
\centering
\includegraphics[width=\textwidth,height=\supfigheight,keepaspectratio]{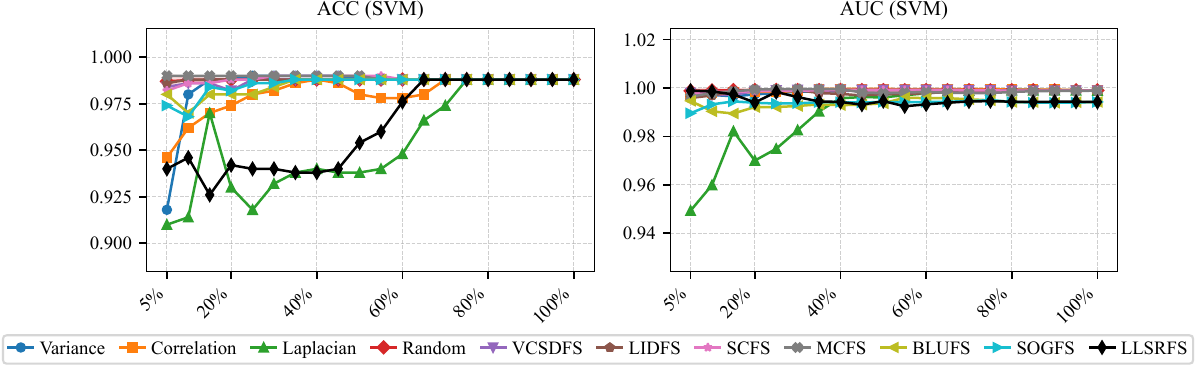}
\caption{ACC and AUC of the SVM classifier on the pixraw10P dataset (5\% to 100\% of the features).}\label{fig:supplot-svm-100-pixraw10P}
\supback
\end{figure}
\begin{figure}[p]
\centering
\includegraphics[width=\textwidth,height=\supfigheight,keepaspectratio]{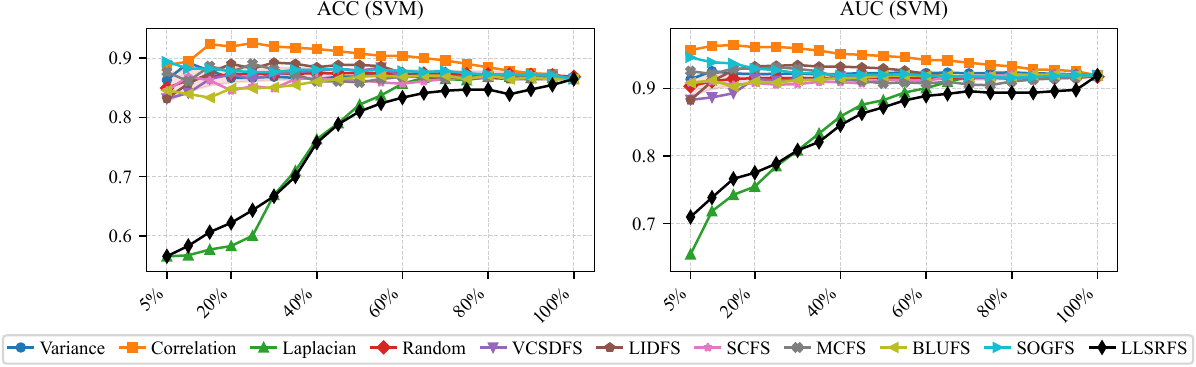}
\caption{ACC and AUC of the SVM classifier on the Prostate\_GE dataset (5\% to 100\% of the features).}\label{fig:supplot-svm-100-Prostate_GE}
\supback
\end{figure}
\begin{figure}[p]
\centering
\includegraphics[width=\textwidth,height=\supfigheight,keepaspectratio]{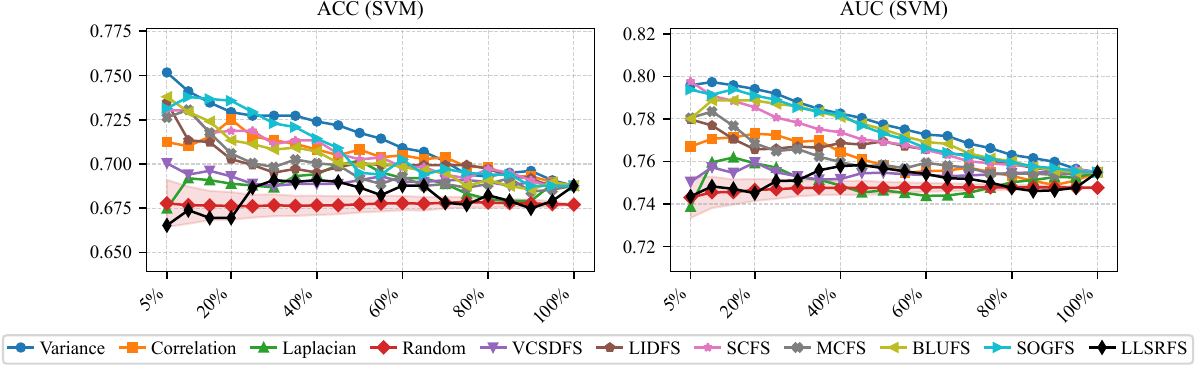}
\caption{ACC and AUC of the SVM classifier on the SMK\_CAN\_187 dataset (5\% to 100\% of the features).}\label{fig:supplot-svm-100-SMK_CAN_187}
\supback
\end{figure}
\begin{figure}[p]
\centering
\includegraphics[width=\textwidth,height=\supfigheight,keepaspectratio]{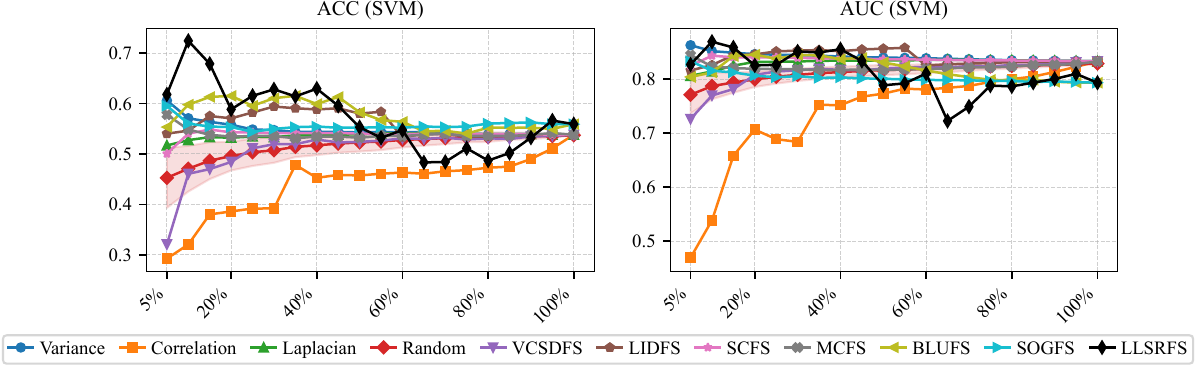}
\caption{ACC and AUC of the SVM classifier on the TOX\_171 dataset (5\% to 100\% of the features).}\label{fig:supplot-svm-100-TOX_171}
\supback
\end{figure}
\begin{figure}[p]
\centering
\includegraphics[width=\textwidth,height=\supfigheight,keepaspectratio]{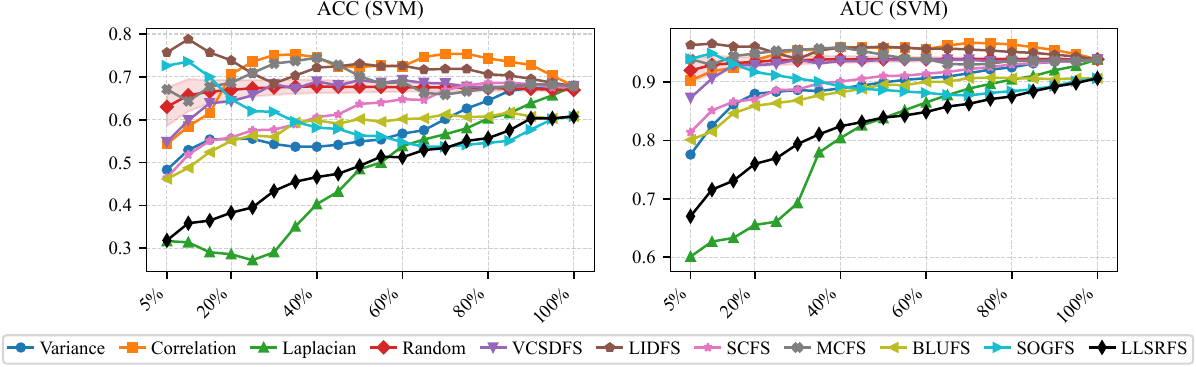}
\caption{ACC and AUC of the SVM classifier on the warpAR10P dataset (5\% to 100\% of the features).}\label{fig:supplot-svm-100-warpAR10P}
\supback
\end{figure}
\begin{figure}[p]
\centering
\includegraphics[width=\textwidth,height=\supfigheight,keepaspectratio]{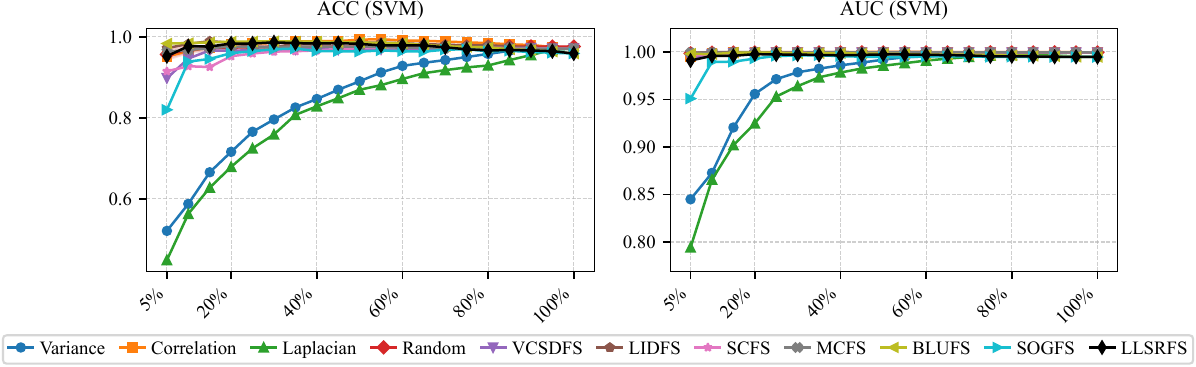}
\caption{ACC and AUC of the SVM classifier on the warpPIE10P dataset (5\% to 100\% of the features).}\label{fig:supplot-svm-100-warpPIE10P}
\supback
\end{figure}
\begin{figure}[p]
\centering
\includegraphics[width=\textwidth,height=\supfigheight,keepaspectratio]{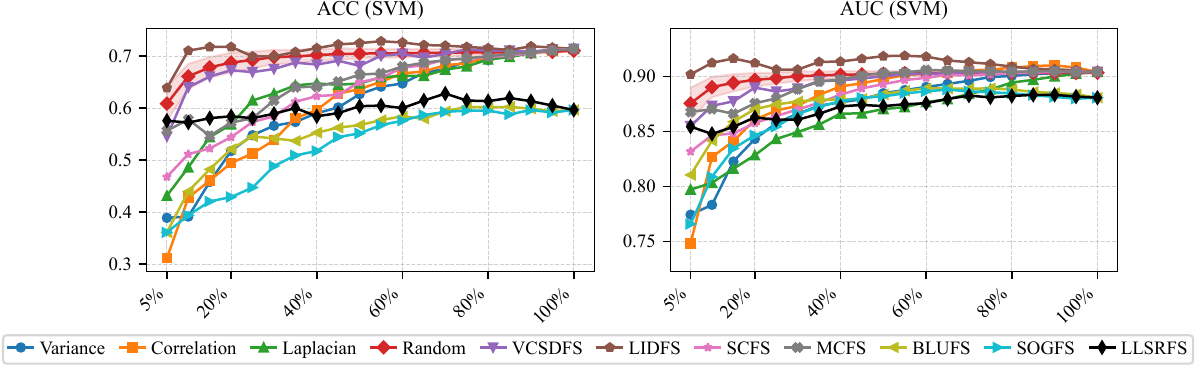}
\caption{ACC and AUC of the SVM classifier on the Yale dataset (5\% to 100\% of the features).}\label{fig:supplot-svm-100-Yale}
\supback
\end{figure}
\clearpage

\begin{figure}[p]
\centering
\pdfbookmark[3]{Random forest}{supplot-bm-rf}
\pdfbookmark[4]{0.5\% to 10\%}{supplot-bm-rf-10}
\includegraphics[width=\textwidth,height=\supfigheight,keepaspectratio]{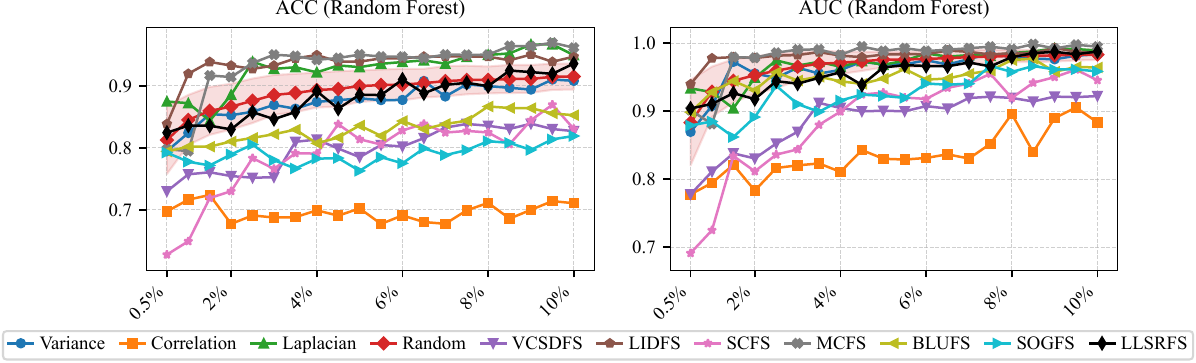}
\caption{ACC and AUC of the random forest classifier on the ALLAML dataset (0.5\% to 10\% of the features).}\label{fig:supplot-rf-10-ALLAML}
\supback
\end{figure}
\begin{figure}[p]
\centering
\includegraphics[width=\textwidth,height=\supfigheight,keepaspectratio]{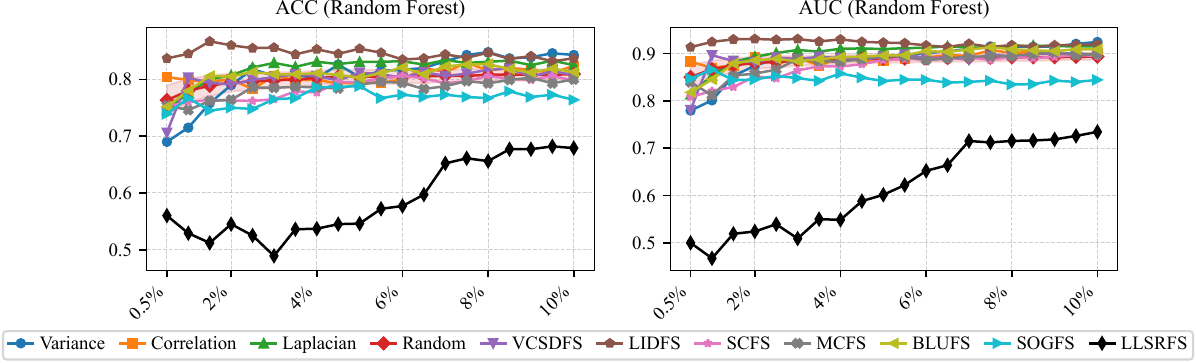}
\caption{ACC and AUC of the random forest classifier on the arcene dataset (0.5\% to 10\% of the features).}\label{fig:supplot-rf-10-arcene}
\supback
\end{figure}
\begin{figure}[p]
\centering
\includegraphics[width=\textwidth,height=\supfigheight,keepaspectratio]{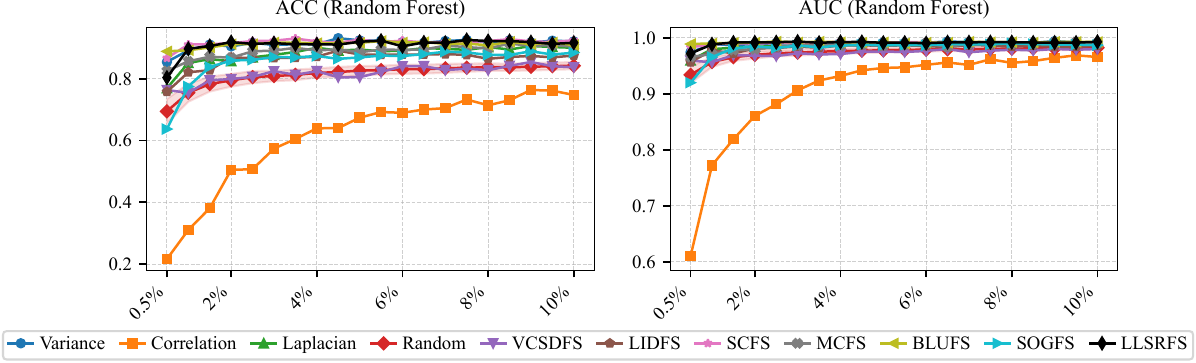}
\caption{ACC and AUC of the random forest classifier on the Carcinom dataset (0.5\% to 10\% of the features).}\label{fig:supplot-rf-10-Carcinom}
\supback
\end{figure}
\begin{figure}[p]
\centering
\includegraphics[width=\textwidth,height=\supfigheight,keepaspectratio]{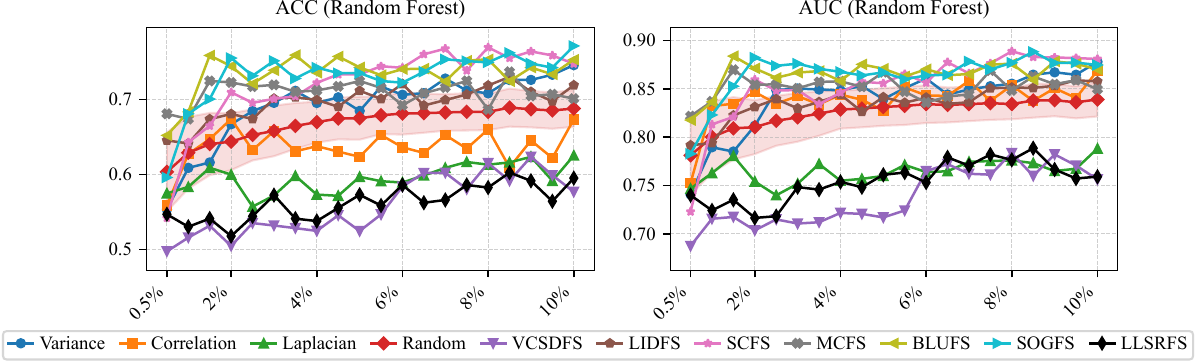}
\caption{ACC and AUC of the random forest classifier on the CLL\_SUB\_111 dataset (0.5\% to 10\% of the features).}\label{fig:supplot-rf-10-CLL_SUB_111}
\supback
\end{figure}
\begin{figure}[p]
\centering
\includegraphics[width=\textwidth,height=\supfigheight,keepaspectratio]{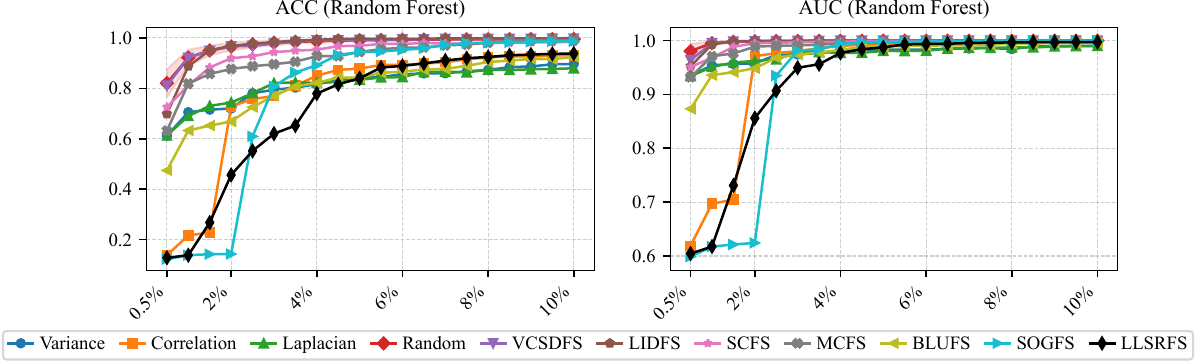}
\caption{ACC and AUC of the random forest classifier on the COIL20 dataset (0.5\% to 10\% of the features).}\label{fig:supplot-rf-10-COIL20}
\supback
\end{figure}
\begin{figure}[p]
\centering
\includegraphics[width=\textwidth,height=\supfigheight,keepaspectratio]{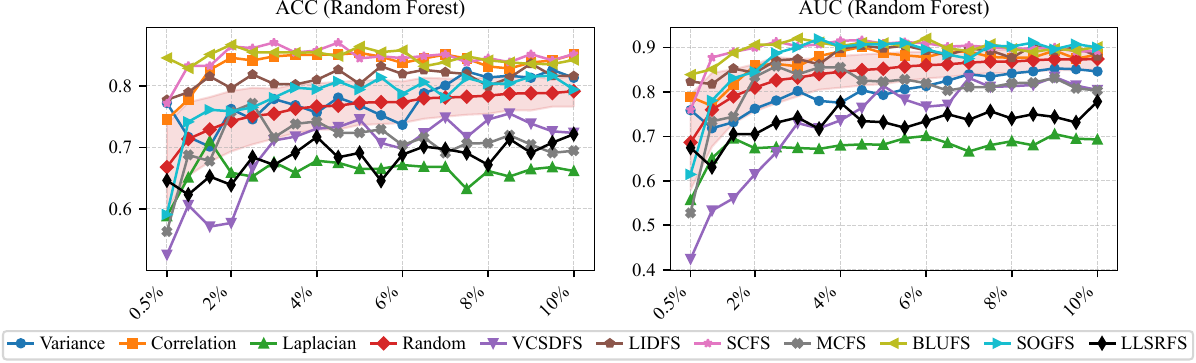}
\caption{ACC and AUC of the random forest classifier on the colon dataset (0.5\% to 10\% of the features).}\label{fig:supplot-rf-10-colon}
\supback
\end{figure}
\begin{figure}[p]
\centering
\includegraphics[width=\textwidth,height=\supfigheight,keepaspectratio]{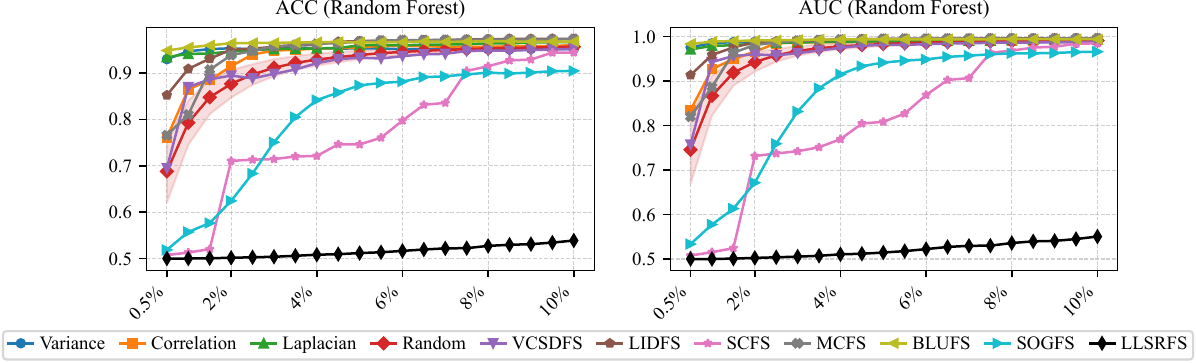}
\caption{ACC and AUC of the random forest classifier on the gisette dataset (0.5\% to 10\% of the features).}\label{fig:supplot-rf-10-gisette}
\supback
\end{figure}
\begin{figure}[p]
\centering
\includegraphics[width=\textwidth,height=\supfigheight,keepaspectratio]{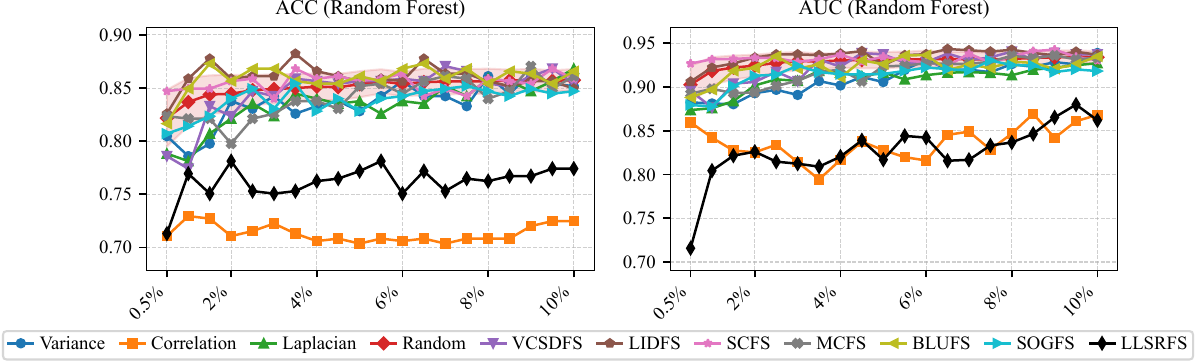}
\caption{ACC and AUC of the random forest classifier on the GLI\_85 dataset (0.5\% to 10\% of the features).}\label{fig:supplot-rf-10-GLI_85}
\supback
\end{figure}
\begin{figure}[p]
\centering
\includegraphics[width=\textwidth,height=\supfigheight,keepaspectratio]{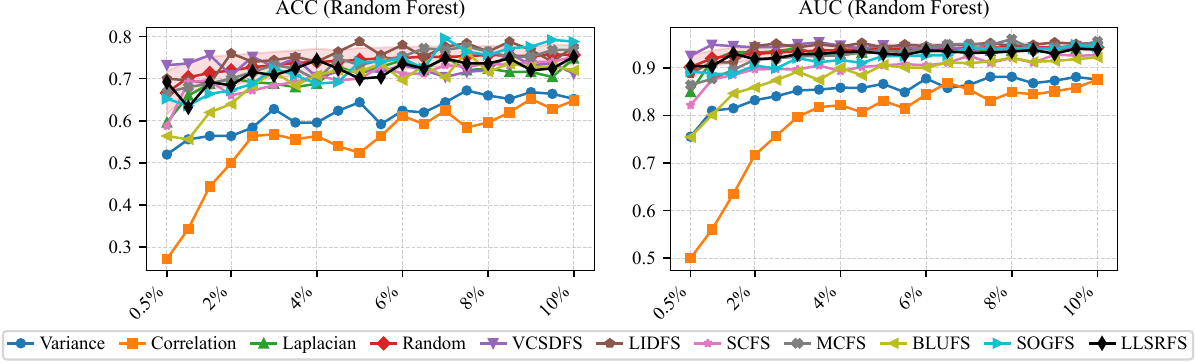}
\caption{ACC and AUC of the random forest classifier on the GLIOMA dataset (0.5\% to 10\% of the features).}\label{fig:supplot-rf-10-GLIOMA}
\supback
\end{figure}
\begin{figure}[p]
\centering
\includegraphics[width=\textwidth,height=\supfigheight,keepaspectratio]{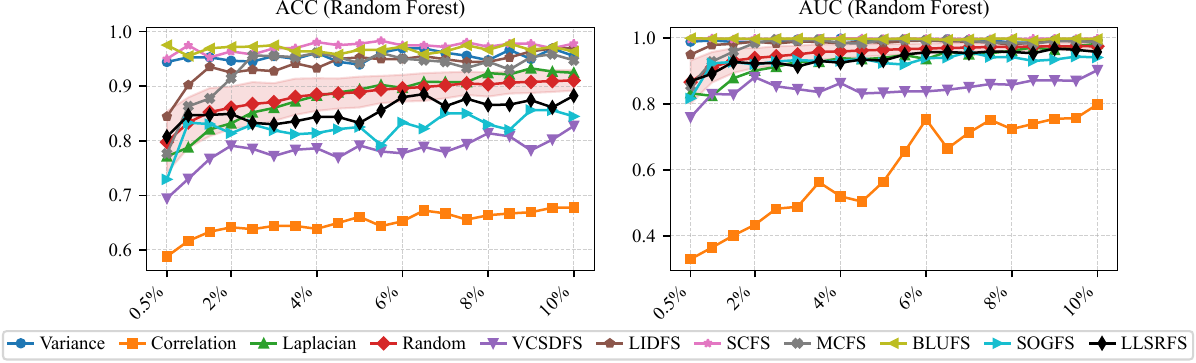}
\caption{ACC and AUC of the random forest classifier on the leukemia dataset (0.5\% to 10\% of the features).}\label{fig:supplot-rf-10-leukemia}
\supback
\end{figure}
\begin{figure}[p]
\centering
\includegraphics[width=\textwidth,height=\supfigheight,keepaspectratio]{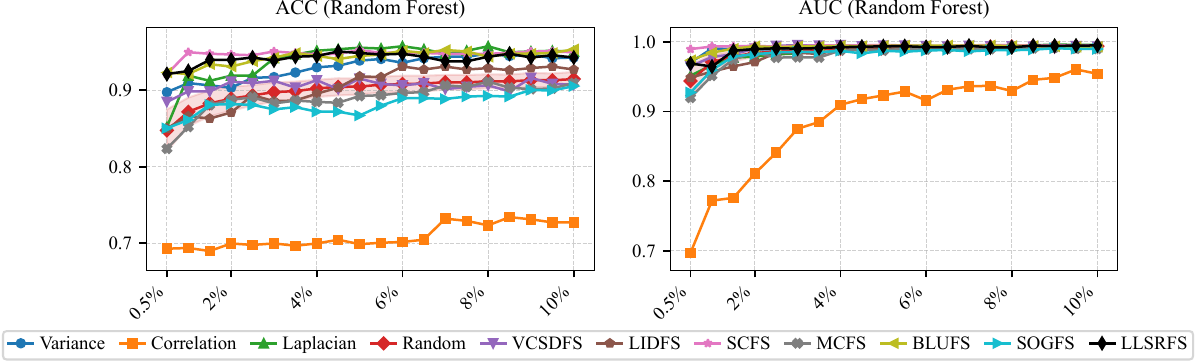}
\caption{ACC and AUC of the random forest classifier on the lung dataset (0.5\% to 10\% of the features).}\label{fig:supplot-rf-10-lung}
\supback
\end{figure}
\begin{figure}[p]
\centering
\includegraphics[width=\textwidth,height=\supfigheight,keepaspectratio]{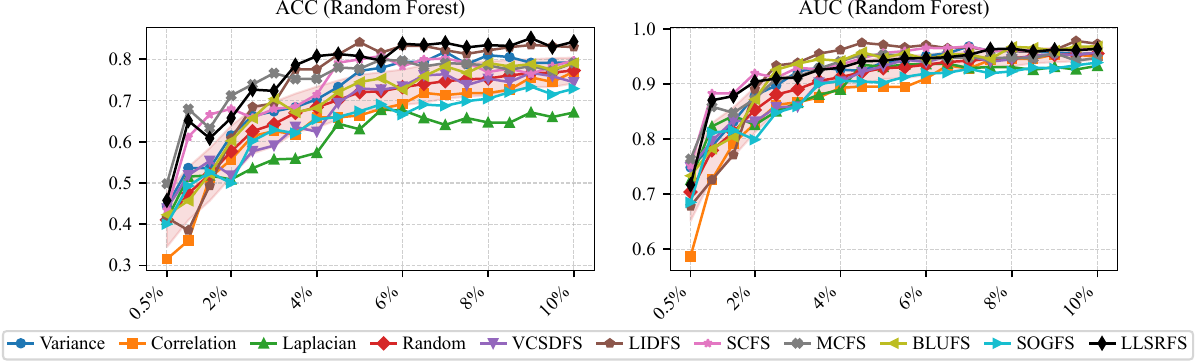}
\caption{ACC and AUC of the random forest classifier on the lung\_discrete dataset (0.5\% to 10\% of the features).}\label{fig:supplot-rf-10-lung_discrete}
\supback
\end{figure}
\begin{figure}[p]
\centering
\includegraphics[width=\textwidth,height=\supfigheight,keepaspectratio]{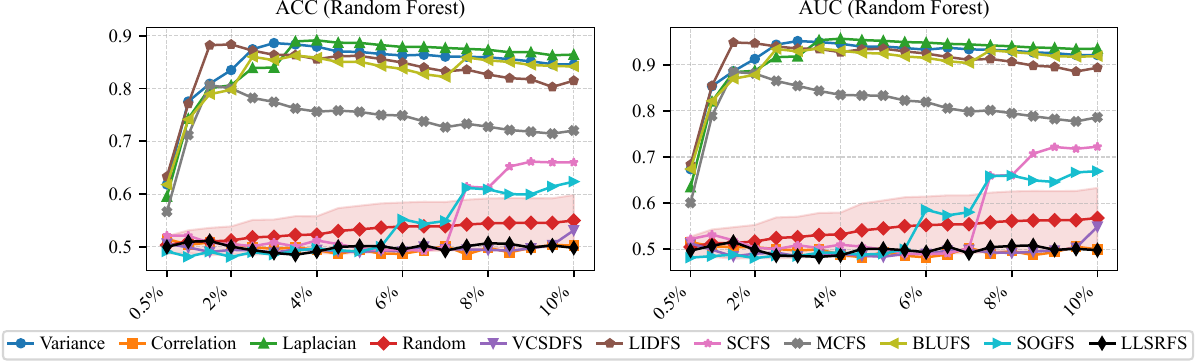}
\caption{ACC and AUC of the random forest classifier on the madelon dataset (0.5\% to 10\% of the features).}\label{fig:supplot-rf-10-madelon}
\supback
\end{figure}
\begin{figure}[p]
\centering
\includegraphics[width=\textwidth,height=\supfigheight,keepaspectratio]{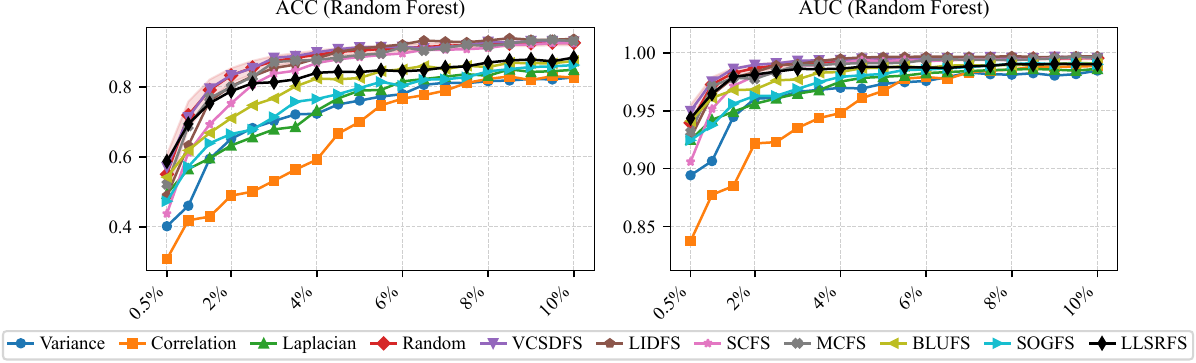}
\caption{ACC and AUC of the random forest classifier on the ORL dataset (0.5\% to 10\% of the features).}\label{fig:supplot-rf-10-ORL}
\supback
\end{figure}
\begin{figure}[p]
\centering
\includegraphics[width=\textwidth,height=\supfigheight,keepaspectratio]{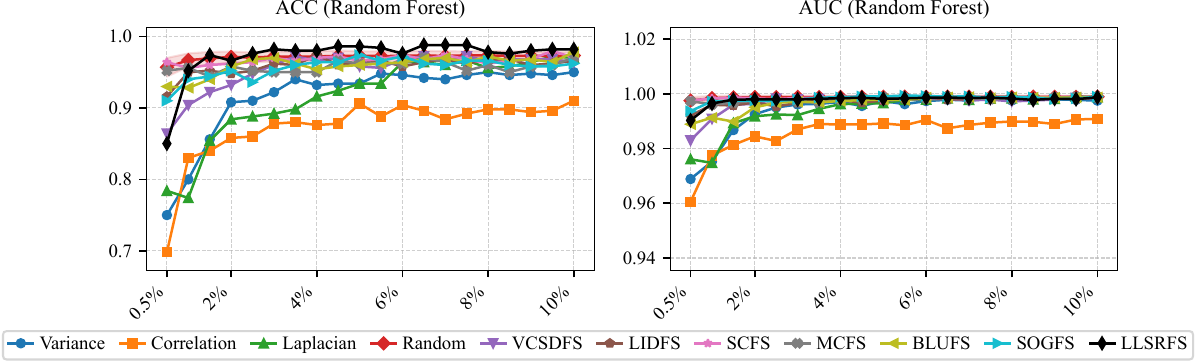}
\caption{ACC and AUC of the random forest classifier on the orlraws10P dataset (0.5\% to 10\% of the features).}\label{fig:supplot-rf-10-orlraws10P}
\supback
\end{figure}
\begin{figure}[p]
\centering
\includegraphics[width=\textwidth,height=\supfigheight,keepaspectratio]{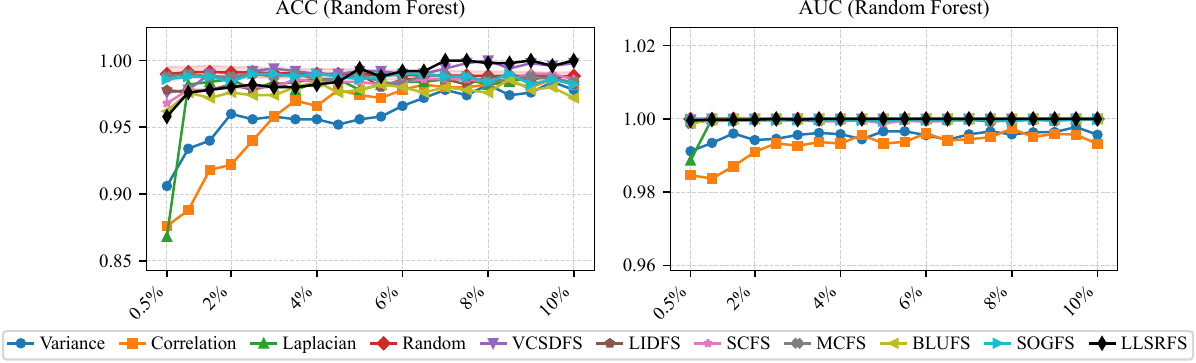}
\caption{ACC and AUC of the random forest classifier on the pixraw10P dataset (0.5\% to 10\% of the features).}\label{fig:supplot-rf-10-pixraw10P}
\supback
\end{figure}
\begin{figure}[p]
\centering
\includegraphics[width=\textwidth,height=\supfigheight,keepaspectratio]{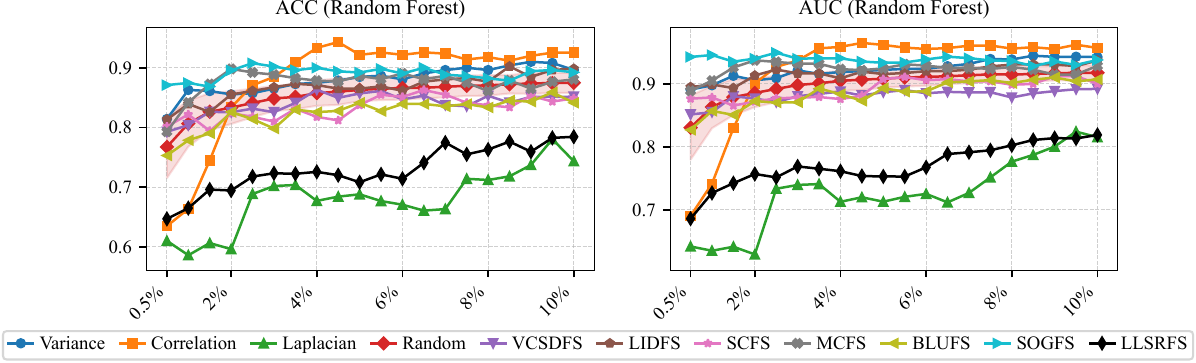}
\caption{ACC and AUC of the random forest classifier on the Prostate\_GE dataset (0.5\% to 10\% of the features).}\label{fig:supplot-rf-10-Prostate_GE}
\supback
\end{figure}
\begin{figure}[p]
\centering
\includegraphics[width=\textwidth,height=\supfigheight,keepaspectratio]{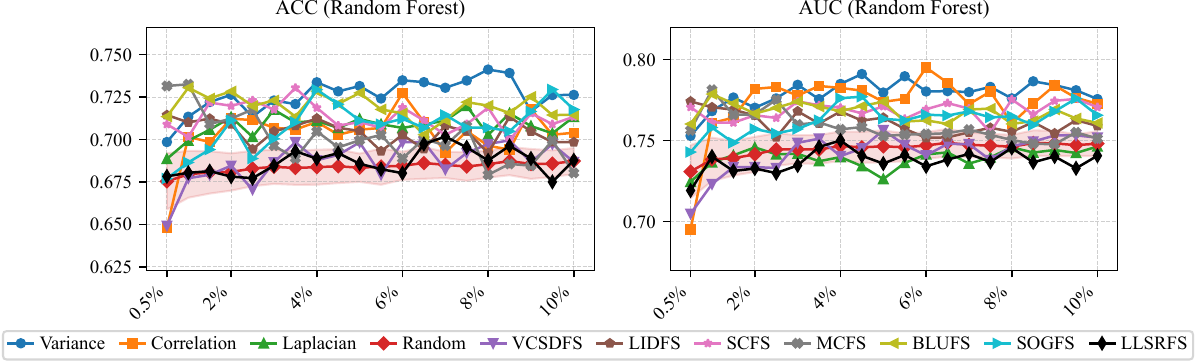}
\caption{ACC and AUC of the random forest classifier on the SMK\_CAN\_187 dataset (0.5\% to 10\% of the features).}\label{fig:supplot-rf-10-SMK_CAN_187}
\supback
\end{figure}
\begin{figure}[p]
\centering
\includegraphics[width=\textwidth,height=\supfigheight,keepaspectratio]{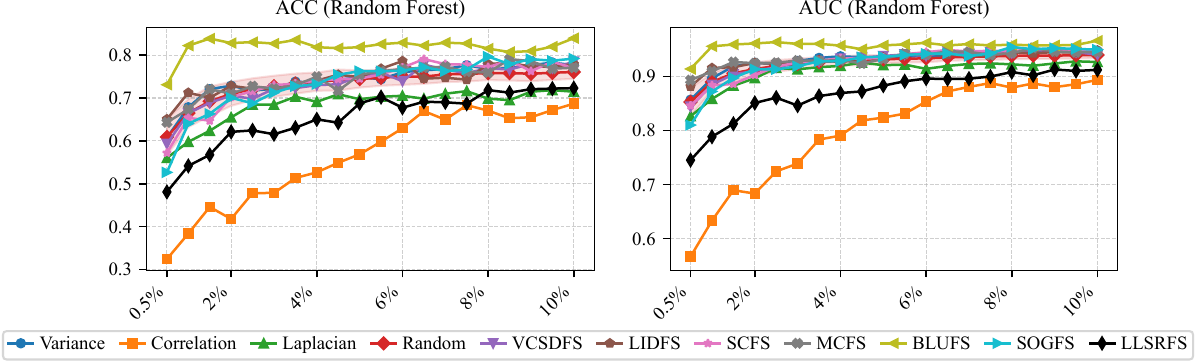}
\caption{ACC and AUC of the random forest classifier on the TOX\_171 dataset (0.5\% to 10\% of the features).}\label{fig:supplot-rf-10-TOX_171}
\supback
\end{figure}
\begin{figure}[p]
\centering
\includegraphics[width=\textwidth,height=\supfigheight,keepaspectratio]{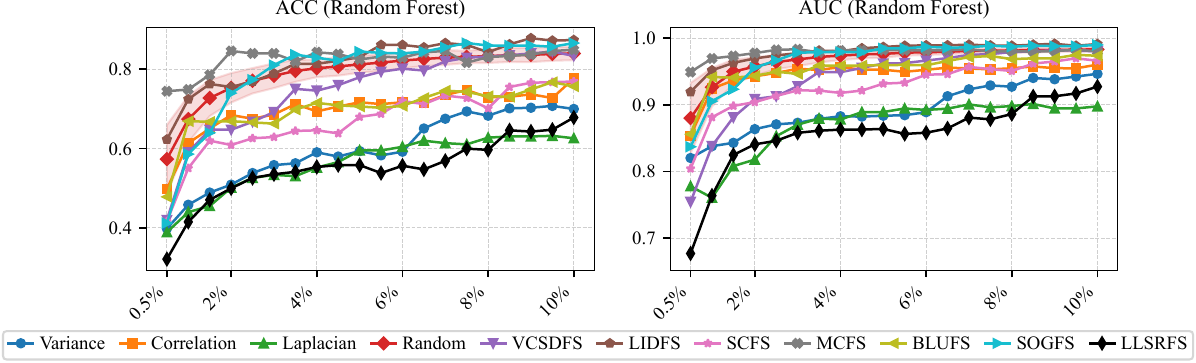}
\caption{ACC and AUC of the random forest classifier on the warpAR10P dataset (0.5\% to 10\% of the features).}\label{fig:supplot-rf-10-warpAR10P}
\supback
\end{figure}
\begin{figure}[p]
\centering
\includegraphics[width=\textwidth,height=\supfigheight,keepaspectratio]{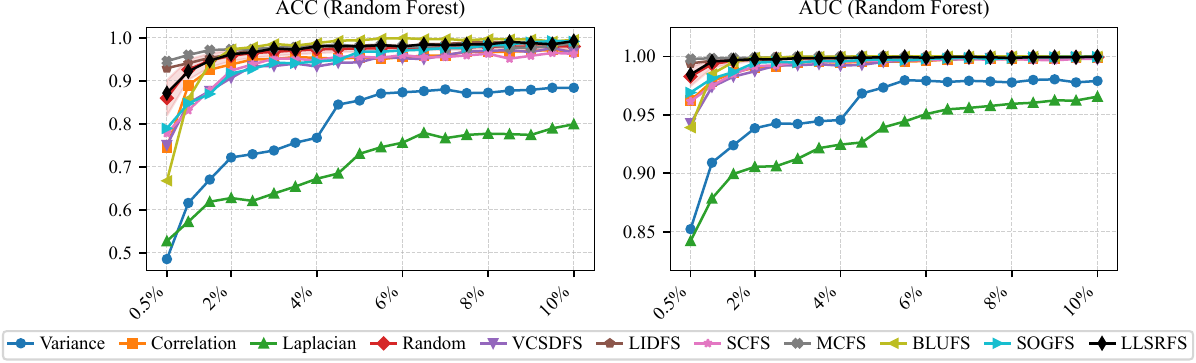}
\caption{ACC and AUC of the random forest classifier on the warpPIE10P dataset (0.5\% to 10\% of the features).}\label{fig:supplot-rf-10-warpPIE10P}
\supback
\end{figure}
\begin{figure}[p]
\centering
\includegraphics[width=\textwidth,height=\supfigheight,keepaspectratio]{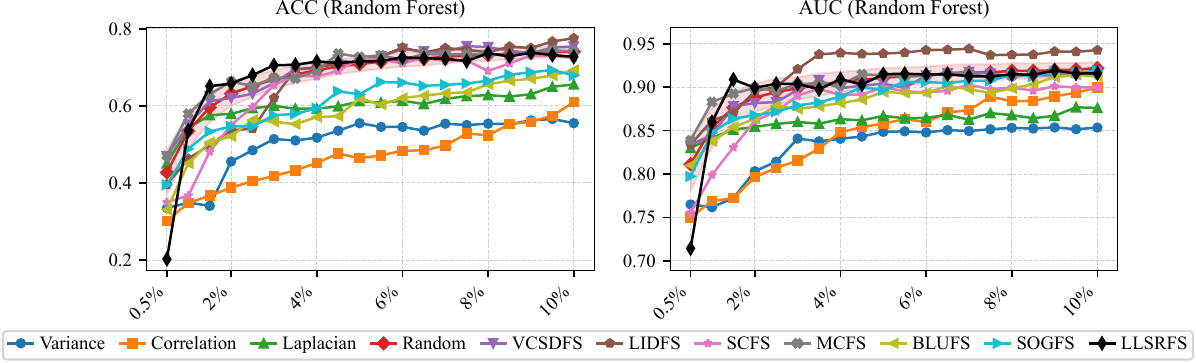}
\caption{ACC and AUC of the random forest classifier on the Yale dataset (0.5\% to 10\% of the features).}\label{fig:supplot-rf-10-Yale}
\supback
\end{figure}
\begin{figure}[p]
\centering
\pdfbookmark[4]{5\% to 100\%}{supplot-bm-rf-100}
\includegraphics[width=\textwidth,height=\supfigheight,keepaspectratio]{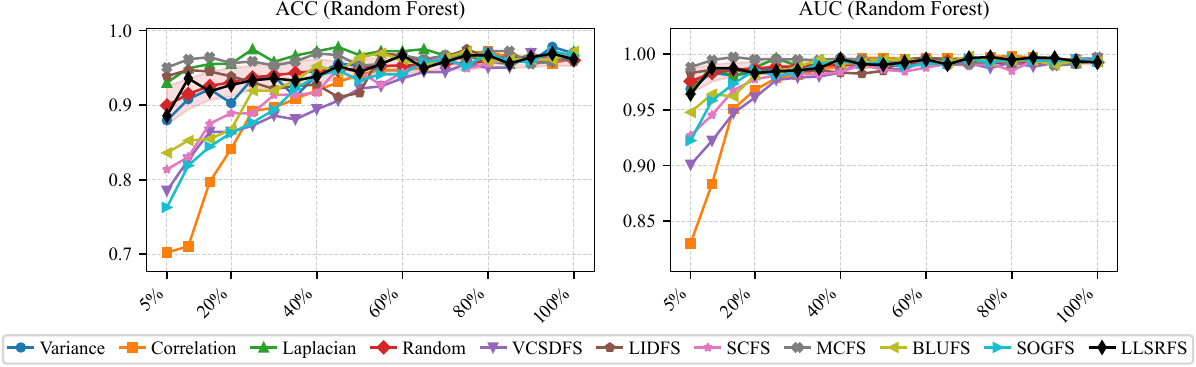}
\caption{ACC and AUC of the random forest classifier on the ALLAML dataset (5\% to 100\% of the features).}\label{fig:supplot-rf-100-ALLAML}
\supback
\end{figure}
\begin{figure}[p]
\centering
\includegraphics[width=\textwidth,height=\supfigheight,keepaspectratio]{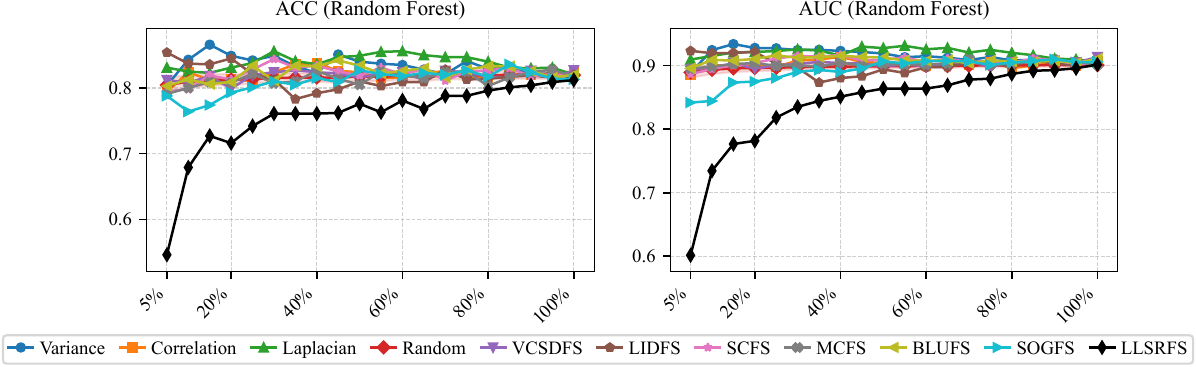}
\caption{ACC and AUC of the random forest classifier on the arcene dataset (5\% to 100\% of the features).}\label{fig:supplot-rf-100-arcene}
\supback
\end{figure}
\begin{figure}[p]
\centering
\includegraphics[width=\textwidth,height=\supfigheight,keepaspectratio]{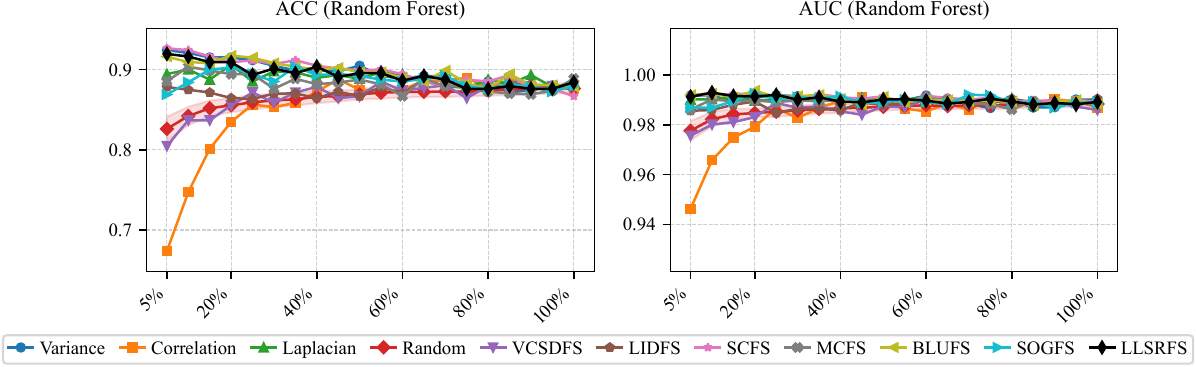}
\caption{ACC and AUC of the random forest classifier on the Carcinom dataset (5\% to 100\% of the features).}\label{fig:supplot-rf-100-Carcinom}
\supback
\end{figure}
\begin{figure}[p]
\centering
\includegraphics[width=\textwidth,height=\supfigheight,keepaspectratio]{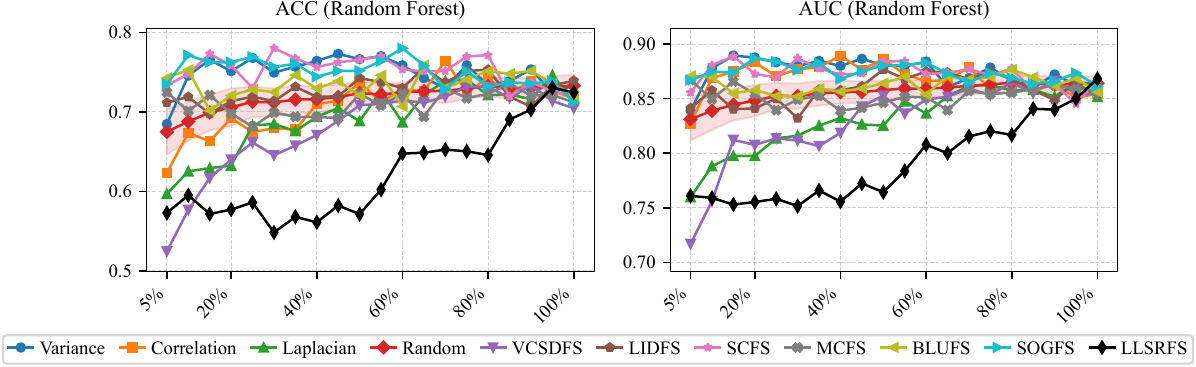}
\caption{ACC and AUC of the random forest classifier on the CLL\_SUB\_111 dataset (5\% to 100\% of the features).}\label{fig:supplot-rf-100-CLL_SUB_111}
\supback
\end{figure}
\begin{figure}[p]
\centering
\includegraphics[width=\textwidth,height=\supfigheight,keepaspectratio]{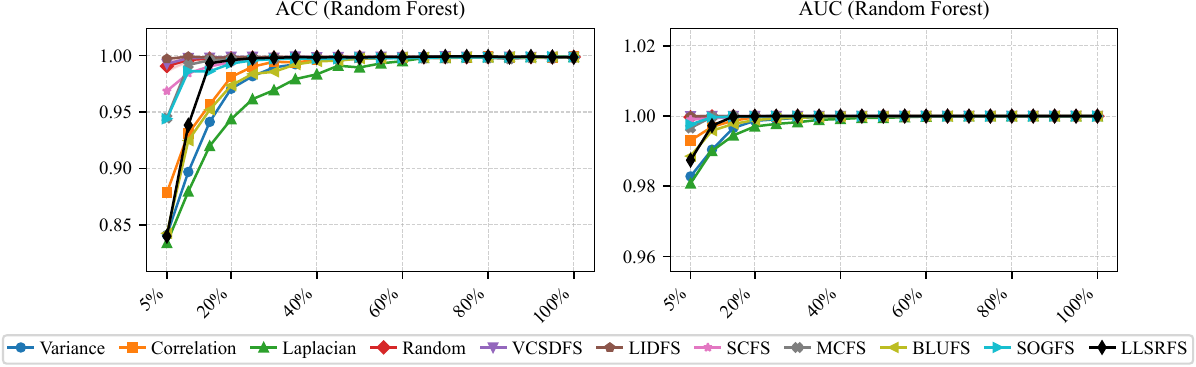}
\caption{ACC and AUC of the random forest classifier on the COIL20 dataset (5\% to 100\% of the features).}\label{fig:supplot-rf-100-COIL20}
\supback
\end{figure}
\begin{figure}[p]
\centering
\includegraphics[width=\textwidth,height=\supfigheight,keepaspectratio]{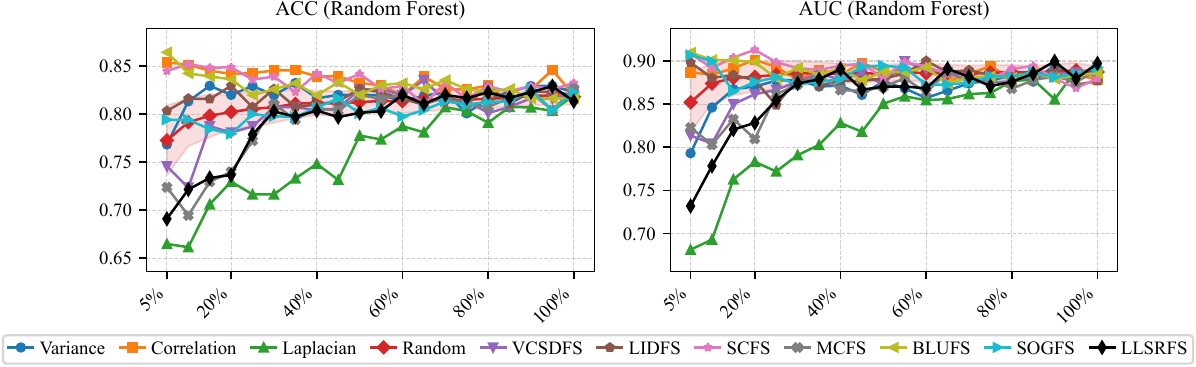}
\caption{ACC and AUC of the random forest classifier on the colon dataset (5\% to 100\% of the features).}\label{fig:supplot-rf-100-colon}
\supback
\end{figure}
\begin{figure}[p]
\centering
\includegraphics[width=\textwidth,height=\supfigheight,keepaspectratio]{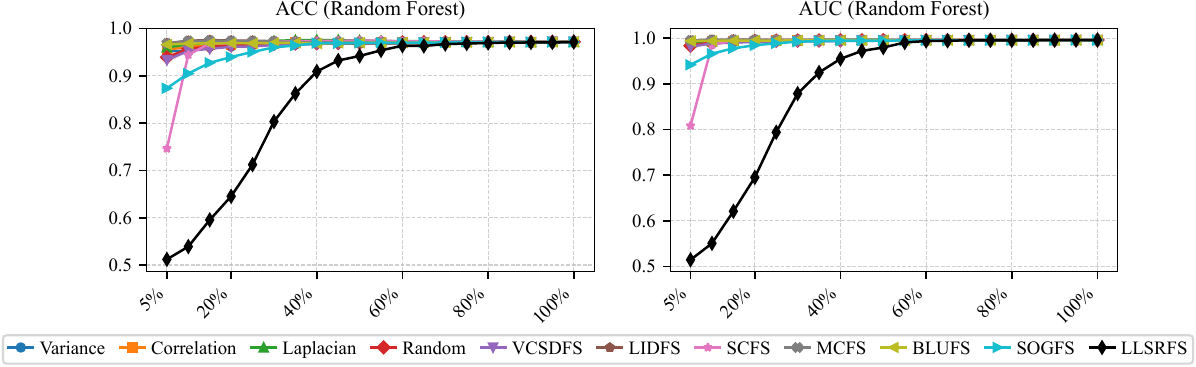}
\caption{ACC and AUC of the random forest classifier on the gisette dataset (5\% to 100\% of the features).}\label{fig:supplot-rf-100-gisette}
\supback
\end{figure}
\begin{figure}[p]
\centering
\includegraphics[width=\textwidth,height=\supfigheight,keepaspectratio]{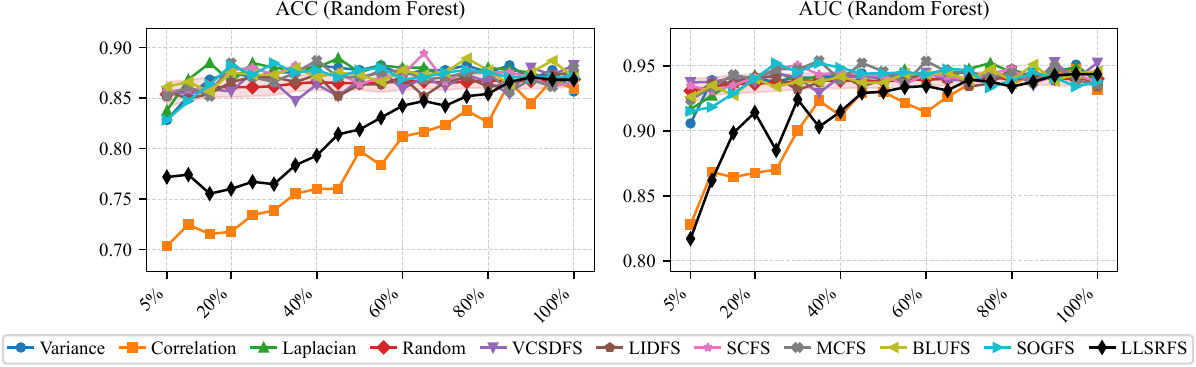}
\caption{ACC and AUC of the random forest classifier on the GLI\_85 dataset (5\% to 100\% of the features).}\label{fig:supplot-rf-100-GLI_85}
\supback
\end{figure}
\begin{figure}[p]
\centering
\includegraphics[width=\textwidth,height=\supfigheight,keepaspectratio]{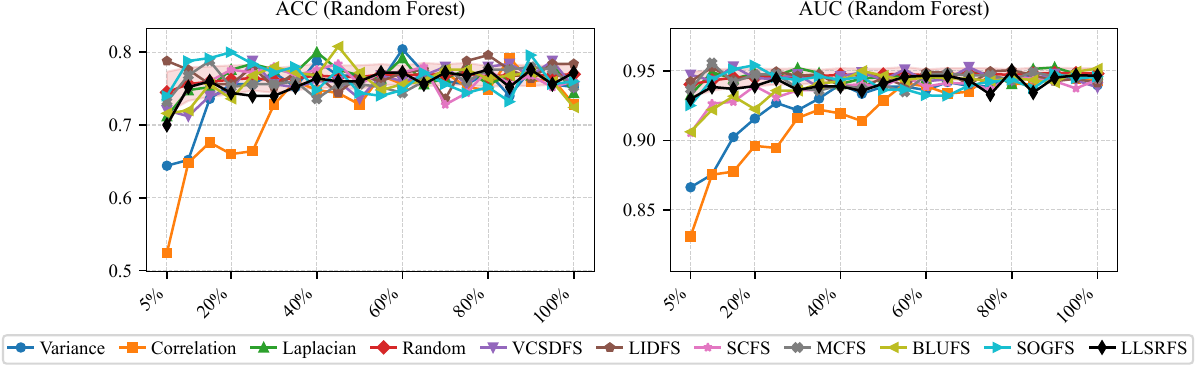}
\caption{ACC and AUC of the random forest classifier on the GLIOMA dataset (5\% to 100\% of the features).}\label{fig:supplot-rf-100-GLIOMA}
\supback
\end{figure}
\begin{figure}[p]
\centering
\includegraphics[width=\textwidth,height=\supfigheight,keepaspectratio]{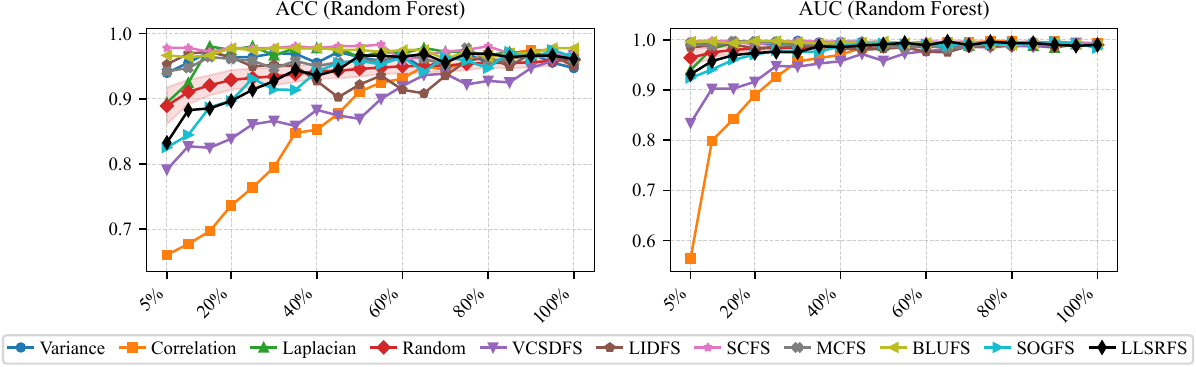}
\caption{ACC and AUC of the random forest classifier on the leukemia dataset (5\% to 100\% of the features).}\label{fig:supplot-rf-100-leukemia}
\supback
\end{figure}
\begin{figure}[p]
\centering
\includegraphics[width=\textwidth,height=\supfigheight,keepaspectratio]{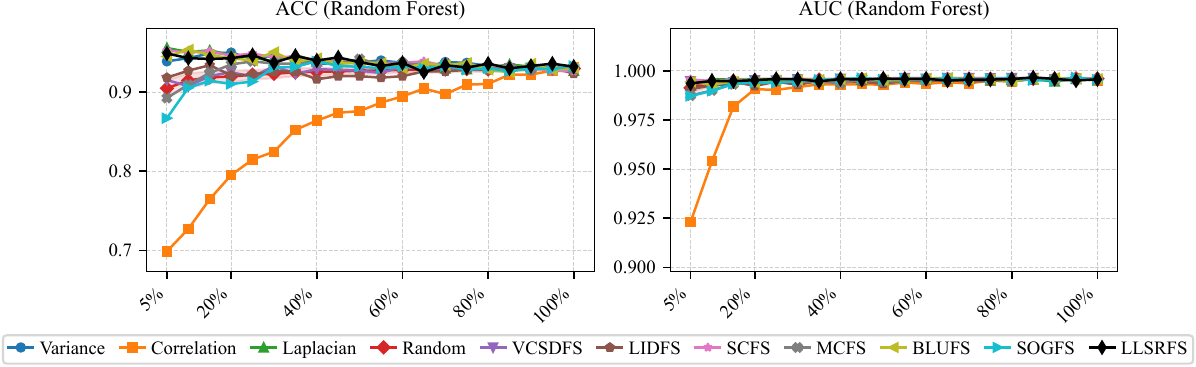}
\caption{ACC and AUC of the random forest classifier on the lung dataset (5\% to 100\% of the features).}\label{fig:supplot-rf-100-lung}
\supback
\end{figure}
\begin{figure}[p]
\centering
\includegraphics[width=\textwidth,height=\supfigheight,keepaspectratio]{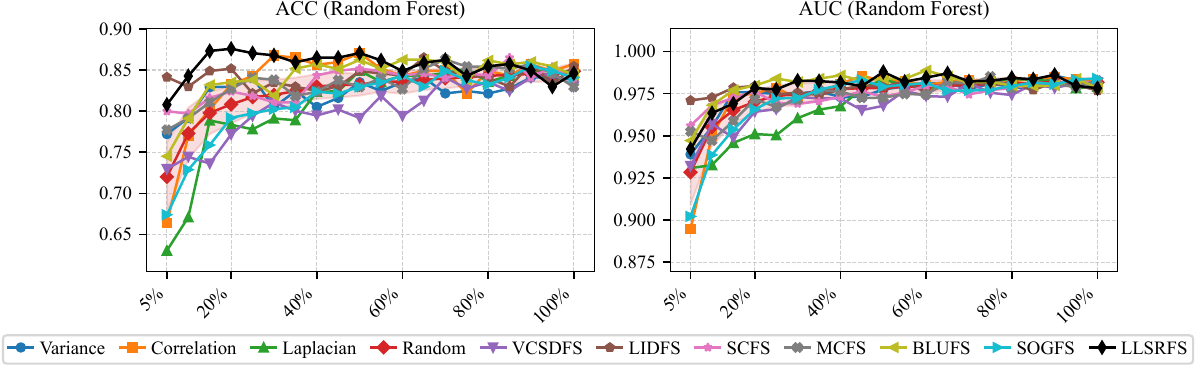}
\caption{ACC and AUC of the random forest classifier on the lung\_discrete dataset (5\% to 100\% of the features).}\label{fig:supplot-rf-100-lung_discrete}
\supback
\end{figure}
\begin{figure}[p]
\centering
\includegraphics[width=\textwidth,height=\supfigheight,keepaspectratio]{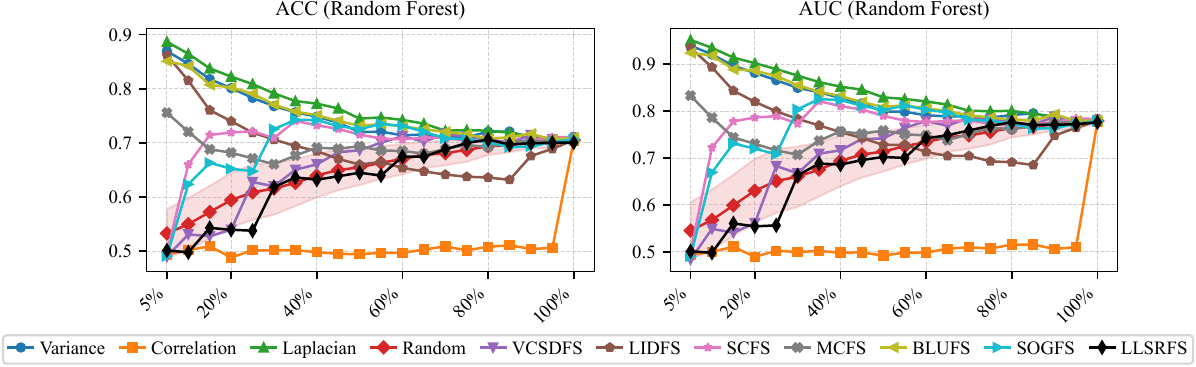}
\caption{ACC and AUC of the random forest classifier on the madelon dataset (5\% to 100\% of the features).}\label{fig:supplot-rf-100-madelon}
\supback
\end{figure}
\begin{figure}[p]
\centering
\includegraphics[width=\textwidth,height=\supfigheight,keepaspectratio]{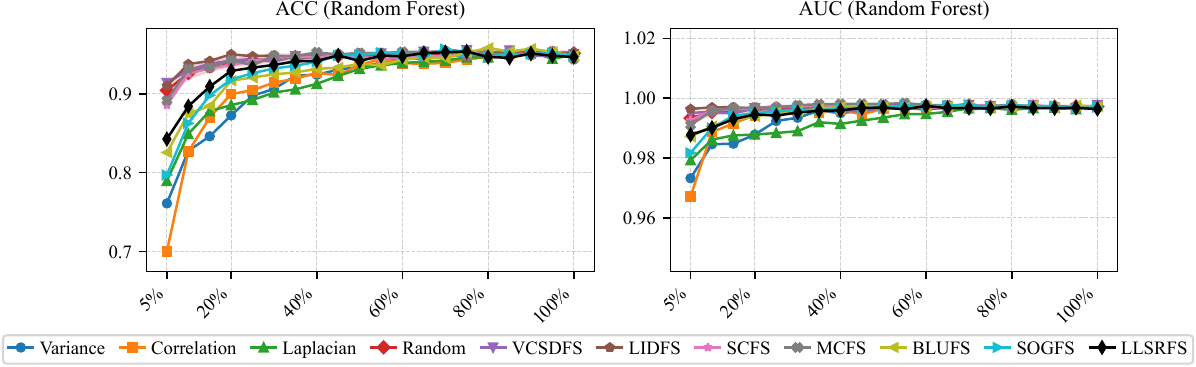}
\caption{ACC and AUC of the random forest classifier on the ORL dataset (5\% to 100\% of the features).}\label{fig:supplot-rf-100-ORL}
\supback
\end{figure}
\begin{figure}[p]
\centering
\includegraphics[width=\textwidth,height=\supfigheight,keepaspectratio]{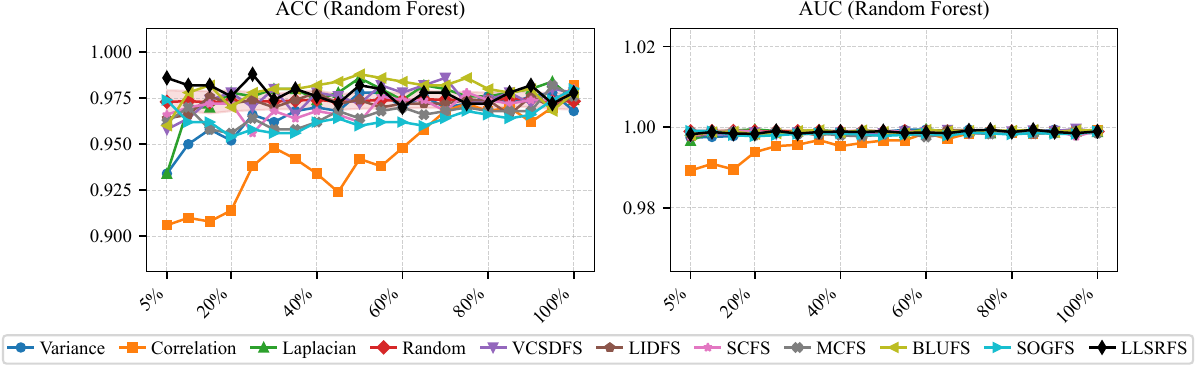}
\caption{ACC and AUC of the random forest classifier on the orlraws10P dataset (5\% to 100\% of the features).}\label{fig:supplot-rf-100-orlraws10P}
\supback
\end{figure}
\begin{figure}[p]
\centering
\includegraphics[width=\textwidth,height=\supfigheight,keepaspectratio]{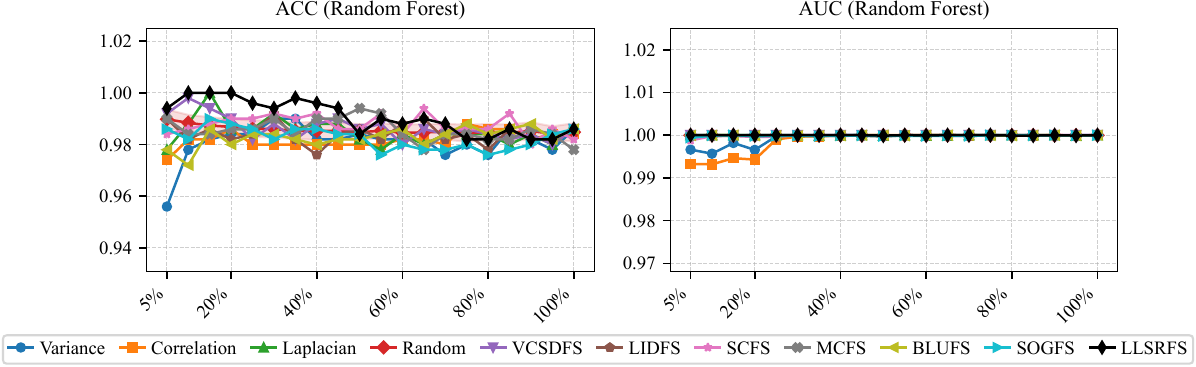}
\caption{ACC and AUC of the random forest classifier on the pixraw10P dataset (5\% to 100\% of the features).}\label{fig:supplot-rf-100-pixraw10P}
\supback
\end{figure}
\begin{figure}[p]
\centering
\includegraphics[width=\textwidth,height=\supfigheight,keepaspectratio]{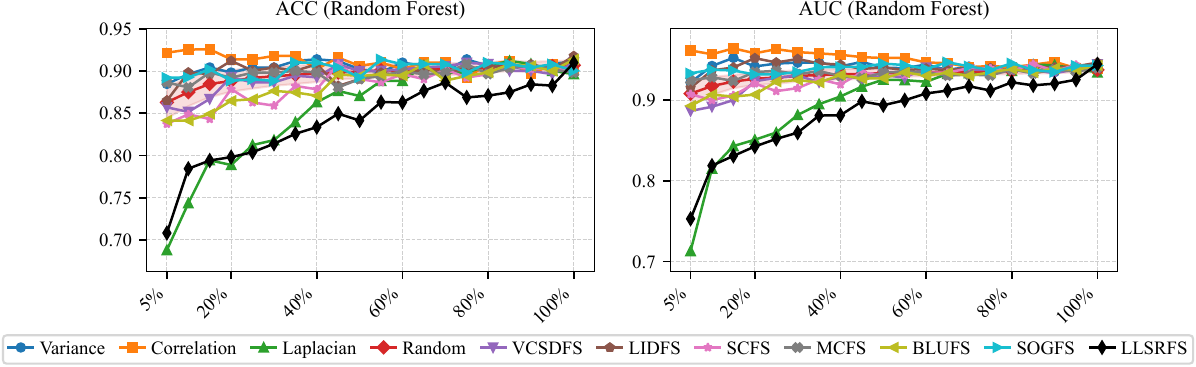}
\caption{ACC and AUC of the random forest classifier on the Prostate\_GE dataset (5\% to 100\% of the features).}\label{fig:supplot-rf-100-Prostate_GE}
\supback
\end{figure}
\begin{figure}[p]
\centering
\includegraphics[width=\textwidth,height=\supfigheight,keepaspectratio]{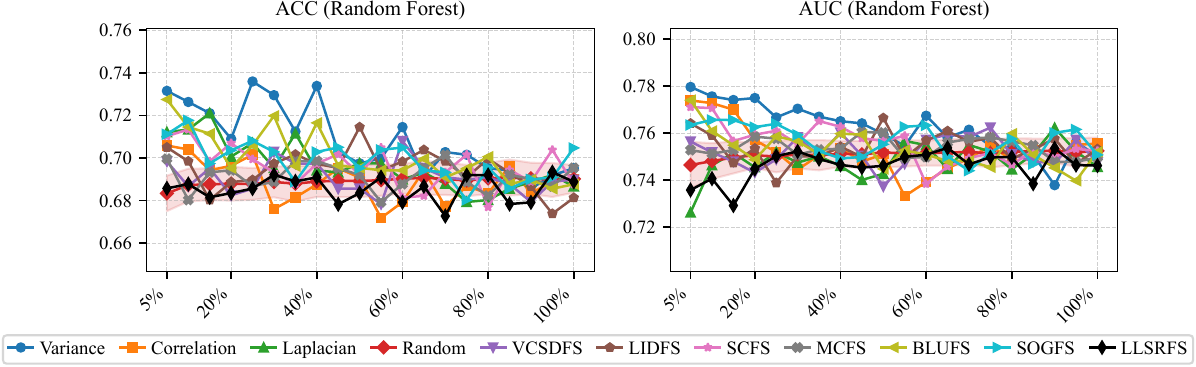}
\caption{ACC and AUC of the random forest classifier on the SMK\_CAN\_187 dataset (5\% to 100\% of the features).}\label{fig:supplot-rf-100-SMK_CAN_187}
\supback
\end{figure}
\begin{figure}[p]
\centering
\includegraphics[width=\textwidth,height=\supfigheight,keepaspectratio]{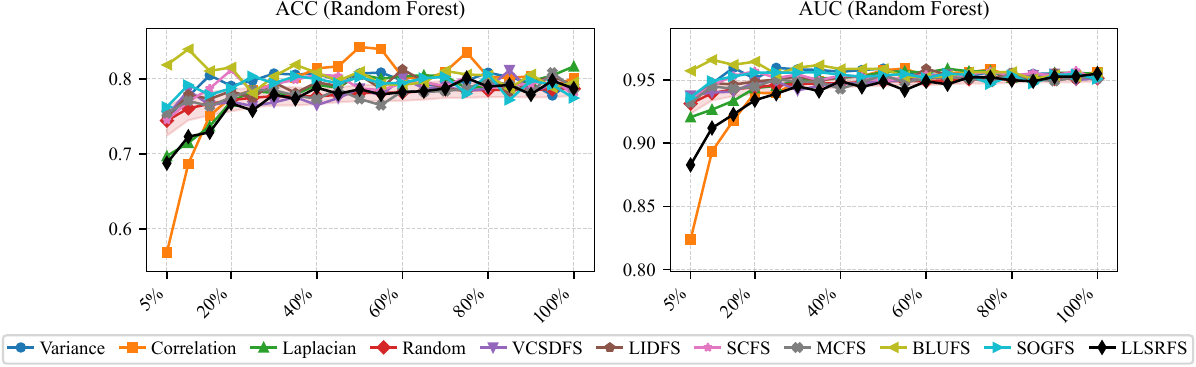}
\caption{ACC and AUC of the random forest classifier on the TOX\_171 dataset (5\% to 100\% of the features).}\label{fig:supplot-rf-100-TOX_171}
\supback
\end{figure}
\begin{figure}[p]
\centering
\includegraphics[width=\textwidth,height=\supfigheight,keepaspectratio]{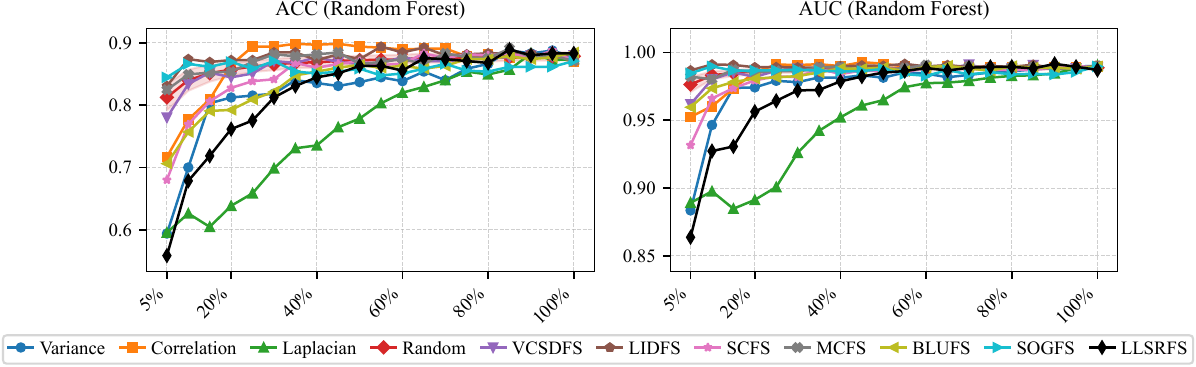}
\caption{ACC and AUC of the random forest classifier on the warpAR10P dataset (5\% to 100\% of the features).}\label{fig:supplot-rf-100-warpAR10P}
\supback
\end{figure}
\begin{figure}[p]
\centering
\includegraphics[width=\textwidth,height=\supfigheight,keepaspectratio]{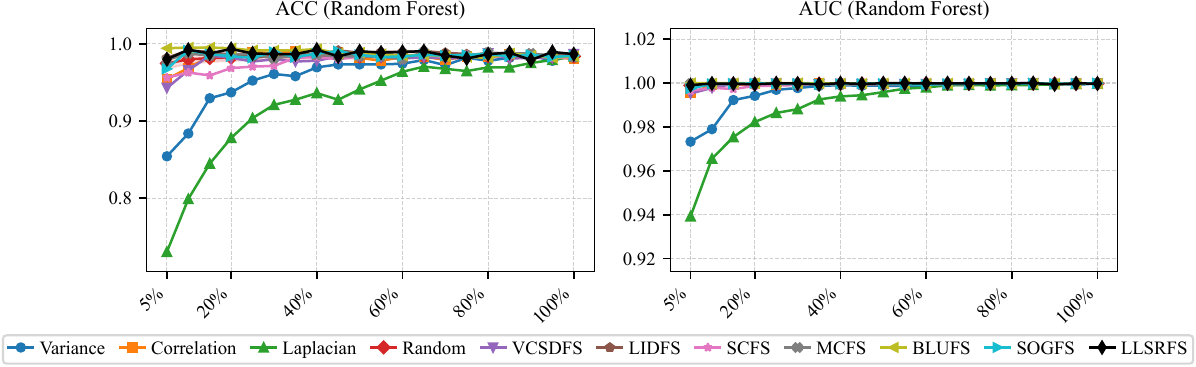}
\caption{ACC and AUC of the random forest classifier on the warpPIE10P dataset (5\% to 100\% of the features).}\label{fig:supplot-rf-100-warpPIE10P}
\supback
\end{figure}
\begin{figure}[p]
\centering
\includegraphics[width=\textwidth,height=\supfigheight,keepaspectratio]{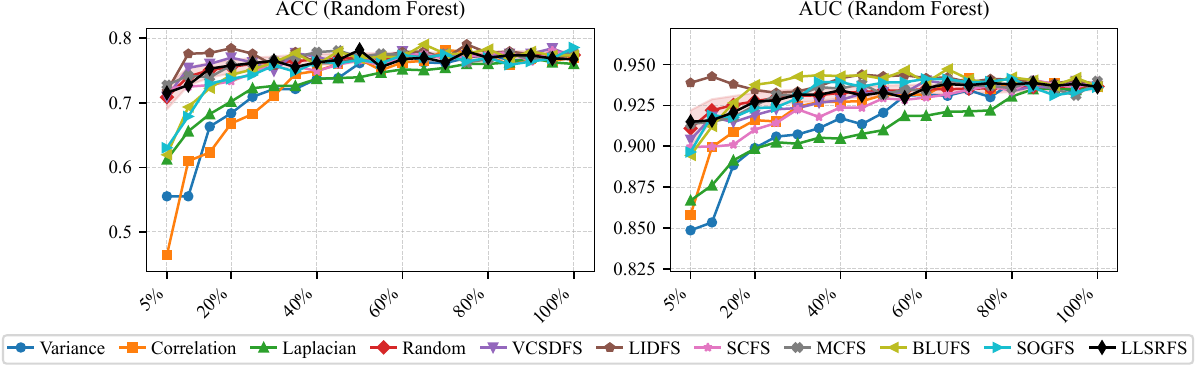}
\caption{ACC and AUC of the random forest classifier on the Yale dataset (5\% to 100\% of the features).}\label{fig:supplot-rf-100-Yale}
\supback
\end{figure}
\clearpage

\begin{figure}[p]
\centering
\pdfbookmark[3]{XGBoost}{supplot-bm-xgb}
\pdfbookmark[4]{0.5\% to 10\%}{supplot-bm-xgb-10}
\includegraphics[width=\textwidth,height=\supfigheight,keepaspectratio]{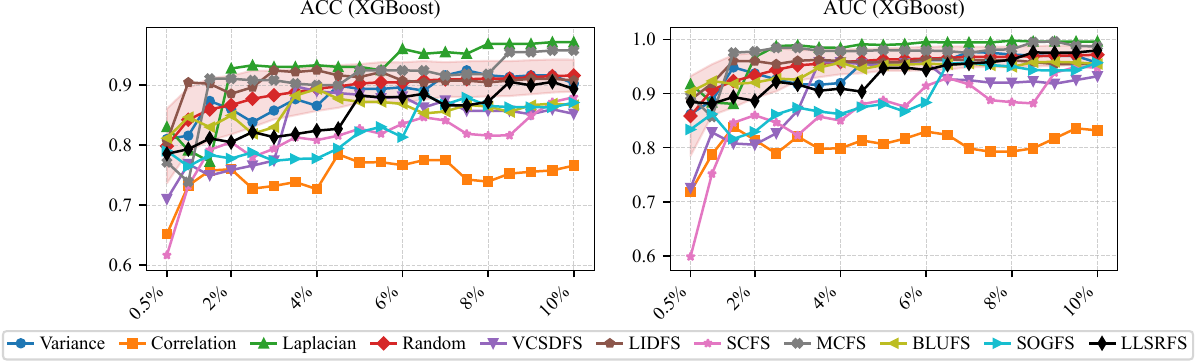}
\caption{ACC and AUC of the XGBoost classifier on the ALLAML dataset (0.5\% to 10\% of the features).}\label{fig:supplot-xgb-10-ALLAML}
\supback
\end{figure}
\begin{figure}[p]
\centering
\includegraphics[width=\textwidth,height=\supfigheight,keepaspectratio]{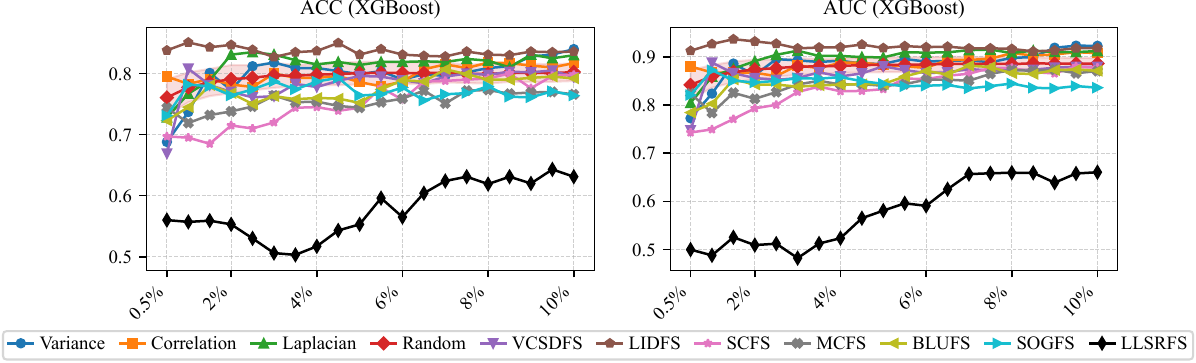}
\caption{ACC and AUC of the XGBoost classifier on the arcene dataset (0.5\% to 10\% of the features).}\label{fig:supplot-xgb-10-arcene}
\supback
\end{figure}
\begin{figure}[p]
\centering
\includegraphics[width=\textwidth,height=\supfigheight,keepaspectratio]{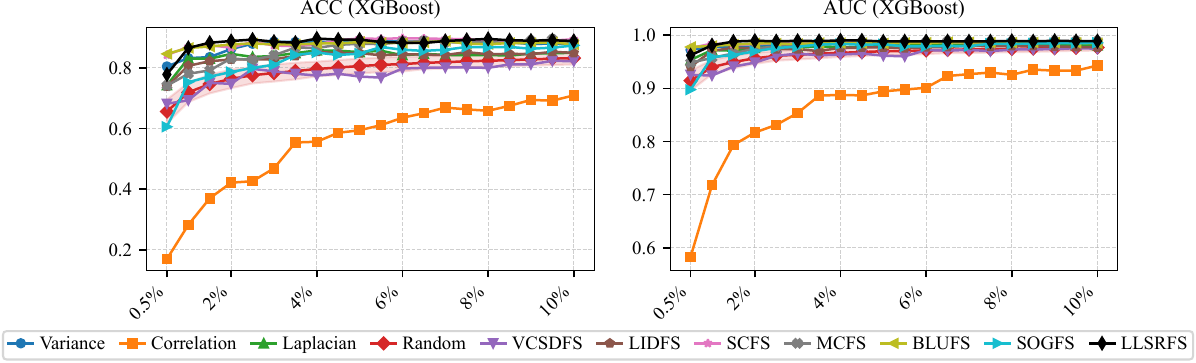}
\caption{ACC and AUC of the XGBoost classifier on the Carcinom dataset (0.5\% to 10\% of the features).}\label{fig:supplot-xgb-10-Carcinom}
\supback
\end{figure}
\begin{figure}[p]
\centering
\includegraphics[width=\textwidth,height=\supfigheight,keepaspectratio]{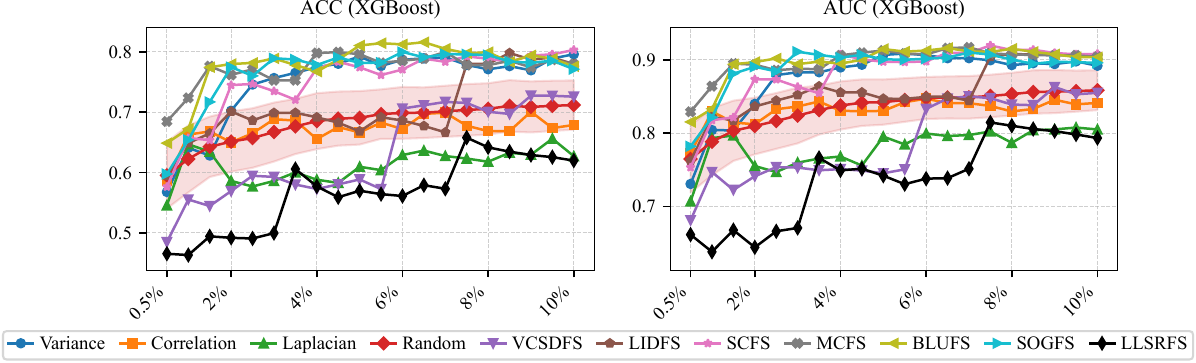}
\caption{ACC and AUC of the XGBoost classifier on the CLL\_SUB\_111 dataset (0.5\% to 10\% of the features).}\label{fig:supplot-xgb-10-CLL_SUB_111}
\supback
\end{figure}
\begin{figure}[p]
\centering
\includegraphics[width=\textwidth,height=\supfigheight,keepaspectratio]{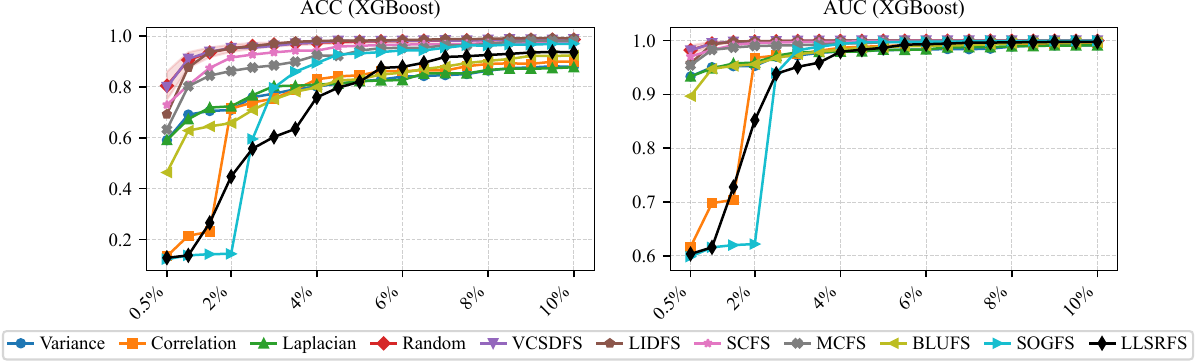}
\caption{ACC and AUC of the XGBoost classifier on the COIL20 dataset (0.5\% to 10\% of the features).}\label{fig:supplot-xgb-10-COIL20}
\supback
\end{figure}
\begin{figure}[p]
\centering
\includegraphics[width=\textwidth,height=\supfigheight,keepaspectratio]{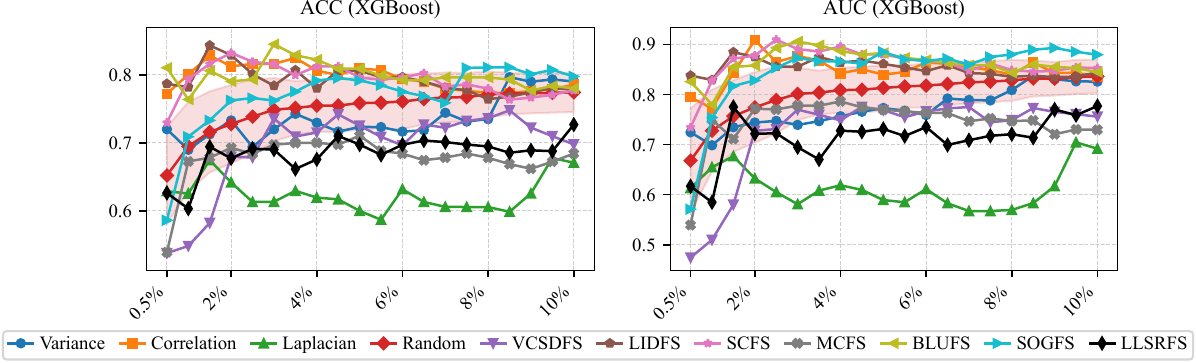}
\caption{ACC and AUC of the XGBoost classifier on the colon dataset (0.5\% to 10\% of the features).}\label{fig:supplot-xgb-10-colon}
\supback
\end{figure}
\begin{figure}[p]
\centering
\includegraphics[width=\textwidth,height=\supfigheight,keepaspectratio]{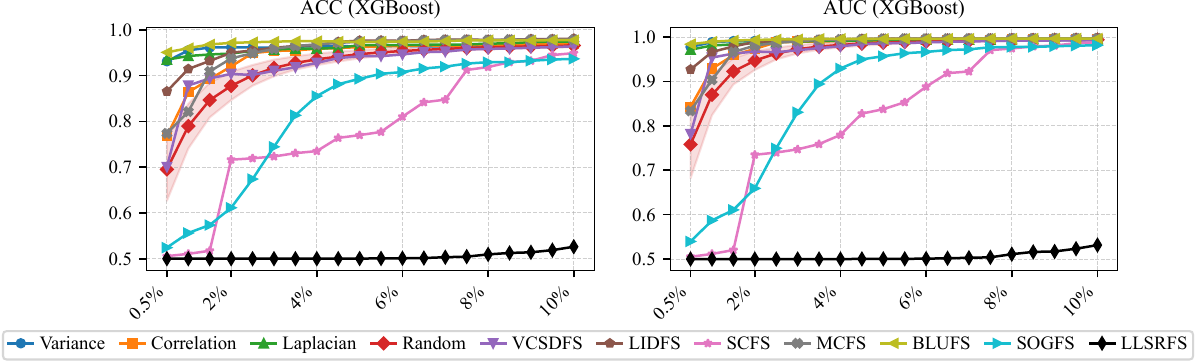}
\caption{ACC and AUC of the XGBoost classifier on the gisette dataset (0.5\% to 10\% of the features).}\label{fig:supplot-xgb-10-gisette}
\supback
\end{figure}
\begin{figure}[p]
\centering
\includegraphics[width=\textwidth,height=\supfigheight,keepaspectratio]{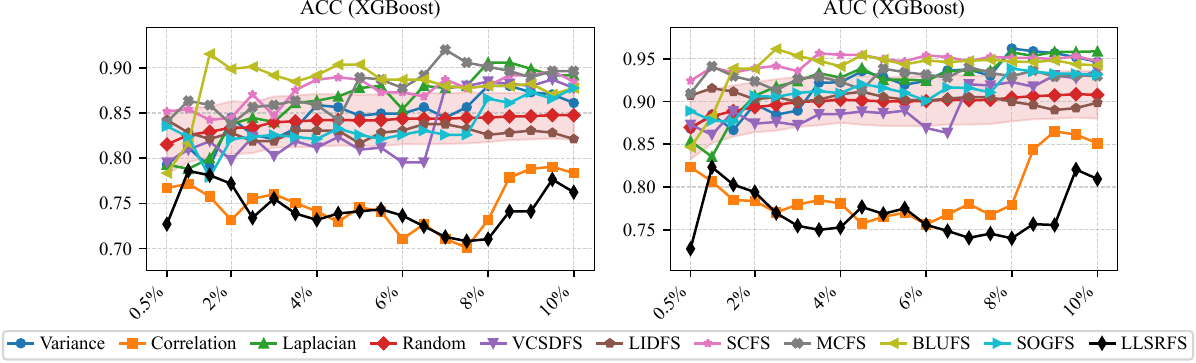}
\caption{ACC and AUC of the XGBoost classifier on the GLI\_85 dataset (0.5\% to 10\% of the features).}\label{fig:supplot-xgb-10-GLI_85}
\supback
\end{figure}
\begin{figure}[p]
\centering
\includegraphics[width=\textwidth,height=\supfigheight,keepaspectratio]{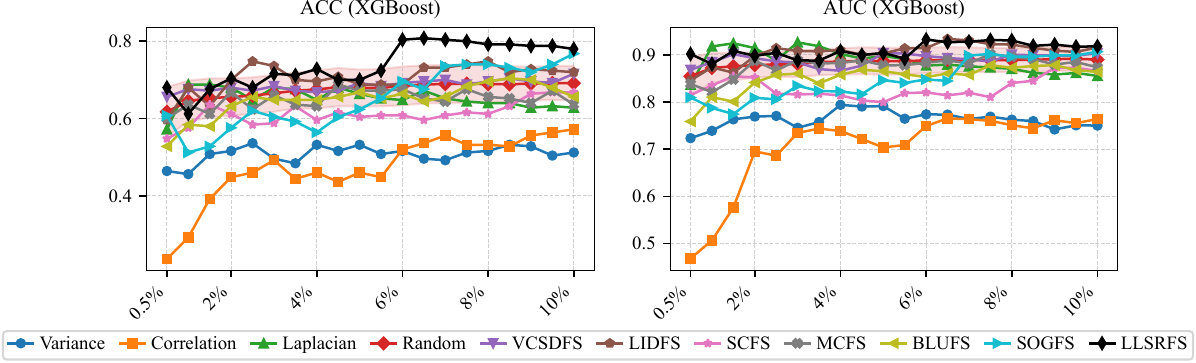}
\caption{ACC and AUC of the XGBoost classifier on the GLIOMA dataset (0.5\% to 10\% of the features).}\label{fig:supplot-xgb-10-GLIOMA}
\supback
\end{figure}
\begin{figure}[p]
\centering
\includegraphics[width=\textwidth,height=\supfigheight,keepaspectratio]{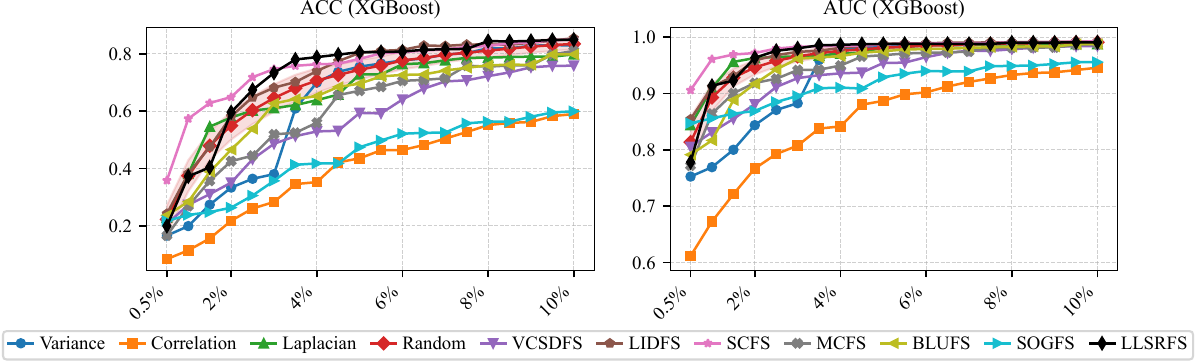}
\caption{ACC and AUC of the XGBoost classifier on the Isolet dataset (0.5\% to 10\% of the features).}\label{fig:supplot-xgb-10-Isolet}
\supback
\end{figure}
\begin{figure}[p]
\centering
\includegraphics[width=\textwidth,height=\supfigheight,keepaspectratio]{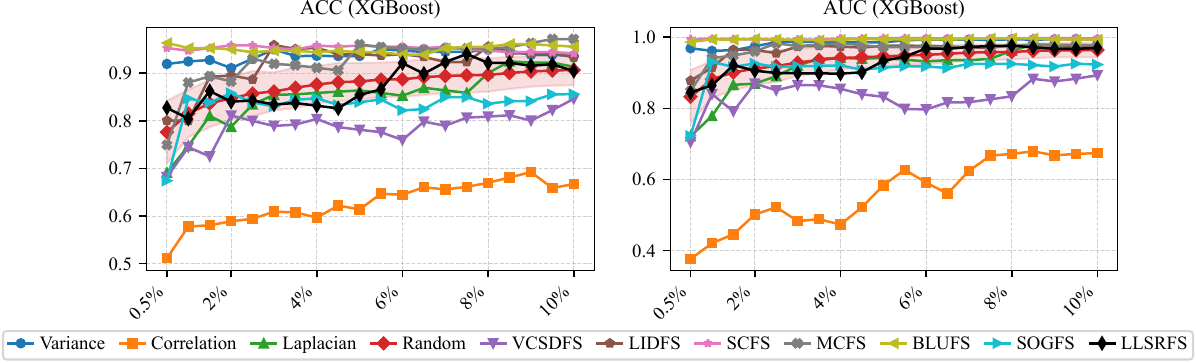}
\caption{ACC and AUC of the XGBoost classifier on the leukemia dataset (0.5\% to 10\% of the features).}\label{fig:supplot-xgb-10-leukemia}
\supback
\end{figure}
\begin{figure}[p]
\centering
\includegraphics[width=\textwidth,height=\supfigheight,keepaspectratio]{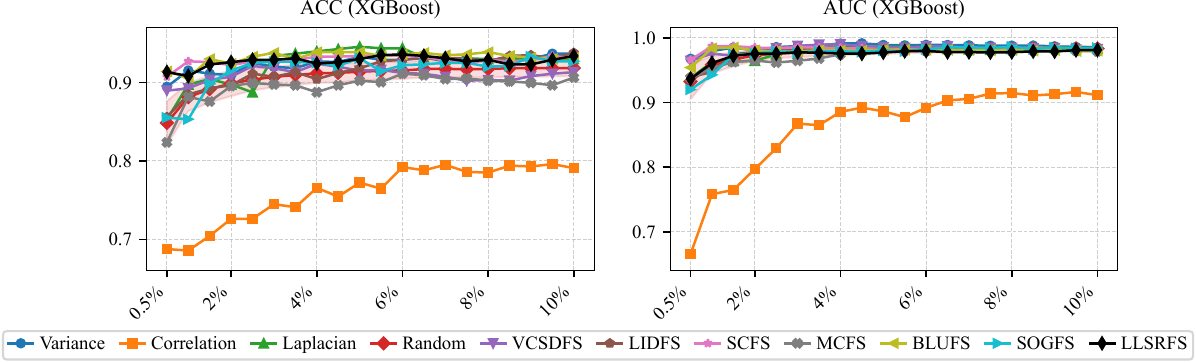}
\caption{ACC and AUC of the XGBoost classifier on the lung dataset (0.5\% to 10\% of the features).}\label{fig:supplot-xgb-10-lung}
\supback
\end{figure}
\begin{figure}[p]
\centering
\includegraphics[width=\textwidth,height=\supfigheight,keepaspectratio]{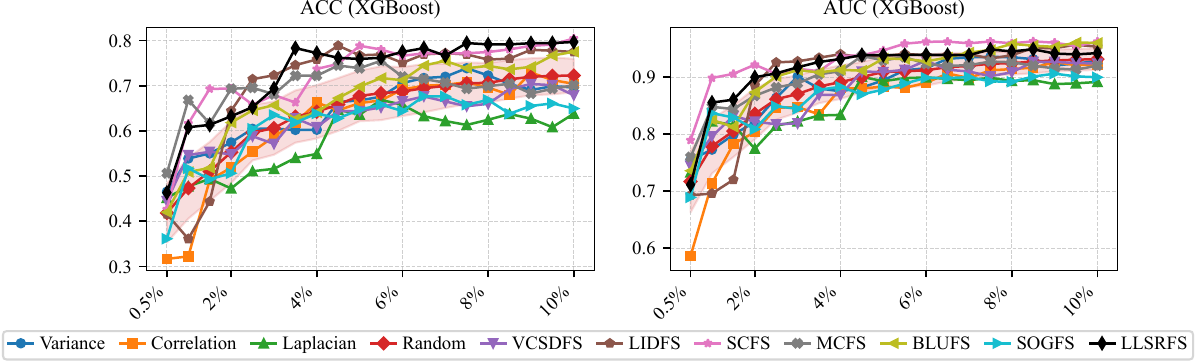}
\caption{ACC and AUC of the XGBoost classifier on the lung\_discrete dataset (0.5\% to 10\% of the features).}\label{fig:supplot-xgb-10-lung_discrete}
\supback
\end{figure}
\begin{figure}[p]
\centering
\includegraphics[width=\textwidth,height=\supfigheight,keepaspectratio]{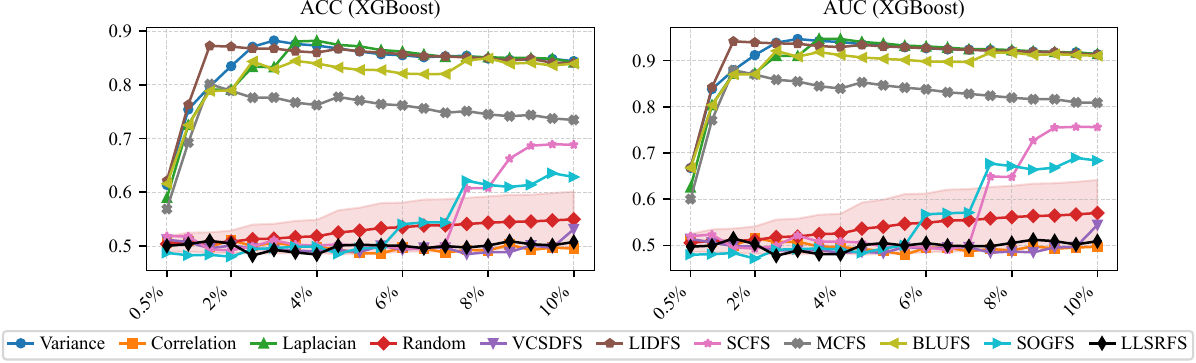}
\caption{ACC and AUC of the XGBoost classifier on the madelon dataset (0.5\% to 10\% of the features).}\label{fig:supplot-xgb-10-madelon}
\supback
\end{figure}
\begin{figure}[p]
\centering
\includegraphics[width=\textwidth,height=\supfigheight,keepaspectratio]{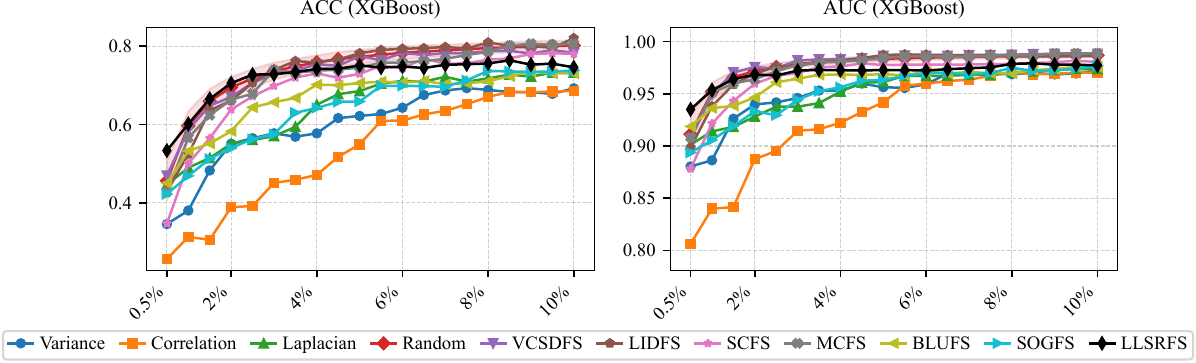}
\caption{ACC and AUC of the XGBoost classifier on the ORL dataset (0.5\% to 10\% of the features).}\label{fig:supplot-xgb-10-ORL}
\supback
\end{figure}
\begin{figure}[p]
\centering
\includegraphics[width=\textwidth,height=\supfigheight,keepaspectratio]{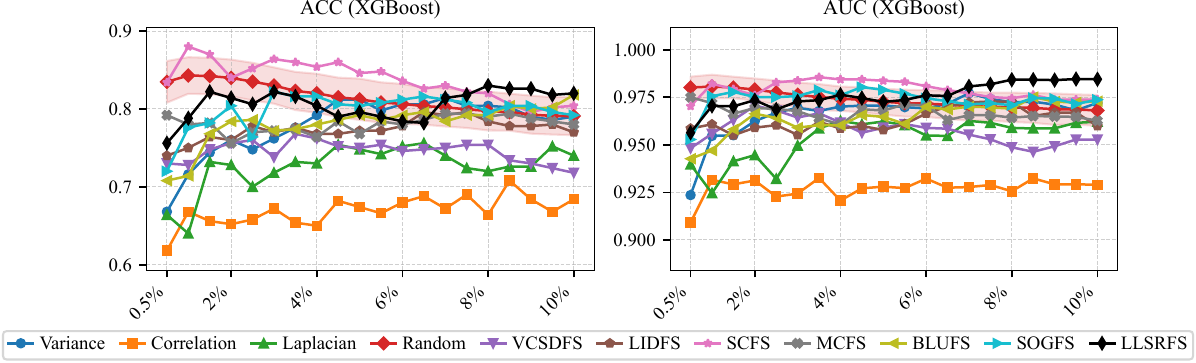}
\caption{ACC and AUC of the XGBoost classifier on the orlraws10P dataset (0.5\% to 10\% of the features).}\label{fig:supplot-xgb-10-orlraws10P}
\supback
\end{figure}
\begin{figure}[p]
\centering
\includegraphics[width=\textwidth,height=\supfigheight,keepaspectratio]{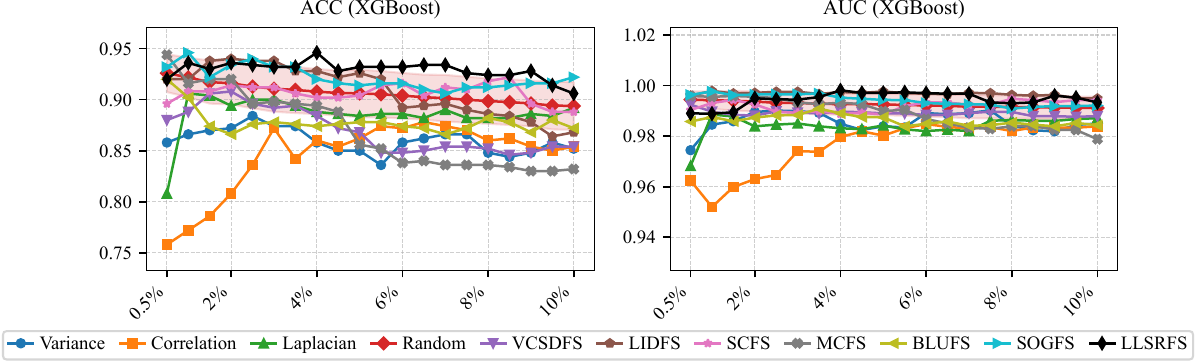}
\caption{ACC and AUC of the XGBoost classifier on the pixraw10P dataset (0.5\% to 10\% of the features).}\label{fig:supplot-xgb-10-pixraw10P}
\supback
\end{figure}
\begin{figure}[p]
\centering
\includegraphics[width=\textwidth,height=\supfigheight,keepaspectratio]{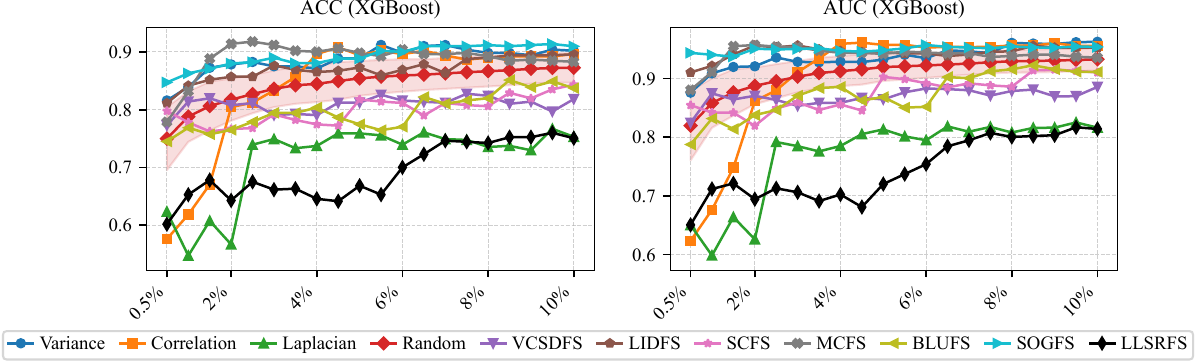}
\caption{ACC and AUC of the XGBoost classifier on the Prostate\_GE dataset (0.5\% to 10\% of the features).}\label{fig:supplot-xgb-10-Prostate_GE}
\supback
\end{figure}
\begin{figure}[p]
\centering
\includegraphics[width=\textwidth,height=\supfigheight,keepaspectratio]{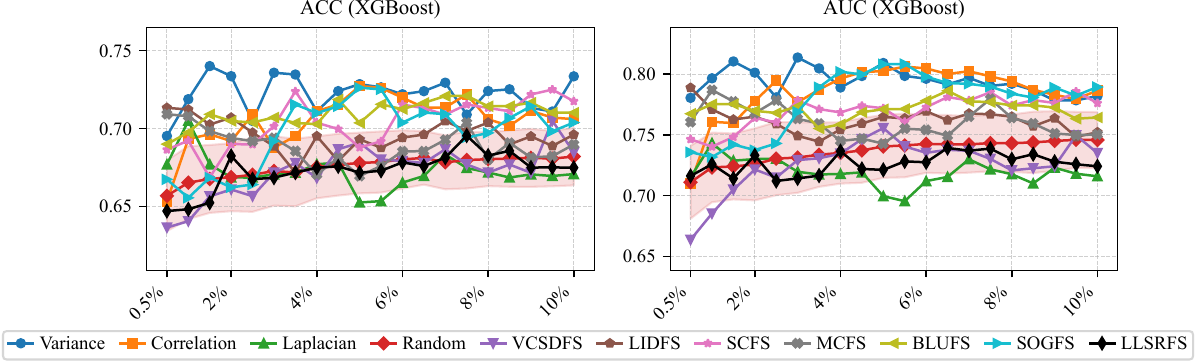}
\caption{ACC and AUC of the XGBoost classifier on the SMK\_CAN\_187 dataset (0.5\% to 10\% of the features).}\label{fig:supplot-xgb-10-SMK_CAN_187}
\supback
\end{figure}
\begin{figure}[p]
\centering
\includegraphics[width=\textwidth,height=\supfigheight,keepaspectratio]{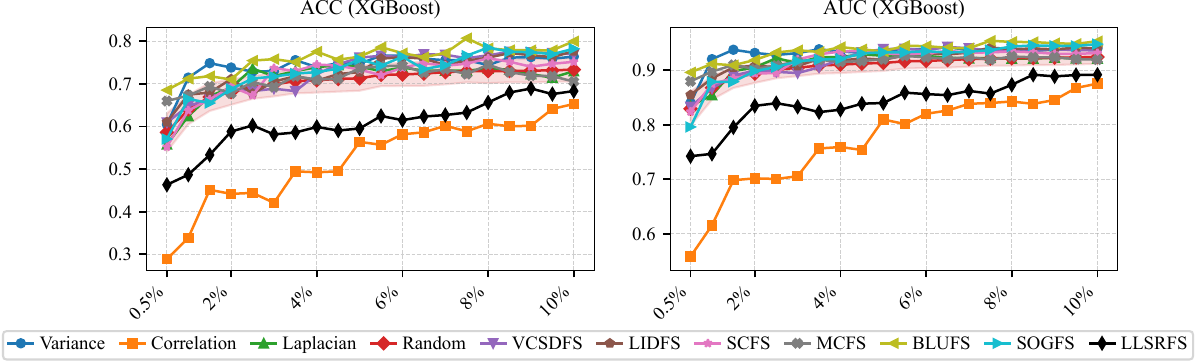}
\caption{ACC and AUC of the XGBoost classifier on the TOX\_171 dataset (0.5\% to 10\% of the features).}\label{fig:supplot-xgb-10-TOX_171}
\supback
\end{figure}
\begin{figure}[p]
\centering
\includegraphics[width=\textwidth,height=\supfigheight,keepaspectratio]{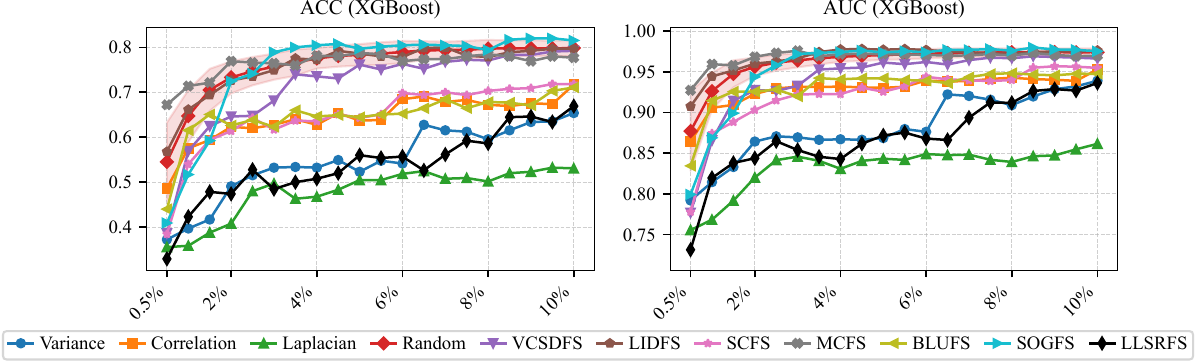}
\caption{ACC and AUC of the XGBoost classifier on the warpAR10P dataset (0.5\% to 10\% of the features).}\label{fig:supplot-xgb-10-warpAR10P}
\supback
\end{figure}
\begin{figure}[p]
\centering
\includegraphics[width=\textwidth,height=\supfigheight,keepaspectratio]{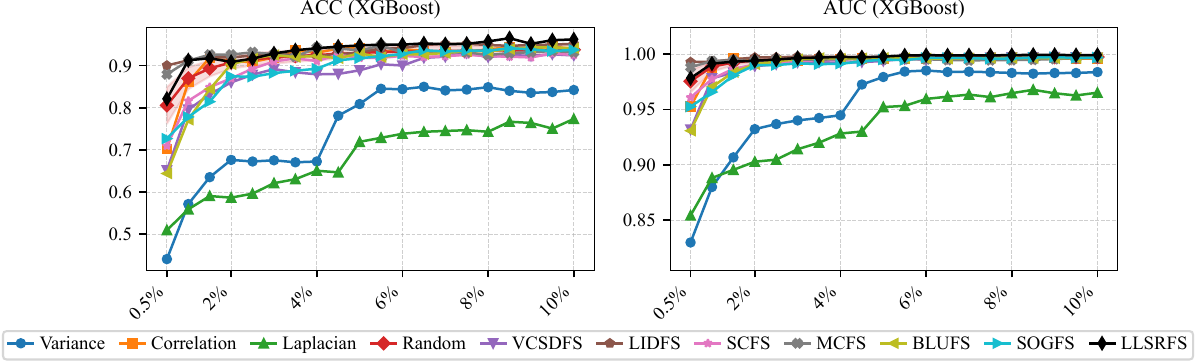}
\caption{ACC and AUC of the XGBoost classifier on the warpPIE10P dataset (0.5\% to 10\% of the features).}\label{fig:supplot-xgb-10-warpPIE10P}
\supback
\end{figure}
\begin{figure}[p]
\centering
\includegraphics[width=\textwidth,height=\supfigheight,keepaspectratio]{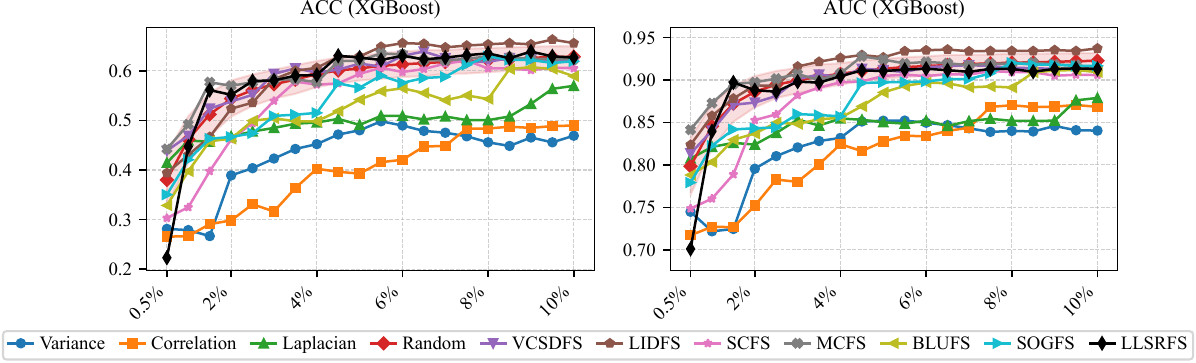}
\caption{ACC and AUC of the XGBoost classifier on the Yale dataset (0.5\% to 10\% of the features).}\label{fig:supplot-xgb-10-Yale}
\supback
\end{figure}
\begin{figure}[p]
\centering
\pdfbookmark[4]{5\% to 100\%}{supplot-bm-xgb-100}
\includegraphics[width=\textwidth,height=\supfigheight,keepaspectratio]{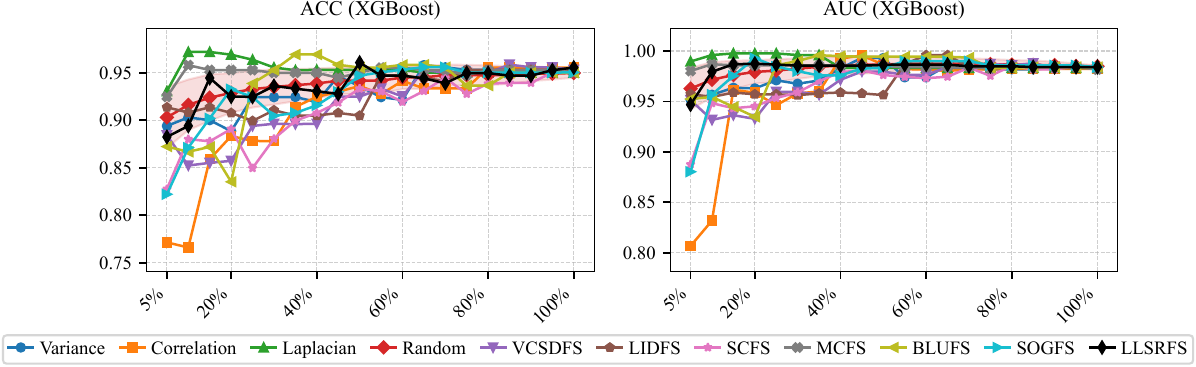}
\caption{ACC and AUC of the XGBoost classifier on the ALLAML dataset (5\% to 100\% of the features).}\label{fig:supplot-xgb-100-ALLAML}
\supback
\end{figure}
\begin{figure}[p]
\centering
\includegraphics[width=\textwidth,height=\supfigheight,keepaspectratio]{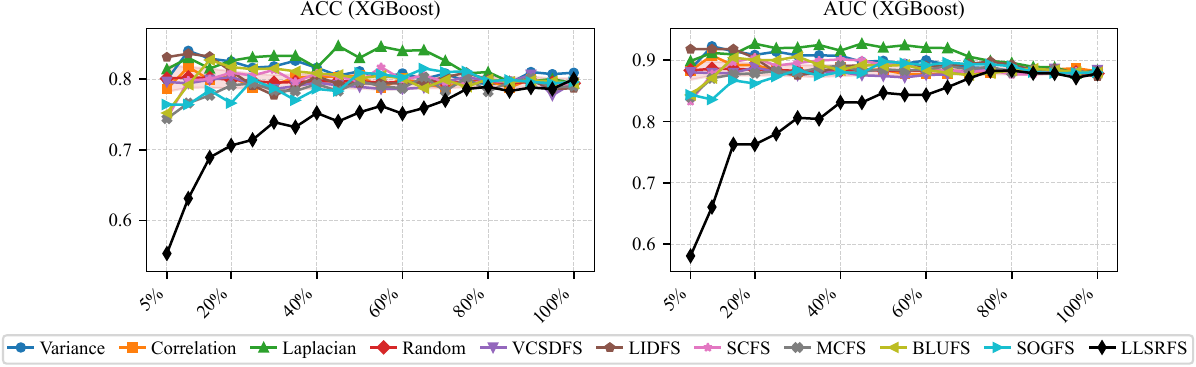}
\caption{ACC and AUC of the XGBoost classifier on the arcene dataset (5\% to 100\% of the features).}\label{fig:supplot-xgb-100-arcene}
\supback
\end{figure}
\begin{figure}[p]
\centering
\includegraphics[width=\textwidth,height=\supfigheight,keepaspectratio]{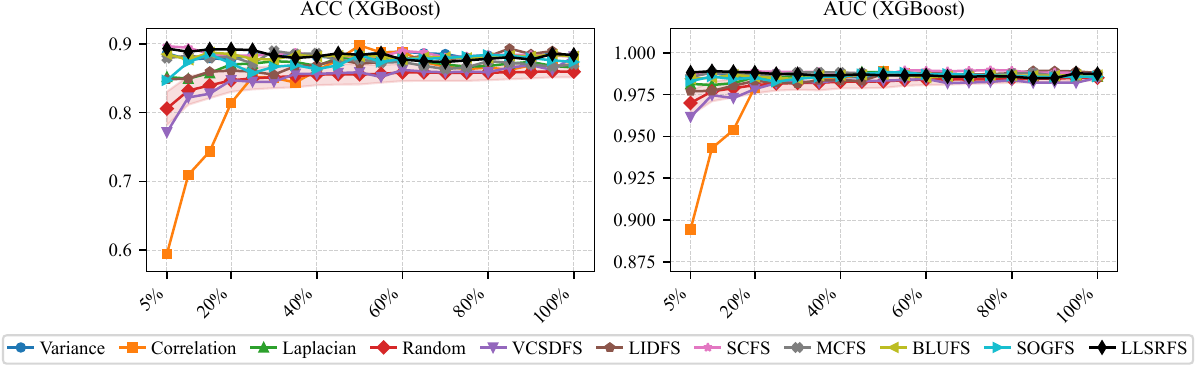}
\caption{ACC and AUC of the XGBoost classifier on the Carcinom dataset (5\% to 100\% of the features).}\label{fig:supplot-xgb-100-Carcinom}
\supback
\end{figure}
\begin{figure}[p]
\centering
\includegraphics[width=\textwidth,height=\supfigheight,keepaspectratio]{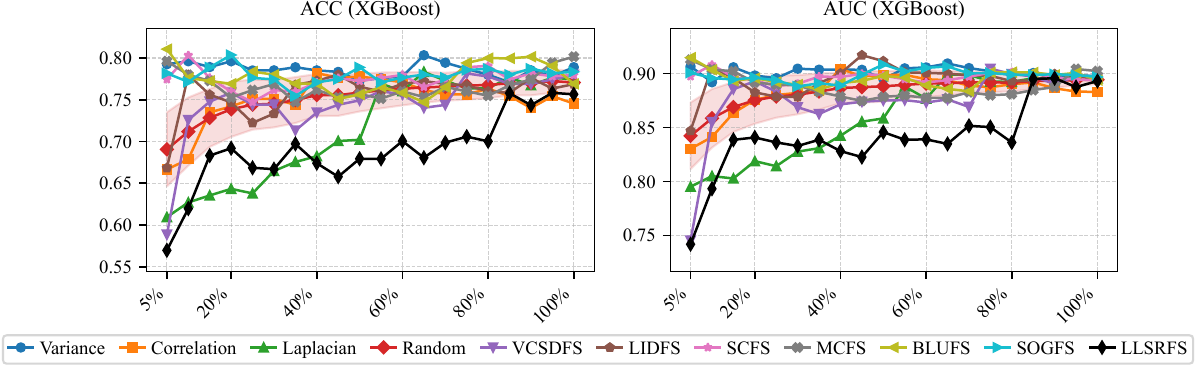}
\caption{ACC and AUC of the XGBoost classifier on the CLL\_SUB\_111 dataset (5\% to 100\% of the features).}\label{fig:supplot-xgb-100-CLL_SUB_111}
\supback
\end{figure}
\begin{figure}[p]
\centering
\includegraphics[width=\textwidth,height=\supfigheight,keepaspectratio]{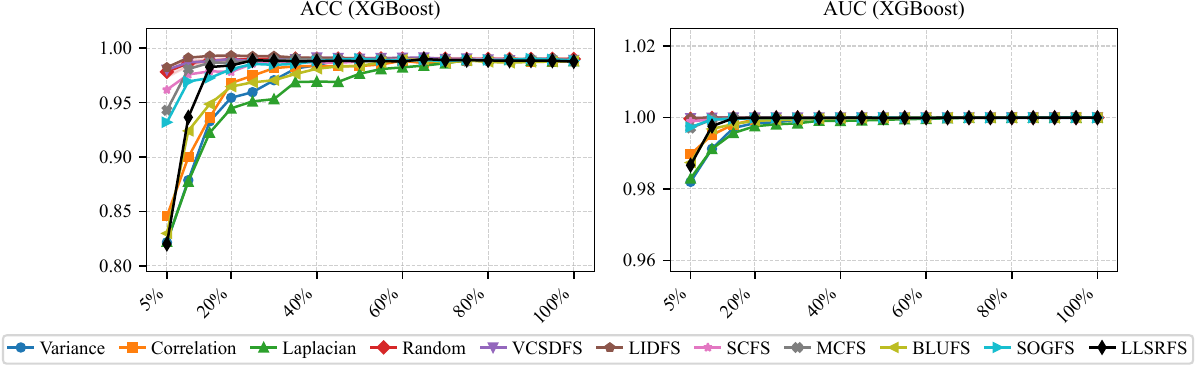}
\caption{ACC and AUC of the XGBoost classifier on the COIL20 dataset (5\% to 100\% of the features).}\label{fig:supplot-xgb-100-COIL20}
\supback
\end{figure}
\begin{figure}[p]
\centering
\includegraphics[width=\textwidth,height=\supfigheight,keepaspectratio]{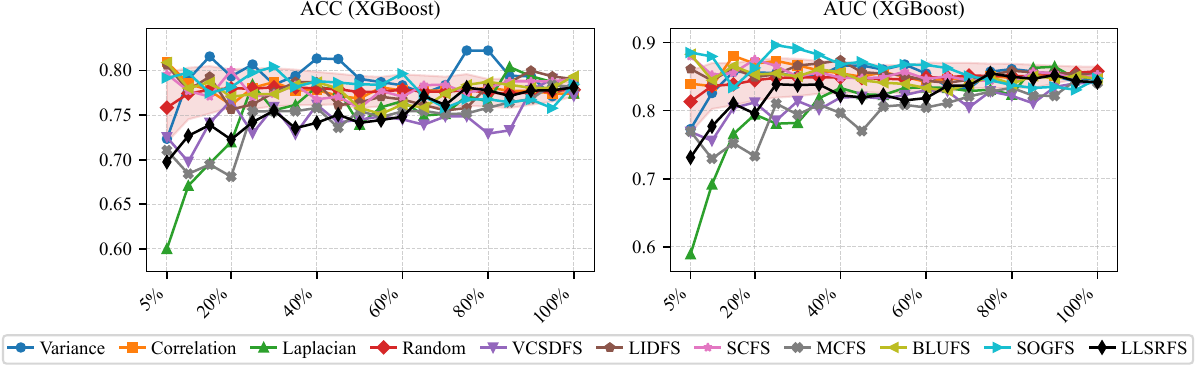}
\caption{ACC and AUC of the XGBoost classifier on the colon dataset (5\% to 100\% of the features).}\label{fig:supplot-xgb-100-colon}
\supback
\end{figure}
\begin{figure}[p]
\centering
\includegraphics[width=\textwidth,height=\supfigheight,keepaspectratio]{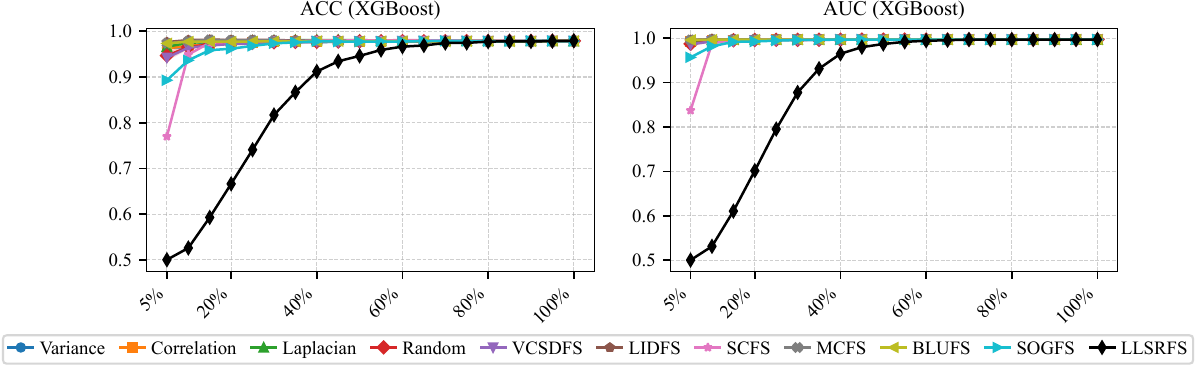}
\caption{ACC and AUC of the XGBoost classifier on the gisette dataset (5\% to 100\% of the features).}\label{fig:supplot-xgb-100-gisette}
\supback
\end{figure}
\begin{figure}[p]
\centering
\includegraphics[width=\textwidth,height=\supfigheight,keepaspectratio]{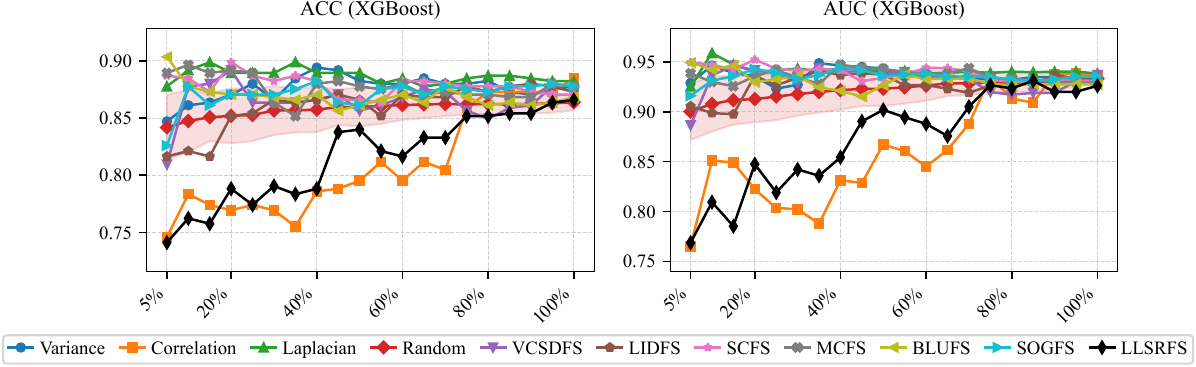}
\caption{ACC and AUC of the XGBoost classifier on the GLI\_85 dataset (5\% to 100\% of the features).}\label{fig:supplot-xgb-100-GLI_85}
\supback
\end{figure}
\begin{figure}[p]
\centering
\includegraphics[width=\textwidth,height=\supfigheight,keepaspectratio]{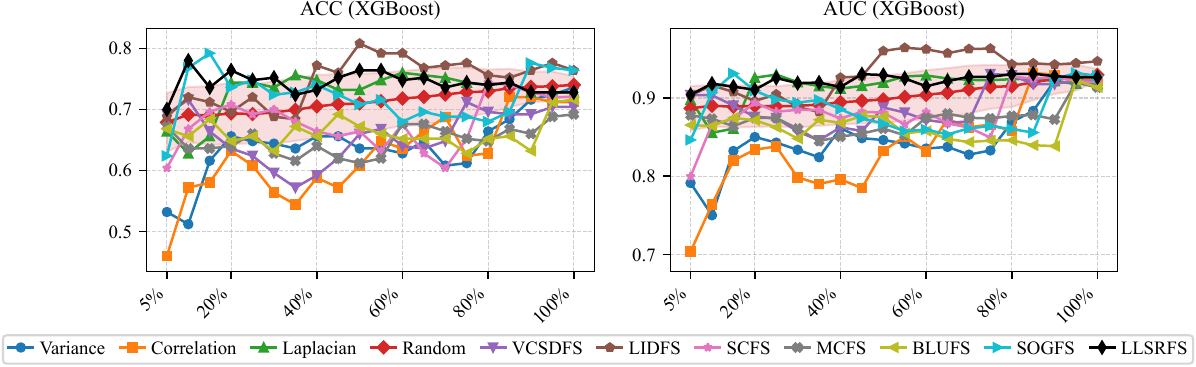}
\caption{ACC and AUC of the XGBoost classifier on the GLIOMA dataset (5\% to 100\% of the features).}\label{fig:supplot-xgb-100-GLIOMA}
\supback
\end{figure}
\begin{figure}[p]
\centering
\includegraphics[width=\textwidth,height=\supfigheight,keepaspectratio]{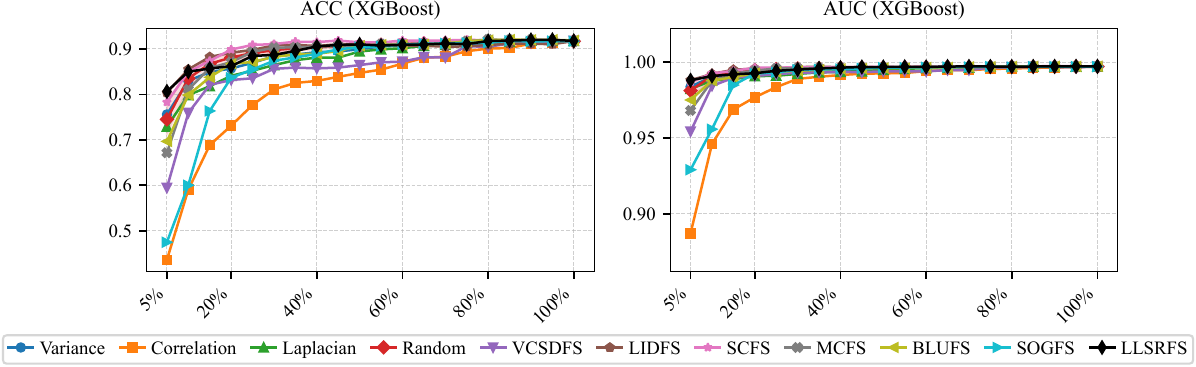}
\caption{ACC and AUC of the XGBoost classifier on the Isolet dataset (5\% to 100\% of the features).}\label{fig:supplot-xgb-100-Isolet}
\supback
\end{figure}
\begin{figure}[p]
\centering
\includegraphics[width=\textwidth,height=\supfigheight,keepaspectratio]{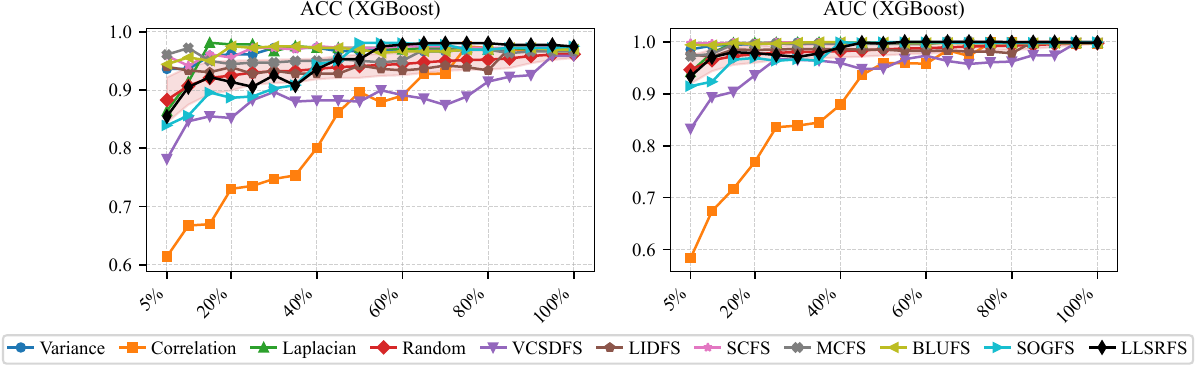}
\caption{ACC and AUC of the XGBoost classifier on the leukemia dataset (5\% to 100\% of the features).}\label{fig:supplot-xgb-100-leukemia}
\supback
\end{figure}
\begin{figure}[p]
\centering
\includegraphics[width=\textwidth,height=\supfigheight,keepaspectratio]{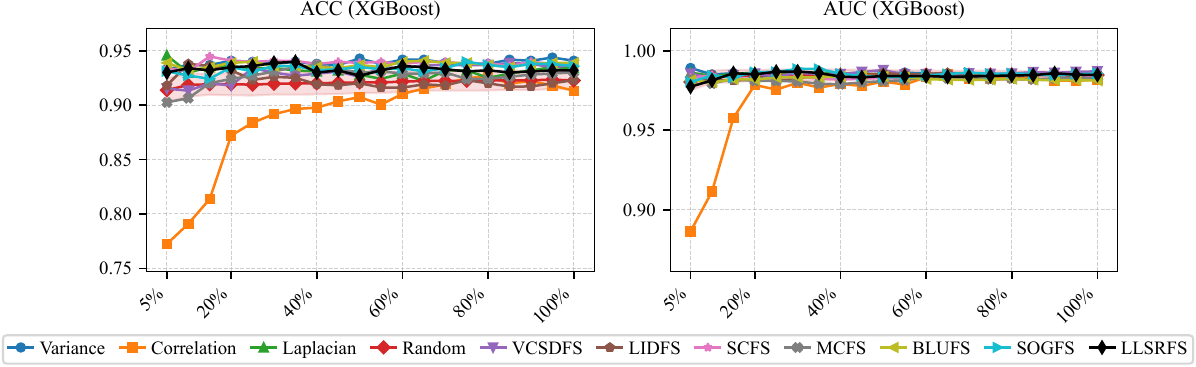}
\caption{ACC and AUC of the XGBoost classifier on the lung dataset (5\% to 100\% of the features).}\label{fig:supplot-xgb-100-lung}
\supback
\end{figure}
\begin{figure}[p]
\centering
\includegraphics[width=\textwidth,height=\supfigheight,keepaspectratio]{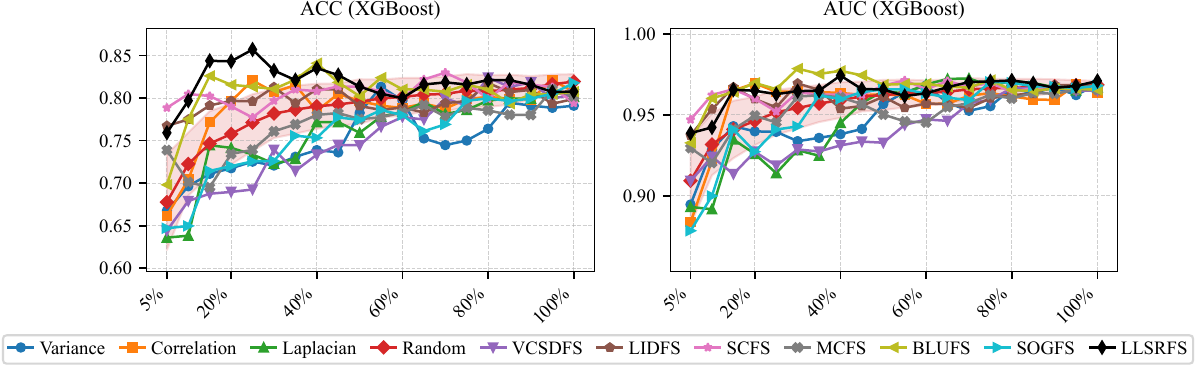}
\caption{ACC and AUC of the XGBoost classifier on the lung\_discrete dataset (5\% to 100\% of the features).}\label{fig:supplot-xgb-100-lung_discrete}
\supback
\end{figure}
\begin{figure}[p]
\centering
\includegraphics[width=\textwidth,height=\supfigheight,keepaspectratio]{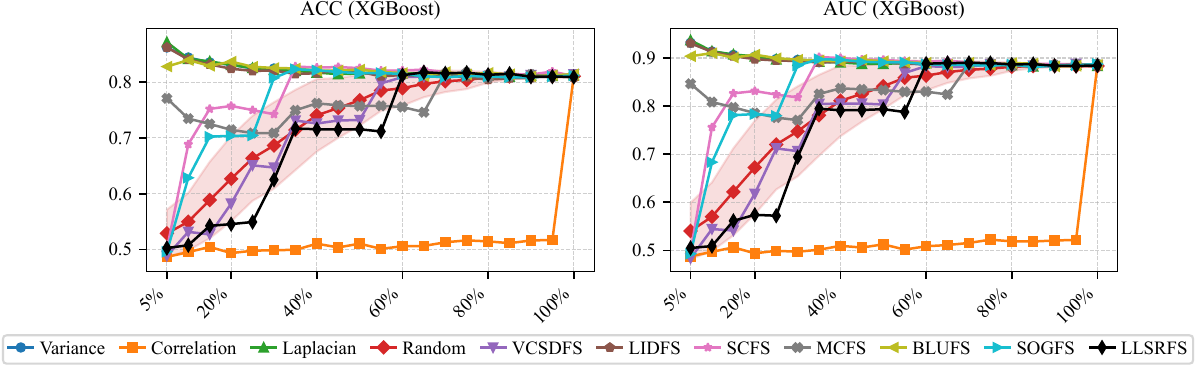}
\caption{ACC and AUC of the XGBoost classifier on the madelon dataset (5\% to 100\% of the features).}\label{fig:supplot-xgb-100-madelon}
\supback
\end{figure}
\begin{figure}[p]
\centering
\includegraphics[width=\textwidth,height=\supfigheight,keepaspectratio]{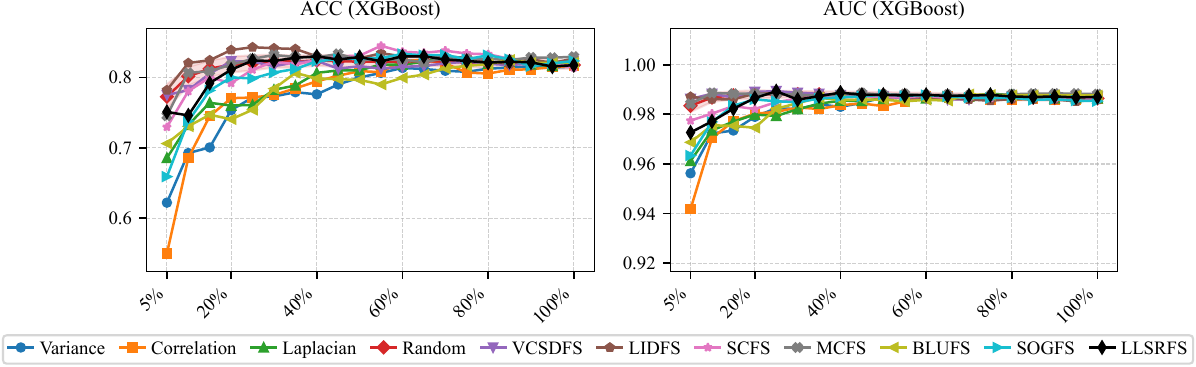}
\caption{ACC and AUC of the XGBoost classifier on the ORL dataset (5\% to 100\% of the features).}\label{fig:supplot-xgb-100-ORL}
\supback
\end{figure}
\begin{figure}[p]
\centering
\includegraphics[width=\textwidth,height=\supfigheight,keepaspectratio]{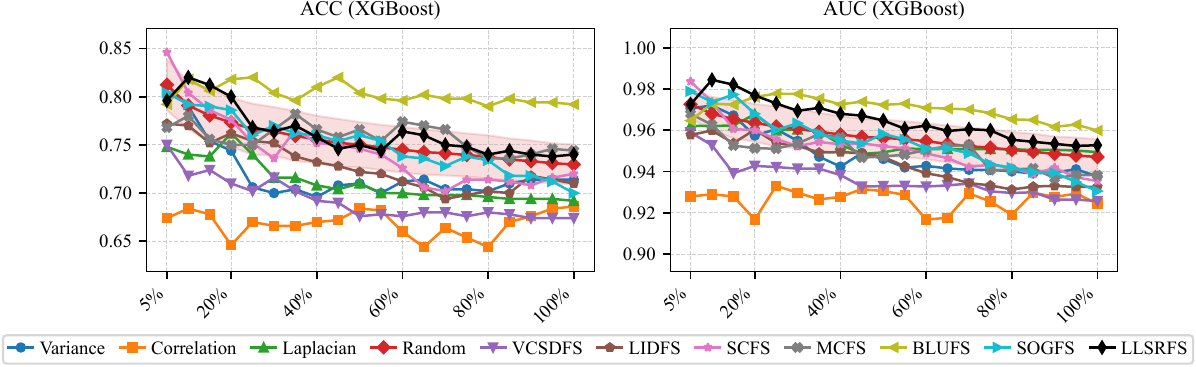}
\caption{ACC and AUC of the XGBoost classifier on the orlraws10P dataset (5\% to 100\% of the features).}\label{fig:supplot-xgb-100-orlraws10P}
\supback
\end{figure}
\begin{figure}[p]
\centering
\includegraphics[width=\textwidth,height=\supfigheight,keepaspectratio]{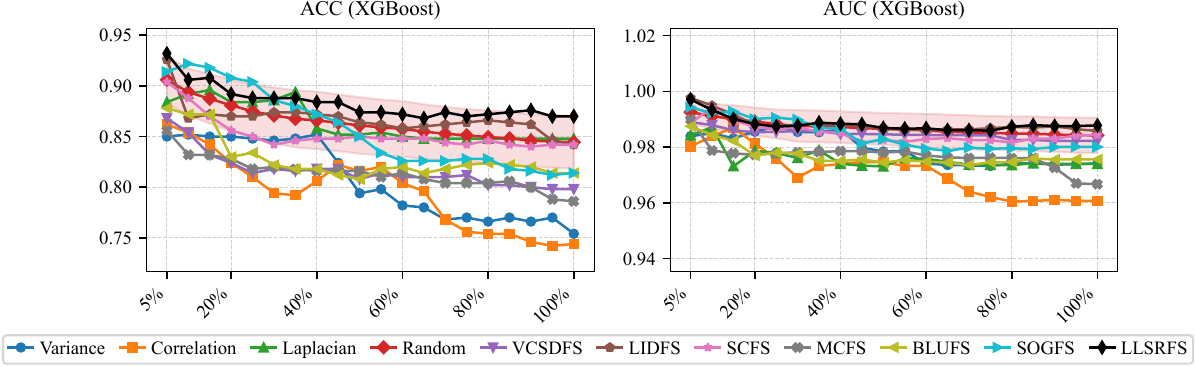}
\caption{ACC and AUC of the XGBoost classifier on the pixraw10P dataset (5\% to 100\% of the features).}\label{fig:supplot-xgb-100-pixraw10P}
\supback
\end{figure}
\begin{figure}[p]
\centering
\includegraphics[width=\textwidth,height=\supfigheight,keepaspectratio]{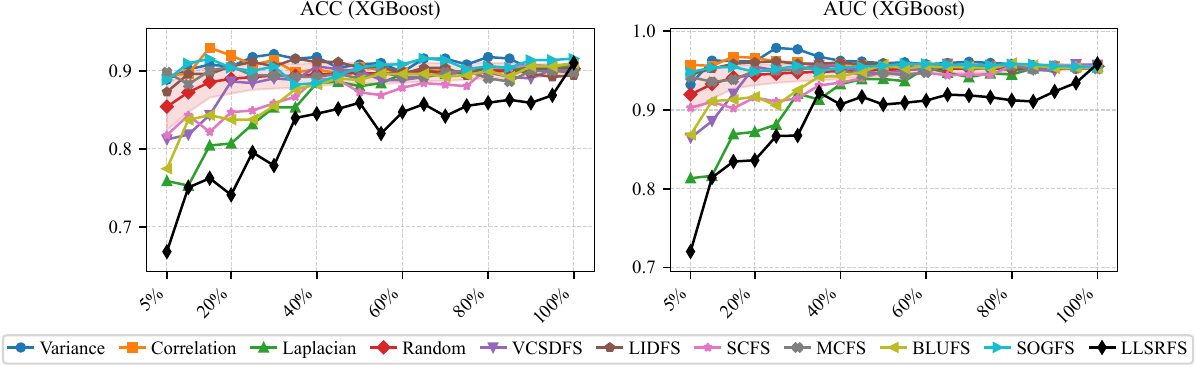}
\caption{ACC and AUC of the XGBoost classifier on the Prostate\_GE dataset (5\% to 100\% of the features).}\label{fig:supplot-xgb-100-Prostate_GE}
\supback
\end{figure}
\begin{figure}[p]
\centering
\includegraphics[width=\textwidth,height=\supfigheight,keepaspectratio]{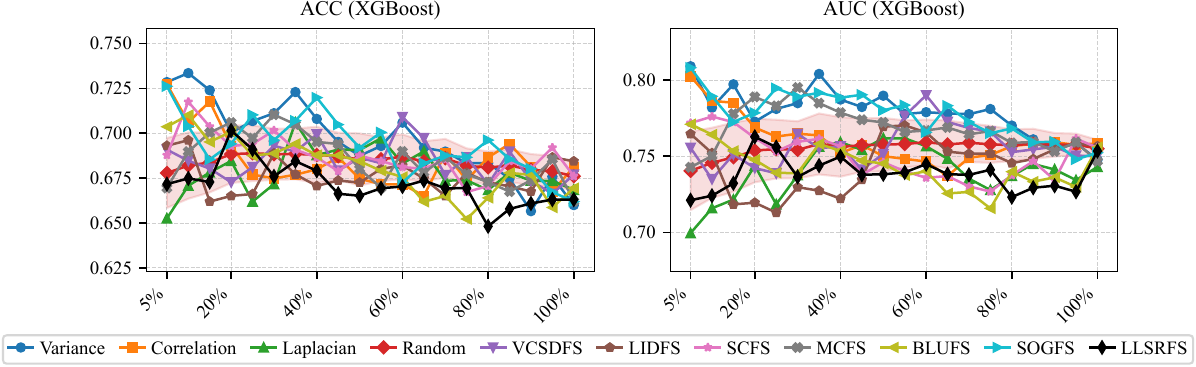}
\caption{ACC and AUC of the XGBoost classifier on the SMK\_CAN\_187 dataset (5\% to 100\% of the features).}\label{fig:supplot-xgb-100-SMK_CAN_187}
\supback
\end{figure}
\begin{figure}[p]
\centering
\includegraphics[width=\textwidth,height=\supfigheight,keepaspectratio]{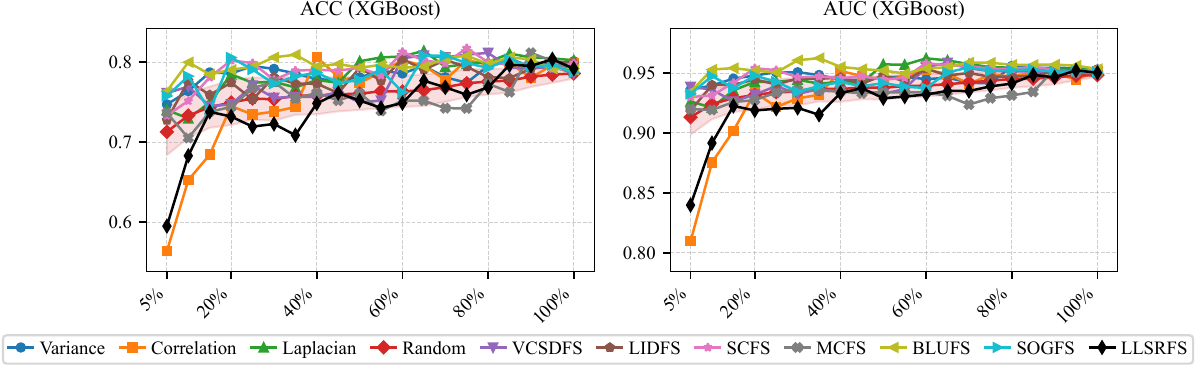}
\caption{ACC and AUC of the XGBoost classifier on the TOX\_171 dataset (5\% to 100\% of the features).}\label{fig:supplot-xgb-100-TOX_171}
\supback
\end{figure}
\begin{figure}[p]
\centering
\includegraphics[width=\textwidth,height=\supfigheight,keepaspectratio]{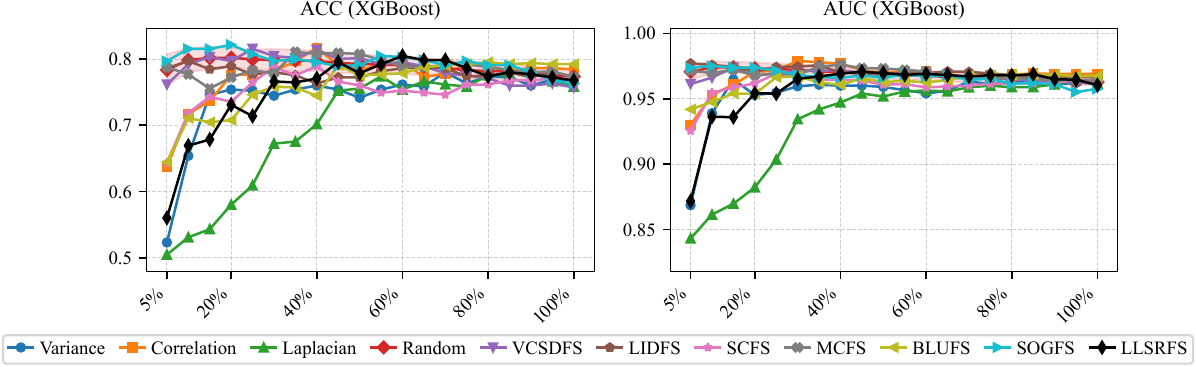}
\caption{ACC and AUC of the XGBoost classifier on the warpAR10P dataset (5\% to 100\% of the features).}\label{fig:supplot-xgb-100-warpAR10P}
\supback
\end{figure}
\begin{figure}[p]
\centering
\includegraphics[width=\textwidth,height=\supfigheight,keepaspectratio]{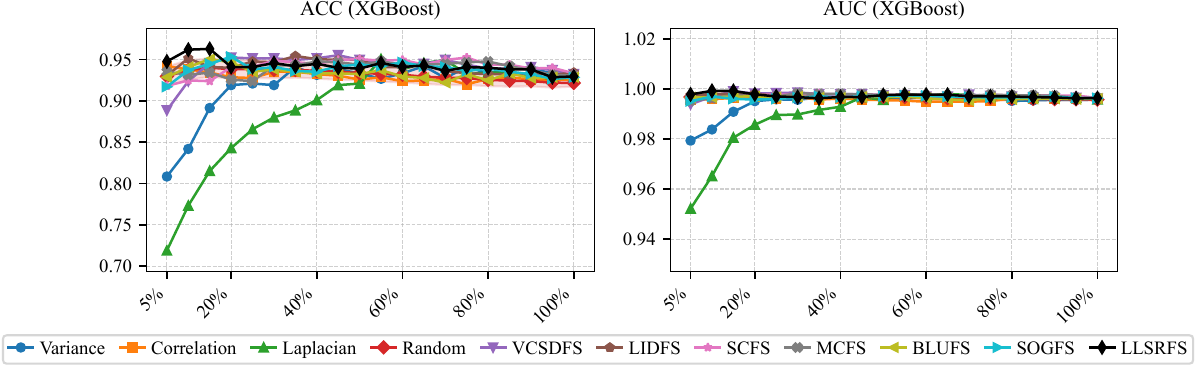}
\caption{ACC and AUC of the XGBoost classifier on the warpPIE10P dataset (5\% to 100\% of the features).}\label{fig:supplot-xgb-100-warpPIE10P}
\supback
\end{figure}
\begin{figure}[p]
\centering
\includegraphics[width=\textwidth,height=\supfigheight,keepaspectratio]{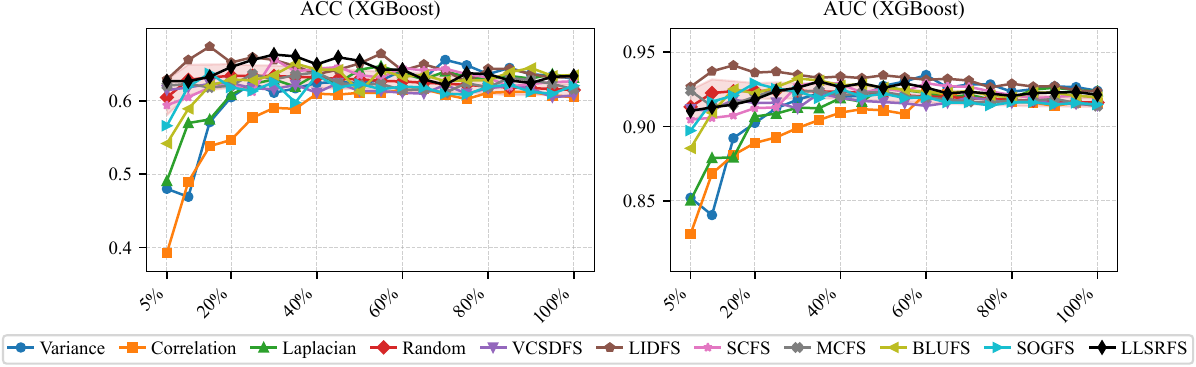}
\caption{ACC and AUC of the XGBoost classifier on the Yale dataset (5\% to 100\% of the features).}\label{fig:supplot-xgb-100-Yale}
\supback
\end{figure}
\clearpage
\end{document}